%% file: AANAS.tex
\PassOptionsToPackage{square,comma,numbers,sort&compress}{natbib}
\documentclass{article}

\usepackage[final]{neurips_2020}

\usepackage[utf8]{inputenc} 
\usepackage[T1]{fontenc}    
\usepackage[colorlinks,bookmarks=false,urlcolor=black]{hyperref}
\usepackage{url}            
\usepackage{booktabs}       
\usepackage{amsfonts}       
\usepackage{nicefrac}       
\usepackage{microtype}      

\usepackage{amsfonts}       
\usepackage{algorithm}
\usepackage{algorithmic}
\usepackage{amsmath}
\usepackage{graphicx}
\usepackage{multirow}
\usepackage{bm}
\usepackage{color}
\usepackage{amsthm}
\usepackage{tabularx}
\usepackage{wrapfig}
\usepackage{footnote}
\usepackage{threeparttable}
\usepackage{tablefootnote}
\makesavenoteenv{tabular}
\makesavenoteenv{table}
\usepackage{caption}
\usepackage{fancyhdr}
\usepackage{amssymb}
\usepackage{pifont}
\usepackage{xr}
\usepackage{wrapfig}
\usepackage{setspace} 

\usepackage{exscale}
\usepackage{relsize}
\usepackage{xr}

\newcommand{\comment}[1]{}

\newcommand{\et}{\emph{et al.}}
\newcommand{\ie}{\emph{i.e.}}
\newcommand{\eg}{\emph{e.g.}}
\input{MACPdef-ams_manu.tex}
\input{MACPdef_manu.tex}

\title{Theory-Inspired Path-Regularized Differential Network Architecture Search}

\author{ 
	Pan Zhou  \quad  \normalsize{Caiming Xiong}  \quad    \normalsize{Richard Socher} \quad   \normalsize{Steven C.H. Hoi}\\
	{Salesforce Research } \\
	{ \texttt \tt \{pzhou, cxiong, rsocher, shoi\}@salesforce.com}  
}

\begin{document}

\maketitle

\begin{abstract}
	
	Despite its high search efficiency, differential architecture search (DARTS) often selects network architectures with dominated skip connections which lead to performance degradation. However, theoretical understandings on this issue remain absent, hindering the development of more advanced methods in a principled way. In this work, we solve this problem by theoretically analyzing the effects of various types of operations, \eg~convolution, skip connection and zero operation, to the  network  optimization.  
	We prove that the architectures with more skip connections can converge faster  than the other candidates,  and thus are selected by DARTS. This result, for the first time, theoretically and explicitly reveals the impact of skip connections to fast network optimization and its competitive advantage over other types of operations in DARTS. Then we propose a theory-inspired path-regularized DARTS that consists of two key modules: (i) a differential group-structured sparse binary gate introduced for each operation to avoid unfair competition among operations, and (ii) a path-depth-wise regularization used to incite search exploration for deep architectures that often converge slower than shallow ones as shown in our theory and are not well explored during search. Experimental results on image classification tasks validate its advantages.  
\end{abstract}

\vspace{-1em}
\section{Introduction}\label{introduction}
\vspace{-0.4em}
Network architecture search (NAS)~\cite{zoph2016neural} is an effective approach for automating  network architecture design, with many successful applications witnessed to  image recognition~\cite{zoph2018learning,pham2018efficient,real2019regularized,tan2019mnasnet,liu2018darts} and language modeling~\cite{zoph2016neural,liu2018darts}. The methodology of NAS is to automatically search for a directed graph and its  edges  from a huge search space. Unlike expert-designed architectures which  require  substantial efforts from  experts by trial  and error, the automatic principle in NAS greatly alleviates   these design efforts  and possible design bias brought by  experts which could  prohibit achieving better performance.  Thanks to these advantages, NAS has been widely devised via reinforcement learning (RL) and  evolutionary algorithm (EA),   and    achieved  promising results in many applications, \eg~classification~\cite{zoph2018learning,real2019regularized}.

\begin{figure}[tb]
	\begin{center}
		\setlength{\tabcolsep}{0.0pt} 
		\begin{tabular}{c}
			{\hspace{-2pt}}
			\includegraphics[width=1\linewidth]{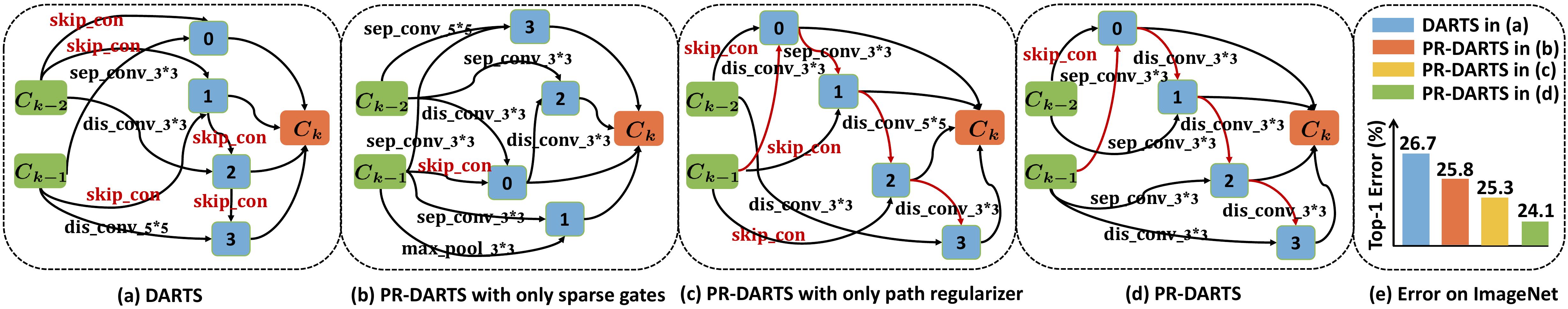}\\ 
		\end{tabular}
	\end{center}
	\vspace{-0.7em}
	\caption{Illustration of selected normal cells by DARTS and PR-DARTS. By comparison, the group-structured  sparse gates in PR-DARTS (b) well alleviate  unfair operation competition and overcome  the dominated-skip-connection  issue in DARTS (a); path-depth-wise regularization in PR-DARTS (c)  helps rectify  cell-selection-bias to shallow cells; PR-DARTS (d) combines these two complementary components and  well alleviates the above two issues,   testified by the results in (e).  
	}
	\label{illustrationcomponents}
	\vspace{-1.0em}
\end{figure}

DARTS~\cite{liu2018darts} is a recently developed  leading approach. Different from RL and EA based methods~\cite{zoph2016neural,zoph2018learning,pham2018efficient,real2019regularized} that discretely optimize  architecture parameters, DARTS converts the  operation selection for each edge in the directed graph into continuously   weighting a fixed set of operations. In this way, it  can optimize the architecture parameters via gradient descent and greatly reduces the high search cost in RL and EA approaches.  However, as observed in  the literatures~\cite{chen2019progressive,chu2019fair,arber2019understanding,liang2019darts} and Fig.~\ref{illustrationcomponents} (a), this differential  NAS family, including DARTS and its variants~\cite{dong2019searching,wu2019fbnet}, typically selects many skip connections which dominate over other types of operations in the network graph. Consequently,  the searched networks are observed to have unsatisfactory performance.  To alleviate this issue, some empirical techniques are developed, \eg~operation-level dropout~\cite{chen2019progressive}, fair operation-competing loss~\cite{chu2019fair}.  But no attention has been paid to developing theoretical understandings for why   skip connections dominate other types of operations in DARTS.  The theoretical answer to this question is important  not only for better understanding DARTS, but also for inspiring new insights for DARTS algorithm improvement.

\noindent{\textbf{Contributions.}}   In this work,  we address the above fundamental question and contribute to derive some new results, insights and alternatives for DARTS. Particularly, we provide rigorous theoretical  analysis for the dominated skip connections in DARTS.  Inspired by our theory, we then propose a new alternative of DARTS which can search  networks without dominated skip connections and achieves state-of-the-art classification performance. Our main contributions  are highlighted below.

Our first contribution is proving that DARTS prefers  to skip connection more than other types of operations, \eg~convolution and zero operation, in the search phase, and tends to search favor skip-connection-dominated networks as shown in Fig.~\ref{illustrationcomponents} (a). Formally,  in the search phase,    DARTS  first fixes architecture parameter $\betam$ which determines the operation weights in the graph to optimize the  network parameter $\Wm$ by minimizing  training loss $F_{\mbox{\tiny{train}}}(\Wm\!, \betaii{}{})$ via gradient descent, and then  uses the validation loss $F_{\mbox{\tiny{val}}}(\Wm\!, \betaii{}{})$ to  optimize $\betam$ via gradient descent. We prove that when  optimizing $F_{\mbox{\tiny{train}}}(\Wm\!, \betaii{}{})$,   the convergence rate at each iteration depends on the weights of skip connections much heavier than other types of operations, \eg~convolution,   meaning that the more skip connections the faster convergence.  Since training and validation data come from the same distribution which means $\EE[F_{\mbox{\tiny{train}}}(\Wm\!, \betaii{}{})]\!=\!\EE[F_{\mbox{\tiny{val}}}(\Wm\!, \betaii{}{})]$,  more skip connections can also faster decay $F_{\mbox{\tiny{val}}}(\Wm\!, \betaii{}{})$ in expectation. So when updating architecture parameter $\betam$, DARTS will tune the weights of skip connections larger to faster decay  validation loss, and meanwhile, will  tune the weights of other  operations smaller since all types of operations on one edge share a softmax distribution. Accordingly, skip connections gradually dominate the  network graph.  To our best knowledge, this is the first theoretical result that explicitly shows heavier dependence of the convergence rate of  NAS  algorithm on   skip connections, explaining the dominated skip connections in DARTS due to their   optimization advantages.

Inspired by our theory, we further develop the path-regularized DARTS (PR-DARTS) as a novel alternative to alleviate unfair  competition between skip connection and other types of operations  in DARTS.   To this end, we define a   
group-structured sparse binary gate implemented by   Bernoulli distribution for  each operation. These gates independently determine whether   their corresponding operations are used in the graph. Then we divide all operations in the graph into   skip connection group and non-skip connection group, and independently regularize the gates in these two groups to be sparse via a hard threshold function.   This group-structured sparsity  penalizes the  skip connection group heavier than another group to rectify the  competitive advantage of skip connections over other operations as shown in Fig.~\ref{illustrationcomponents} (b), and globally and  gradually  prunes  unnecessary connections in the search phase to reduce  the  pruning information loss after searching.   
More importantly, we introduce a path-depth-wise regularization which encourages large activation probability of gates along  the  long paths in the network graph  and thus  incites more search  exploration to deep graphs illustrated by Fig.~\ref{illustrationcomponents} (c). As our theory shows that gradient descent can faster optimize shallow and wide networks  than deep and thin ones, this path-depth-wise regularization can rectify the competitive advantage of shallow network over deep one.   
So PR-DARTS can  search  performance-oriented networks instead of fast-convergence-oriented networks and achieves better performance testified by Fig.~\ref{illustrationcomponents} (e).

\section{Related Work}

DARTS~\cite{liu2018darts} has gained much attention recently thanks to its high search  efficiency~\cite{chen2019progressive,chu2019fair,arber2019understanding,liang2019darts,dong2019searching,wu2019fbnet,cai2018proxylessnas,xie2018snas,xu2019pc,yang2020cars,guo2020breaking}.  It relaxes a  discrete search space to a continuous one via continuously weighting the operations, and then employs  gradient descent algorithm to select promising candidates. In this way, it significantly improves the search efficiency over RL and EA based NAS approaches~\cite{zoph2016neural,zoph2018learning,pham2018efficient,real2019regularized}.  
But the selected networks by DARTS have dominated skip connections which lead  to unsatisfactory performance~\cite{chen2019progressive,chu2019fair,arber2019understanding,liang2019darts}.  To solve this issue, Chen~\et~\cite{chen2019progressive}  introduced operation-level dropout~\cite{srivastava2014dropout}  to regularize  skip connection.  Chu~\et~\cite{chu2019fair} used independent sigmoid function for weighting each operation to avoid  operation  competition, and  designed a new loss to independently push the operation weights to zero or one.  In contrast, our PR-DARTS uses  binary gate for each operation and then imposes  group-structured and path-depth-wise regularizations to alleviate  the fast-convergence-oriented search issue  in DARTS.

The intrinsic theoretical reasons for the dominated skip connection in DARTS are rarely investigated though heavily desired.   Zela~\et~\cite{arber2019understanding}  empirically analyzed the poor generalization performance of the selected architectures by DARTS from the argument of sharp and flat minima.  
Shu~\et~\cite{shu2019understanding} studied general NAS and showed that NAS prefers to shallow and wide networks since these networks have more smooth landscape empirically and smaller gradient variance which both boost training speed. But they did not reveal any relation between skip connections and convergence behaviors. Differently, we explicitly show the role of weights of different operations in determining the convergence rate in network optimization, revealing the intrinsic reasons for the dominated skip connections in DARTS.  
 
\section{Theoretical Analysis for DARTS}\label{analysispart}
 
In this section, we first recall the formulation of DARTS, and then theoretically analyze the intrinsic reasons for the  dominated skip connections in DARTS by analyzing its convergence behaviors.
 
\subsection{Formulation of DARTS} 
 
DARTS~\cite{liu2018darts} searches cells which are used to stack  the full network architecture. A cell is organized as a directed acyclic graph with $h$ nodes  $\{\Xnii{l}{}\}_{l=0}^{h-1}$.  Typically, the graph contains two input nodes $\Xnii{0}{}$ and $\Xnii{1}{}$ respectively defined as the outputs of two previous cells,  and has one output node $\Xnii{h-\!1}{}$ giving by concatenating  all intermediate  nodes $\Xnii{l}{}$.   Each intermediate  node $\Xnii{l}{}$   connects with all previous nodes $\Xnii{s}{}\ (0\!\leq\!s\!<\!l)$ via a continuous operation-weighting     strategy, namely 
\begin{equation}\label{equationoperation} 
{\Xnii{l}{} = \sum\nolimits_{0\leq s <l} \sum\nolimits_{t=1}^{r}   \alphanii{l}{s,t} O_{t}\big(\Xnii{s}{} \big) \quad \text{with}\quad  \alphanii{l}{s,t}= \exp(\betanii{l}{s,t})/\sum\nolimits_{t=1}^{r} \exp(\betanii{l}{s,t})},
\end{equation}
where the operation $O_{t} $ comes  from the operation set $\Om=\{O_{t}\}_{t=1}^{r}$, including zero operation, skip connection, convolution, etc. In this way, the architecture search problem becomes efficiently learning continuous architecture parameter  $\betaii{}{}=\{\betanii{l}{s,t}\}_{l,s,t}$ via optimizing the following bi-level model
\begin{equation}\label{dartsmodel} 
\min\nolimits_{\betam}\ F_{\mbox{\tiny{val}}}(\Wm^*(\betaii{}{}), \betaii{}{}),\qquad \mbox{s.t.}\ \Wm^*(\betaii{}{})=\argmin\nolimits_{\Wm}\ F_{\mbox{\tiny{train}}}(\Wm, \betaii{}{}),
\end{equation}
where $F_{\mbox{\tiny{train}}}$ and  $F_{\mbox{\tiny{val}}}$ respectively denote the loss on the training and validation datasets, $\Wm$ is the  network parameters in the graph, \eg~convolution parameters.   Then DARTS optimizes the architecture parameter $\betam$ and the network parameter $\Wm$ by alternating gradient descent. After learning $\betam$, DARTS prunes the dense graph according to the  weight  $\alphanii{l}{s,t}$ in Eqn.~\eqref{equationoperation} to obtain compact cells.

Despite  its much higher search efficiency over RL and EA based  methods,  DARTS typically selects a cell with dominated skip connections, leading to unsatisfactory performance~\cite{chen2019progressive,chu2019fair,arber2019understanding,liang2019darts}.  But  there is no  rigorously theoretical analysis  that explicitly justifies  why DARTS tends to favor skip connections. The following section  attempts to solve this issue by analyzing the convergence behaviors of DARTS.

\subsection{Analysis Results for DARTS}\label{DARTSanalysis}

For analysis, we detail the cell structures in  DARTS. Let  input  be $\Xm\in \Rs{\bar{\m}\times \bar{\p}}$ where $\bar{\m}$ and $\bar{\p}$ are respectively  the channel number and dimension of input. Typically,  one  needs to resize the input  to a target size {$\m\times \p$} via a convolution layer with parameter   $\Wmii{(0)}{}\in\Rs{m\times \kc \bar{\m}}$ (kernel size {$\kc\times \kc$})
\begin{equation}\label{convolutiondefinition} 
\Xnii{0}{} = \convo{\Wnii{0}{}, \Xm} \in\Rs{m\times p}\quad \text{with}\quad \convo{\Wm; \Xm}= \tau \sigmai{\Wm \Phi(\Xm)},
\end{equation} 
and  then feed it into the subsequent layers. The convolution operation $\convs$  performs convolution and then  nonlinear mapping via activation function {$\sigma$}.  The scaling factor {$\tau$} equals to $\frac{1}{\sqrt{\bar{m}}}$ when channel number in $\convs$  is {$\bar{m}$}. It  is introduced to simplify the notations in our analysis and does not affect convergence behaviors of DARTS.   For notation simplicity,  we assume stride {$\sco$}~$\!=\!1$ and padding zero {$\pc$}~$\!=\!\frac{\kc-1}{2}$ to  make  the same sizes of output and input. Given a matrix $ \Zm\!\in\! \Rs{\m\times \p}$,   $\Phi(\Zm)$  is defined as   
\begin{equation*} 
\Phi(\Zm) \!=\!\!\! \begin{bmatrix} \Zm_{1,-\pc+1: \pc+1}^\top  &\!\!\!\! \Zm_{1,-\pc+2:\pc+2}^\top  &\!\!\!\! \cdots &\!\!\!\!\Zm_{1, p-\pc: p+\pc}^\top  \\
\Zm_{2,-\pc+1: \pc+1}^\top   &\!\!\!\!\Zm_{2,-\pc+2:\pc+2}^\top  &\!\!\!\! \cdots &\!\!\!\! \Zm_{2, p-\pc: p+\pc}^\top  \\
\vdots & \vdots & \ddots &\vdots\\
\Zm_{m,-\pc+1: \pc+1}^\top   &\!\!\!\! \Zm_{m,-\pc+2:\pc+2}^\top  &\!\!\!\!  \cdots &\!\!\!\! \Zm_{m, p-\pc: p+\pc}^\top  \\
\end{bmatrix} \!\!\!\in\! \Rs{\kc m\times p},
\end{equation*}
where $\Zm_{i,t} =0\  (t \leq 0$ or $ t>p)$.  
Then the conventional convolution can be computed as $\Wm \Phi(\Xm)$ where each row in $\Wm$ denotes a conventional  kernel.   Note, for other convolutions, e.g.  depth-wise separable convolution, our analysis framework still holds and can derive very similar  results.  Now we are ready to define the subsequent layers in the cell:
\begin{equation}\label{connections}
{\Xnii{l}{}  \!=\! \sum\nolimits_{s=0}^{l-1}    \big(\alphanii{l}{s,1}\zeroo{\Xm} \!+\!\alphanii{l}{s,2} \skipo{\Xm} \!+\! \alphanii{l}{s,3} \convo{\Wnii{l}{s};\Xnii{s}{}}\big)  \! \in\!\Rs{m\times p}\ (l\!=\!1,\cdots\!,h-1), }
\end{equation}
where zero operation $\zeroo{\Xm}=\bm{0}$ and skip connection   $ \skipo{\Xm}  =\Xm$,  $\alphaii{(l)}{s,t}$ is given in~\eqref{equationoperation}.  In this work, we consider three representative operations, \ie~zero, skip connection and convolution, and ignore pooling operation since  it reveals the same behaviors as convolution, namely both being dominated by skip connections~\cite{chen2019progressive,chu2019fair,arber2019understanding}.  
Next, we feed concatenation of all intermediate nodes into a linear layer to obtain the prediction {$\ui{i}$} of the {$i$}-th sample $\Xmi{i}$ and then obtain a mean squared loss:
\begin{equation}\label{loss} 
{F(\Wm, \betam) = \frac{1}{2n}\sum\nolimits_{i=1}^{n} (\ui{i}-\ymi{i})^2 \quad \text{with}\quad \ui{i}=  \sum\nolimits_{s=0}^{h-1}  \langle \Wmi{s}, \Xnii{s}{i} \rangle \in\Rs{}, }
\end{equation}
where $\Xnii{s}{i}$ denotes the {$s$}-th feature node for sample $\Xmi{i}$, $ \{\Wmi{s}\}_{s=0}^{h-1}$ denote  the parameters for the linear layer.    $F(\Wm, \betam)$ becomes $F_{\mbox{\tiny{train}}}(\Wm, \betam)$ ($F_{\mbox{\tiny{val}}}(\Wm, \betam)$) when samples come  from training dataset (validation dataset).   
Subsequently, we   analyze the effects of various types of  operations to the convergence behaviors of  $F_{\mbox{\tiny{train}}}(\Wm\!, \betam)$ when optimize the network parameter $\Wm$ via gradient descent: 
\begin{equation}\label{updation} 
\Wmii{(l)}{s}\!(k\!+\!1) \!=\! \Wmii{(l)}{s}\!(k)-\eta \nabla_{\!\Wmii{(l)}{s}\!(k)}\! F_{\mbox{\tiny{train}}}(\Wm\!,\! \betam)\ (\forall l,\! s),\ \ \Wmi{s}(k\!+\!1) \!=\!\Wmi{s}(k) -\eta \nabla_{\!\Wmi{s}(k)} F_{\mbox{\tiny{train}}}(\Wm\!,\! \betam)\ (\forall s),
\end{equation}
where {$\eta$} is the learning rate. We use gradient descent instead of stochastic gradient descent, since gradient descent is expectation version of stochastic one and can reveal similar convergence behaviors.  For  analysis, we first introduce mild assumptions widely used in stochastic optimization~\cite{Rakhlin12,zhou2019Riemannian,zhou2018HSGDHT,zhou2020hybird} and network  analysis~\cite{du2018gradient,du2018gradient22,allen2018convergence,tian2017analytical,zhou2018analysiscnns,zhou2018analysisdnn,zhou2019metalearning,zhou2020sgd}.   
\begin{assum} \label{activationassumption}
	Assume the activation function {$\sigma$}  is {$\mu$}-Lipschitz  and {$\rho$}-smooth.  That is, for {$\forall x_1, x_2$}, {$\sigma$}  satisfies {$|\sigmai{x_1} -\sigmai{x_2}|\leq \mu |x_1-x_2|$} and   {$|\sigma'(x_1) -\sigma’(x_2)|\leq \rho |x_1-x_2|$}. Moreover, we assume that{$\sigma$}$(0)$ can be upper bounded, and {$\sigma$} is analytic and is not a polynomial function. \vspace{-0.3em}
\end{assum}
\begin{assum}\label{initilizationassumption}
	Assume the initialization of the convolution  parameters ($\Wmii{(l)}{s}$) and the linear mapping parameters ($\Wmi{s}$)  are drawn from Gaussian distribution  $\N(\bm{0}, \Imm)$.\vspace{-0.1em}
\end{assum}
\begin{assum}\label{sampleassumption}
	Suppose the samples $\{\Xmi{i}\}_{i=1}^n$ are  normalized such that $\|\Xmi{i}\|_F= 1$. Moreover, they are not parallel, namely $\vect{\Xmi{i}}\notin \text{span}(\vect{\Xmi{j}})$ for all $i\neq j$, where $\vect{\Xmi{i}}$  vectorizes  $\Xmi{i}$.\vspace{-0.5em}
\end{assum}
Assumption~\ref{activationassumption} is mild, since most differential activation functions, \eg~softplus and sigmoid, satisfy it. The Gaussian assumption on initial  parameters  in Assumption~\ref{initilizationassumption} is used in practice. We assume  Gaussian variance to be one for notation simplicity in analysis, but our technique is applicable to any constant variance.  The normalization and non-parallel conditions in Assumption~\ref{sampleassumption}  are satisfied in practice, as normalization is a data preprocess and samples in a  dataset are often not restrictively parallel.  Based on assumptions, we summarize our  result in Theorem~\ref{mainconvergence3} with proof in Appendix~\ref{proofofmainconvergence3}.

\begin{thm}\label{mainconvergence3} 
	Suppose Assumptions~\ref{activationassumption}, \ref{initilizationassumption} and~\ref{sampleassumption} hold.  Let {$c$}~$\!=\!\left(1\!+\!\alphaii{}{2} \!+\! 2\alphaii{}{3}  \mu \sqrt{\kc} \cwo \right)^{h}$, $\alphaii{}{2}\!=\! \max_{s,l}\alphanii{l}{s,2}$ and $\alphaii{}{3}\!=\!\max_{s,l} \alphanii{l}{s,3}$.	If   $m\!\geq\!  \frac{c_m \mu^2  }{\lambda^2}   \left[ \rho p^2 n^2\log(n/\delta) \!+\! c^2  \kc^2  \cwo^2/n   \right]$ and $\eta \!\leq\!  \frac{c_\eta \lambda}{\sqrt{m} \mu^4   h^{3}  \kc^2 c^4}$, 
	where {$\cwo$}, {$ c_m$},  {$c_\eta$} are  constants, $\lambda$ is given below. Then when fixing  architecture parameterize $\alpham $ in~\eqref{equationoperation}  and  optimizing  network parameter $\Wm$ via gradient descent~\eqref{updation}, with probability at least $1-\delta$ we have 
	\begin{equation*} 
	F_{\mbox{\tiny{train}}}(\Wm(k+1), \betam)   \leq \left( 1 - \eta  \lambda/4\right)F_{\mbox{\tiny{train}}}(\Wm(k), \betam) \quad (\forall k\geq 1),
	\end{equation*} 
	where $\lambda 
	= \frac{3c_{\sigma}}{4} \lambda_{\min}(\Km)\sum_{s=0}^{h-2}(\alphanii{h-1}{s,3} )^2
	\prod_{t=0}^{s-1}  (\alphanii{s}{t,2})^2 
	$,  the positive constant {$c_\sigma$} only depends on {$\sigma$} and  input data,   $\lambda_{\min}({\Km})=\min_{i,j} \lambda_{\min}(\Km_{ij}) $ is larger than zero  in which  $\lambda_{\min}(\Km_{ij}) $ is  the smallest eigenvalue of $\Km_{ij} =  \begin{bmatrix}
	\Xmi{i}^{\top} \Xmi{j}, \Xmi{i}^{\top} \Xmi{j};
	\Xmi{j}^{\top} \Xmi{i}, \Xmi{j}^{\top} \Xmi{j} 
	\end{bmatrix}$.  
\end{thm}

Theorem~\ref{mainconvergence3} shows that for an architecture-fixed over-parameterized network, when using gradient descent  to optimize the network parameter $\Wm$, one can expect the convergence of the algorithm which is consistent with prior deep learning optimization work~\cite{du2018gradient,du2018gradient22,allen2018convergence,tian2017analytical}. More importantly, the convergence rate at each iteration depends on the network architectures which is parameterized by $\alpham$.

Specifically,  for each  factor $\lambda_s\!=\!(\alphanii{h-1}{s,3} )^2\prod_{t=0}^{s-1}  (\alphanii{s}{t,2})^2$ in the factor $\lambda$, it is induced by the connection path $\Xnii{0}{}\!\rightarrow\!\Xnii{1}{}\!\rightarrow\!\cdots \!\rightarrow\! \Xnii{s}{}\!\rightarrow\!\Xnii{h-1}{}$. By observing  $\lambda_s$, one can find that (1) for the connections before node $\Xnii{s}{}$,  it depends on the weights  $\alphanii{s}{t,2}$ of skip connections  heavier than  convolution and zero operation, and (2) for the direct connection between  $\Xnii{s}{}$ and $\Xnii{h-1}{}$, it relies on convolution  weight $\alphanii{h}{s,3}$ heavier than the weights of other type operations.  For observation (1), it can be intuitively understood: as shown in~\cite{he2016deep,he2016identity,orhan2017skip,balduzzi2017shattered}, skip connection  often provides larger gradient flow than the parallel convolution  and zero connection  and thus greatly benefits  faster convergence of networks, since skip connection maintains primary information flow, while  convolution  only learns the residual information and zero operation does not delivery any information.  
So convolution and zero operations have  negligible contribution to information flow and thus their weights do not occur in $\prod_{t=0}^{s-1}  (\alphanii{s}{t,2})^2$ of $\lambda_s$. For observation (2), as the path $\Xnii{0}{}\!\rightarrow\!\Xnii{1}{}\!\rightarrow\!\cdots \!\rightarrow\! \Xnii{s}{}$ is shared for all subsequent layers, it prefers  skip connection more to maintain information flow, while for  the private  connection between  $\Xnii{s}{}$ and $\Xnii{h-1}{}$ which is not shared since $\Xnii{h-1}{}$ is the last node, it  relies on learnable  convolution  more heavily than non-parameterized operations, since learnable operations have parameter to learn and can reduce the loss.  For the theoretical reasons for observations (1) and (2),  the skip connection in the shared  path   can improve the singularity of network  Gram matrix more than other types of operations, where the singularity directly determines the convergence rate, while the learnable convolution  in private path can benefit the Gram matrix singularity  much more. See details in Appendix~\ref{Proofoflinearconvergene}. The weight $\alphanii{l}{s,3}$ of zero operation does not occur in $\lambda$, as it does not delivery any information. 
 
\begin{wrapfigure}{r}{0.30\linewidth}
	\vspace{-1.5em}
	\begin{center}
		\setlength{\tabcolsep}{0.0pt}  
		\begin{tabular}{cc}
			\includegraphics[width=\linewidth]{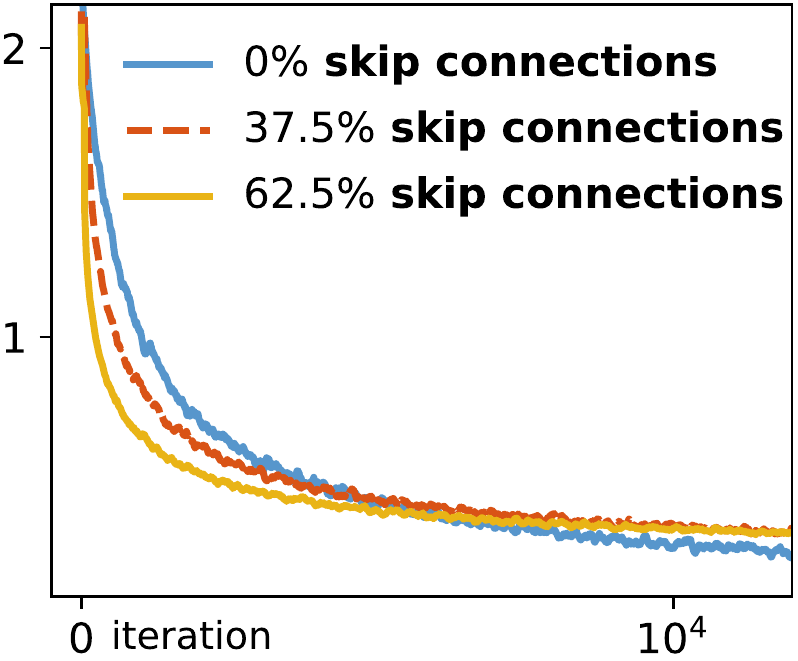}  \\ 
		\end{tabular}
	\end{center}
	\vspace{-1.0em}
	\caption{Effects of skip connections to convergence rate of network.}
	\label{skipfaster}
	\vspace{-2.0em}
\end{wrapfigure}

Now we     analyze why the selected cell has dominated skip connections.   The above analysis shows that  the convergence rate when optimizing $F_{\mbox{\tiny{train}}}(\Wm,\betam)$ depends on the weights of skip connections heavier than other weights  in the shared connection path which dominates  the connections of a cell.  So larger weights of skip connections often give faster loss decay of $F_{\mbox{\tiny{train}}}(\Wm,\betam)$. Consider the samples for training and validation come from the same distribution which means  $\EE [F_{\mbox{\tiny{train}}}(\Wm\!,\betam)]\!=\!\EE[F_{\mbox{\tiny{val}}}(\Wm\!,\betam)]$, larger weights of skip connections can also  faster reduce $F_{\mbox{\tiny{val}}}(\Wm)$ in expectation, which accords with the empirical observations in Fig.~\ref{skipfaster} and the observations in~\cite{arber2019understanding}.   In Fig.~\ref{skipfaster}, we first set all operations in   NAS cell (normal and reduction cells, see details in Sec.~\ref{experiments})  as convolution 
($3\!\times \!3$), and randomly select $0\%$, $37.5\%$ and $62.5\%$ operations as  skip connections. Next, we stack  
8 NAS cells to build a network and train  on CIFAR10 with  same settings. Fig.~\ref{skipfaster} shows  that more skip connections gives faster convergence.    
So when optimizing   $\alpham$  via optimizing $\betam$ in $F_{\mbox{\tiny{val}}}(\Wm\!,\betam)$,  
DARTS will  tune   weights of most skip connections   
larger to faster reduce  $F_{\mbox{\tiny{val}}}(\Wm\!,\betam)$.   As the weights of three operations  on one edge share a softmax distribution in~\eqref{equationoperation}, increasing   one operation weight means reducing  other operation weights.    Thus,   skip connections  gradually dominate over other types of operations for most connections in the cell.    So when pruning  operations according to their weights,  most of skip connections are preserved while most of  other  operations are pruned. This explains the dominated skip connections in the cell selected by DARTS.

\section{Path-Regularized Differential Network Architecture Search}\label{methodpart}
The proposed method consists of two main components, \ie~group-structured  sparse stochastic gate for each operation and path-depth-wise regularization on gates, which are introduced below in turn.

\subsection{Group-structured Sparse Operation Gates}
The analysis in Sec.~\ref{DARTSanalysis} shows that skip connection has superior competing advantages over other types of operations when they share  one softmax distribution.  To resolve this issue, we introduce independent stochastic gate  for each operation between two nodes to avoid the direct competition between skip connection and other operations. Specifically, we define a stochastic binary gate $\gnii{l}{s,t}$ for the $t$-th operation between nodes $\Xnii{s}{}$ and $\Xnii{l}{}$, where $\gnii{l}{s,t} \!\sim\! \text{Bernoulli}\big( \!\exp(\betanii{l}{s,t})/(1\!+\!\exp(\betanii{l}{s,t})\big)$. Then at each iteration, we sample gate $\gii{(l)}{s,t}$   from its   Bernoulli distribution and compute each node as 
\begin{equation}\label{conenctionwaypcdarts} 
{\Xnii{l}{} = \sum\nolimits_{1\leq i <l} \sum\nolimits_{t=1}^{r} \gnii{l}{s,t} O_{t}\big(\Xnii{i}{} \big)}.
\end{equation}
Since the discrete sampling of $\gnii{l}{s,t}$ is not differentiable, we use Gumbel technique~\cite{dyson1962statistical,maddison2014sampling} to approximate $\gnii{l}{s,t}$ as $\gnbii{l}{s,t}=\Theta\big((\ln \delta - \ln(1-\delta) + \betanii{l}{s,t})/\tau\big)$ where $\Theta$ denotes sigmoid function, $\delta\sim \mbox{Uniform}(0,1)$.  For temperature {$\tau$},   when {$\tau$}$\rightarrow 0$ the approximated distribution $\gnbii{l}{s,t}$ recovers Bernoulli distribution and is  non-smooth, while when {$\tau$}$\rightarrow +\infty$,   the approximated distribution becomes very smooth. In this way, the gradient can be back-propagated through $\gnbii{l}{s,t}$ to the network parameter $\Wm$.

If there is no   regularization on the independent gates, then there are two issues. The first one is that the selected cells would have   large weights for most operations. This is because (1) as shown in Theorem~\ref{mainconvergence3}, increasing operation weights can lead to faster convergence rate; (2) increasing weights of any operations can strictly reduce or maintain the  loss which is formally stated in Theorem~\ref{lossproperty}. Let   $t_{\mbox{\tiny{skip}}}$ and $t_{\mbox{\tiny{conv}}}$  respectively be     the indexes of skip connection and convolution in the operation set $\Om$.

\begin{thm}\label{lossproperty}
	Assume the weights in DARTS model~\eqref{dartsmodel} is replaced with the independent gates $\gnii{l}{s,t}$. \\
	(1) Increasing the value of $\gnii{l}{s,t}$ of the operations, including zero operation, skip connection, pooling, and convolution with any kernel size,  can reduce or maintain the loss $F_{\mbox{\tiny{val}}}(\Wm^*(\betam), \betam)$ in \eqref{dartsmodel}. \\
	(2) Suppose the assumptions in Theorem~\ref{mainconvergence3} hold. With  probability at least $1-\delta$,  increasing   $\gnii{l}{s,t_{\mbox{\tiny{skip}}}}\ (0\!\leq \!$~{$s$}~$\!<\!l\!<h-1)$  of skip connection or   $\gnii{h-1}{s,t_{\mbox{\tiny{conv}}}}\ (0\!\leq \!$~{$s$}~$\!<\!h-1)$ of convolution with increment {$\epsilon$} can reduce the  loss $F_{\mbox{\tiny{val}}}(\Wm^*(\betam), \betam)$ in \eqref{dartsmodel} to $F_{\mbox{\tiny{val}}}(\Wm^*(\betam), \betam)- C${$\epsilon$} in expectation, where $C$ is a positive constant.\vspace{-0.5em}
\end{thm}

See its  proof in Appendix~\ref{Proofoflossproperty}. 
Theorem~\ref{lossproperty} shows that DARTS with independent  gates would tune the weights of most operations large to obtain faster convergence and smaller loss, leading to dense cells and thus  performance degradation when pruning these large weights.  The second issue is that independent gates cannot encourage  benign competition and cooperation among operations, as Theorem~\ref{lossproperty} shows most operations tend to increase their weights. Considering the performance degradation caused by  pruning  dense cells,  benign competition and cooperation among operations  are necessary for gradually pruning unnecessary   operations to obtain relatively   sparse selected cells.

To resolve these two issues,  we impose group-structured sparsity regularization on the   stochastic gates. Following~\cite{louizos2017learning} we stretch $\gnbii{l}{s,t}$ from the range $[0,1]$ to $[${$a,b$}$]$ via rescaling $\gntii{l}{s,t} \!=\!$~{$a$}$+(${$b\!-a$}$)\gnbii{l}{s,t}$,   where {$a$}~$\!<\!0$ and {$b$}~$\!>\!1$ are two constants. Then we feed   $\gntii{l}{s,t}$ into a hard threshold gate to obtain  the gate   $\gnii{l}{s,t}\!=\! \min(1,\max(0, \gntii{l}{s,t}))$.  In this way, the gate   $\gnii{l}{s,t}$ enjoys good properties, \eg~exact zero values and computable activation probability ($\Pro(\gnii{l}{s,t}\!\neq\! 0$), which are formally stated in Theorem~\ref{gateproperty}.
\begin{thm}\label{gateproperty}
	For each stochastic gate $\gii{(l)}{s,t}$, it satisfies $\gnii{l}{s,t}= 
	0$ when $\gntii{l}{s,t}\in(0,-\frac{a}{b-a}]$; $\gnii{l}{s,t}= 
	\gntii{l}{s,t}$ when $\gntii{l}{s,t}\in(-\frac{a}{b-a}, \frac{1- a}{b-a}]$; $\gnii{l}{s,t}= 
	1$ when $ \gntii{l}{s,t}\in(\frac{1- a}{b-a},1]$. Moreover, $ \Pro(\gnii{l}{s,t}\neq 0) =\Theta(\betanii{l}{s,t} - \tau \ln \frac{-a}{b} ).$ \vspace{-0.5em}
\end{thm}
See its   proof in Appendix~\ref{proofofgateproperty}.  Theorem~\ref{gateproperty} shows that the  gate $\gnii{l}{s,t}$ can achieve exact zero, which  can reduce  information loss caused by pruning at the end of search. Next based on the activation probability   $\Pro(\gii{(l)}{s,t} \!\neq\!0)$ in Theorem~\ref{gateproperty}, we design group-structured sparsity regularizations. We collect all skip connections in the cell as a skip-connection  group  and take the remaining   operations into non-skip-connection group. Then we   compute the average  activation probability of these two groups:
\begin{equation*} 
{\LL_{\mbox{\tiny{skip}}}(\betam) \!=\!\zeta \sum_{l=1}^{h-1}  \sum_{s=0}^{l-1}   \Theta \Big(\betaii{(l)}{s,t_{\mbox{\tiny{skip}}}} \!\!- \!\tau \ln \!\frac{-a}{b} \Big),\    \LL_{\mbox{\tiny{non-skip}}}(\betam) \!=\! \frac{\zeta}{r-1}\!\sum_{l=1}^{h-1}  \sum_{s=0}^{l-1}  \sum_{1\leq t \leq r, t \neq t_{\mbox{\tiny{skip}}}}\! \!\!\! \!\!\Theta \Big(\betaii{(l)}{s,t}\! -\! \tau \ln \!\frac{-a}{b} \Big)\!,}
\end{equation*}
where $\zeta\!=\!\frac{2}{h(h-1)}$. Then we respectively regularize  $\LL_{\mbox{\tiny{skip}}}$ and $\LL_{\mbox{\tiny{non-skip}}}$ by two different regularization constants $\lambda_1$ and $\lambda_2$ ($\lambda_1\!>\!\lambda_2$ in experiments).  This group-structured sparsity has three benefits: (1) penalizing skip connections heavier than other types of operations can rectify the competitive advantage of skip connections over other operations and avoids skip-connection-dominated cell;   (2) sparsity regularizer gradually and automatically prunes redundancy and unnecessary connections which reduces the information loss of pruning at the end of search;   (3)  sparsity regularizer defined on the whole cell can encourage global competition and cooperation of all operations in the cell, which differs from DARTS that only introduces local competition among the operations between two nodes.

\begin{wrapfigure}{r}{0.39\linewidth} 
	\vspace{-1.8em}
	\begin{center}
		\setlength{\tabcolsep}{0.1pt}  
		\begin{tabular}{c}
			\includegraphics[width=\linewidth]{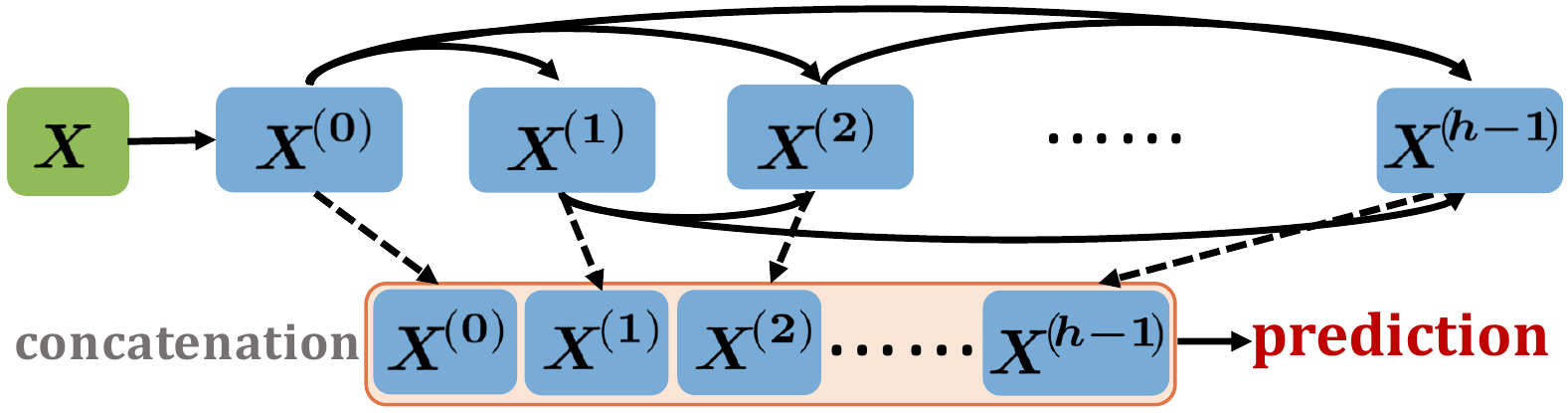}\vspace{-0.2em}\\
			{\small (a)}\vspace{-0.14em}\\
			\includegraphics[width=\linewidth]{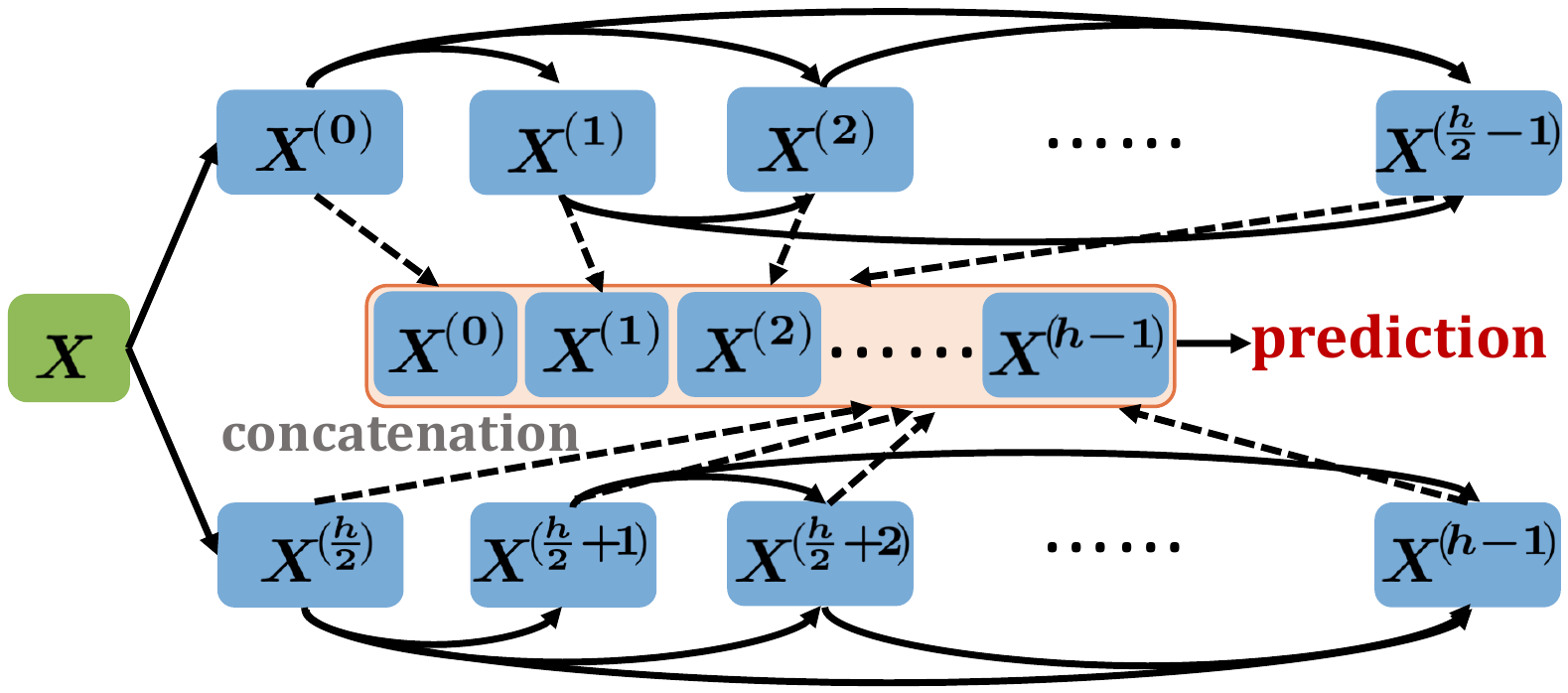}\vspace{-0.18em}\\
			{\small (b)}\vspace{-0.2em}\\
		\end{tabular}
	\end{center}
	\vspace{-1.1em}
	\caption{Illustration of a deep  cell (a)  and a shallow   cell (b).}
	\label{deepandshallow}
	\vspace{-1.2em}
\end{wrapfigure}	
\subsection{Path-depth-wise  Regularizer on Operation Gates} 
Except for the above advantages,  independent sparse gates also introduce one issue: they prohibit the method to select deep cells.  Without dominated skip connections in the cell,  other types of operations, \eg~zero operation, become freer and are widely used.  Accordingly, the search algorithm can easily transform a deep cell to a shallow  cell whose intermediate nodes  connect with input nodes via  
skip connections and whose intermediate neighboring nodes are not connected via zero operations.  
Meanwhile,  gradient descent algorithm prefers shallow cells than deep ones, as shallow cells often have more smooth landscapes and can be faster optimized. So these two factors together lead to a  bias of   search algorithm to shallow cells.  
Here we provide an example to prove the faster convergence of shallow cells. Suppose $\Xnii{l}{} (l\!=\!0,\!$ $ \cdots\!,h\!-\!1)$ are in two branches  in Fig.~\ref{deepandshallow} (b):   nodes $\Xnii{0}{} $ to $\Xnii{\frac{h}{2}-1}{}$ are in one branch with input $\Xm$ and they are connected via~\eqref{conenctionwaypcdarts}, and  $\Xnii{l}{}\ (l\!=\!\frac{h}{2},\cdots,h\!-\!1)$ are in another branch with input $\Xm$ and connection~\eqref{conenctionwaypcdarts}. Next, similar to DARTS we use all intermediate nodes to obtain a  squared loss in~\eqref{loss}.   Then we show  in Theorem~\ref{fasterconvergenceloss}  that   the shallow cell \textsf{B}  in Fig.~\ref{deepandshallow} (b) enjoys much faster convergence than the deep cell \textsf{A} in Fig.~\ref{deepandshallow} (a).  Note for cell \textsf{B}, when its node $\Xnii{h/2}{}$ connects with  node $\Xnii{l}{} (l<h/2-1)$, we have very similar results.     

\begin{thm}\label{fasterconvergenceloss}
	Suppose the assumptions in Theorem~\ref{mainconvergence3} hold and for each  $\gnii{l}{s,t}\ (0\!\leq\! s <\! l \!\leq \!h-1)$ in deep cell \textsf{A},  it has the same value in  shallow cell \textsf{B}  if it occurs in \textsf{B}. When  optimizing  $\Wm$ in $F_{\mbox{\tiny{train}}}(\Wm,\betam)$ via gradient descent~\eqref{updation},  both  losses of  cells \textsf{A} and \textsf{B} obey $F_{\mbox{\tiny{train}}}(\Wm(k\!+\!1), \betam) \!  \leq \!\left( 1 \!-\! \eta  \lambda'/4\right)F_{\mbox{\tiny{train}}}(\Wm(k), \betam)$, where $\lambda'$ in \textsf{A} is defined as   $\lambda_{\mbox{\tiny{\textsf{A}}}}
	= \frac{3c_{\sigma}}{4} \lambda_{\min}(\Km)\ \sum_{s=0}^{h-2}(\gnii{h-1}{s,3} )^2
	\prod_{t=0}^{s-1}  (\gnii{s}{t,2})^2  $, while  $\lambda'$  in \textsf{B} becomes  $\lambda_{\mbox{\tiny{\textsf{B}}}} $ and obeys $\lambda_{\mbox{\tiny{\textsf{B}}}} \geq \lambda_{\mbox{\tiny{\textsf{A}}}} + 
	\frac{3c_{\sigma}}{4} \lambda_{\min}(\Km) \sum_{s=h/2}^{h-1} (\gnii{h-1}{s,3} )^2
	\prod_{t=h/2}^{s-1}   (\gnii{s}{t,2})^2 > \lambda_{\mbox{\tiny{\textsf{A}}}} 
	$. \vspace{-0.3em}
\end{thm}

See its  proof in Appendix~\ref{Proofoffasterconvergenceloss}. 
Theorem~\ref{fasterconvergenceloss} shows that when using gradient descent  to optimize the inner-level   loss $F_{\mbox{\tiny{train}}}(\Wm,\betam)$ equipped with independent gates,  shallow  cells  can  faster reduce the  loss  $F_{\mbox{\tiny{train}}}(\Wm,\betam)$ than deep cells. As   training and validation data come from the same distribution which means  $\EE [F_{\mbox{\tiny{train}}}(\Wm\!,\betam)]\!=\!\EE[F_{\mbox{\tiny{val}}}(\Wm\!,\betam)]$,  shallow  cells  reduce $F_{\mbox{\tiny{val}}}(\Wm,\betam)$ faster in expectation which accords with the theoretical and empirical results in~\cite{sankararaman2019impact}.  So it is likely that to avoid deep cells, search algorithm would connect  intermediate nodes with  input nodes and cut the connection between  neighboring  nodes via zero operation, which is indeed illustrated by Fig.~\ref{illustrationcomponents}  (b). But it leads to cell-selection bias in the search phase, as some cells  that   fast decay the loss $F_{\mbox{\tiny{val}}}(\Wm,\betam)$ at the current iteration  have   competitive advantage over  other cells that reduce $F_{\mbox{\tiny{val}}}(\Wm,\betam)$ slowly currently  but can achieve superior final performance. This prohibits us to search good cells.  

\makebox[\textwidth][s]{To resolve this cell-selection bias, we propose a path-depth-wise regularization to   rectify the unfair}

competition between shallow and deep cells. From Theorem~\ref{gateproperty},  the  probability that  $\Xnii{l}{}$ and $\Xnii{l+1}{}$ are connected by parameterized  operations $\Om_{\mbox{\tiny{p}}}$,  \eg~various types of convolutions, is $\Pro_{l,l+1}(\betam)$ $= \! \sum_{\Om_t \in \Om_{\mbox{\tiny{p}}}} \!\Theta\big(\betanii{l+1}{l,t}\! -\! \tau \ln \frac{-a}{b} \big)$.  So  the probability that all neighboring nodes $\Xnii{l}{}$ and $\Xnii{l+1}{}$  ($l=0,\cdots,h-1$) are connected via  operations $\Om_{\mbox{\tiny{p}}}$, namely, the probability of the path of depth $h$,  is 
\begin{equation}\label{pathregularization} 
{ \LL_{\mbox{\tiny{path}}}(\betam) =\prod\nolimits_{l=1}^{h-1}\Pro_{l,l+1}(\betam) =\prod\nolimits_{l=1}^{h-1} \sum\nolimits_{O_t \in \Om_{p}} \Theta\big(\betaii{(l+1)}{l,t} - \tau \ln \frac{-a}{b} \big).}
\end{equation}
Here we do not consider  skip connection, zero and pooling operations, as they indeed make  a network shallow. To rectify the   competitive advantage of shallow cells over deep ones, we impose path-depth-wised regularization $- \LL_{\mbox{\tiny{path}}}(\betam)$ on the stochastic gates to encourage more exploration to deep cells   and then decide the depth of cells instead of greedily choosing shallow cell at the beginning of   search.

Now we are ready to define our proposed PR-DARTS model  as follows:
\begin{equation*}\label{PCdartsmodel} 
\min_{\betam} F_{\mbox{\tiny{val}}}(\Wm^*\!(\betaii{}{}), \betaii{}{}) + \lambda_{1} \LL_{\mbox{\tiny{skip}}}(\betam)  +  \lambda_{2} \LL_{\mbox{\tiny{non-skip}}}(\betam)   - \lambda_{3} \LL_{\mbox{\tiny{path}}}(\betam),\   \mbox{s.t.} \Wm^*\!(\betaii{}{})\!=\!\argmin\nolimits_{\Wm} F_{\mbox{\tiny{train}}}(\Wm, \betaii{}{}),
\end{equation*}
where 
$\Wm$ denotes network parameters, $\betaii{}{}$ denotes the parameters for the stochastic gates. Similar to  DARTS,  we  alternatively update  parameters $\Wm$ and  $\betaii{}{}$ via gradient descent. See optimization details in   Algorithm~\ref{searchalgorithm} of  Appendix~\ref{lamgugetask}.   After searching, following DARTS, we  prune   redundancy connections according to the activation probability in Theorem~\ref{gateproperty} to obtain more compact cells.

\vspace{-0.8em}
\section{Experiments}\label{experiments}
\vspace{-0.4em}
Here   we evaluate  PR-DARTS on  classification task and  compare it with  representative state-of-the-art NAS approaches, including RL based NAS, EA based NAS and differential NAS methods. Code is available at \url{https://panzhous.github.io/}.
 
\textbf{Datasets.}   
CIAFR10~\cite{krizhevsky2009learning} and CIFAR100~\cite{krizhevsky2009learning} contain 50K training and 10K test images which are of size $32\times32$ and   distribute over  10 classes in CIFAR10 and 100 classes in CIFAR100.  ImageNet~\cite{russakovsky2015imagenet} has  1.28M training and 50K test images  which  roughly equally distribute over 1K object categories.  

\textbf{Implementations.}  
For searching, each cell contains two input nodes (outputs of two previous cells), four intermediate nodes and one output node (concatenation of all intermediate nodes). Then we stack $k$ cells  for search. The $k/3$- and $2k/3$-th cells are reduction cells in which all operations have a stride of two, and the remaining cells are normal cells with operation stride of one.  Reduction cells share the same architecture and normal cells also have the same architecture.  The operation set $\Om$  has eight choices: zero operation, skip connection, $3\!\times\! 3$ and $5\!\times\! 5$ separable convolutions, $3\!\times\! 3$ and $5\!\times \!5$ dilated separable convolutions, $3\!\times \!3$ average pooling and $3\!\times \!3$ max pooling.  For fairness, all above settings follow the convention~\cite{zoph2016neural,zoph2018learning,real2019regularized,liu2018darts}. 
For each cell,  we use the input node which is the output of the previous cell to construct the path-depth-wise regularization in \eqref{pathregularization}  as illustrated by Fig.~\ref{illustrationcomponents} (c).  

\begin{savenotes}
	\begin{table*}[tp]
		\begin{threeparttable}[b]
			\caption{Classification errors ($\%$) on CIFAR10 (C10) and CIFAR100 (C100).\vspace{-0.5em}}
			\setlength{\tabcolsep}{5.9pt}  
			\renewcommand{\arraystretch}{0.7} 
			\label{CIFARcomparsion} \begin{center}
				{ \fontsize{8.1}{3}\selectfont{
						\begin{tabularx}{\textwidth}{c|cc ccc cc}\toprule
							\multirow{2}{*}{\textbf{Architecture}} &\multicolumn{2}{c}{ \textbf{Test Error ($\%$)}} & \textbf{Params} &  \textbf{Search Cost} & \textbf{Search space} & \textbf{Search}  \\
							& \textbf{C10} & \textbf{C100} &  \textbf{(M)}	 & \textbf{(GPU-days)} & \textbf{$\#$Ops$/$zero} & \textbf{method}\\
							\midrule 
							DenseNet-BC~\cite{huang2017densely} &3.46 & 17.18& 25.6 & --- &---& manual  \\ 
							\midrule
							NASNet-A + cutout~\cite{zoph2018learning} & 2.65 & ---& 3.3  & 1800& 13 & RL\\ 
							AmoebaNet-B + cutout~\cite{real2019regularized}  & 2.55 & ---& 2.8  & 3150& 19 & evolution\\ 
							PNAS~\cite{liu2018progressive} & 3.41 & ---& 3.2 & 225& 8 & SMBO\\
							ENAS + cutout~\cite{pham2018efficient}& 2.89 & ---& 4.6  & 0.5 & 6 & RL\\
							\midrule
							DARTS (first-order) + cutout~\cite{liu2018darts} &3.00 & 17.76 & 3.3  & 1.5 & 7 & gradient-based\\
							DARTS (second-order) + cutout~\cite{liu2018darts} &2.76 & 17.54 & 3.3   & 4.0& 7& gradient-based\\ 
							SNAS (moderate) + cutout~\cite{xie2018snas} &2.85& --- & 2.8 & 1.5 &7& gradient-based\\ 
							P-DARTS +  cutout~\cite{chen2019progressive} &2.50& 16.55 & 3.4 & 0.3&7 & gradient-based\\
							BayesNAS +  cutout~\cite{zhou2019bayesnas} &2.81& --- & 3.4  & 0.18 &7& gradient-based\\
							PC-DARTS +  cutout~\cite{xu2019pc} &2.81& --- & 3.6 & 0.13 &7& gradient-based\\
							GDAS + cutout~\cite{dong2019searching} &2.93& --- & 3.4 & 0.21 &7& gradient-based\\
							Fair DARTS + cutout~\cite{chu2019fair} &2.54& --- & 2.8  & 0.4 &7& gradient-based\\
							\midrule 
							PR-DARTS + cutout &  2.32   &  16.45  & 3.4   & 0.17 & 7 & gradient-based  \\ 
							\bottomrule 
						\end{tabularx}
				}}
			\end{center}
			\vspace{-0.2em}
		\end{threeparttable}
	\end{table*}
	\vspace{-0.8em}
\end{savenotes}

\vspace{-0.2em}
\subsection{Results on CIFAR}\label{CIFAR}
\vspace{-0.3em}
In the search phase, following~\cite{liu2018darts} we stack 8 cells with channel number 16. We divide  50K training samples in CIFAR10 into two equal-sized training and validation datasets.   In PR-DARTS, we set   $\lambda_1\!=\!0.01$, $\lambda_2\!=\!0.005$, and $\lambda_3\!=\!0.005$ for regularization.  Then we train the network 200 epochs with mini-batch size 128. For acceleration, per iteration, we follow~\cite{dong2019searching} and randomly select only two operations on each edge to update.  We respectively use   SGD and ADAM~\cite{kingma2014adam} to optimize parameters $\Wm$ and $\betam$ with detailed settings in Appendix~\ref{lamgugetask}.   
We set temperature  {$\tau$}~$\!=\!10$ and  linearly reduce it to 0.1, {$a$}~$\!=-0.1$ and {$b$}~$\!=\!1.1$. For pruning on each node,  we compare the gate activation probabilities of all non-zero  operations collected from all  previous nodes and retain top two operations~\cite{liu2018darts} .

For evaluation on CIFAR10 and CIFAR100, we set channel number 36 and then stack  18 normal cells and 2 reduction cells (the 7- and 14-th cells) to build a large network. We train the network 600 epochs with a mini-batch size of 128 from scratch.  See detailed settings of  SGD   in Appendix~\ref{lamgugetask}.  We also use drop-path with probability 0.2 and   cutout~\cite{devries2017improved} with length 16, for regularization.  

Table~\ref{CIFARcomparsion} summarizes the classification results on CIFAR10 and CIFAR100.  In merely 0.17 GPU-days on Tesla V100, PR-DARTS respectively achieves $2.31\%$ and $16.45\%$ classification errors on CIAR10 and CIFAR100, with both search time and accuracy significantly surpassing the DARTS baseline. By comparison, PR-DARTS  also consistently outperforms other NAS approaches, including differential NAS (\eg~P-DARTS, PC-DARTS),  RL based NAS (\eg~NASNet), as well as EA based NAS (\eg Amobdanet).   
These results demonstrate the superiority and transferability of the selected cells by  PR-DARTS. As shown in Fig.~\ref{illustrationcomponents},  this advantage comes from the group-structured binary gates and path-depth-wise regularization in PR-DARTS which can well alleviate unfair operation competition and cell-selection bias to shallow cells which are not well considered in the compared   NAS methods.  
Fair DARTS  imposes independent sigmoid distribution and zero-one loss for each operation,  which actually does not encourage the important global operation competition and cooperation.  
PR-DARTS runs faster over DARTS, because (1) the sparsity regularization prunes unnecessary connections as illustrated in Fig.~\ref{cells} in Appendix~\ref{lamgugetask}, and thus reduces the costs; and (2) following~\cite{dong2019searching} we randomly select only two operations instead of eight operations between two nodes to update per iteration, also helping reducing cost.   
Note, Proxyless NAS~\cite{cai2018proxylessnas} reports an error rate of 2.08$\%$ on CIAFR10, but it performs architecture search on the tree-structured  PyramidNet~\cite{cai2018path} which is much complex protocol than the DARTS search space in this work, and requires much longer time (4 GPU-days) for search. 

For \textit{ablation study}, Fig.~\ref{illustrationcomponents} shows the individual benefits of the two complementary components, group-structured binary gates and path-depth-wise regularization, in PR-DARTS. See details in  Fig.~\ref{illustrationcomponents}. 
Due to space limit, Appendix~\ref{lamgugetask} investigates  the \textit{effects of regularization parameters}  $\lambda_1\!\sim\!\lambda_3$ to the  performance of PR-DARTS.   The results show the stable performance of PR-DARTS on CIAFR10 when tuning these parameters in a relatively large range, and thus testify the robustness of PR-DARTS.

\begin{savenotes}
	\begin{table*}[tp]
		\begin{threeparttable}[b]
			\caption{\!Classification errors ($\%$) on ImageNet {\small(all methods use the cells searched on CIFAR10).}\vspace{-0.5em}} 
			\setlength{\tabcolsep}{6.0pt}  
			\renewcommand{\arraystretch}{0.7} 
			\label{Imagenetresults} \begin{center}
				{ \fontsize{8.1}{3}\selectfont{
						\begin{tabularx}{\textwidth}{c|cc cccc cc}\toprule
							\multirow{2}{*}{\textbf{Architecture}} &\multicolumn{2}{c}{ \textbf{Test Error ($\%$)}} & \textbf{Params} & \textbf{$\bm{\times +}$}	&  \textbf{Search Cost} & \textbf{Search space} & \textbf{Search}  \\
							& \textbf{Top-1} & \textbf{Top-5} & \textbf{(M)} & \textbf{(M)}	 & \textbf{(GPU-days)} & \textbf{$\#$Ops$/$zero}  & \textbf{method}\\
							\midrule 
							MobileNet~\cite{howard2017mobilenets} &29.4 & 10.5& 4.2 &569& --- &---& manual  \\  
							ShuffleNet2$\times$(v2)~\cite{ma2018shufflenet} &25.1& ---& $\sim$5 &591& --- &---& manual  \\ 
							\midrule
							NASNet-A~\cite{zoph2018learning} & 26.0 & 8.4& 5.3 & 564 & 1800& 13 & RL\\ 
							AmoebaNet-C~\cite{real2019regularized}  & 24.3 & 7.6& 6.4 & 570 & 3150& 19 & evolution\\ 
							PNAS~\cite{liu2018progressive} & 25.8 & 8.1& 5.1 & 588 & 225& 8 & SMBO\\ 
							MnaNet-92~\cite{tan2019mnasnet} & 25.2 & 8.0& 4.4 & 388 & --- & hierarchical & RL\\
							\midrule 
							DARTS (second-order)~\cite{liu2018darts} &26.7 & 8.7 & 4.7 &  574& 4.0& 7& gradient-based\\
							SNAS (mild)~\cite{xie2018snas} &27.3& 9.2 & 4.3 & 522& 1.5 &7& gradient-based\\ 
							P-DARTS~\cite{chen2019progressive} &24.4& 7.4  & 4.9 &557& 0.3&7 & gradient-based\\
							BayesNAS~\cite{zhou2019bayesnas} &26.5& 8.9 & 3.9 & --- & 0.18 &7& gradient-based\\
							PC-DARTS~\cite{xu2019pc} &25.1 & 7.8 & 5.3 & 586& 0.13 &7& gradient-based\\
							GDAS~\cite{dong2019searching} &26.0& 8.5 & 5.3 & 581& 0.21 &7& gradient-based\\
							Fair DARTS~\cite{chu2019fair} &24.9& 7.5& 4.8  & 541 &  0.4  &7& gradient-based\\
							\midrule
							PR-DARTS  & 24.1 & 7.3 & 4.98 & 543 & 0.17 & 7 & gradient-based  \\  
							\bottomrule
						\end{tabularx} 
				}}
			\end{center}
			\vspace{-1.0em}
		\end{threeparttable}
	\end{table*}
\end{savenotes}

\vspace{-0.5em}
\subsection{Results on ImageNet}
\vspace{-0.3em}
We further evaluate the transferability of the cells selected on CIFAR10 by testing them on more challenging ImageNet. Following DARTS, we rescale input size to  $224\times 224$. We stack three convolutional layers,12 normal cells and 2 reduction cells (channel number  48) to build a large network, and train it 250 epochs with mini-batch size 128.  See detailed settings of  SGD   in Appendix~\ref{lamgugetask}.

Table~\ref{Imagenetresults} reports the  results on ImageNet. One can observe that PR-DARTS consistently outperforms the compared state-of-the-art approaches. In particular, it respectively improves DARTS by  $2.4\%$  and  $1.4\%$  on top-1  and top 5 accuracies. These results demonstrate the superior transferability of the cells selected by PR-DARTS behind which the potential reasons have been discussed in Sec.~\ref{CIFAR}. 

\vspace{-0.7em}
\section{Conclusion}
\vspace{-0.3em}
In this work,  for the first time we theoretically explicitly show the benefits of more skip connections to fast network optimization in  DARTS,  explaining the dominated skip connections in the selected cells by DARTS. Then inspired by our theory, we propose PR-DARTS to improve DARTS by using group-structured binary gates and path-depth-wise regularization to alleviate unfair operation competition and cell-selection bias to  shallow cells. Experimental results validated the  advantages of PR-DARTS.

\newpage

\section*{Broader Impacts}
\label{sec:impact} 
This work advances network architecture search (NAS) in both theoretical performance analysis and practical algorithm design. As NAS can automatically design state-of-the-art architectures,   this work alleviates substantial efforts from domain experts for effective architecture design, and could also help develop more intelligent algorithms. But NAS  still needs an expert-designed search space which may have bias and prohibit NAS development.  So automatically designing search space is desirable.

{
\small
\bibliographystyle{unsrt}
\bibliography{referen}
}

 \newpage
 
 \appendix

\section{Structure of This Document}
	This supplementary document contains the technical proofs of convergence results and some additional experimental results of the main draft  entitled ``Theory-Inspired Path-Regularized Differential Network Architecture Search''. It is structured as follows. In Appendix~\ref{lamgugetask}, we provides more experimental results and  details, including the robustness investigation of PR-DARTS to regularization parameters,  effects of group-structured sparse regularization to gate activate probability, and training algorithms and details of PR-DARTS.  	Appendix~\ref{notations} summarizes the notations throughout this document and also provides the existing auxiliary theories and lemmas for subsequent analysis.  Then Appendix~\ref{proofofanalysispart} gives the proofs of the main results in Sec.~\ref{analysispart}, namely  Theorem~\ref{mainconvergence3}, by first introducing  auxiliary theories and lemmas for subsequent analysis whose proofs are deferred to Appendix~\ref{proofofAuxiliaryLemmas}.  Next, in Appendix~\ref{proofofmethodpart} we presents the results in Sec.~\ref{methodpart}, including Thoerems~\ref{lossproperty}, \ref{gateproperty} and \ref{fasterconvergenceloss}.  Finally, Appendix~\ref{proofofAuxiliaryLemmas} provides the proofs for auxiliary theories and lemmas  in Appendix~\ref{proofofanalysispart}.

\section{More Experimental Results and Details}\label{lamgugetask}
Due to space limitation, we defer more experimental  results and  details   to this appendix. Here we first  investigate robustness   of PR-DARTS to regularization parameters. Then we present effects of group-structured sparse regularization to gate activate probability,  and also show the reduction cell of PR-DARTS on CIFAR10.   
Next, we introduce the training algorithm of PR-DARTS, and finally present more setting details of optimizers for searching architectures and retraining from scratch.  
\subsection{Robustness to Regularization Parameters}
\begin{figure}[h] 
	\begin{center}
		\setlength{\tabcolsep}{0.1pt} 
		\begin{tabular}{ccc}
			\includegraphics[width=0.33\linewidth]{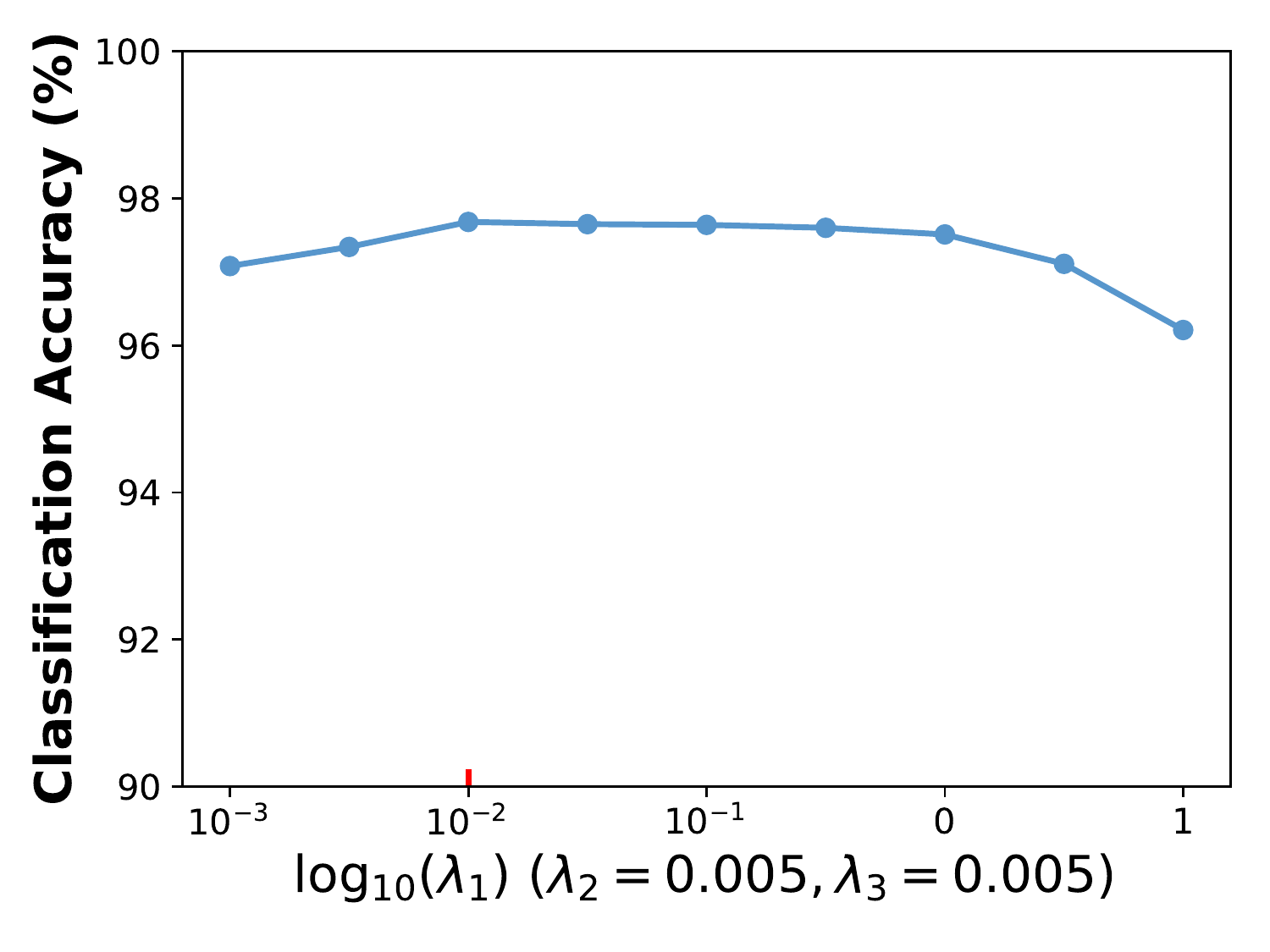}&
			\includegraphics[width=0.33\linewidth]{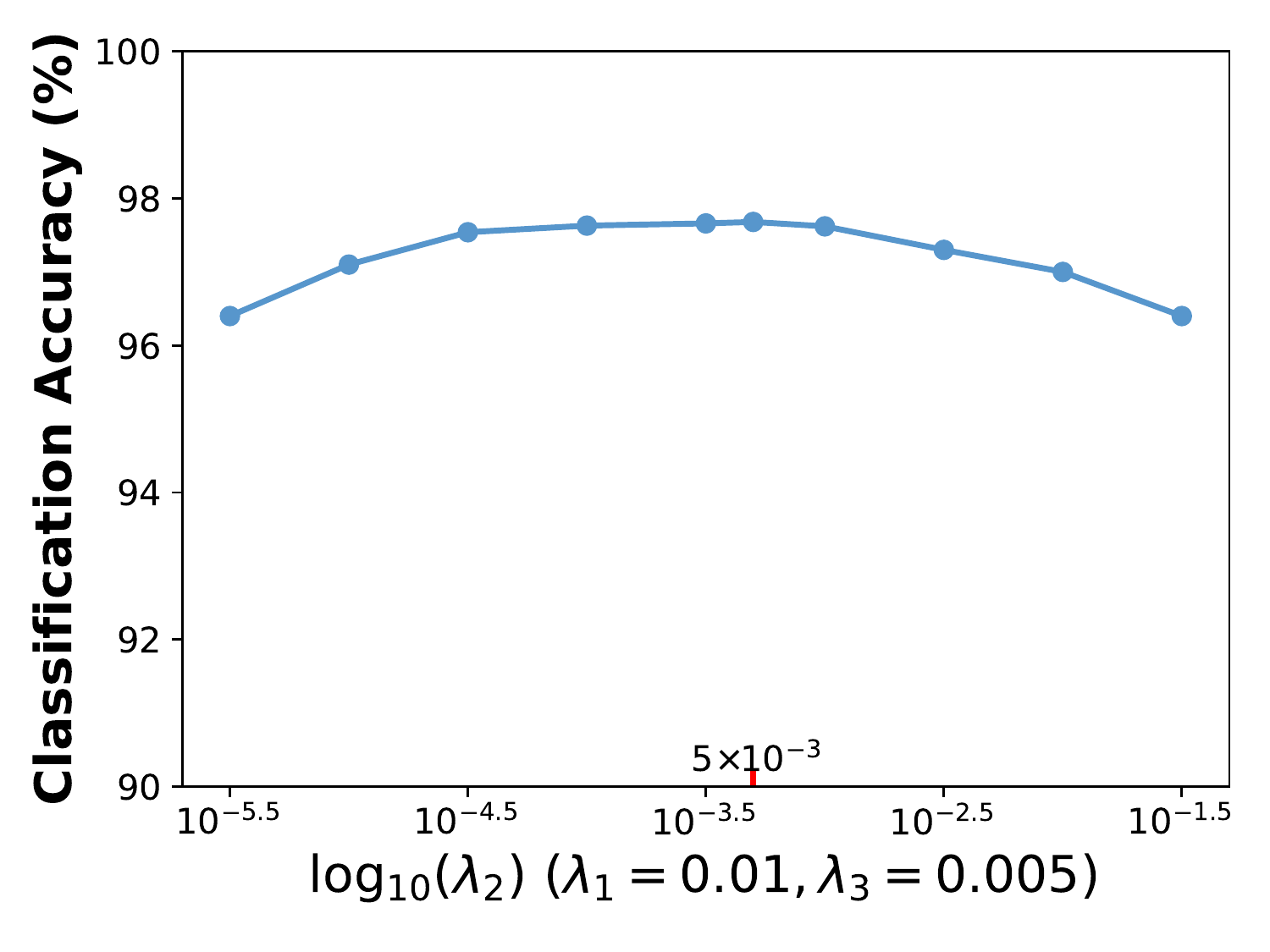}&
			\includegraphics[width=0.33\linewidth]{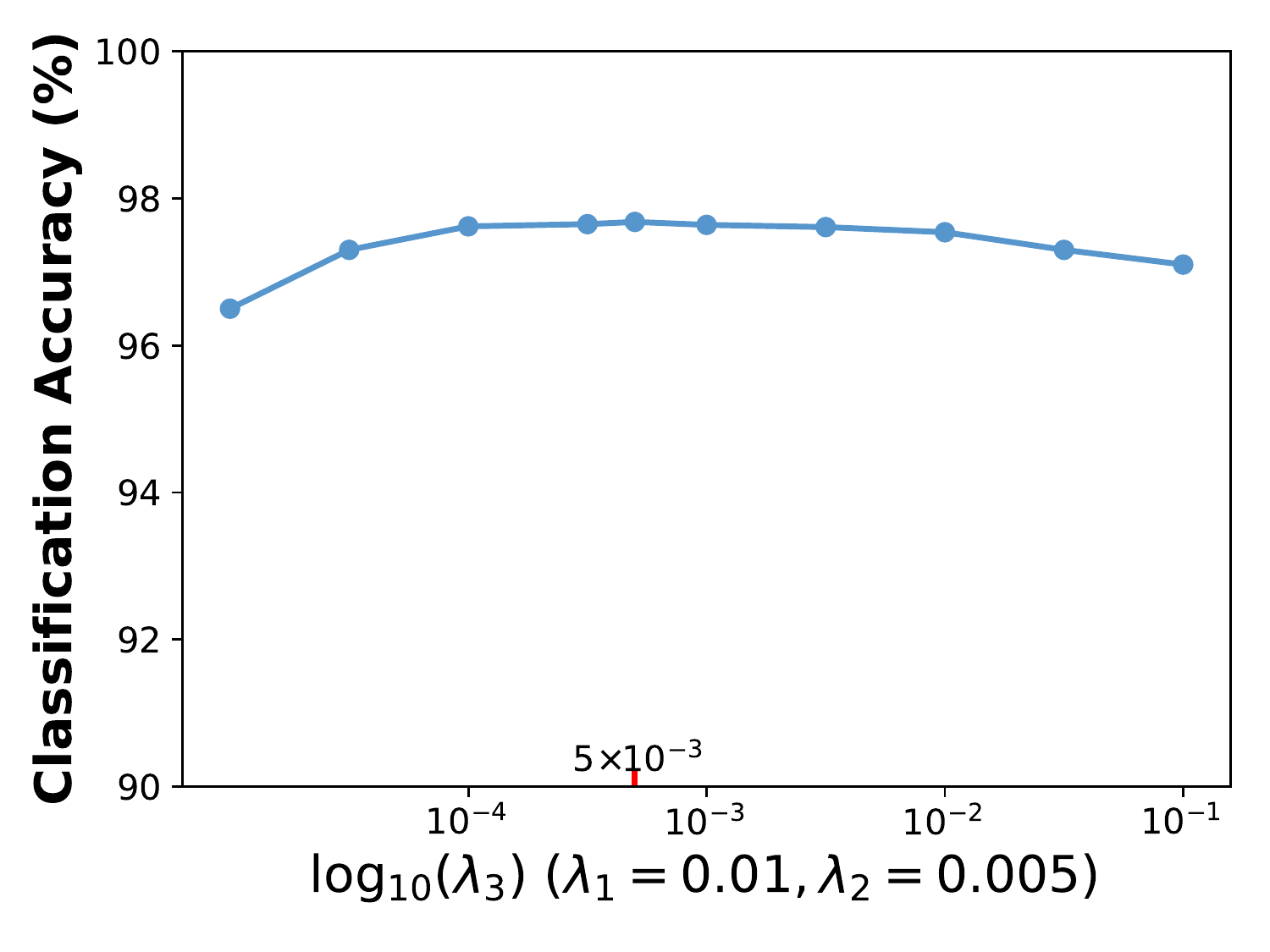}
		\end{tabular}
	\end{center}
	\vspace{-0.6em}
	\caption{Effects of regularization parameters $\lambda_1\sim \lambda_3$ to the  performance of PR-DARTS.}
	\label{robust}
\end{figure}

Fig.~\ref{robust} reports the effects of regularization parameters $\lambda_1\sim \lambda_3$ to the  performance of PR-DARTS. Due to the high training cost,  we  fix two regularization parameters and then investigate the third one.  From Fig.~\ref{robust}, one can observe that for each $\lambda$ ($\lambda_1$ or $\lambda_2$ or $\lambda_3$), when tuning it in a relatively large range, \eg~$\lambda_{1}\in[10^{-2}, 1]$, $\lambda_{2}\in[10^{-4.5} ,  10^{-2.5}]$ and $\lambda_{3}\in[10^{-4},  10^{-1.5}]$,   PR-DARTS has relatively stable  performance on CIFAR10. This  testifies the robustness of PR-DARTS to   regularization parameters.

\subsection{Effects of Group-Structured Sparse Regularization to Gate Activate Probability}
\begin{figure}
	\begin{center}
		\setlength{\tabcolsep}{0.0pt} 
		\begin{tabular}{ccc}
			\includegraphics[width=0.275\linewidth]{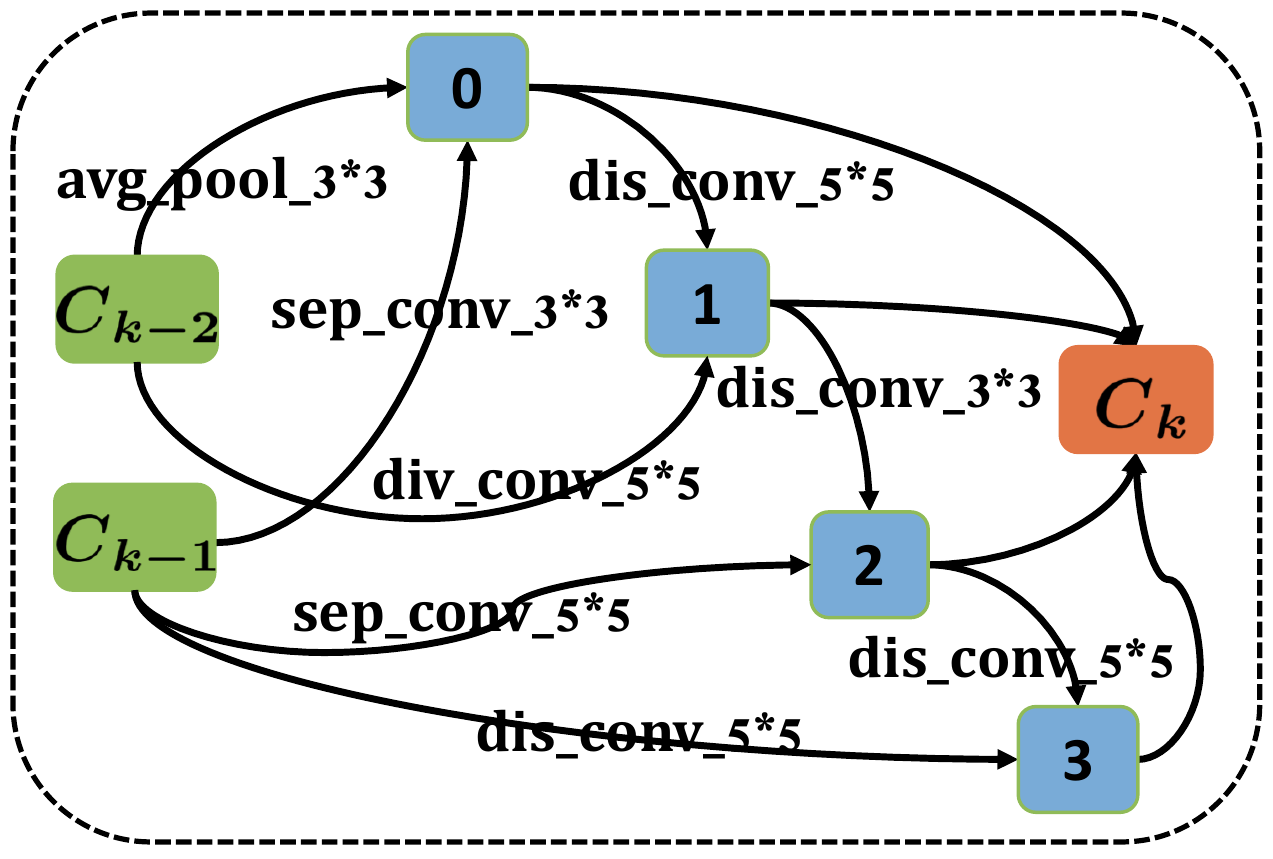}& 
			\includegraphics[width=0.35\linewidth]{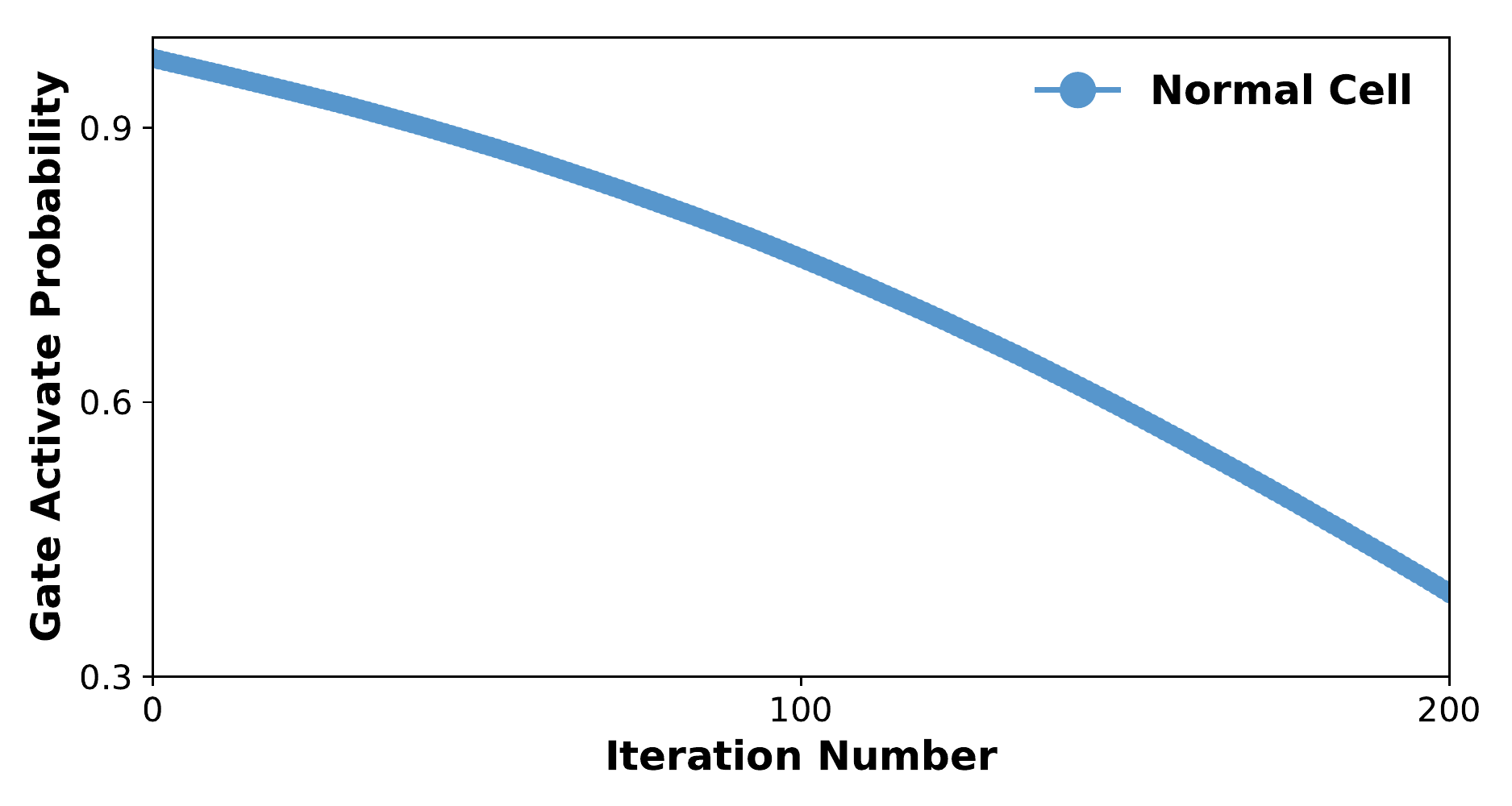}&
			\includegraphics[width=0.35\linewidth]{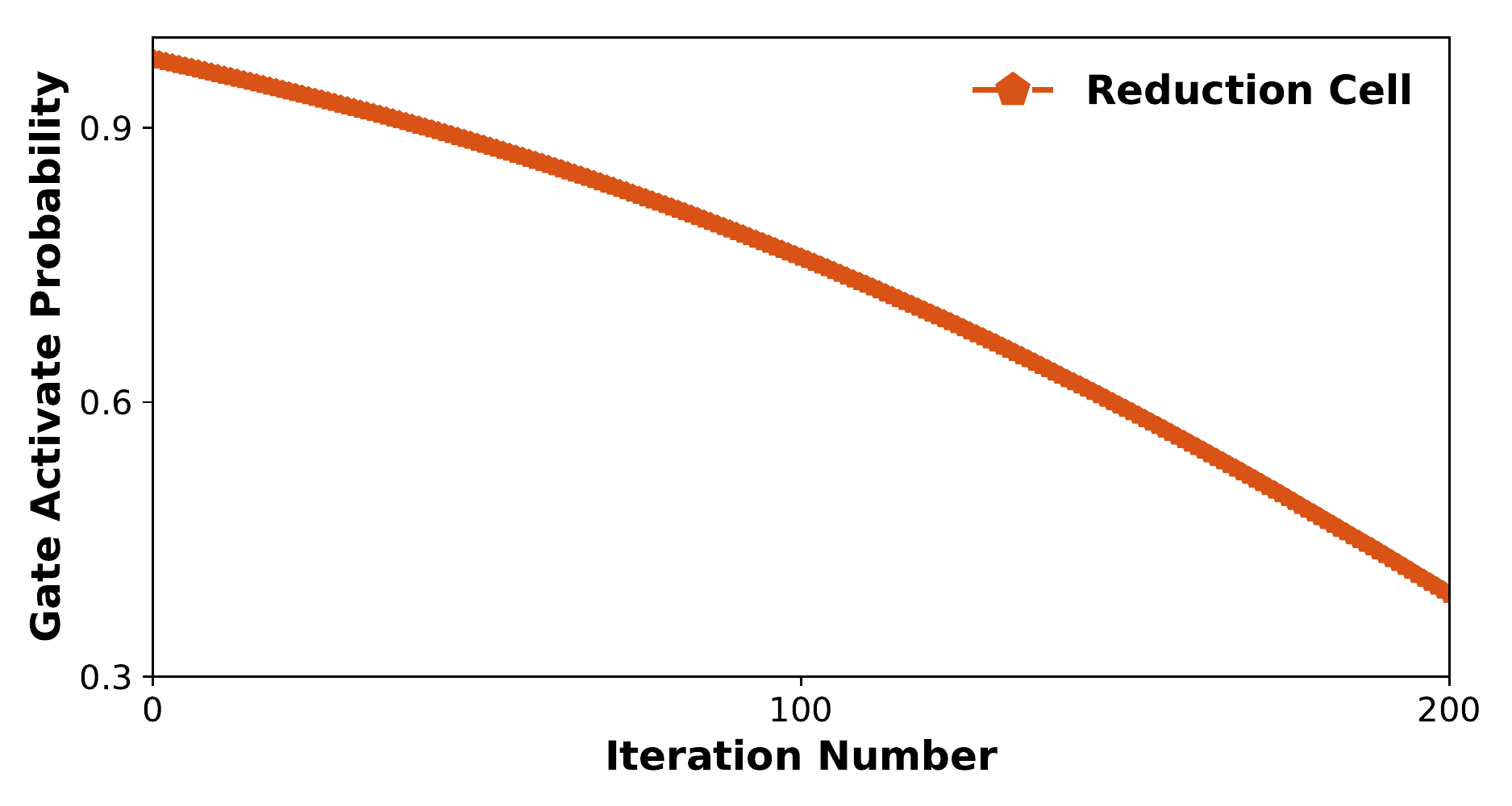}\\
			{\small(a) reduction cell on CIRAR10} & \multicolumn{2}{c}{\small(b) gate activate probability of normal  and reduction cells} \\ 
		\end{tabular}
	\end{center}
	\vspace{-0.5em}
	\caption{Visualization of search results. (a) denotes the selected reduction cell on CIRAR10. The normal cell is displayed in Fig.~\ref{illustrationcomponents} in the manuscript. (b) shows the gate activate probability of normal cell and reduction cell in PR-DARTS. }
	\label{cells}
\end{figure}

Here we first  display the selected reduction cell on CIRAR10  in Fig.~\ref{cells} (a). The normal cell selected on CIFAR10  is displayed in Fig.~\ref{illustrationcomponents} in the manuscript. 

Next, we also report the average gate activate probability in the normal and reduction cells in Fig.~\ref{cells} (b).   At the beginning of the search, we initialize the activation probability of each gate to be one. This is because  (1) as shown in Theorem~\ref{gateproperty}, the activation probability of the gate $\gnii{l}{s,t}$ is $ \Pro(\gnii{l}{s,t}\neq 0) =\Theta(\betanii{l}{s,t} - \tau \ln \frac{-a}{b} )$; (2) we set $a=-0.1, b =1.1, \betanii{l}{s,t} =0.5$ and initialize $\tau=10$ which leads to  $ \Pro(\gnii{l}{s,t}\neq 0) =\Theta(\betanii{l}{s,t} - \tau \ln \frac{-a}{b} )\approx 1$.  In this way, all gates will be well explored. With along more iterations, the group structured sparsity regularization encourages competition and cooperation among all operations to improve the performance, and  also prunes redundancy and unnecessary  connections in the cells as well. To  measure the overall sparsity of the normal cell, we compute its  overall average activation probability $ \frac{1}{|\mathcal{G}|}\sum_{\gnii{l}{s,t}\in\mathcal{G}} \Pro(\gnii{l}{s,t}\neq 0)$, where the gate set $\mathcal{G}$ collects all the operation gate in the normal cell. Similarly, we can compute the average activation probability of gates in the reduction cell.   As shown in Fig.~\ref{cells} (b), for both normal and reduction cells, their average gate activate probability becomes smaller with along more iterations. This  indicates the activation probability of the gates on redundancy and unnecessary  connections becomes smaller, which means that   sparsity regularizer gradually and automatically  prunes redundancy and unnecessary connections which reduces the information loss of pruning at the end of search. Moreover, this  sparsity regularizer defined on the whole cell  can   encourage global competition and cooperation of all operations in the cell, which  differs from DARTS that only introduces local competition among the operations between two nodes.  Actually, sparse cell also can reduce the computation cost and boost the search efficiency. 
 
\subsection{Algorithm Framework of PR-DARTS}
In this subsection, we introduce the training algorithm of PR-DARTS in details.  Same as  DARTS,  we  alternatively update  the network parameter $\Wm$ and the architecture parameter  $\betaii{}{}$ via gradient descent which is detailed in  Algorithm~\ref{searchalgorithm}.   For notation in Algorithm~\ref{searchalgorithm},  $F_{\B_{\mbox{\tiny{train}}}}\!(\Wm,\betaii{}{})= \frac{1}{|\B_{\mbox{\tiny{train}}}|}\sum_{(\xm,\ym)\in\B_{\mbox{\tiny{train}}} } f(\Wm,\betaii{}{};(\Xm,\ym))$  denotes the training loss on mini-batch $\B_{\mbox{\tiny{train}}}$. Similarly, the  loss $F_{\B_{\mbox{\tiny{val}}}}(\Wm,\betaii{}{})$ denotes the validation loss on mini-batch $\B_{\mbox{\tiny{val}}}$. When we compute the gradient $\nabla_{\!\betaii{}{}}  F_{\B_{\mbox{\tiny{train}}}}(\Wm,\betaii{}{})$, we ignore the second-order Hessian to accelerate the computation which is the same as first-order DARTS.  
\begin{algorithm}[H]
	\caption{Searching Algorithm for PR-DARTS} 
	\label{searchalgorithm}
	\setstretch{0.8}
	\renewcommand{\baselinestretch}{0.6}
	\begin{algorithmic} 
		\STATE {\bfseries Input:} training dataset $\D_{\mbox{\tiny{train}}}$ and validation dataset $\D_{\mbox{\tiny{val}}}$, mini-batch size $b$, learning rate $\eta$.
		\WHILE{ not convergence}
		\STATE  sample mini-batch $\B_{\mbox{\tiny{train}}}$   from $\D_{\mbox{\tiny{train}}}$  to update $\Wm$ by gradient descent $\Wm\!=\!\Wm\!-\!\eta \nabla_{\!\Wm}\! F_{\B_{\mbox{\tiny{train}}}}\!(\Wm,\betaii{}{}).$
		\STATE  sample mini-batch $\B_{\mbox{\tiny{val}}}$   from $\D_{\mbox{\tiny{val}}}$  to update $\betaii{}{}$ by gradient descent $\betaii{}{}\!=\!\betaii{}{}-\eta\nabla_{\!\betaii{}{}}  F_{\B_{\mbox{\tiny{val}}}}(\Wm,\betaii{}{}).$
		\ENDWHILE
		\STATE {\bfseries Output: $\betaii{}{}$}  
	\end{algorithmic}
\end{algorithm}

\subsection{Algorithm Parameter Settings}

\textbf{CIFAR10 and CIAFR100.}  In the search phase, following~DARTS, we use  momentum SGD to optimize network parameter $\Wm$, with an initial learning rate $0.025$ (annealed down to zero via cosine decay~\cite{loshchilov2016sgdr}), a momentum of 0.9, and a weight decay of  $3\times 10^{-4}$. Architecture parameter $\betaii{}{}$ is updated by ADAM~\cite{kingma2014adam} with a learning rate of  $3\times 10^{-4}$ and a weight decay of $10^{-3}$.  For evaluation on CIFAR10 and CIFAR100, we use  momentum SGD  with an initial learning $0.025$ (cosine decayed to zero), a momentum of 0.9, a weight decay of $3\times 10^{-4}$, and  gradient norm clipping parameter 5.0. 

\textbf{ImageNet.} 
We  evaluate the transfer ability of the  cells selected on CIFAR10 by testing them on   ImageNet. Following DARTS,  we use   momentum  SGD  with an initial learning $0.025$ (cosine decayed to zero), a momentum of 0.9,  a weight decay of $3\!\times\! 10^{-4}$,  and  gradient norm clipping parameter 5.0.

\section{Notation and Preliminarily}\label{notations}
\subsection{Notations}
In this document, we use $\Xmii{(l)}{i}(k)$ to denote the output $\Xmii{(l)}{i}$ of the $i$-th sample in the $l$-th layer at the $k$-th iteration. For brevity, we usually ignore the notation $(k)$ and $i$ and use $\Xmii{(l)}{}$ to denote the output $\Xmii{(l)}{}$ of any sample $\Xmi{i}\ (\forall i=1,\cdots,n)$ in the $l$-th layer at any iteration. We use $\Omegam=\{\Wmii{(0)}{}, \Wmii{(1)}{0}, \Wmii{(2)}{0}, \Wmii{(2)}{1}, \cdots,  \Wmii{(l)}{0},\cdots, \Wmii{(l)}{l-1}, \cdots, \Wmii{h-1}{0},\cdots,\Wmii{(h-1)}{h-2}, \Umi{0},\cdots,\Umi{h-1}\}$ to denote the set of all $\frac{h(h+3)}{2}$ learnable matrix parameters, including the convolution parameters $\Wmii{(l)}{s}$ and the linear mapping parameters $\Umi{s}$. Let $\Omegam_i$ denote the $i$-th matrix parameters in $\Omegam$, \eg~$\Omegam_1=\Wmii{(0)}{}$. For notation simplicity, here we assume the input size is $m\times p$ to avoid using $\bar{m} \times \bar{p}$.  The operation $\vect{\Xm}$ vectorizes the matrix $\Xm$.

Then we define the loss 
\begin{equation*}
\begin{split}
F(\Omegam)=\frac{1}{2n} \|\ymm - \hmm(k)\|_2^2 = \frac{1}{2n} \sum_{i=1}^{n}  (\ymi{i}-\hii{i})^2= \frac{1}{n}\sum_{i=1}^n \ell_i,
\end{split}
\end{equation*}
where $\umi{}(k) = [\ui{1}(k); \ui{2}(k); \cdots, \ui{n}(k) ]\in\Rs{n}$ denotes the prediction at the $k$-th iteration, $\ymm= [\ymi{1}; \ymi{2}; \cdots, \ymi{n}]\in\Rs{n}$ is the labels for the $n$ samples $\{\Xmi{i}\}_{i=1}^n$, and $\ell_i = (\ymi{i}-\hii{i})^2$ denotes the individual loss of the $i$-th sample $\Xmi{i}$.

Then for brevity, $\ell(\Omegam)$ and $\ell_i(\Omegam)$ respectively denote  the losses when feeding the input $(\Xm,\ym)$ and  $(\Xmi{i},\ymi{i})$. Then we denote the gradient of $\ell(\Omegam)$ with respect to  all learnable parameters $\Omegam$ as 
\begin{equation*}
\begin{split}
\nabla_{\Omegam} \ell(\Omegam) = & \left[\vect{\frac{\partial \ell }{\partial \Wmii{(0)}{}} }; \left\{\vect{\frac{\partial \ell }{\partial \Wmii{(l)}{s}} }\right\}_{0\leq l \leq h-1, 0\leq s \leq l-1};  \left\{\vect{\frac{\partial \ell }{\partial \Umi{s}}}\right\}_{0\leq s\leq h-1}   \right] ,
\end{split}
\end{equation*}
where the $\vect{\Xm}$ operation vectorizes the matrix $\Xm$ into vector.  Here we also let $ \nabla_{\Omegami{i}}  \ell(\Omegam) $ denotes the gradient of $ \ell(\Omegam)$ with the $i$-th matrix parameter, e.g. $\nabla_{\Omegami{1}}  \ell(\Omegam)  = \vect{\frac{\partial \ell }{\partial \Wmii{(0)}{}} }$. Therefore, $\nabla_{\Omegam}  F(\Omegam)  = \frac{1}{n} \sum_{i=1}^{n}\nabla_{\Omegam}  \partial \ell_i(\Omegam) $ where $\ell_i(\Omegam)$ is the loss given input $(\Xmi{i},\ymi{i})$. In this way, we can define the Gram matrix $\Gmii{}{}(k)\in\Rs{n\times n}$ at the $k$-th iteration in which its $(i,j)$-th  entry is defined as 
\begin{equation*}
\begin{split}
\Gmii{ij}{}(k) =  \langle \nabla_{\Omegam}  \ell_i(\Omegam(k)), \nabla_{\Omegam}  \ell_j(\Omegam(k))   \rangle, 
\end{split}
\end{equation*}
where $\nabla_{\Omegam}  \ell_i(\Omegam(k)) $ denote the gradient of the loss $\ell_i$ on the $i$-th sample $(\Xmi{i},\ymi{i})$ with respect to all parameter $\Omegam$ at the $k$-th iteration. We often ignore the notation $k$ and use $\Gmii{}{}$ to denote the Gram matrix that does not depend on iteration number $k$. 

According to the definitions, we have 
\begin{equation*}
\begin{split}
& \Gmii{ij}{}(k) =  \langle \nabla_{\Omegam}  \ell_i(\Omegam(k)), \nabla_{\Omegam}  \ell_j(\Omegam(k))   \rangle = \sum_{t=1}^{\frac{h(h+3)}{2}} \langle \nabla_{\Omegami{t}}  \ell_i(\Omegam(k)), \nabla_{\Omegami{t}}  \ell_j(\Omegam(k))   \rangle\\
= &\left\langle \frac{\partial \ell_i}{\partial \Wmii{(0)}{}(k)}, \frac{\partial \ell_j}{\partial \Wmii{(0)}{}(k)}\right\rangle + \sum_{l=1}^{h-1} \sum_{s=0}^{l-1} \left\langle \frac{\partial \ell_i}{\partial \Wmii{(l)}{s}(k)},  \frac{\partial \ell_j}{\partial \Wmii{(l)}{s}(k)}\right\rangle + \sum_{s=0}^{h-1} \left\langle \frac{\partial \ell_i}{\partial \Umi{s}(k)}, \frac{\partial \ell_j}{\partial \Umi{s}(k)}\right\rangle\\
\end{split}
\end{equation*}
For brevity, we let 
\begin{equation*}
\begin{split}
\Gmbii{ij}{0}(k) \!=\!\left\langle\! \frac{\partial \ell_i}{\partial \Wmii{(0)}{}(k)}, \frac{\partial \ell_j}{\partial \Wmii{(0)}{}(k)}\!\right\rangle, \  \Gmii{ij}{ls}(k) \!=\! \left\langle\! \frac{\partial \ell_i}{\partial \Wmii{(l)}{s}(k)},  \frac{\partial \ell_j}{\partial \Wmii{(l)}{s}(k)}\!\right\rangle, \  \Gmii{ij}{s}(k) \!=\! \left\langle \!\frac{\partial \ell_i}{\partial \Umi{s}(k)}, \frac{\partial \ell_j}{\partial \Umi{s}(k)}\!\right\rangle.
\end{split}
\end{equation*}
Therefore, we have 
\begin{equation*}
\begin{split}
\Gmii{ij}{}(k) = \Gmbii{ij}{0}(k)  + \sum_{l=1}^{h-1} \sum_{s=0}^{l-1} \Gmii{ij}{ls}(k)  + \sum_{s=0}^{h-1}  \Gmii{ij}{s}(k),\quad \Gmii{}{}(k) = \Gmbii{}{0}(k)  + \sum_{l=1}^{h-1} \sum_{s=0}^{l-1} \Gmii{}{ls}(k)  + \sum_{s=0}^{h-1}  \Gmii{}{s}(k).
\end{split}
\end{equation*}

Finally, since we need to compute the gradient. Here we define an operation for computing the gradient for convolution operation.  	  For back-propagate,  we define the inverse operation of  $\Phi(\Xm)$ as $\convb{\frac{1}{\tau}\Phi(\Xm)}=\Xm \in \Rs{\m\times \p}$.  For the $(i,j)$-th entry in $\convb{\Xm}$, it equals to the sum of all $\Xm_{i,j}$ in $\Phi(\Xm)$.

\subsection{Auxiliary Lemmas}\label{AuxiliaryLemmas}
\begin{lem}~\cite{saw1984chebyshev}[Chebyshev's inequality]\label{Chebyshevinequality}
	For any variable $x$, we have 
	\begin{equation*}
	\begin{split}
	\Pro\left( |x-\EE[x]|\geq a \right) \leq \frac{\Var(x)}{a^2},
	\end{split}
	\end{equation*}
	where $a$ is a positive constant, $\Var(x)$ denotes the variance of $x$. 
\end{lem}

\begin{lem}\label{boundmatrix}\cite{du2018gradient}
	Given a set of matrices $\{\Am_i,\Bm_i\}$ with proper sizes, if $\|\Am_i\|_2\leq a_i$ and $\|\Bm_i\|_2\leq a_i$ and $\|\Am_i -\Bm_i\|_F \leq b_i a_i$, we have 
	\begin{equation*}
	\left\|\prod_{i=1}^n \Am_i - \prod_{i=1}^n \Bm_i\right\|_F \leq \left(\sum_{i=1}^n b_i\right) \prod_{i=1}^n a_i.
	\end{equation*}
\end{lem}

\begin{lem}\cite{hwang2004cauchy}[Cauchy Interlace Theorem] Let $\Am$ be a Hermitian matrix of order $n$ and let $\Bm$ be a principal submatrix of $\Am$ of order $n-1$. If $\lambda_n\leq \lambda_{n-1} \leq \cdots \leq \lambda_{1}$ lists the eigenvalues of $\Am$ and $\mu_{n} \leq \mu_{n-1}\leq \cdots \leq \mu_{2}$ the eigenvalues of $\Bm$, then  $\lambda_n \leq \mu_n \leq \lambda_{n-1} \leq \mu_{n-1} \cdots \leq \lambda_{2} \leq \mu_{2} \leq \lambda_{1}$. 
\end{lem}

\begin{lem}\cite{Alessandro2017}[Chi-Square Variable Bound] \label{chisquare}
	Let $x$ be chi-square variable with $n$ degree of freedom. Then for any $t>0$, it holds 
	\begin{equation*}
	\Pro\left(x-n\geq 2 \sqrt{nt } +2 t\right) \leq \exp(-t),\quad \text{and} \quad 
	\Pro\left(x-n\leq -2 \sqrt{nt} \right) \leq \exp(-t).
	\end{equation*}
\end{lem}

\begin{lem}\label{sdaffdasfreawsdf}\cite{du2018gradient}
	Suppose $\sigma$ is analytic and not a polynomial function. Consider data $\{\Xmi{i=1}^n\}_{i=1}^n$ are not parallel, namely $\vect{\Xmi{i}}\notin \text{span}(\vect{\Xmi{j}})$ for all $i\neq j$, Then the smallest eigenvalue the matrix $\Gm$ which is defined as 
	\begin{equation*}
	\Gm(\Xm)_{ij}= \EE_{\Wm\sim\mathcal{N}(0,\Imm)} \ \sigma(\langle \Wm, \Xmi{i} \rangle)  \sigma(\langle \Wm, \Xmi{j}\rangle) 
	\end{equation*}   
	is larger than zero, namely $\lambda_{\min}(\Gm)>0$. 
\end{lem}

\begin{lem}\label{sdaffdasfasfdafdreawsdf}\cite{du2018gradient}
	Suppose $\sigma$ is analytic and not a polynomial function. Consider data $\{\Xmi{i=1}^n\}_{i=1}^n$ are not parallel, namely $\vect{\Xmi{i}}\notin \text{span}(\vect{\Xmi{j}})$ for all $i\neq j$, Then the smallest eigenvalue the matrix $\Gm$ which is defined as 
	\begin{equation*}
	\Gm(\Xm)_{ij}= \EE_{\Wm\sim\mathcal{N}(0,\Imm)} \ \sigma'(\langle \Wm, \Xmi{i} \rangle)  \sigma'(\langle \Wm, \Xmi{j}\rangle) 
	\end{equation*}   
	is larger than zero, namely $\lambda_{\min}(\Gm)>0$. 
\end{lem}

\begin{lem}\label{distributionbound}\cite{du2018gradient}
	Suppose the activation function $\sigma(\cdot)$ satisfies Assumption~\ref{activationassumption}. Suppose there exists $c>0$ such that 
	\begin{equation*}
	\Am=\begin{bmatrix}
	a_1^2&\rho a_1b_1\\
	\rho_1 a_1 b_1 &b_1^2
	\end{bmatrix} \succ 0,\qquad \Bm=\begin{bmatrix}
	a_2^2&\rho_2 a_2b_2\\
	\rho a_2 b_2 &b_2^2
	\end{bmatrix} \succ 0,
	\end{equation*}
	where the parameter satisfies $1/c\leq x\leq c $ in which $x$ could be  $a_1$, $a_2$, $b_1$, $b_2$. Let $g(\Am)=\EE_{(u,v)\sim\mathcal{N}(0,\Am)}\sigma(u)\sigma(v)$. Then we have 
	\begin{equation*}
	|g(\Am) - g(\Bm)|\leq c\|\Am-\Bm\|_F \leq 2C\|\Am-\Bm\|_{\infty},
	\end{equation*}
	where $C$ is a constant that only depends on $c$ and the Lipschitz and smooth parameter of $\sigma(\cdot)$.
\end{lem}

\section{Proofs of Results in Sec.~\ref{analysispart}}\label{proofofanalysispart}
\subsection{Proof of Theorem~\ref{mainconvergence3}}\label{proofofmainconvergence3}

Suppose Assumptions~\ref{activationassumption}, \ref{initilizationassumption} and~\ref{sampleassumption} hold.  To prove our main results, namely the results in Theorem~\ref{mainconvergence3}, we have two steps.   In the first step, from Lemma~\ref{mainconvergence2}, we have that  if $m$ and $\eta$ satisfy
\begin{equation*}
m\geq   \frac{c_m'   c^2 \rho  \kc^2  \cwo^2 \mu^2}{\lambda^2 n},\quad 
\eta\leq  \frac{c_\eta' \lambda}{\sqrt{m} \mu^4     h^{3}  \kc^2 c^4},
\end{equation*}
where $c_m'$ and $c_\eta'$ are two constants, $c=\left(1+\alphaii{}{2} + 2\alphaii{}{3}  \mu \sqrt{\kc} \cwo \right)^{h}$, $\alphaii{}{2}=\max_{s,l}\alphaii{(l)}{s,2}$ and $\alphaii{}{3}=\max_{s,l} \alphaii{(l)}{s,3}$. Then with probability at least $1-\delta/2$ we have 
\begin{equation*}
\begin{split}
\|\ymm - \hmm(k)\|_2^2   \leq \left( 1 - \frac{\eta  \lambda_{\min}\left( \Gm(0) \right)}{4}\right) \|\ymm - \hmm(k-1)\|_2^2 .  
\end{split}
\end{equation*}  
where $k$ denotes the iteration number, $ \lambda_{\min}\left( \Gm(0)\right)$ denotes the smallest eigenvalue of the Gram matrix $ \Gm(0)$ at the initialization.  For this part, we prove it in Appendix~\ref{Proofoflinearconvergene}. 

In the second step, we will prove that the smallest eigenvalue of can be lower bounded. Specifically,  we prove this results in Lemma~\ref{sdaffdasfasfdafdreawsdfadad}:  if $m\geq \frac{c_4\mu^2 p^2 n^2\log(n/\delta)}{\lambda^2}$, it holds that with probability at least $1-\delta/2$, the smallest eigenvalue the matrix $\Gm$ satisfies 
\begin{equation*}
\begin{split}
\lambda_{\min}\left(\Gmii{}{}(0)\right) 
\geq \frac{3c_{\sigma}}{4} \sum_{s=0}^{h-1}(\alphaii{(h)}{s,3} )^2
\left(\prod_{t=0}^{s-1}  (\alphaii{(s)}{t,2})^2\right)   
\lambda_{\min}(\Km).
\end{split}
\end{equation*}
where $\lambda=3c_{\sigma} \sum_{s=0}^{h-1}(\alphaii{(h)}{s,3} )^2
\left(\prod_{t=0}^{s-1}  (\alphaii{(s)}{t,2})^2\right)   
\lambda_{\min}(\Km)$, $c_\sigma$ is a constant that only depends on $\sigma$ and the input data,    $\lambda_{\min}({\Km})=\min_{i,j} \lambda_{\min}(\Km_{ij}) $ is larger than zero  in which  $\lambda_{\min}(\Km_{ij}) $ is the the smallest eigenvalue of $\Km_{ij} =  \begin{bmatrix}
\Xmi{i}^{\top} \Xmi{j}, \Xmi{i}^{\top} \Xmi{j}\\
\Xmi{j}^{\top} \Xmi{i}, \Xmi{j}^{\top} \Xmi{j}\\
\end{bmatrix}$ .  
Appendix~\ref{ProofofLowerbound} provides the proof for this result.

Finally, we combine these results in the above two steps and can obtain  that  if   $m\!\geq\!  \frac{c_m \mu^2  }{\lambda^2}   \left[ \rho p^2 n^2\log(n/\delta) \!+\! c^2  \kc^2  \cwo^2/n   \right]$ and $\eta \!\leq\!  \frac{c_\eta \lambda}{\sqrt{m} \mu^4   h^{3}  \kc^2 c^4}$, 
where $\cwo, c_m, c_\eta$ are  constants, with probability at least $(1-\delta/2)^2>1-\delta$,  we have 
\begin{equation*}
\|\ymm - \hmm(k)\|_2^2   \leq \left( 1 - \eta  \lambda/4\right) \|\ymm - \hmm(k-1)\|_2^2 \quad (\forall k\geq 1),
\end{equation*} 
where $\lambda 
= \frac{3c_{\sigma}}{4} \lambda_{\min}(\Km)\sum_{s=0}^{h-2}(\alphanii{h-1}{s,3} )^2
\prod_{t=0}^{s-1}  (\alphanii{s}{t,2})^2 
$,  the positive constant $c_\sigma$ only depends on $\sigma$ and  input data. On the other hand, we have 
$$F_{\mbox{\tiny{train}}}(\Wm(k+1), \betam) =\frac{1}{2n}\|\ymm - \hmm(k+1)\|_2^2,  $$
then we can obtain the desired results in Theorem~\ref{mainconvergence3}. 
Please refer to the proof details in Appendix~\ref{Proofoflinearconvergene} and \ref{ProofofLowerbound} for the above two steps respectively.

Note that our proof framework is similar to~\cite{du2018gradient}. But there are essential differences. The main difference is that here our network architecture is much complex (e.g. each layer connects all the previous layers) and each edge in our network also involves more operations, including zero operation, skip operation and convolution operation, which requires bounding many terms in this work differently and  more elaborately. 

For the following proofs, Appendix~\ref{AuxiliaryLemmasfdsafdas} provides the auxiliary lemmas for the proofs for Step 1 and Step 2. Then Appendix~\ref{Proofoflinearconvergene} and \ref{ProofofLowerbound} respectively present the proof details in Step 1 and Step 2.

\subsection{Auxiliary Lemmas}\label{AuxiliaryLemmasfdsafdas}
\begin{lem}\label{gradientcomputation}
	The gradient of the loss $\ell=\frac{1}{2}(u-y)^2$ with parameter and temporary output can be written as follows:
	\begin{equation*}
	\begin{split}
	\frac{\partial \ell }{\partial \Xmii{(l)}{}}
	=&  (\ui{} -\ymi{} ) \Umi{l} +\sum_{s=l+1}^{h-1}\left(\alphaii{(s)}{l,2}\frac{\partial \ell }{\partial \Xmii{(s)}{}}  + \alphaii{(s)}{l,3}\tau   \convb{(\Wmii{(s)}{l})^{\top} \left(\sigma'\left( \Wmii{(s)}{l} \Phi(\Xmii{(l)}{})\right) \odot \frac{\partial \ell }{\partial \Xmii{(s)}{}}\right)}  \right),\\
	& \qquad  \qquad \qquad \qquad \qquad \qquad \qquad \qquad \qquad \qquad \qquad \qquad	\ (0\leq l \leq h-1, 0 \leq s \leq l-1),\\
	\frac{\partial \ell }{\partial \Xmii{}{}}= &    \tau   \convb{(\Wmii{(0)}{})^{\top} \left(\sigma'\left( \Wmii{(0)}{} \Phi(\Xmii{}{})\right) \odot \frac{\partial \ell }{\partial \Xmii{(0)}{}}\right)} \in\Rs{m\times p},\\
	\frac{\partial \ell }{\partial \Wmii{(l)}{s}}= &  \alphaii{(l)}{s,3}\tau  \Phi(\Xmii{(s)}{})    \left(\sigma'\left( \Wmii{(l)}{s} \Phi(\Xmii{(s)}{})\right) \odot \frac{\partial \ell }{\partial \Xmii{(l)}{}}\right)^{\top}  \in\Rs{m\times p} \ (0\leq l \leq h-1, 0 \leq s \leq l-1),\\
	\frac{\partial \ell }{\partial \Wmii{(0)}{}}= &   \tau  \Phi(\Xmii{}{})    \left(\sigma'\left( \Wmii{(0)}{} \Phi(\Xmii{}{})\right) \odot \frac{\partial \ell }{\partial \Xmii{(0)}{}}\right)^{\top}  \in\Rs{m\times p},\\
	\frac{\partial \ell }{\partial \Umi{s}}= &    (\ui{} -\ymi{} ) \Xmii{(s)}{} \in\Rs{m\times p},\\
	\end{split}
	\end{equation*}
	where $\odot$ denotes the dot product, $\frac{\partial \ell }{\partial \Xmii{(l)}{}}  \in\Rs{m\times p}$. 
\end{lem}
See its proof in Appendix~\ref{proofgradientcomputation}.

\begin{lem}\label{gradientcomputation2}
	The gradient of the network output $\ui{}$ with respect to the output and convolution parameter can be written as follows:
	\begin{equation*}
	\begin{split}
	\frac{\partial \ui{} }{\partial \Xmii{(l)}{}}
	=& \taum \Umi{l} +\sum_{s=l+1}^{h-1}\left(\alphaii{(s)}{l,2}\frac{\partial \ui{} }{\partial \Xmii{(s)}{}}  + \alphaii{(s)}{l,3}\tau   \convb{(\Wmii{(s)}{l})^{\top} \left(\sigma'\left( \Wmii{(s)}{l} \Phi(\Xmii{(l)}{})\right) \odot \frac{\partial \ui{} }{\partial \Xmii{(s)}{}}\right)}  \right),\\
	& \qquad  \qquad \qquad \qquad \qquad \qquad \qquad \qquad \qquad \qquad \qquad \qquad	\ (0\leq l \leq h-1, 0 \leq s \leq l-1),\\
	\frac{\partial \ui{} }{\partial \Xmii{}{}}= &    \tau   \convb{(\Wmii{(0)}{})^{\top} \left(\sigma'\left( \Wmii{(0)}{} \Phi(\Xmii{}{})\right) \odot \frac{\partial \ui{}  }{\partial \Xmii{(0)}{}}\right)} \in\Rs{m\times p},\\
	\frac{\partial \ui{} }{\partial \Wmii{(l)}{s}}= &  \alphaii{(l)}{s,3}\tau  \Phi(\Xmii{(s)}{})    \left(\sigma'\left( \Wmii{(l)}{s} \Phi(\Xmii{(s)}{})\right) \odot \frac{\partial \ui{} }{\partial \Xmii{(l)}{}}\right)^{\top}  \in\Rs{m\times p} \ (0\leq l \leq h-1, 0 \leq s \leq l-1),\\
	\frac{\partial \ui{} }{\partial \Wmii{(0)}{}}= &   \tau  \Phi(\Xmii{}{})    \left(\sigma'\left( \Wmii{(0)}{} \Phi(\Xmii{}{})\right) \odot \frac{\partial \ui{} }{\partial \Xmii{(0)}{}}\right)^{\top}  \in\Rs{m\times p},\\
	\frac{\partial \ui{} }{\partial \Umi{s}}= &  \taum  \Xmii{(s)}{} \in\Rs{m\times p},	\ ( 0 \leq s \leq h-1),\\
	\end{split}
	\end{equation*}
	where $\odot$ denotes the dot product and $\frac{\partial u }{\partial \Xmii{(l)}{}}  \in\Rs{m\times p}$. 
\end{lem}
See its proof in Appendix~\ref{proofgradientcomputation2}.

\begin{lem}\label{boundofinitialization}
	Suppose Assumptions~\ref{activationassumption}, \ref{initilizationassumption} and~\ref{sampleassumption} hold. Given a constant $\delta \in (0,1)$, assume $m\geq \frac{16c_1 n p^2}{c^2\delta}$, where $c_1= \sigma^4(0) + 4  |\sigma^3(0)| \mu \sqrt{2/\pi} + 8|\sigma(0)| \mu^3 \sqrt{2/\pi} +32 \mu^4$ and $c=\EE_{\omega \sim \Nn(0,\frac{1}{\sqrt{p}})}\left[  \sigma^2(  \omega) \right]$.  Suppose $\Wmii{(l)}{s} (0)\leq \sqrt{\m} \cwo\ \forall 0\leq l\leq h, 0\leq s\leq l-1$.  Then with probability at least $1-\delta/4$, we have 
	\begin{equation*}
	\begin{split}
	\frac{1}{\cxo} \leq \|\Xmii{(l)}{} (0)\|_F  \leq \cxo.  
	\end{split}
	\end{equation*}
	where $\cxo\geq 1$ is a constant.
\end{lem}
See its proof in Appendix~\ref{proofofboundofinitialization}.

\begin{lem}\label{boundofmidoutput}
	Suppose Assumptions~\ref{activationassumption}, \ref{initilizationassumption} and~\ref{sampleassumption} hold. Assume   $\|\Wmii{l}{s}(0)\|_2 \leq \sqrt{\m}   \cwo$,  $ \|\Wmii{l}{s}(k) - \Wmii{l}{s}(0)\|_F \leq  \sqrt{\m} \taum \rc$.  Then  for $\forall l$, we have 
	\begin{equation*}
	\begin{split}
	& \|\Xmii{(l)}{} (k)-\Xmii{(l)}{} (0)\|_F 
	\leq  \left(1+ \alphaii{}{2}+\alphaii{}{3}\mu\sqrt{\kc} \left( \taum \rc + \cwo\right)  \right)^{l} \taum \mu \sqrt{\kc} \rc,\\
	&\left\| \Wmii{(l)}{s}(k)  \Phi(\Xmii{(s)}{}(k) ) -  \Wmii{(l)}{s}(0)  \Phi(\Xmii{(s)}{}(0) )   \right\|_F \leq \frac{1}{\alphai{3}}
	\left(1+ \alphaii{}{2}+\alphaii{}{3}\mu\sqrt{\kc} \left( \taum \rc + \cwo\right)  \right)^{l}  \taum \sqrt{\kc m} \rc,
	\end{split}
	\end{equation*} 
	where $\alphaii{}{2}=\max_{s,l}\alphaii{(l)}{s,2}$ and $\alphaii{}{3}=\max_{s,l} \alphaii{(l)}{s,3}$, and $\cxo\geq 1$ is given in Lemma~\ref{boundofinitialization}.
\end{lem}

See its proof in Appendix~\ref{Proofofboundofmidoutput}

\begin{lem}\label{boundgradient}
	Suppose Assumptions~\ref{activationassumption}, \ref{initilizationassumption} and~\ref{sampleassumption} hold. Assume $\frac{1}{\sqrt{n}}     \left\|\um(t)-\ymm  \right\|_F =  c_y$ and $ \left\| \Umi{h}(t) \right\|_F  \leq  c_u$, $ \|\Wmii{(s)}{l}(t) - \Wmii{(s)}{l}(0)\|_F \leq \sqrt{m} r$, and $ \|\Wmii{(s)}{l}(0)\|_F  \leq \sqrt{m}\cwo$.  Then  for $\forall l$, we have 
	\begin{equation*}
	\begin{split}
	\frac{1}{n}\sum_{i=1}^{n} \left\|\frac{\partial \ell }{\partial \Xmii{(l)}{i}(t)}\right\|_F 
	\leq  \left(1+\alphaii{}{2} + \alphaii{}{3}  \mu \sqrt{\kc}(\taum r +\cwo) \right)^{l}  \taum c_y c_u,
	\end{split}
	\end{equation*}	
	where $\alphaii{}{2}=\max_{s,l}\alphaii{(l)}{s,2}$ and $\alphaii{}{3}=\max_{s,l} \alphaii{(l)}{s,3}$.  
\end{lem}
See its proof in Appendix~\ref{proofofboundgradient}.

\begin{lem}\label{boundofallweights}
	Suppose Assumptions~\ref{activationassumption}, \ref{initilizationassumption} and~\ref{sampleassumption} hold.  Assume $\|\ymm-\um(t)\|_2^2\leq (1-\frac{\eta \lambda}{2})^{t} \|\ymm-\um(0)\|_2^2$ holds for $t=1,\cdots,k$.    Then  by setting \begin{equation*}
	\rcc= \frac{8\cxo\|\ymm-\um(0)\|_2}{\lambda  \sqrt{mn}} \max\left(1,  2 \left(1+\alphaii{}{2} + 2\alphaii{}{3} \mu \sqrt{\kc} \cwo \right)^{l} \alphaii{(l)}{s,3} \mu \sqrt{\kc} \cwo    \right) \leq \cwo,
	\end{equation*}	
	we have that for any $s=1,\cdots, k+1$,  
	\begin{equation*} 
	\begin{split}
	&\|\Wmii{(0)}{}(t) - \Wmii{(0)}{}(0)\|_F  \leq \sqrt{\m}\taum \rcc,\quad \|\Wmii{(l)}{s}(t) - \Wmii{(l)}{s}(0)\|_F  \leq \sqrt{\m} \taum \rcc,\quad \|\Umi{s}(t) - \Umi{s}(0)\|_F  \leq \sqrt{\m} \taum \rcc,\\
	&	\|\Wmii{(0)}{}(t+1) - \Wmii{(0)}{}(t)\|_F=\eta \left\| \frac{\partial F(\Omega)}{\partial \Wmii{(0)}{}(t)}\right\|_F \leq  \frac{4 c \taum \eta  \mu   \cxo    \cwo \sqrt{\kc }}{\sqrt{n}}\left\|\um(t)-\ymm  \right\|_2,\\
	&\|\Wmii{(l)}{s}(t+1) - \Wmii{(l)}{s}(t)\|_F= \eta \left\| \frac{\partial F(\Omega)}{\partial \Wmii{(l)}{s}(t)}\right\|_F\leq  \frac{4 c \taum \eta \alphaii{(l)}{s,3} \mu   \cxo    \cwo \sqrt{\kc }}{\sqrt{n}}\left\|\um(t)-\ymm  \right\|_2,\\
	& \|\Umi{s}(t+1) - \Umi{s}(t)\|_F=\eta \left\| \frac{\partial F(\Omega)}{\partial \Umi{s}(t)}\right\|_F\leq    \frac{2\eta \taum \cxo}{ \sqrt{n}} \|\um(t)-\ymm\|_2 ,\\
	\end{split}
	\end{equation*}	
	where $c= \left(1+\alphaii{}{2} + 2\alphaii{}{3} \mu \sqrt{\kc} \cwo  \right)^{l}$ with $\alphaii{}{2}=\max_{s,l}\alphaii{(l)}{s,2}$ and $\alphaii{}{3}=\max_{s,l} \alphaii{(l)}{s,3}$.  
\end{lem}
See its proof in Appendix~\ref{proofofboundofallweights}.

\begin{lem}\label{boundedtempoutpsss}
	Suppose Assumptions~\ref{activationassumption}, \ref{initilizationassumption} and~\ref{sampleassumption} hold.   Then we have 
	\begin{equation*}
	\begin{split}
	&\left\|\Xmii{(l)}{}(k+1) - \Xmii{(l)}{}(k)\right\|_F \\
	\leq&  \left(1+\alphaii{}{2} +2\sqrt{\kc } \cwo   \alphaii{}{3}   \mu \right)^l \left(1 +     \frac{2  (\alphaii{}{3})^2     \cxo  }{(\alphaii{}{2} +2\sqrt{\kc } \cwo   \alphaii{}{3}   \mu )\sqrt{n}}  \right)  \frac{  4 c  \taum   \tau \eta  \mu^2   \cxo    \cwo  \kc }{\sqrt{n}}    \left\|\um(k)-\ymm  \right\|_F,
	\end{split}
	\end{equation*}
	where  $\alphaii{}{2}=\max_{s,l}\alphaii{(l)}{s,2}$ and $\alphaii{}{3}=\max_{s,l} \alphaii{(l)}{s,3}$.
\end{lem}
See its proof in Appendix~\ref{proofofboundedtempoutpsss}.

\begin{lem}\label{boundedtempoutpsss222}
	Suppose Assumptions~\ref{activationassumption}, \ref{initilizationassumption} and~\ref{sampleassumption} hold.   Then we have 
	\begin{equation*}
	\begin{split}
	\left\|\Wmii{(0)}{}(k)\right\|_F  \leq 2\sqrt{m} \cwo, \quad	\left\|\Wmii{(l)}{s}(k)\right\|_F  \leq2\sqrt{m} \cwo, \quad \left\|\Umi{s}(k)\right\|_F  \leq 2\sqrt{m} \cwo.
	\end{split}
	\end{equation*}
	If $\rcc$ in Lemma~\ref{boundofallweights} satisfies $ \rcc 
	\leq \frac{\cxo}{\left(1+ \alphaii{}{2}+2\alphaii{}{3}\mu\sqrt{\kc} \cwo   \right)^{l}  \mu \sqrt{\kc} }$ which can be achieved by using large $m$, then we have 
	\begin{equation*}  
	\begin{split}
	\left\| \Xmii{(l)}{i} (k)\right\|_F \leq   2\cxo,
	\end{split}
	\end{equation*}	
	where   $\alphaii{}{2}=\max_{s,l}\alphaii{(l)}{s,2}$ and $\alphaii{}{3}=\max_{s,l} \alphaii{(l)}{s,3}$.  
\end{lem}
See its proof in Appendix~\ref{proofofboundedtempoutpsss222}.
 
\begin{lem} \label{boundX}
	Suppose Assumptions~\ref{activationassumption}, \ref{initilizationassumption} and~\ref{sampleassumption} hold.  Then we have 
	\begin{equation*} 
	\begin{split}
	\| \Xmii{(0)}{i}(k)  - \Xmii{(0)}{i}(0) \|_F\leq  \mu \sqrt{\kc }\rcc,\quad 
	\| \Xmii{(l)}{i}(k)  - \Xmii{(l)}{i}(0) \|_F
	\leq      c ( 1 +2 \alphai{3}     \cxo   ) \mu \sqrt{\kc } \taum \rcc,
	\end{split}
	\end{equation*}
	where  $c= \left(1+\alphaii{}{2} + 2\alphaii{}{3} \mu \sqrt{\kc} \cwo  \right)^{l}$ with $\alphaii{}{2}=\max_{s,l}\alphaii{(l)}{s,2}$ and $\alphaii{}{3}=\max_{s,l} \alphaii{(l)}{s,3}$.   Here $\rcc$ is given in Lemma~\ref{boundofallweights}.
\end{lem}
See its proof in Appendix~\ref{proofpfboundX}.

\begin{lem}\label{boundfinaloutput}
	Suppose Assumptions~\ref{activationassumption}, \ref{initilizationassumption} and~\ref{sampleassumption} hold.   
	\begin{equation*} 
	\begin{split}
	|   \ui{i} (k) - \ui{i}(0)| \leq  2 \sqrt{m} h  \left( \cxo +\cwo c ( 1 +2 \alphai{3}     \cxo   ) \mu \sqrt{\kc } \right) \rcc,
	\end{split}
	\end{equation*}
	where $c= \left(1+\alphaii{}{2} + 2\alphaii{}{3} \mu \sqrt{\kc} \cwo  \right)^{l}$ with $\alphaii{}{2}=\max_{s,l}\alphaii{(l)}{s,2}$ and $\alphaii{}{3}=\max_{s,l} \alphaii{(l)}{s,3}$.  Here $\rcc$ is given in Lemma~\ref{boundofallweights}. Besides, we have 
	\begin{equation*}
	\begin{split}
	\left\| \frac{\partial \ell }{\partial \Xmii{(l)}{i}(k)} - \frac{\partial \ell }{\partial \Xmii{(l)}{i}(0)}   \right\|_F
	\leq    c_1 c\alphai{3}   \cwo^2\cxo  \rho   \kc m \taum \rcc,
	\end{split}
	\end{equation*} 	
	where   $c_1$ is a constant. 
\end{lem}
See its proof in Appendix~\ref{proofofboundfinaloutput}.

\begin{lem}\label{initialbound}
	Suppose Assumption~\ref{initilizationassumption} holds.   Then with probability at least $1-\delta/4$, it holds  
	\begin{equation*}
	\begin{cases}
	\|\Wmii{0}{}\|_F \leq \sqrt{\m} \cwo,\\
	\|\Wmii{(l)}{s} (0)\|_F\leq \sqrt{\m} \cwo\ (\forall 0\leq l\leq h-1, 0\leq s\leq l-1),\\
	\|\Umi{s} (0)\|_F\leq \sqrt{\m} \cwo\ (\forall 0\leq s\leq h-1).
	\end{cases}
	\end{equation*}
\end{lem}

See its proof in Appendix~\ref{proofoinitialbound}.

\subsection{Step 1 Linear Convergence of $\|\ymm - \hmm(k)\|_2^2 $}\label{Proofoflinearconvergene}
Here we first present our results and then provides their proofs. 
\begin{lem}\label{mainconvergence}
	Suppose Assumptions~\ref{activationassumption}, \ref{initilizationassumption} and~\ref{sampleassumption} hold.   
	If  $m$ and $\eta $ satisfy
	\begin{equation*}
	\begin{cases}
	m\geq  \frac{c_1 \rho\kc^2 \cwo^2 \|\ymm-\um(0)\|_2^2}{\lambda^2   n} \left(1+ \alphaii{}{2}+2\alphaii{}{3}\mu\sqrt{\kc} \cwo   \right)^{2h},\\ \eta\leq  \frac{c_2 \lambda}{\sqrt{m} \mu^4     \cwo^4 \cxo^2 h^{3}  \kc^2 \left(1+\alphaii{}{2} +2\sqrt{\kc } \cwo   \alphaii{}{3}   \mu \right)^{4h}},
	\end{cases}
	\end{equation*}
	where $c_1$ and $c_2$ are two constants and $\lambda$ is smallest eigenvalue of the Gram matrix $\Gm(t)\ (t=1,\cdots,k-1)$, then with probability at least $1-\delta/2$ we have 
	\begin{equation*}
	\begin{split}
	\|\ymm - \hmm(k)\|_2^2   \leq \left( 1 - \frac{\eta \lambda}{2}\right) \|\ymm - \hmm(k-1)\|_2^2  \leq \left( 1 - \frac{\eta \lambda}{2}\right)^{k}\|\ymm - \hmm(0)\|_2^2.  
	\end{split}
	\end{equation*} 
\end{lem}

See its proof in Appendix~\ref{proofofmainconvergence}.

\begin{lem}\label{boundofeigenvalue}
	Suppose Assumptions~\ref{activationassumption}, \ref{initilizationassumption} and~\ref{sampleassumption} hold.  
	If  $m$  satisfy
	\begin{equation*}
	\begin{split} 
	m\geq \frac{c_3 \alphai{3}^2 \mu^2 \kc    \cxo^2 c^2 }{\lambda^2  n  },
	\end{split}
	\end{equation*}
	where $c_3$ is a constant, $c=\left(1+\alphaii{}{2} + 2\alphaii{}{3}  \mu \sqrt{\kc} \cwo \right)^{h}$, $\alphaii{}{2}=\max_{s,l}\alphaii{(l)}{s,2}$ and $\alphaii{}{3}=\max_{s,l} \alphaii{(l)}{s,3}$, then we have 
	\begin{equation*}
	\begin{split}
	\left\| \Gmii{}{}(k) -  \Gmii{}{}(0)  \right\|_2 \leq\frac{\eta  \lambda_{\min}\left( \Gm(0) \right)}{2},
	\end{split}
	\end{equation*}
	where $ \lambda_{\min}\left( \Gm(0) \right)$ is the smallest eigenvalue of $\Gmii{}{}(0)  $. 
\end{lem}

See its proof in Appendix~\ref{proofofboundofeigenvalue}.

\begin{lem} \label{mainconvergence2}Suppose Assumptions~\ref{activationassumption}, \ref{initilizationassumption} and~\ref{sampleassumption} hold. 	If  $m$ and $\eta$ satisfy
	\begin{equation*}
	\begin{cases}
	m\geq   \frac{c_m'   c^2 \rho  \kc^2  \cwo^2 \mu^2}{\lambda^2 n},\\
	\eta\leq  \frac{c_\eta' \lambda}{\sqrt{m} \mu^4     h^{3}  \kc^2 c^4},
	\end{cases}
	\end{equation*}
	where $c_m$ and $c_\eta$ are two constants, $c=\left(1+\alphaii{}{2} + 2\alphaii{}{3}  \mu \sqrt{\kc} \cwo \right)^{h}$, $\alphaii{}{2}=\max_{s,l}\alphaii{(l)}{s,2}$ and $\alphaii{}{3}=\max_{s,l} \alphaii{(l)}{s,3}$. Then with probability at least $1-\delta$ we have 
	\begin{equation*}
	\begin{split}
	\|\ymm - \hmm(k)\|_2^2   \leq \left( 1 - \frac{\eta  \lambda_{\min}\left( \Gm(0) \right)}{4}\right) \|\ymm - \hmm(k-1)\|_2^2 \leq \left( 1 - \frac{\eta  \lambda_{\min}\left( \Gm(0) \right)}{4}\right)^{k}\|\ymm - \hmm(0)\|_2^2.  
	\end{split}
	\end{equation*} 
\end{lem}
See its proof in Appendix~\ref{proofofmainconvergence2}.

\subsubsection{Proof of Lemma~\ref{mainconvergence}}\label{proofofmainconvergence}

\begin{proof}
	Here we use mathematical  induction to prove the result. For $k=0$,  the results in Theorem~\ref{mainconvergence} holds. Then we assume for $j=1,\cdots,k$, it holds 
	\begin{equation*}
	\begin{split}
	\|\ymm - \hmm(j)\|_2^2  \leq \left( 1 - \frac{\eta \lambda}{2}\right) \|\ymm - \hmm(j-1)\|_2^2 \leq \left( 1 - \frac{\eta \lambda}{2}\right)^j\|\ymm - \hmm(0)\|_2^2\quad (j=1,\cdots,k).  
	\end{split}
	\end{equation*}
	Then we need to prove $j=k+1$ still holds. Our proof has four steps. In the first step, we establish the relation between $\|\ymm - \hmm(j)\|_2^2\leq \|\ymm - \hmm(j)\|_2^2 + H_1 +H_2$. Then in the second, third and fourth steps, we bound the terms $H_1$, $H_2$, $H_3$ respectively. Finally, we combine results to obtain the desired result. 
	
	\textbf{Step 1. Establishing relation between $\|\ymm - \hmm(j)\|_2^2\leq \|\ymm - \hmm(j)\|_2^2 + H_1 +H_2 + H_3$. } 
	
	According to the definition, we can obtain 
	\begin{equation*}
	\begin{split}
	\|\ymm - \hmm(k+1)\|_2^2 =& \|\ymm - \hmm(k) + \hmm(k) - \hmm(k+1)\|_2^2\\
	=& \|\ymm - \hmm(k)\|_2^2 +2 \langle \ymm - \hmm(k), \hmm(k) - \hmm(k+1) \rangle  + \| \hmm(k) - \hmm(k+1)\|_2^2.
	\end{split}
	\end{equation*}
	Then for brevity, $\ell(\Omegam)$ and $\ell_i(\Omegam)$ respectively denote  the losses when feeding the input $(\Xm,\ym)$ and  $(\Xmi{i},\ymi{i})$. Then as introduced in Sec.~\ref{notations}, we denote the gradient of $\ell(\Omegam)$ with respect to  all learnable parameters $\Omegam$ as 
	\begin{equation*}
	\begin{split}
	\nabla_{\Omegam} \ell(\Omegam) = & \left[\vect{\frac{\partial \ell }{\partial \Wmii{(0)}{}} }; \left\{\vect{\frac{\partial \ell }{\partial \Wmii{(l)}{s}} }\right\}_{0\leq l \leq h-1, 0\leq s \leq l-1};  \left\{\vect{\frac{\partial \ell }{\partial \Umi{s}}}\right\}_{0\leq s\leq h-1}   \right] .
	\end{split}
	\end{equation*} 
	Based on the above definitions, when we use gradient descent algorithm to update the variables with learning rate $\eta$, we have
	\begin{equation*}
	\begin{split}
	\hmi{i}(k+1) - \hmi{i}(k) 
	= & \hmi{i}\left(\Omegam(k) - \eta \nabla_{\Omegam} F(\Omegam(k)) \right) - \hmi{i}(\Omegam(k)) \\
	= &- \int_{t=0}^{\eta} \left\langle \nabla_{\Omegam} F(\Omegam(k)) ,   \nabla_{\Omegam}  \hmi{i}\left(\Omegam(k) - s  \nabla_{\Omegam} F(\Omegam(k)) \right) \right\rangle d t 
	= \Deltamii{1}{i}(k)  +  \Deltamii{2}{i}(k) ,
	\end{split}
	\end{equation*}
	where 
	\begin{equation*}
	\begin{split}
	\Deltamii{1}{i}(k) = &- \int_{t=0}^{\eta} \left\langle \nabla_{\Omegam} F(\Omegam(k)) ,   \nabla_{\Omegam}  \hmi{i}\left(\Omegam(k)   \right) \right\rangle d t \\
	\Deltamii{2}{i}(k) = &\int_{t=0}^{\eta} \left\langle \nabla_{\Omegam} F(\Omegam(k)) ,   \nabla_{\Omegam}  \hmi{i}\left(\Omegam(k)   \right) - \nabla_{\Omegam}  \hmi{i}\left(\Omegam(k) - t  \nabla_{\Omegam} F(\Omegam(k)) \right) \right\rangle d t .
	\end{split}
	\end{equation*}
	
	Then we define two important notations:
	\begin{equation*}
	\begin{split}
	\Deltamii{1}{}(k) = [\Deltamii{1}{1}(k) ; \Deltamii{1}{2}(k) ;  \cdots; \Deltamii{1}{n}(k) ]\in\Rs{n},\qquad 
	\Deltamii{2}{}(k) = [\Deltamii{2}{1}(k) ; \Deltamii{2}{2}(k) ;  \cdots; \Deltamii{2}{n}(k) ]\in\Rs{n}.
	\end{split}
	\end{equation*}
	
	In this way, we have $\um(k+1) - \um(k) = \Deltamii{1}{}(k)  + \Deltamii{2}{}(k) $. Now we consider 
	\begin{equation*}
	\begin{split}
	\Deltamii{1}{i}(k) = & - \int_{s=0}^{\eta} \left\langle \nabla_{\Omegam} F(\Omegam(k)) ,   \nabla_{\Omegam}  \hmi{i}\left(\Omegam(k)   \right) \right\rangle \\
	= & - \eta \left\langle \nabla_{\Omegam} F(\Omegam(k)) ,   \nabla_{\Omegam}  \hmi{i}\left(\Omegam(k)   \right) \right\rangle \\
	= & - \frac{\eta}{n} \sum_{j=1}^{n} (\ymi{j}-\hmi{j})\left\langle \nabla_{\Omegam}  \hmi{j}\left(\Omegam(k)   \right)) ,   \nabla_{\Omegam}  \hmi{i}\left(\Omegam(k)   \right) \right\rangle\\
	= & - \frac{\eta}{n} \sum_{j=1}^{n} (\ymi{j}-\hmi{j}) \sum_{t=1}^{(h+1)(\frac{h}{2}+1)}\left\langle \nabla_{\Omegami{t}}  \hmi{j}\left(\Omegam(k)   \right)) ,   \nabla_{\Omegami{t}}  \hmi{i}\left(\Omegam(k)   \right) \right\rangle.
	\end{split}
	\end{equation*}
	Let $\Gmii{ij}{t}(k)=\left\langle \nabla_{\Omegami{t}}  \hmi{j}\left(\Omegam(k)   \right)) ,   \nabla_{\Omegami{t}}  \hmi{i}\left(\Omegam(k)   \right) \right\rangle$.  
	In this way, we have $\Gm(k) = \sum_{t=1}^{(h+1)(\frac{h}{2}+1)}\Gmii{}{t}$. Then $\Deltamii{1}{}(k)$ can be formulated as follows:
	\begin{equation*}
	\begin{split}
	\Deltamii{1}{}(k) = -\eta \Gm(k) (\hmm(k) - \ym).
	\end{split}
	\end{equation*}
	In this way, we can compute 
	\begin{equation*}
	\begin{split}
	2 \langle \ymm - \hmm(k), \hmm(k) - \hmm(k+1) \rangle = & -2 \langle \ymm - \hmm(k),  \Deltamii{1}{}(k)  + \Deltamii{2}{}(k) \rangle\\
	= &   -2 \eta  (\hmm(k) - \ym)^\top \Gm(k) (\hmm(k) - \ym)  -2 \langle \ymm - \hmm(k),    \Deltamii{2}{}(k) \rangle\\
	\end{split}
	\end{equation*}
	
	Therefore, we can decompose $\|\ymm - \hmm(k+1)\|_2^2 $ into 
	\begin{equation}\label{afdsarqwrfdwqda}
	\begin{split}
	&\|\ymm - \hmm(k+1)\|_2^2  \\
	=& \|\ymm - \hmm(k)\|_2^2 +2 \langle \ymm - \hmm(k), \hmm(k) - \hmm(k+1) \rangle  + \| \hmm(k) - \hmm(k+1)\|_2^2\\
	=& \|\ymm - \hmm(k)\|_2^2   -2 \eta  (\hmm(k) - \ym)^\top \Gm(k) (\hmm(k) - \ym)  -2 \langle \ymm - \hmm(k),    \Deltamii{2}{}(k) \rangle+ \| \hmm(k) - \hmm(k+1)\|_2^2\\
	\leq & \|\ymm - \hmm(k)\|_2^2   -2 \eta  (\hmm(k) - \ym)^\top \Gm(k) (\hmm(k) - \ym)   +2 \| \ymm - \hmm(k)\|_2 \|  \Deltamii{2}{}(k) \|_2+ \| \hmm(k) - \hmm(k+1)\|_2^2.
	\end{split}
	\end{equation}
	Let $H_1= -2 \eta  (\hmm(k) - \ym)^\top \Gm(k) (\hmm(k) - \ym)   $,  $H_2=2 \| \ymm - \hmm(k)\|_2 \|  \Deltamii{2}{}(k) \|_2$ and $H_3= \| \hmm(k) - \hmm(k+1)\|_2^2$. The remaining task is to upper bound $H_1\sim H_3$. 
	
	\textbf{Step 2. Bound of $H_1$}. 
	
	To bound $H_1$, we can easily to bound it as follows:
	\begin{equation*}
	\begin{split}
	H_1= -2 \eta  (\hmm(k) - \ym)^\top \Gm(k) (\hmm(k) - \ym) 
	\leq    -2 \eta  \lambda \|\hmm(k) - \ym\|_2^2,
	\end{split}
	\end{equation*}
	where $\lambda=\min_{k} \ \lambda_{\min}(\Gm(k)). $

	\textbf{Step 3. Bound of $H_2$}.  
	
	In this step, we aim to bound $H_2=2 \| \ymm - \hmm(k)\|_2 \|  \Deltamii{2}{}(k) \|_2$ by bounding  $ \|\Deltamii{2}{i}(k)\|_2$. According to the definition, we have 
	\begin{equation*}
	\begin{split}
	\Deltamii{2}{i}(k) = & \int_{t=0}^{\eta} \left\langle \nabla_{\Omegam} F(\Omegam(k)) ,   \nabla_{\Omegam}  \hmi{i}\left(\Omegam(k)   \right) - \nabla_{\Omegam}  \hmi{i}\left(\Omegam(k) - s  \nabla_{\Omegam} F(\Omegam(k)) \right) \right\rangle d t  \\
	\leq & \eta \max_{t\in[0,\eta]} \left\| \nabla_{\Omegam} F(\Omegam(k))\right\|_F   \left\| \nabla_{\Omegam}  \hmi{i}\left(\Omegam(k)   \right) - \nabla_{\Omegam}  \hmi{i}\left(\Omegam(k) - t  \nabla_{\Omegam} F(\Omegam(k)) \right) \right\|_F.\\
	\end{split}
	\end{equation*}
	In this way, we need to bound    $\max_{t\in[0,\eta]}    \left\| \nabla_{\Omegam}  \hmi{i}\left(\Omegam(k)   \right) - \nabla_{\Omegam}  \hmi{i}\left(\Omegam(k) - t  \nabla_{\Omegam} F(\Omegam(k)) \right) \right\|_F$ and  $\left\| \nabla_{\Omegam} F(\Omegam(k))\right\|_F$. 
	
	\textbf{Step 3.1  Bound of $\left\| \nabla_{\Omegam} F(\Omegam(k))\right\|_F$ in $H_2$}.  
	According to the definition, we have 
	\begin{equation*}
	\begin{split}
	\|\nabla_{\Omegam} F(\Omegam(k)) \|_F\leq  & \sum_{t=1}^{(h+1)(h/2+1)}\left\| \nabla_{\Omegami{t}} F(\Omegam(k))\right\|_F \\
	= &\left\| \frac{\partial F(\Omega)}{\partial \Wmii{(0)}{}(k)}\right\|_F + \sum_{l=0}^{h-1} \sum_{s=0}^{l-1} \left\| \frac{\partial F(\Omega)}{\partial \Wmii{(l)}{s}(k)}\right\|_F + \sum_{s=0}^{h-1} \left\| \frac{\partial F(\Omega)}{\partial \Umi{s}(k)}\right\|_F\\
	\led{172} & \left( h + 2c    \mu        \cwo \sqrt{\kc } \left(1+  \sum_{l=0}^{h-1} \sum_{s=0}^{l-1}\alphaii{(l)}{s,3} \right)     \right)   \frac{2    \taum \cxo}{\sqrt{n}}\left\|\um(t)-\ymm  \right\|_2,
	\end{split}
	\end{equation*}
	where \ding{172} holds by using Lemma~\ref{boundofallweights} with $c= \left(1+\alphaii{}{2} + 2\alphaii{}{3} \mu \sqrt{\kc} \cwo  \right)^{l}$, $\alphaii{}{2}=\max_{s,l}\alphaii{(l)}{s,2}$ and $\alphaii{}{3}=\max_{s,l} \alphaii{(l)}{s,3}$ since Lemma~\ref{boundofallweights}   proves 
	\begin{equation*} 
	\begin{split}
	& \left\| \frac{\partial F(\Omega)}{\partial \Wmii{(0)}{}(t)}\right\|_F\! \leq\!  \frac{4 c  \taum  \mu   \cxo    \cwo \sqrt{\kc }}{\sqrt{n}}\left\|\um(t)\!-\!\ymm  \right\|_2,\ \left\| \frac{\partial F(\Omega)}{\partial \Wmii{(l)}{s}(t)}\right\|_F\!\!\leq\!  \frac{4 c \taum \alphaii{(l)}{s,3} \mu   \cxo    \cwo \sqrt{\kc }}{\sqrt{n}}\left\|\um(t)\!-\!\ymm  \right\|_2,\\
	&\left\| \frac{\partial F(\Omega)}{\partial \Umi{s}(t)}\right\|_F\leq    \frac{2\taum \cxo}{ \sqrt{n}} \|\um(t)-\ymm\|_2 ,\\
	\end{split}
	\end{equation*}

	\textbf{Step 3.2  Bound of $\left\| \nabla_{\Omegam}  \hmi{i}\left(\Omegam(k)   \right) - \nabla_{\Omegam}  \hmi{i}\left(\Omegam(k) - t  \nabla_{\Omegam} F(\Omegam(k)) \right) \right\|_F$ in $H_2$}.  
	
	For brevity, let $\Omegam(k,t) = \Omegam(k) - t  \nabla_{\Omegam} F(\Omegam(k)) $. In this way, we can bound 
	\begin{equation*}
	\begin{split}
	&\left\| \nabla_{\Omegam}  \hmi{i}\left(\Omegam(k)   \right) - \nabla_{\Omegam}  \hmi{i}\left(\Omegam(k,t)) \right) \right\|_F  
	\leq   \sum_{o=1}^{(h+1)(h/2+1)}\left\|  \nabla_{\Omegami{o}}  \hmi{i}\left(\Omegam(k)   \right) - \nabla_{\Omegami{o}}  \hmi{i}\left(\Omegam(k,s)\right) \right\|_F \\
	= &\left\| \frac{\partial \ui{i}}{\partial \Wmii{(0)}{}(k)} - \frac{\partial \ui{i}}{\partial \Wmii{(0)}{}(k,t)}\right\|_F + \!\sum_{l=0}^{h-1} \!\sum_{s=0}^{l-1} \!\!\left\| \frac{\partial \ui{i}}{\partial \Wmii{(l)}{s}(k)} - \frac{\partial \ui{i}}{\partial \Wmii{(l)}{s}(k,t)}\right\|_F\!\! +\! \sum_{s=0}^{h-1} \left\| \frac{\partial \ui{i}}{\partial \Umi{s}(k)} - \frac{\partial \ui{i}}{\partial \Umi{s}(k,t)}\right\|_F.
	\end{split}
	\end{equation*}
	In the following, we will bound each term. We first look at $\left\| \frac{\partial \ui{i}}{\partial \Umi{s}(k)} - \frac{\partial \ui{i}}{\partial \Umi{s}(k,t)}\right\|_F$. By using Lemma~\ref{gradientcomputation}, we have $\frac{\partial \ui{i}}{\partial \Umi{s}(k)} =\Xmii{(l)}{i}(k) $. Therefore, we can obtain
	\begin{equation}\label{afdasfwrfdw}
	\begin{split}
	\left\| \frac{\partial \ui{i}}{\partial \Umi{s}(k)} - \frac{\partial \ui{i}}{\partial \Umi{s}(k,t)}\right\|_F = &  \left\| \Xmii{(l)}{i}(k) - \Xmii{(l)}{i}(k,t)\right\|_F =t  \left\| \frac{\partial F(\Omegam)}{\partial \Xmii{(l)}{i}(k) }  \right\|_F\\
	\leq & t \frac{1}{n} \sum_{i=1}^{n}\left\| \frac{\partial \ell_i}{\partial \Xmii{(l)}{i}(k) }  \right\|_F \led{172} \eta \left(1+\alphaii{}{2} + 2\alphaii{}{3} \mu \sqrt{\kc} \cwo \right)^{l}  \taum c_y c_u,
	\end{split}
	\end{equation}
	where \ding{172} holds since in Lemma~\ref{boundofallweights}, we have show  
	\begin{equation} \label{conditionasda}
	\begin{split}
	\max\left(\|\Wmii{(0)}{}(t) - \Wmii{(0)}{}(0)\|_F , \|\Wmii{(l)}{s}(t) - \Wmii{(l)}{s}(0)\|_F , \|\Umi{s}(t) - \Umi{s}(0)\|_F \right) \!\leq\! \sqrt{\m} \taum \rcc\!\leq\! \sqrt{\m} \cwo,
	\end{split}
	\end{equation}	
	which allows us to 
	use Lemma~\ref{boundgradient} which shows 
	\begin{equation} \label{conditiadafdonasda}
	\begin{split}
	\frac{1}{n} \sum_{i=1}^{n}\!\left\| \frac{\partial \ell_i}{\partial \Xmii{(l)}{i}(k) }  \right\|_F\! \leq\!   \left(1\!+\!\alphaii{}{2} \!+\! \alphaii{}{3} \mu \sqrt{\kc}(\taum\rcc +\cwo) \right)^{l} \taum c_y c_u\!\leq\! \left(1+\alphaii{}{2} + 2\alphaii{}{3} \mu \sqrt{\kc} \cwo \right)^{l}  \taum c_y c_u,
	\end{split}
	\end{equation}	
	where  parameters $\frac{1}{\sqrt{n}}   \left\|\um(t)-\ymm  \right\|_2 =  c_y$ and $ \left\| \Umi{h}(t) \right\|_F  \leq  c_u$,  $\alphaii{}{2}=\max_{s,l}\alphaii{(l)}{s,2}$ and $\alphaii{}{3}=\max_{s,l} \alphaii{(l)}{s,3}$.  Moreover, from Lemma~\ref{boundofallweights}, we have $ \left\| \Umi{h}(t) \right\|_F \leq  \left\| \Umi{h}(t) -   \Umi{h}(0)  \right\|_F  +  \left\| \Umi{h}(0) \right\|_F\leq 2\sqrt{m} \cwo $.  In this way, we have
	\begin{equation*}
	\begin{split}
	& \sum_{s=1}^{h} \left\| \frac{\partial \ui{i}}{\partial \Umi{s}(k)} - \frac{\partial \ui{i}}{\partial \Umi{s}(k,t)}\right\|_F  \leq   \eta \taum h \left(1+\alphaii{}{2} + 2\alphaii{}{3} \mu \sqrt{\kc}\cwo \right)^{l}  \sqrt{m}\cwo\frac{1}{\sqrt{n}}   \left\|\um(t)-\ymm  \right\|_2\\
	\leq &  \eta \taum h \left(1+\alphaii{}{2} + 2\alphaii{}{3} \mu \sqrt{\kc} \cwo \right)^{l}  \sqrt{m}\cwo\frac{1}{\sqrt{n}}   \left(1- \frac{\eta \lambda}{2}\right)^{t/2} \left\|\um(0)-\ymm  \right\|_2= \eta c_1,
	\end{split}
	\end{equation*} 	
	where $c_1=\taum  h \left(1+\alphaii{}{2} + 2\alphaii{}{3} \mu \sqrt{\kc}\cwo\right)^{l}  \sqrt{m}\cwo\frac{1}{\sqrt{n}}   \left(1- \frac{\eta \lambda}{2}\right)^{t/2} \left\|\um(0)-\ymm  \right\|_F$ is a constant. 
	
	Then we  consider $\left\| \frac{\partial \ui{i}}{\partial \Wmii{(l)}{s}(k)} - \frac{\partial \ui{i}}{\partial \Wmii{(l)}{s}(k,t)}\right\|_F  $ as follows:
	\begin{equation*}
	\begin{split}
	&\left\| \frac{\partial \ui{i}}{\partial \Wmii{(l)}{s}(k)} - \frac{\partial \ui{i}}{\partial \Wmii{(l)}{s}(k,t)}\right\|_F 
	\!	= \!   \alphaii{(l)}{s,3}\tau  \left[\left\|  \Phi(\Xmii{(s)}{i}(k))  \!\!  \left(\!\sigma'\left( \Wmii{(l)}{s}(k) \Phi(\Xmii{(s)}{i}(k))\right) \odot \frac{\partial \ui{i} }{\partial \Xmii{(l)}{i}(k)}\right)^{\top}  \right.\right. \\ 
	& \qquad\qquad\qquad\qquad\qquad \left.\left.-\Phi(\Xmii{(s)}{i}(k,t))    \left(\sigma'\left( \Wmii{(l)}{s}(k,t) \Phi(\Xmii{(s)}{i}(k,t))\right) \odot \frac{\partial \ui{i} }{\partial \Xmii{(l)}{i}(k,t)}\right)^{\top}  \right\|_F \right]\\
	&\qquad \qquad\qquad\qquad\qquad  \led{172}   \alphaii{(l)}{s,3}\tau \frac{a_1 a_2 (b_1+ b_2)}{\max(a_1,a_2)} ,
	\end{split}
	\end{equation*}
	where \ding{172} uses Lemma~\ref{boundmatrix}. For parameters $a_1, a_2, b_1, b_2$ satisfies 
	\begin{equation*}
	\begin{split}
	&a_1 =\max\left(\left\|\Phi(\Xmii{(s)}{i}(k)) \right\|_2, \left\|\Phi(\Xmii{(s)}{i}(k,t)) \right\|_2\right) \leq \sqrt{\kc} \max\left(\left\| \Xmii{(s)}{i}(k) \right\|_2, \left\|\Xmii{(s)}{i}(k,t) \right\|_2\right),\\
	&a_2 \!=\!\max\left(\left\|\sigma'\!\!\left( \Wmii{(l)}{s}(k) \Phi(\Xmii{(s)}{i}(k))\right) \!\odot\! \frac{\partial \ui{i} }{\partial \Xmii{(l)}{i}(k)}\right\|_2, \left\|\sigma'\left( \Wmii{(l)}{s}(k,t) \Phi(\Xmii{(s)}{i}(k,t))\right) \!\odot \!\frac{\partial \ui{i} }{\partial \Xmii{(l)}{i}(k,t)}\right\|_2\right)\!,\\
	&b_1 = \left\|\Phi(\Xmii{(s)}{i}(k))-\Phi(\Xmii{(s)}{i}(k,t)) \right\|_2\leq \sqrt{\kc}\left\|\Xmii{(s)}{i}(k)-\Xmii{(s)}{i}(k,t) \right\|_2,\\
	&b_2 = \left\|\sigma'\left( \Wmii{(l)}{s}(k) \Phi(\Xmii{(s)}{i}(k))\right) \odot \frac{\partial \ui{i} }{\partial \Xmii{(l)}{i}(k)}- \sigma'\left( \Wmii{(l)}{s}(k,t) \Phi(\Xmii{(s)}{i}(k,t))\right) \odot \frac{\partial \ui{i} }{\partial \Xmii{(l)}{i}(k,t)} \right\|_2.
	\end{split}
	\end{equation*}

	In Lemma~\ref{boundofinitialization}, we show that when Eqn.~\eqref{boundofinitialization} holds which is proven in Lemma~\ref{boundofallweights}, then $\|\Xmii{(l)}{i} (0)\|_F\leq \cxo$.  Under Eqn.~\eqref{boundofinitialization}, Lemma~\ref{boundofmidoutput}  shows 
	\begin{equation}\label{adafcsdfcas}
	\|\Xmii{(l)}{i} (k)-\Xmii{(l)}{i} (0)\|_F  
	\leq  \left(1+ \alphaii{}{2}+2\alphaii{}{3}\mu\sqrt{\kc} \cwo   \right)^{l}  \mu \sqrt{\kc}\taum  \rcc \led{172} \cxo,
	\end{equation}
	where \ding{172} holds since in Lemma~\ref{boundofallweights}, we set $m=\Oc{\frac{\rho\kc^2 \cwo^2 \|\ymm-\um(0)\|_2^2}{\lambda^2   n} \left(1+ \alphaii{}{2}+2\alphaii{}{3}\mu\sqrt{\kc} \cwo   \right)^{2h} }$  such that 
	\begin{equation*}
	\begin{split}
	\rcc=& \frac{8\cxo\|\ymm-\um(0)\|_2}{\lambda  \sqrt{mn}} \max\left(1,  2 \left(1+\alphaii{}{2} + 2\alphaii{}{3} \mu \sqrt{\kc} \cwo \right)^{l} \alphaii{(l)}{s,3} \mu \sqrt{\kc} \cwo    \right) \\
	\leq& \frac{\cxo}{\left(1+ \alphaii{}{2}+2\alphaii{}{3}\mu\sqrt{\kc} \cwo   \right)^{l}  \mu \sqrt{\kc} }.
	\end{split}
	\end{equation*}	
	
	By using Lemma~\ref{boundofmidoutput} and Lemma~\ref{boundofinitialization}, we have 
	\begin{equation}\label{adafcsadfcaswre3qwaadadf}
	\begin{split}\|\Xmii{(s)}{}(t)\| \leq  &
	\|\Xmii{(l)}{i} (k)-\Xmii{(l)}{i} (0)\|_F + \|\Xmii{(l)}{i} (0)\|_F\leq 2\cxo.
	\end{split}
	\end{equation}	
	Then by using Eqn.~\eqref{conditiadafdonasda} we upper bound $  \left\|\Xmii{(s)}{i}(k,t) \right\|_2$ as follows:
	\begin{equation*}
	\begin{split}
	\left\|\Xmii{(s)}{i}(k,t) \right\|_2 \leq &  \left\|\Xmii{(s)}{i}(k) -t \frac{\partial F(\Omegam)}{\partial \Xmii{(s)}{i}(k) }\right\|_2\leq \left\|\Xmii{(s)}{i}(k) \right\|_2 + t \frac{1}{n}\sum_{i=1}^{n} \left\| \frac{\partial \ell_i}{\partial \Xmii{(s)}{i}(k) }\right\|_2\\
	\leq &  2\cxo + \eta \taum  \left(1+\alphaii{}{2} + 2\alphaii{}{3} \mu \sqrt{\kc} \cwo \right)^{l}  \sqrt{m}\cwo\frac{1}{\sqrt{n}}   \left\|\um(t)-\ymm  \right\|_F\leq c_2,\\
	\end{split}
	\end{equation*}
	where $c_2= 2\cxo + \eta  \taum \left(1+\alphaii{}{2} + 2\alphaii{}{3} \mu \sqrt{\kc} \cwo \right)^{l}  \sqrt{m}\cwo\frac{1}{\sqrt{n}} \left(1- \frac{\eta \lambda}{2}\right)^{t/2} \left\|\um(0)-\ymm  \right\|_F$ is a constant. 
	In this way, we can upper bound 
	\begin{equation*}
	\begin{split}
	a_1 \leq \sqrt{\kc} \max\left( 2\cwo, c_2\right),\qquad b_1 \led{172} \frac{\sqrt{\kc}c_1 \eta}{h},
	\end{split}
	\end{equation*}
	where \ding{172} uses the results in Eqn.~\eqref{afdasfwrfdw}. Now we try to bound $a_2$ and $b_2$ as follows: 
	\begin{equation*}
	\begin{split}
	a_2 \!=&\!\max\left(\left\|\sigma'\!\!\left( \Wmii{(l)}{s}(k) \Phi(\Xmii{(s)}{i}(k))\right) \!\odot\! \frac{\partial \ui{i} }{\partial \Xmii{(l)}{i}(k)}\right\|_2, \left\|\sigma'\left( \Wmii{(l)}{s}(k,t) \Phi(\Xmii{(s)}{i}(k,t))\right) \!\odot \!\frac{\partial \ui{i} }{\partial \Xmii{(l)}{i}(k,t)}\right\|_2\right)\\
	\leq &\mu \max\left(\left\|  \frac{\partial \ui{i} }{\partial \Xmii{(l)}{i}(k)}\right\|_2, \left\| \frac{\partial \ui{i} }{\partial \Xmii{(l)}{i}(k,t)}\right\|_2\right) \led{172} \mu (1+L) c_1^2\eta^2, 
	\end{split}
	\end{equation*}
	where \ding{172} uses $\left\| \frac{\partial \ui{i} }{\partial \Xmii{(l)}{i}(k,t)}\right\|_2 \leq \left\| \frac{\partial \ui{i} }{\partial \Xmii{(l)}{i}(k,t)}\right\|_F \leq \left\| \frac{\partial \ui{i} }{\partial \Xmii{(l)}{i}(k)} \right\|_F + L \|\Xmii{(l)}{i}(k,t)-\Xmii{(l)}{i}(k)\|_F^2 \led{173} (1+L) c_1^2\eta^2$ where $L$ is the Lipschitz constant of $\frac{\partial \ui{i} }{\partial \Xmii{(l)}{}}$. In \ding{173} we use the results in Eqn.~\eqref{adafcsadfcaswre3qwaadadf}.   Since  $\sigma$ is   $\rho$-smooth and $\ui{}$ is $h$-layered, by computing, we know $L$ is at the order of $\Oc{\beta^h}$ and is a constant. For $b_2$ we can bound it as follows: 
	\begin{equation*}
	\begin{split}
	b_2 \leq \mu \left\|  \frac{\partial \ui{i} }{\partial \Xmii{(l)}{i}(k)}-  \frac{\partial \ui{i} }{\partial \Xmii{(l)}{i}(k,t)} \right\|_2 \leq 2\mu (1+L) c_1^2\eta^2. 
	\end{split}
	\end{equation*}
	Therefore, we can bound 
	\begin{equation*}
	\begin{split}
	\sum_{l=1}^{h} \sum_{s=0}^{l-1} \left\| \frac{\partial \ui{i}}{\partial \Wmii{(l)}{s}(k)} - \frac{\partial \ui{i}}{\partial \Wmii{(l)}{s}(k,t)}\right\|_F   \leq   \tau \frac{a_1 a_2 (b_1+ b_2)}{\max(a_1,a_2)}   \sum_{l=1}^{h} \sum_{s=0}^{l-1} \alphaii{(l)}{s,3} = c_3 \eta,
	\end{split}
	\end{equation*}
	where $\alphaii{}{3}=\max \alphaii{(l)}{s,3}$ and $c_3=\frac{\tau \sqrt{\kc} \max\left( 2\cwo, c_2\right) \mu (1+L) c_1^2\eta^2 }{\max(\sqrt{\kc} \max\left( 2\cwo, c_2\right), \mu (1+L) c_1^2\eta^2, )} \left( \frac{\sqrt{\kc}c_1 }{h}+2\mu (1+L) c_1^2\eta\right)$ is a constant. By using the same method, we can bound 
	\begin{equation*}
	\begin{split}
	&\left\| \frac{\partial \ui{i}}{\partial \Wmii{(0)}{}(k)} - \frac{\partial \ui{i}}{\partial \Wmii{(0)}{}(k,t)}\right\|_F \\
	=&  \tau \left\| \Phi(\Xmii{}{i})    \left(\sigma'\left( \Wmii{(0)}{}(k) \Phi(\Xmii{}{i})\right) \odot \frac{\partial \ui{i} }{\partial \Xmii{(0)}{i}(k)}\right)^{\top} \!\!\!-  \Phi(\Xmii{}{i})    \left(\sigma'\left( \Wmii{(0)}{}(k,t) \Phi(\Xmii{}{i})\right) \odot \frac{\partial \ui{i} }{\partial \Xmii{(0)}{i}(k,t)}\right)^{\top}\right\|_F\\
	\led{172} &  \tau \sqrt{\kc} \left\|  \frac{\partial \ui{i} }{\partial \Xmii{(0)}{i}(k)}  - \frac{\partial \ui{i} }{\partial \Xmii{(0)}{i}(k,t)} \right\|_F \leq 2\mu (1+L) c_1^2\eta^2= c_4\eta,
	\end{split}
	\end{equation*}
	where \ding{172} uses $\|\Phi{(\Xmii{}{i})}\|_F \leq \sqrt{\kc}\|\Xmii{}{i}\|_F\leq \sqrt{\kc}$ and $\sigma$ is 
	$\mu$-Lipschitz, and $c_4 =2\mu (1+L) c_1^2\eta$. By combing the above results, we can further conclude
	\begin{equation*}
	\begin{split}
	\left\| \nabla_{\Omegam}  \hmi{i}\left(\Omegam(k)   \right) - \nabla_{\Omegam}  \hmi{i}\left(\Omegam(k,t)) \right) \right\|_F  
	\leq  (c_1+ c_3 +c_4) \eta =c_5\eta,
	\end{split}
	\end{equation*}
	which further gives
	\begin{equation*}
	\begin{split}
	\Deltamii{2}{i}(k)  
	\leq & \eta \max_{t\in[0,\eta]} \left\| \nabla_{\Omegam} F(\Omegam(k))\right\|_F   \left\| \nabla_{\Omegam}  \hmi{i}\left(\Omegam(k)   \right) - \nabla_{\Omegam}  \hmi{i}\left(\Omegam(k) - t  \nabla_{\Omegam} F(\Omegam(k)) \right) \right\|_F.\\ 
	\leq & \eta^2 c_5  \left( h + 2c    \mu        \cwo \sqrt{\kc } \left(1+  \sum_{l=1}^{h} \sum_{s=0}^{l-1}\alphaii{(l)}{s,3} \right)     \right)   \frac{2 \taum   \cxo}{\sqrt{n}}\left\|\um(t)-\ymm  \right\|_F =\hat{c} \eta^2\left\|\um(t)-\ymm  \right\|_F,
	\end{split}
	\end{equation*}
	where $\hat{c} = c_5  \left( h + 2c    \mu        \cwo \sqrt{\kc } \left(1+  \sum_{l=1}^{h} \sum_{s=0}^{l-1}\alphaii{(l)}{s,3} \right)     \right)   \frac{2    \taum \cxo}{\sqrt{n}}$. Therefore we have 
	
	\textbf{Step 3.3 Upper bound $H_2=2 \| \ymm - \hmm(k)\|_2 \|  \Deltamii{2}{}(k) \|_2$.}    
	By combining the above results, we can bound 
	\begin{equation*}
	\begin{split}
	H_2=2 \| \ymm - \hmm(k)\|_2 \|  \Deltamii{2}{}(k) \|_2\leq \hat{c} \eta^2\left\|\um(t)-\ymm  \right\|_2^2,
	\end{split}
	\end{equation*}
	where $\hat{c}=\Oc{  \frac{\mu \cxo\cwo^2 \sqrt{\kc m} h^3 (1+\alphai{2}+ 2\alphai{3}\mu\sqrt{\kc}\cwo)^h }{n}}$.
	
	\textbf{Step 4. Upper bound $H_3=\| \hmm(k) - \hmm(k+1)\|_2^2 $.}    
	\begin{equation*}
	\begin{split}
	\| \hmm(k) - \hmm(k+1)\|_2^2 =& \sum_{i=1}^{n} \left( \sum_{s=0}^{h-1}  \left(\langle \Umi{s}(k), \Xmii{(l)}{i}(k) \rangle -\langle \Umi{s}(k+1), \Xmii{(l)}{i}(k+1) \rangle \right) \right)^2\\
	\leq & \sqrt{h}  \sum_{i=1}^{n} \sum_{s=0}^{h-1}   \left(\langle \Umi{s}(k), \Xmii{(l)}{i}(k) \rangle -\langle \Umi{s}(k+1), \Xmii{(l)}{i}(k+1) \rangle \right)^2.
	\end{split}
	\end{equation*}
	Now we consider each term:
	\begin{equation*}
	\begin{split}
	& \left(\langle \Umi{s}(k), \Xmii{(l)}{i}(k) \rangle -\langle \Umi{s}(k+1), \Xmii{(l)}{i}(k+1) \rangle \right)^2 \\
	= &  \left(\langle \Umi{s}(k)- \Umi{s}(k+1), \Xmii{(l)}{i}(k+1) \rangle + \langle \Umi{s}(k), \Xmii{(l)}{i}(k) - \Xmii{(l)}{i}(k+1) \rangle\right)^2\\
	\leq &  2 \| \Umi{s}(k)- \Umi{s}(k+1)\|_F^2 \| \Xmii{(l)}{i}(k+1)\|_F^2 + 2\| \Umi{s}(k)\|_F^2  \|\Xmii{(l)}{i}(k) - \Xmii{(l)}{i}(k+1)\|_F^2\\
	\led{172} &  8\cxo^2 \| \Umi{s}(k)- \Umi{s}(k+1)\|_F^2  + 8 m\cwo^2   \|\Xmii{(l)}{i}(k) - \Xmii{(l)}{i}(k+1)\|_F^2\\
	\led{173} & \frac{32\eta^2 \cxo^2}{n} \left[\cxo^2  +   4 c^2    \mu^4     \cwo^4  \kc^2 \left(1+\alphaii{}{2} +2\sqrt{\kc } \cwo   \alphaii{}{3}   \mu \right)^{2l} \left(1 +     \frac{2  (\alphaii{}{3})^2     \cxo  }{(\alphaii{}{2} +2\sqrt{\kc } \cwo   \alphaii{}{3}   \mu )\sqrt{n}}  \right)^2 \right]  \\
	& \cdot \left\|\um(k)-\ymm  \right\|_2^2,
	\end{split}
	\end{equation*}
	where \ding{172} uses $\| \Xmii{(l)}{i}(k+1)\|_F^2\leq 4\cxo^2$ in Eqn.~\eqref{adafcsadfcaswre3qwaadadf}, and the results in Eqn.~\eqref{conditionasda} that $\| \Umi{s}(k)\|_F \leq \| \Umi{s}(k)-\Umi{s}(0)\|_F + \| \Umi{s}(0)\|_F  \leq 2\sqrt{m}\cwo$; \ding{173} holds since (1) in Lemma~\ref{boundofallweights} we have $\|\Umi{s}(t+1) - \Umi{s}(t)\|_F=\eta \left\| \frac{\partial F(\Omega)}{\partial \Umi{s}(t)}\right\|_F\leq    \frac{2\eta \cxo}{ \sqrt{n}} \|\um(t)-\ymm\|_2 $
	where $c= \left(1+\alphaii{}{2} + 2\alphaii{}{3} \mu \sqrt{\kc} \cwo  \right)^{l}$ with $\alphaii{}{2}=\max_{s,l}\alphaii{(l)}{s,2}$ and $\alphaii{}{3}=\max_{s,l} \alphaii{(l)}{s,3}$, and (2) in Lemma~\ref{boundedtempoutpsss} we have 
	\begin{equation*}
	\begin{split}
	&\left\|\Xmii{(l)}{}(k+1) - \Xmii{(l)}{}(k)\right\|_F \\
	\leq&  \left(1+\alphaii{}{2} +2\sqrt{\kc } \cwo   \alphaii{}{3}   \mu \right)^l \left(1 +     \frac{2  (\alphaii{}{3})^2     \cxo  }{(\alphaii{}{2} +2\sqrt{\kc } \cwo   \alphaii{}{3}   \mu )\sqrt{n}}  \right)  \frac{  4 c   \tau \eta  \mu^2   \cxo    \cwo  \kc }{\sqrt{n}}    \left\|\um(k)-\ymm  \right\|_2.
	\end{split}
	\end{equation*}
	
	In this way, we can conclude 
	\begin{equation*}
	\begin{split}
	\| \hmm(k) - \hmm(k+1)\|_2^2 \leq    \eta^2  \tilde{c}  \left\|\um(k)-\ymm  \right\|_2^2,
	\end{split}
	\end{equation*}
	where $\tilde{c}= 32  \cxo^2 h^{1.5}  \left[\cxo^2  +   4 c^2  \mu^4     \cwo^4  \kc^2 \left(1+\alphaii{}{2} +2\sqrt{\kc } \cwo   \alphaii{}{3}   \mu \right)^{2l} \left(1 +     \frac{2  (\alphaii{}{3})^2     \cxo  }{(\alphaii{}{2} +2\sqrt{\kc } \cwo   \alphaii{}{3}   \mu )\sqrt{n}}  \right)^2 \right]=\Oc{ \mu^4     \cwo^4 \cxo^2 h^{1.5}  \kc^2 \left(1+\alphaii{}{2} +2\sqrt{\kc } \cwo   \alphaii{}{3}   \mu \right)^{4l} }$.

	\textbf{Step 5. Upper bound $\|\ymm - \hmm(k+1)\|_2^2 $.}    
	
	In this way, by using Eqn.~\eqref{afdsarqwrfdwqda} we can finally obtain
	\begin{equation*}
	\begin{split}
	&\|\ymm - \hmm(k+1)\|_2^2 \leq \|\ymm - \hmm(k)\|_2^2  + H_1 + H_2 + H_3\\ 
	\led{172} & \|\ymm - \hmm(k)\|_2^2   -2 \eta  \lambda \|\hmm(k) - \ym\|_2^2  +2 \hat{c} \eta^2\left\|\um(t)-\ymm  \right\|_2^2+     \eta^2  \tilde{c}  \left\|\um(k)-\ymm  \right\|_2^2\\
	=&\left( 1 - \eta \lambda+(2 \hat{c}+\tilde{c} ) \eta^2 \right) \|\ymm - \hmm(k)\|_2^2 \\
	\led{173} &\left( 1 - \frac{\eta \lambda}{2}\right)\|\ymm - \hmm(k)\|_2^2   
	\end{split}
	\end{equation*}
	where \ding{172} holds by using $H_1 \leq -2 \eta  \lambda \|\hmm(k) - \ym\|_2^2 $,  $H_2\leq2 \hat{c} \eta^2\left\|\um(t)-\ymm  \right\|_2^2 $ and $H_3\leq \eta^2  \tilde{c}  \left\|\um(k)-\ymm  \right\|_2^2$;  \ding{173} holds by setting $\eta\leq  \frac{  \lambda}{2(2 \hat{c}+\tilde{c} )}=\Oc{\frac{\lambda}{\sqrt{m} \mu^4     \cwo^4 \cxo^2 h^{3}  \kc^2 \left(1+\alphaii{}{2} +2\sqrt{\kc } \cwo   \alphaii{}{3}   \mu \right)^{4l}}}$. The proof is completed. 
\end{proof}

\subsubsection{Proof of Lemma~\ref{boundofeigenvalue}}\label{proofofboundofeigenvalue}

\begin{proof}
	
	According to the definitions in Sec.~\ref{notations}, we can write  	
	\begin{equation*}
	\begin{split}
	\left\| \Gmii{}{}(k) -  \Gmii{}{}(0)  \right\|_2 \leq &\left\|\Gmbii{}{0}(k) - \Gmbii{}{0}(0) \right\|_{2} \!+\! \sum_{l=0}^{h-1} \sum_{s=0}^{l-1} \left\|\Gmii{}{ls}(k) -  \Gmii{}{ls}(0)\right\|_{2} \!+ \!\sum_{s=0}^{h-1} \left\| \Gmii{}{s}(k)-\Gmii{}{s}(0) \right\|_{2}.
	\end{split}
	\end{equation*}
	In this way, we only need to upper bound $\left\|\Gmbii{}{0}(k) - \Gmbii{}{0}(0) \right\|_{2}$, $\left\|\Gmii{}{ls}(k) -  \Gmii{}{ls}(0)\right\|_{2}$ and $\left\| \Gmii{}{s}(k)-\Gmii{}{s}(0) \right\|_{2}$.
 
	\textbf{Step 1. Bound of $\left\| \Gmii{}{s}(k)-\Gmii{}{s}(0) \right\|_{2}$ ($s=0,\cdots,h-1$)}. 
	
	For analysis, we first recall existing results. Lemma~\ref{boundofallweights} shows   
	\begin{equation}  \label{afdcsarfwaefca}
	\begin{split}
	\max\left(\|\Wmii{(0)}{}(t) - \Wmii{(0)}{}(0)\|_F , \|\Wmii{(l)}{s}(t) - \Wmii{(l)}{s}(0)\|_F , \|\Umi{s}(t) - \Umi{s}(0)\|_F \right) \!\leq\! \sqrt{\m} \taum \rcc\!\leq\! \sqrt{\m} \cwo,
	\end{split}
	\end{equation}	
	where $c= \left(1+\alphaii{}{2} + 2\alphaii{}{3} \mu \sqrt{\kc} \cwo  \right)^{l}$ with $\alphaii{}{2}=\max_{s,l}\alphaii{(l)}{s,2}$ and $\alphaii{}{3}=\max_{s,l} \alphaii{(l)}{s,3}$.  Based on this result, Lemma~\ref{boundedtempoutpsss222} shows 
	\begin{equation}\label{fsafdsaf}
	\begin{split}
	\left\|\Wmii{(0)}{}(k)\right\|_F  \leq 2\sqrt{m} \cwo, \	\left\|\Wmii{(l)}{s}(k)\right\|_F  \leq2\sqrt{m} \cwo, \  \left\|\Umi{s}(k)\right\|_F  \leq 2\sqrt{m} \cwo, \ \left\| \Xmii{(l)}{i} (k)\right\|_F \leq   2\cxo.
	\end{split}
	\end{equation}
	Moreover, Lemma~\ref{boundX} shows  
	\begin{equation*} 
	\begin{split}
	\| \Xmii{(0)}{i}(k)  - \Xmii{(0)}{i}(0) \|_F\leq  \mu \sqrt{\kc }\taum\rcc,\quad 
	\| \Xmii{(l)}{i}(k)  - \Xmii{(l)}{i}(0) \|_F
	\leq      c ( 1 +2 \alphai{3}     \cxo   ) \mu \sqrt{\kc }\taum\rcc.
	\end{split}
	\end{equation*}

	To bound $H_s$, we only need to bound each entry in $(\Gmii{}{s}(k) - \Gmii{}{s}(0))$:
	\begin{equation*}
	\begin{split}
	| \Gmii{}{s}(k)& -\Gmii{}{s}(0) |  =\left| \left\langle \frac{\partial \ell_i}{\partial \Umi{s}(k)}, \frac{\partial \ell_j}{\partial \Umi{s}(k)}\right\rangle - \left\langle \frac{\partial \ell_i}{\partial \Umi{s}(0)}, \frac{\partial \ell_j}{\partial \Umi{s}(0)}\right\rangle \right|\\
	=& \left| \left\langle \Xmii{(s)}{i}(k), \Xmii{(s)}{j}(k)\right\rangle - \left\langle \Xmii{(s)}{i}(0), \Xmii{(s)}{j}(0)\right\rangle \right|\\
	\leq & \left| \left\langle \Xmii{(s)}{i}(k) - \Xmii{(s)}{i}(0), \Xmii{(s)}{j}(k)\right\rangle \right|+\left|  \left\langle \Xmii{(s)}{i}(0), \Xmii{(s)}{j}(k) -\Xmii{(s)}{j}(0)\right\rangle \right|\\
	\leq & \left\|\Xmii{(s)}{i}(k) - \Xmii{(s)}{i}(0)\right\|_F \left\| \Xmii{(s)}{j}(k)\right\|_F  +\left\|  \Xmii{(s)}{i}(0)\right\|_F \left\|\Xmii{(s)}{j}(k) -\Xmii{(s)}{j}(0) \right\|_F\\
	\led{172} & 4 \cxo c ( 1 +2 \alphai{3}     \cxo   ) \mu \sqrt{\kc } \taum\rcc,
	\end{split}
	\end{equation*}
	So we can further bound
	\begin{equation*}
	\begin{split}
	\left\| \Gmii{}{s}(k)-\Gmii{}{s}(0) \right\|_{2} \leq \sqrt{n} \left\| \Gmii{}{s}(k)-\Gmii{}{s}(0) \right\|_{\infty} \leq 4 \cxo c ( 1 +2 \alphai{3}     \cxo   ) \mu \sqrt{\kc }\taum \rcc ,\ (1\leq s \leq h).
	\end{split}
	\end{equation*}

	\textbf{Step 2. Bound of $\left\|\Gmii{}{ls}(k) -  \Gmii{}{ls}(0)\right\|_{2} \ (0\leq l \leq h-1, 0\leq s\leq l-1 )$}. 
	
	\textbf{We first consider $l=h-1$, namely bound of $\left\|\Gmii{}{hs}(k) -  \Gmii{}{hs}(0)\right\|_{2} \ ( 0\leq s\leq h-2 )$. } For notation simplicity, we use $h$ to denote $h-1$. In this way, according to   Lemma~\ref{gradientcomputation}, we have 
	\begin{equation*}
	\begin{split}
	\frac{\partial \ui{} }{\partial \Wmii{(h)}{s}}= &  \alphaii{(h)}{s,3}\tau  \Phi(\Xmii{(s)}{})    \left(\sigma'\left( \Wmii{(h)}{s} \Phi(\Xmii{(s)}{})\right) \odot \Umi{h} \right)^{\top} \ (  1 \leq s \leq h-1).
	\end{split}
	\end{equation*}
	Let $\Hmi{i}= \Phi(\Xmii{(s)}{i})$, $\Hmi{i,:t}= [\Hmi{i}]_{:,t}$, $\Hmi{i,tr}= [\Hmi{i}]_{t,r}$, and $\Zmi{i,tr} = (\Wmii{(h)}{s,: r})^\top \Hmi{i,:t} $. In this way, for $1 \leq s \leq h-1$ we can write $\Gmii{ij}{hs}$ as 
	\begin{equation*}
	\begin{split}
	\Gmii{ij}{hs}= &  (\alphaii{(h)}{s,3}\tau)^2 \sum_{r=1}^{m} \left[ \sum_{t=1}^{p} \Umi{h, tr} \Hmi{i, :t} (\sigma'\left( (\Wmii{(h)}{s,: r})^\top \Hmi{i,:t}\right) \right]^{\top} \left[ \sum_{q=1}^{p} \Umi{h, qr} \Hmi{j, :q} (\sigma'\left(  (\Wmii{(h)}{s,: r})^\top \Hmi{j,:q}\right) \right]\\
	= &  (\alphaii{(h)}{s,3}\tau)^2 \sum_{t=1}^{p}  \sum_{q=1}^{p}  \Hmi{i, :t}^\top  \Hmi{j, :q}   \sum_{r=1}^{m}   \Umi{h, tr} \Umi{h, qr} \sigma'\left( \Zmi{i,tr} \right)  \sigma'\left( \Zmi{i,qr} \right)  .
	\end{split}
	\end{equation*}
	Then we can obtain
	\begin{equation*}
	\begin{split}
	& | \Gmii{ij}{hs}(k) -\Gmii{ij}{hs}(0) | \\
	=& ( \alphaii{(h)}{s,3}\tau)^2 \left|\sum_{t=1}^{p}  \sum_{q=1}^{p}  (\Hmi{i, :t}(k))^\top  \Hmi{j, :q}(k)   \sum_{r=1}^{m}   \Umi{h, tr}(k) \Umi{h, qr}(k) \sigma'\left( \Zmi{i,tr}(k) \right)  \sigma'\left( \Zmi{j,qr}(k) \right) \right.\\
	&\qquad  \qquad\quad\qquad \left.- \sum_{t=1}^{p}  \sum_{q=1}^{p}  (\Hmi{i, :t}(k))^\top  \Hmi{j, :q}(k)   \sum_{r=1}^{m}   \Umi{h, tr}(k) \Umi{h, qr}(k) \sigma'\left( \Zmi{i,tr}(k) \right)  \sigma'\left( \Zmi{j,qr}(k) \right) \right|.
	\end{split}
	\end{equation*}
	For brevity, we define $\Ami{1}, \Ami{2}$ and $\Ami{3}$ as follows:
	\begin{equation*}
	\begin{split}
	\Ami{1} =& \left|\sum_{t=1}^{p}  \sum_{q=1}^{p} \left((\Hmi{i, :t}(k))^\top  \Hmi{j, :q}(k)  \!-\! (\Hmi{i, :t}(0))^\top  \Hmi{j, :q}(0)   \right) \sum_{r=1}^{m}   \Umi{h, tr}(0) \Umi{h, qr}(0) \sigma'\left( \Zmi{i,tr}(k) \right)  \sigma'\left( \Zmi{j,qr}(k) \right)   \right|,\\
	\Ami{2} =& \left|\sum_{t=1}^{p}\!  \sum_{q=1}^{p}\!  (\Hmi{i, :t}(0))^\top\!\!  \Hmi{j, :q}(0)  \! \sum_{r=1}^{m}\!  \Umi{h, tr}(0) \Umi{h, qr}(0) \left( \sigma'\!\left( \Zmi{i,tr}(k) \right)  \sigma'\!\left( \Zmi{j,qr}(k) \right)\!  -\! \sigma'\!\left( \Zmi{i,tr}(0) \right)\sigma'\!\left( \Zmi{j,qr}(0) \right)    \right) \right|,\\
	\Ami{3} =& \left|\sum_{t=1}^{p}  \sum_{q=1}^{p}  (\Hmi{i, :t}(0))^\top  \Hmi{j, :q}(0)   \sum_{r=1}^{m}  \left( \Umi{h, tr}(k) \Umi{h, qr}(k) - \Umi{h, tr}(0) \Umi{h, qr}(0) \right) \sigma'\left( \Zmi{i,tr}(k) \right)  \sigma'\left( \Zmi{j,qr}(k) \right)    \right|.
	\end{split}
	\end{equation*}
	Then we have 
	\begin{equation*}
	\begin{split}
	| \Gmii{ij}{hs}(k) -\Gmii{ij}{hs}(0) |   =& ( \alphaii{(h)}{s,3}\tau)^2\left( \Ami{1}+ \Ami{2} +\Ami{3} \right).
	\end{split}
	\end{equation*}
	The remaining work is to upper bound $\Ami{1}$, $\Ami{2}$ and $\Ami{3}$. We first look at $\Ami{1}$:
	\begin{equation*}
	\begin{split}
	\Ami{1} =& \left|\sum_{t=1}^{p}  \!\sum_{q=1}^{p} \!\left(\Hmi{i, :t}(k)^\top\!  \Hmi{j, :q}(k) \! -\! (\Hmi{i, :t}(0))^\top  \Hmi{j, :q}(0)   \right) \!\!\sum_{r=1}^{m} \!  \Umi{h, tr}(0) \Umi{h, qr}(0) \sigma'\!\left( \Zmi{i,tr}(k) \right)  \sigma'\!\left( \Zmi{j,qr}(k) \right)   \right|\\
	\leq & m \mu^2 \cuo^2 \left|\sum_{t=1}^{p}  \sum_{q=1}^{p} \left((\Hmi{i, :t}(k))^\top  \Hmi{j, :q}(k)  - (\Hmi{i, :t}(0))^\top  \Hmi{j, :q}(0)   \right)     \right|\\
	\led{172} & m \mu^2 \cuo^2 \sum_{t=1}^{p}  \sum_{q=1}^{p} \left[\left|(\Hmi{i, :t}(k)- \Hmi{i, :t}(0))^\top  \Hmi{j, :q}(k)     \right| + \left|  (\Hmi{i, :t}(0))^\top  (\Hmi{j, :q}(k)   - \Hmi{j, :q}(0) )      \right|\right]\\
	\leq & m \mu^2 \cuo^2  \sqrt{\sum_{t=1}^{p}  \sum_{q=1}^{p} \| \Hmi{i, :t}(k)- (\Hmi{i, :t}(0)\|_2^2 } \sqrt{\sum_{t=1}^{p}  \sum_{q=1}^{p} \|  \Hmi{j, :q}(k)\|_2^2 } \\
	& + m \mu^2 \cuo^2  \sqrt{\sum_{t=1}^{p}  \sum_{q=1}^{p} \|\Hmi{j, :q}(k)   - \Hmi{j, :q}(0)\|_2^2 } \sqrt{\sum_{t=1}^{p}  \sum_{q=1}^{p} \| \Hmi{i, :t}(0)\|_2^2 } \\
	\leq & m p \mu^2 \cuo^2   \left( \| \Hmi{i}(k)- \Hmi{i}(0)\|_F \|  \Hmi{j}(k)\|_F + \| \Hmi{j}(k)- \Hmi{j}(0)\|_F \|  \Hmi{i}(k)\|_F\right)\\
	\leq & m p \mu^2 \cuo^2   \left( \| \Hmi{i}(k)- \Hmi{i}(0)\|_F \|  \Hmi{j}(k)\|_F + \| \Hmi{j}(k)- \Hmi{j}(0)\|_F \|  \Hmi{i}(k)\|_F\right)\\
	\end{split}
	\end{equation*}
	where \ding{172} holds since  the activation function $\sigmai{\cdot}$  is $\mu$-Lipschitz and $\rho$-smooth and the assumption $\|\Umi{s}\|_{\infty}\leq \cuo$. To bound $\| \Hmi{i}(k)- \Hmi{i}(0)\|_F \|  \Hmi{j}(k)\|_F$, we first recall our existing results. Lemma~\ref{boundX} that 
	\begin{equation*} 
	\begin{split}
	\| \Xmii{(l)}{i}(k)  - \Xmii{(l)}{i}(0) \|_F
	\leq      c ( 1 +2 \alphai{3}     \cxo   ) \mu \sqrt{\kc } \taum \rcc,
	\end{split}
	\end{equation*}
	where  $c= \left(1+\alphaii{}{2} + 2\alphaii{}{3} \mu \sqrt{\kc} \cwo  \right)^{l}$ with $\alphaii{}{2}=\max_{s,l}\alphaii{(l)}{s,2}$ and $\alphaii{}{3}=\max_{s,l} \alphaii{(l)}{s,3}$.   Here $\rcc$ is given in Lemma~\ref{boundofallweights}.   Based on this result, Lemma~\ref{boundedtempoutpsss222} shows  that \eqref{fsafdsaf} holds. 
	So we have 
	\begin{equation}\label{afdafdad}
	\begin{split}
	\| \Hmi{i}(k)- \Hmi{i}(0)\|_F  \leq &  \|  \Phi(\Xmii{(s)}{i}(k)) -  \Phi(\Xmii{(s)}{i}(0))\|_F \leq \sqrt{\kc} \|  \Xmii{(s)}{i}(k) -  \Xmii{(s)}{i}(0)\|_F \\
	\leq &c ( 1 +2 \alphai{3}     \cxo   ) \mu \kc \taum  \rcc, \\
	\|  \Hmi{j}(k)\|_F =&   \|  \Phi(\Xmii{(s)}{j}(k))\|_F\leq  \sqrt{\kc}\|  \Xmii{(s)}{j}(k)\|_F \leq 2 \sqrt{\kc} \cwo,
	\end{split}
	\end{equation}
	which indicates 
	\begin{equation*}
	\begin{split}
	\left( \| \Hmi{i}(k)- \Hmi{i}(0)\|_F \|  \Hmi{j}(k)\|_F + \| \Hmi{j}(k)- \Hmi{j}(0)\|_F \|  \Hmi{i}(k)\|_F\right) \leq 4 c \cwo ( 1 +2 \alphai{3}     \cxo   ) \mu \kc^{1.5} \taum \rcc.
	\end{split}
	\end{equation*}
	Therefore, we can upper bound 
	\begin{equation*}
	\begin{split}
	\Ami{1} \leq 4 c  m p \mu^3 \kc^{1.5} \cuo^2   \cwo  ( 1 +2 \alphai{3}     \cxo   )  \taum  \rcc.
	\end{split}
	\end{equation*}
	
	Then we consider to bound $\Ami{2}$. To begin with, we have 
	\begin{equation*}
	\begin{split}
	& \left| \sigma'\left( \Zmi{i,tr}(k) \right)  \sigma'\left( \Zmi{j,qr}(k) \right)  - \sigma'\left( \Zmi{i,tr}(0) \right)\sigma'\left( \Zmi{j,qr}(0)   \right) \right| \\
	\leq &  \left| (\sigma'\left( \Zmi{i,tr}(k) \right) - \sigma'\left( \Zmi{i,tr}(0) \right))   \sigma'\left( \Zmi{j,qr}(k) \right) \right| + \left|  \sigma'\left( \Zmi{i,tr}(0) \right)(\sigma'\left( \Zmi{j,qr}(k) \right) -\sigma'\left( \Zmi{j,qr}(0)   \right)) \right| \\
	\led{172} &  \mu \left| \sigma'\left( \Zmi{i,tr}(k) \right) - \sigma'\left( \Zmi{i,tr}(0) \right)  \right| + \mu\left| \sigma'\left( \Zmi{j,qr}(k) \right) -\sigma'\left( \Zmi{j,qr}(0)   \right) \right| \\
	\led{173} &  \mu \rho \left|  \Zmi{i,tr}(k) -   \Zmi{i,tr}(0)    \right| + \mu\rho \left|   \Zmi{j,qr}(k)   -  \Zmi{j,qr}(0)    \right|,
	\end{split}
	\end{equation*}
	where \ding{172}  holds since the activation function $\sigmai{\cdot}$  is $\mu$-Lipschitz;  \ding{173} holds since the activation function $\sigmai{\cdot}$  is  $\rho$-smooth.   Therefore, we can upper bound 
	\begin{equation*}
	\begin{split}
	\Ami{2}  
	\leq & \sum_{t=1}^{p}\!  \sum_{q=1}^{p} \! \left|\Hmi{i, :t}(0)^\top \!  \Hmi{j, :q}(0) \right| \!  \sum_{r=1}^{m} \!  \left| \Umi{h, tr}(0) \Umi{h, qr}(0) \right| \\
	&\qquad \qquad \qquad\cdot \left| \left( \sigma'\left( \Zmi{i,tr}(k) \right)  \sigma'\left( \Zmi{j,qr}(k) \right)  - \sigma'\left( \Zmi{i,tr}(0) \right)\sigma'\left( \Zmi{j,qr}(0) \right)    \right) \right|\\
	\leq & \mu\rho\! \sum_{t=1}^{p} \! \sum_{q=1}^{p} \!\left|(\Hmi{i, :t}(0))^\top \! \Hmi{j, :q}(0) \right| \! \sum_{r=1}^{m} \! \left| \Umi{h, tr}(0) \Umi{h, qr}(0) \right| \left[ \left|  \Zmi{i,tr}(k) \!- \!  \Zmi{i,tr}(0)    \right|\! +\! \left|   \Zmi{j,qr}(k)  \! - \! \Zmi{j,qr}(0)    \right| \right]\\
	\leq & \mu\rho \sqrt{ \sum_{t=1}^{p} \! \sum_{q=1}^{p} \!\left\||\Hmi{i, :t}(0)\right\|_2^2 \left\|  \Hmi{j, :q}(0) \right\|_2^2 } \left[ \sqrt{\sum_{t=1}^{p} \! \sum_{q=1}^{p}\! \left( \sum_{r=1}^{m} \! \left| \Umi{h, tr}(0) \Umi{h, qr}(0) \right|  \left|  \Zmi{i,tr}(k) \!- \!  \Zmi{i,tr}(0)    \right| \right)^2  } \right.  \\
	&\qquad \qquad  \qquad \qquad \qquad \qquad \left. + \sqrt{\sum_{t=1}^{p}  \sum_{q=1}^{p} \left( \sum_{r=1}^{m}  \left| \Umi{h, tr}(0) \Umi{h, qr}(0) \right|  \left|  \Zmi{j,qr}(k) -   \Zmi{j,qr}(0)    \right| \right)^2  } \right]\\
	\leq & \mu\rho \cuo \sqrt{m} \left\|\Hmi{i}(0)\right\|_F \left\|  \Hmi{j}(0) \right\|_F\cdot\\
	&\qquad \qquad \qquad \left[ \sqrt{\sum_{t=1}^{p}  \sum_{q=1}^{p} \sum_{r=1}^{m}   \left|  \Zmi{i,tr}(k) -   \Zmi{i,tr}(0)    \right|^2  } + \sqrt{\sum_{t=1}^{p}  \sum_{q=1}^{p} \sum_{r=1}^{m}   \left|  \Zmi{j,tr}(k) -   \Zmi{j,tr}(0)    \right|^2  }  \right]  \\ 
	\leq & \mu\rho \cuo \sqrt{m p} \left\|\Hmi{i}(0)\right\|_F \left\|  \Hmi{j}(0) \right\|_F \left[   \left\|  \Zmi{i}(k) -   \Zmi{i}(0)    \right\|_F  + \left\|  \Zmi{j}(k) -   \Zmi{j}(0)    \right\|_F \right].
	\end{split}
	\end{equation*}
	From Eqn.~\eqref{afdafdad}, we have $
	\|  \Hmi{j}(k)\|_F   \leq 2 \sqrt{\kc} \cwo$. Lemma~\ref{boundofallweights} shows  that Eqn.~\eqref{afdcsarfwaefca} holds. 
	Based on this result and the fact that $\rcc\leq \cwo$,   Lemma~\ref{boundofmidoutput} shows 
	\begin{equation*}
	\begin{split}
	&\left\| \Wmii{(l)}{s}(k)  \Phi(\Xmii{(s)}{}(k) ) -  \Wmii{(l)}{s}(0)  \Phi(\Xmii{(s)}{}(0) )   \right\|_F \leq \frac{c}{\alphai{3}}
	\sqrt{\kc m} \taum \rcc.
	\end{split}
	\end{equation*} 
	Therefore we can bound 
	\begin{equation*}
	\begin{split}
	\Ami{2}  
	\leq &  \frac{8 c m \kc^{1.5} \cwo^2 \mu\rho \cuo \sqrt{p}  \taum \rcc}{\alphai{3}}.
	\end{split}
	\end{equation*}
	Now we bound $\Ami{3}$ as follows:
	\begin{equation*}
	\begin{split}
	\Ami{3} =& \left|\! \sum_{t=1}^{p} \!  \sum_{q=1}^{p} \!  (\Hmi{i, :t}(0))^\top  \Hmi{j, :q}(0) \!   \sum_{r=1}^{m} \!  \left( \Umi{h, tr}(k) \Umi{h, qr}(k)\!  -\!  \Umi{h, tr}(0) \Umi{h, qr}(0) \right) \sigma'\! \left( \Zmi{i,tr}(k) \right)  \sigma'\! \left( \Zmi{j,qr}(k) \right)    \right|\\
	\leq & \mu^2 \left|\sum_{t=1}^{p}  \sum_{q=1}^{p}  (\Hmi{i, :t}(0))^\top  \Hmi{j, :q}(0)   \sum_{r=1}^{m}  \left( \Umi{h, tr}(k) \Umi{h, qr}(k) - \Umi{h, tr}(0) \Umi{h, qr}(0) \right)     \right|\\
	\leq & \mu^2 \! \sum_{t=1}^{p} \!  \sum_{q=1}^{p}\!  \left|  (\Hmi{i, :t}(0))^\top \! \Hmi{j, :q}(0) \right| \!  \sum_{r=1}^{m} \!  \left( |\Umi{h, tr}(k) \! -\!  \Umi{h, tr}(0) | | \Umi{h, qr}(k)| \! +\!  |\Umi{h, tr}(0) | | \Umi{h, qr}(k)\!  -\! \Umi{h, qr}(0)| \right)   \\
	\leq & \mu^2 \! \sum_{t=1}^{p} \!  \sum_{q=1}^{p} \! \left|  (\Hmi{i, :t}(0))^\top\!   \Hmi{j, :q}(0) \right|    \left( \|\Umi{h, t:}(k) \! -\!  \Umi{h, t:}(0) \|_2 \| \Umi{h, q:}(k)\|_2 \! \! + \|\Umi{h, t:}(0) \|_2 \| \Umi{h, qr}(k) \! -\! \Umi{h, qr}(0)\|_2 \right)   \\
	\leq & \mu^2 \sqrt{ \sum_{t=1}^{p}  \sum_{q=1}^{p} \left\||\Hmi{i, :t}(0)\right\|_2^2 \left\|  \Hmi{j, :q}(0) \right\|_2^2 } \left[  \sqrt{ \sum_{t=1}^{p}  \sum_{q=1}^{p}\|\Umi{h, t:}(k) - \Umi{h, t:}(0) \|_2 \| \Umi{h, q:}(k)\|_2 }  \right.\\
	&  \qquad \qquad \qquad \qquad \qquad \qquad \qquad \qquad \qquad \qquad \left. +   \sqrt{ \sum_{t=1}^{p}  \sum_{q=1}^{p}\|\Umi{h, t:}(k) - \Umi{h, t:}(0) \|_2 \| \Umi{h, q:}(k)\|_2 }   \right]  \\
	\leq & \mu^2   \left\||\Hmi{i}(0)\right\|_F \left\|  \Hmi{j}(0) \right\|_F  \left[  \|\Umi{h}(k) - \Umi{h}(0) \|_F \| \Umi{h}(k)\|_F    + \|\Umi{h}(k) - \Umi{h}(0) \|_F \| \Umi{h}(k)\|_F   \right]\\
	\led{172} &    8\kc \mu^2 \cwo^3    m \taum \rcc,
	\end{split}
	\end{equation*}
	where \ding{172} holds by using  Eqn.s~\eqref{afdcsarfwaefca}, \eqref{fsafdsaf},  \eqref{afdafdad}.
	
	By combining the above results, we have that for $s=0,\cdots, h-1$
	\begin{equation*}
	\begin{split}
	|\Gmii{}{hs}(k) -&  \Gmii{}{hs}(0) \|_{2} \leq  \sqrt{n} | \Gmii{ij}{hs}(k) -\Gmii{ij}{hs}(0) |_{\infty} \\
	\leq &  4(\alphaii{(h)}{s,3} )^2\kc \mu \cwo n^{0.5 } \taum\rcc  \left(  c   p \mu^2 \kc^{0.5} \cuo^2      ( 1 +2 \alphai{3}     \cxo   )     + \frac{2 c  \kc^{0.5} \cwo  \rho \cuo \sqrt{p}   }{\alphai{3}} +2  \mu \cwo^2    \right).
	\end{split}
	\end{equation*}

	\textbf{Then we consider $1\leq l<h$, namely bound of $H_{ls} \ ( 0\leq s\leq h-1 )$. } For brevity, let $\Bmi{i}(k) = \frac{\partial \ell }{\partial \Xmii{(l)}{i}(k)}$.  Here we use the same strategy as above.  Let 
	\begin{equation*}
	\begin{split}
	\Ami{1} \!=& \!\sum_{t=1}^{p} \! \sum_{q=1}^{p}\! \left((\Hmi{i, :t}(k))^\top \! \Hmi{j, :q}(k) \! -\! (\Hmi{i, :t}(0))^\top  \Hmi{j, :q}(0)   \right) \!\sum_{r=1}^{m} \!  \Bmi{i, tr}(0) \Bmi{j, qr}(0) \sigma'\!\left( \Zmi{i,tr}(k) \right)  \sigma'\!\left( \Zmi{j,qr}(k) \right)\!,\\
	\Ami{2}\! =& \! \sum_{t=1}^{p} \! \sum_{q=1}^{p} \! \Hmi{i, :t}(0)^\top \! \Hmi{j, :q}(0)  \! \sum_{r=1}^{m} \!  \Bmi{i, tr}(0) \Bmi{j, qr}(0)\left( \sigma'\!\left( \Zmi{i,tr}(k) \right)  \sigma'\!\left( \Zmi{j,qr}(k) \right)  \!- \!\sigma'\!\left( \Zmi{i,tr}(0) \right)\sigma'\!\left( \Zmi{j,qr}(0) \right)    \right)\!,\\
	\Ami{3,ij} \!=&  \sum_{t=1}^{p}  \sum_{q=1}^{p}  (\Hmi{i, :t}(0))^\top  \Hmi{j, :q}(0)   \sum_{r=1}^{m}  \left(  \Bmi{i, tr}(k) \Bmi{j, qr}(k) -  \Bmi{i, tr}(0) \Bmi{j, qr}(0) \right) \sigma'\left( \Zmi{i,tr}(k) \right)  \sigma'\left( \Zmi{j,qr}(k) \right) .
	\end{split}
	\end{equation*}
	By assuming $\|\Bmi{i}(k) \|_{\infty} \leq \cgo$, we can use the same method to bound $\Ami{1}$ and $\Ami{2}$ as follows:
	\begin{equation*}
	\begin{split}
	|\Ami{1} | \leq 4 c  m p \mu^3 \kc^{1.5} \cgo^2   \cwo  ( 1 +2 \alphai{3}     \cxo   )  \taum \rcc, \quad |\Ami{2} |  
	\leq  \frac{8 c m \kc^{1.5} \cwo^2 \mu\rho \cgo \sqrt{p} \taum \rcc}{\alphai{3}}.
	\end{split}
	\end{equation*}
	Then we need to carefully bound $\Ami{3}$:
	\begin{equation*}
	\begin{split}
	|\Ami{3,ij}| \!=&\! \left|\!\sum_{t=1}^{p} \!\sum_{q=1}^{p} \! (\Hmi{i, :t}(0))^\top\!\!  \Hmi{j, :q}(0) \!  \sum_{r=1}^{m} \! \left(  \Bmi{i, tr}(k) \Bmi{j, qr}(k) \!- \! \Bmi{i, tr}(0) \Bmi{j, qr}(0)  \right) \sigma'\!\left( \Zmi{i,tr}(k) \right)  \sigma'\!\left( \Zmi{j,qr}(k) \right)    \right|\\
	\leq & \mu^2 \left|\sum_{t=1}^{p}  \sum_{q=1}^{p}  (\Hmi{i, :t}(0))^\top  \Hmi{j, :q}(0)   \sum_{r=1}^{m}  \left( \Bmi{i, tr}(k) \Bmi{j, qr}(k) -  \Bmi{i, tr}(0) \Bmi{j, qr}(0)  \right)     \right|\\
	\leq & \mu^2 \!\sum_{t=1}^{p} \! \sum_{q=1}^{p}\! \left|  (\Hmi{i, :t}(0))^\top \!\! \Hmi{j, :q}(0) \right|  \!\sum_{r=1}^{m} \! \left( | \Bmi{i, tr}(k) \!- \! \Bmi{i, tr}(0) | |  \Bmi{j, qr}(k)| \!+ \!| \Bmi{i, tr}(0) |  \Bmi{j, qr}(k) \!-\! \Bmi{j, qr}(0)| \right)   \\
	\leq & \mu^2 \!\sum_{t=1}^{p}  \!\sum_{q=1}^{p}\! \left|  (\Hmi{i, :t}(0))^\top\!\!  \Hmi{j, :q}(0) \right|    \left( \|\Bmi{i, t:}(k) \!- \! \Bmi{i, t:}(0) \|_2 \| \Bmi{j, q:}(k)\|_2 \!+\! \| \Bmi{i, t:}(0)\|_2 \| \Bmi{j, q:}(k) \!- \!\Bmi{j, q:}(0)\|_2 \right)   \\
	\leq & \mu^2 \sqrt{ \sum_{t=1}^{p}  \sum_{q=1}^{p} \left\||\Hmi{i, :t}(0)\right\|_2^2 \left\|  \Hmi{j, :q}(0) \right\|_2^2 } \left[  \sqrt{ \sum_{t=1}^{p}  \sum_{q=1}^{p} \|\Bmi{i, t:}(k) -  \Bmi{i, t:}(0) \|_2^2 \| \Bmi{j, q:}(k)\|_2^2 }  \right.\\
	&\qquad  \qquad \qquad  \qquad \qquad \qquad \qquad \qquad \left. +   \sqrt{ \sum_{t=1}^{p}  \sum_{q=1}^{p} \| \Bmi{i, t:}(0)\|_2^2 \| \Bmi{j, q:}(k) - \Bmi{j, q:}(0)\|_2^2  }   \right]  \\
	\leq & \mu^2   \left\||\Hmi{i}(0)\right\|_F \left\|  \Hmi{j}(0) \right\|_F  \left[  \|\Bmi{i}(k) - \Bmi{i}(0) \|_F \| \Bmi{j}(k)\|_F    + \|\Bmi{j}(k) - \Bmi{j}(0) \|_F \| \Bmi{i}(0)\|_F   \right]\\
	\led{172} &   4 \mu^2 \cwo^2   \left[  \|\Bmi{i}(k) - \Bmi{i}(0) \|_F \| \Bmi{j}(k)\|_F    + \|\Bmi{j}(k) - \Bmi{j}(0) \|_F \| \Bmi{i}(0)\|_F   \right] ,
	\end{split}
	\end{equation*}
	where \ding{172} holds by using   Eqn.s~\eqref{afdcsarfwaefca}, \eqref{fsafdsaf},  \eqref{afdafdad}. Then when for $c_y=\frac{1}{\sqrt{n}} \|\umi{t} -\ymm \|_2$ and $c_u=\|\Umi{t}\|_F$,  Lemma~\ref{boundgradient} shows 
	\begin{equation*}
	\begin{split}
	\frac{1}{n}\sum_{i=1}^{n} \left\|\frac{\partial \ell }{\partial \Xmii{(l)}{i}(t)}\right\|_F 
	\leq & \left(1+\alphaii{}{2} + \alphaii{}{3}  \mu \sqrt{\kc}(\taum r +\cwo) \right)^{l}  \taum c_y c_u \\
	\led{172} &  2c  \taum \sqrt{m}\cwo   \left(1-\frac{\eta \lambda}{2}\right)^{t/2}\|\umi{0} -\ymm \|_2 ,
	\end{split}
	\end{equation*}	
	where $c=\left(1+\alphaii{}{2} + 2\alphaii{}{3}  \mu \sqrt{\kc} \cwo \right)^{l}$, $\alphaii{}{2}=\max_{s,l}\alphaii{(l)}{s,2}$ and $\alphaii{}{3}=\max_{s,l} \alphaii{(l)}{s,3}$. \ding{172} holds since $c_u=\|\Umi{t}\|_F\leq \|\Umi{t}-\Umi{0}\|_F + \|\Umi{0}\|_F\leq \sqrt{m} (\taum \rcc + \cwo)\leq 2\sqrt{m}\cwo$ and $\|\umi{t} -\ymm \|_2\leq \left(1-\frac{\eta \lambda}{2}\right)^{t/2}\|\umi{0} -\ymm \|_2$ in Theorem~\ref{mainconvergence}.  Lemma~\ref{boundfinaloutput}
	proves 
	\begin{equation*}
	\begin{split}
	\left\| \frac{\partial \ell }{\partial \Xmii{(l)}{i}(k)} - \frac{\partial \ell }{\partial \Xmii{(l)}{i}(0)}   \right\|_F
	\leq    c_1 c\alphai{3}   \cwo^2\cxo  \rho   \kc m \taum \rcc,
	\end{split}
	\end{equation*} 	
	where   $c_1$ is a constant. The remaining work is to bound 
	\begin{equation*}
	\begin{split}
	\|\Bmi{i}(k) - \Bmi{i}(0) \|_F \| \Bmi{j}(k)\|_F   \leq &   c_1 c\alphai{3}   \cwo^2\cxo  \rho   \kc m \taum \rcc \| \Bmi{j}(k)\|_F. 
	\end{split}
	\end{equation*}
	In this way, we have  
	\begin{equation*}
	\begin{split}
	\|\Ami{3} \|_1 \leq & \sum_{j=1}^{n}\sum_{i=1}^{n} \|\Ami{3,ij}|   \leq    4 \mu^2 \cwo^2 c_1 c\alphai{3}   \cwo^2\cxo  \rho   \kc m \taum \rcc  \sum_{j=1}^{n}\sum_{i=1}^{n} \left(\|\Bmi{j}(k)\|_F + \Bmi{i}(k)\|_F\right) \\
	\leq & 8 c_1 n \mu^2  c^2 \alphai{3}   \cwo^5\cxo  \rho   \kc m^{1.5}  \rcc     \left(1-\frac{\eta \lambda}{2}\right)^{t/2}\|\umi{0} -\ymm \|_2.
	\end{split}
	\end{equation*}
	
	Then combining all above results gives
	\begin{equation*}
	\begin{split}
	\left\|\Gmii{}{hs}(k) -  \Gmii{}{hs}(0)\right\|_{2} = &  ( \alphaii{(h)}{s,3}\tau)^2 \left\|\Am_1 + \Am_2 +\Am_3\right\|_2 \leq  ( \alphaii{(h)}{s,3}\tau)^2\left(  \left\|\Am_1 \right\|_2+ \|\Am_2\|_2 +\left\| \Am_3\right\|_2 \right) \\
	\leq &  ( \alphaii{(h)}{s,3}\tau)^2\sqrt{n} \left(\left\|\Am_1 \right\|_{\infty}+  \|\Am_2\|_{\infty} +    \left\| \Am_3\right\|_1 \right) \\
	\leq & 4(\alphaii{(h)}{s,3} )^2\kc \mu \cwo n^{0.5 } \taum\rcc  \left(  c   p \mu^2 \kc^{0.5} \cuo^2      ( 1 +2 \alphai{3}     \cxo   )     + \frac{2 c  \kc^{0.5} \cwo  \rho \cuo \sqrt{p}   }{\alphai{3}}     \right)\\
	&+8(\alphaii{(h)}{s,3} )^2 n c_1  \mu^2  c^2 \alphai{3}   \cwo^5\cxo  \rho   \kc m^{0.5}  \rcc     \left(1-\frac{\eta \lambda}{2}\right)^{t/2}\|\umi{0} -\ymm \|_2.
	\end{split}
	\end{equation*}
	
	In this way, we only need to upper bound $\left\|\Gmii{}{0}(k) - \Gmii{}{0}(0) \right\|_{2}$, $\left\|\Gmii{}{ls}(k) -  \Gmii{}{ls}(0)\right\|_{2}$ and $\left\| \Gmii{}{s}(k)-\Gmii{}{s}(0) \right\|_{2}$.

	\textbf{Step 3. Bound of $\left\|\Gmbii{}{0}(k) - \Gmbii{}{0}(0) \right\|_{2}$.} 
	
	Here we use the same method when we bound $\left\|\Gmii{}{ls}(k) - \Gmii{}{ls}(0) \right\|_{2}$ to bound $\left\|\Gmii{}{0}(k) - \Gmii{}{0}(0) \right\|_{2}$. Let $\Hmi{i}= \Phi(\Xmii{}{i})$, $\Hmi{i,:t}= [\Hmi{i}]_{:,t}$, $\Hmi{i,tr}= [\Hmi{i}]_{t,r}$,   $\Zmi{i,tr} = (\Wmii{(0)}{s,: r})^\top \Hmi{i,:t} $ and $\Bmi{i}(k) = \frac{\partial \ell }{\partial \Xmii{(l)}{i}(k)}$. In this way, for $1 \leq s \leq h-1$ we can write $\Gmii{ij}{hs}$ as  
	Then we define 
	\begin{equation*}
	\begin{split}
	\Ami{1} \!=& \!\sum_{t=1}^{p} \! \sum_{q=1}^{p} \!\left((\Hmi{i, :t}(k))^\top \!\! \Hmi{j, :q}(k)  \!-\! (\Hmi{i, :t}(0))^\top\!\!  \Hmi{j, :q}(0)   \right) \!\sum_{r=1}^{m} \!  \Bmi{i, tr}(0) \Bmi{j, qr}(0) \sigma'\!\left( \Zmi{i,tr}(k) \right)  \sigma'\!\left( \Zmi{j,qr}(k) \right)   ,\\
	\Ami{2} \!=& \! \sum_{t=1}^{p} \! \sum_{q=1}^{p} \! (\Hmi{i, :t}(0))^\top\!\!  \Hmi{j, :q}(0) \!  \sum_{r=1}^{m} \!  \Bmi{i, tr}(0) \Bmi{j, qr}(0)\left( \sigma'\!\left( \Zmi{i,tr}(k) \right)  \sigma'\!\left( \Zmi{j,qr}(k) \right) \! -\! \sigma'\left( \Zmi{i,tr}(0) \right)\sigma'\left( \Zmi{j,qr}(0) \right)    \right)  ,\\
	\Ami{3,ij} \!=&  \sum_{t=1}^{p}  \sum_{q=1}^{p}  (\Hmi{i, :t}(0))^\top  \Hmi{j, :q}(0)   \sum_{r=1}^{m}  \left(  \Bmi{i, tr}(k) \Bmi{j, qr}(k) -  \Bmi{i, tr}(0) \Bmi{j, qr}(0) \right) \sigma'\left( \Zmi{i,tr}(k) \right)  \sigma'\left( \Zmi{j,qr}(k) \right) .
	\end{split}
	\end{equation*}
	Then by using the same method, we can prove 
	\begin{equation*}
	\begin{split}
	\left\|\Gmbii{}{0}(k) - \Gmbii{}{0}(0)\right\|_{2} = &  \tau^2 \left\|\Am_1 + \Am_2 +\Am_3\right\|_2 \leq  ( \alphaii{(h)}{s,3}\tau)^2\left(  \left\|\Am_1 \right\|_2+ \|\Am_2\|_2 +\left\| \Am_3\right\|_2 \right) \\
	\leq &  \tau^2\sqrt{n} \left(\left\|\Am_1 \right\|_{\infty}+  \|\Am_2\|_{\infty} +    \left\| \Am_3\right\|_1 \right) \\
	\leq & 4\kc \mu \cwo n^{0.5 } \taum\rcc  \left(  c   p \mu^2 \kc^{0.5} \cuo^2      ( 1 +2 \alphai{3}     \cxo   )     + \frac{2 c  \kc^{0.5} \cwo  \rho \cuo \sqrt{p}   }{\alphai{3}}     \right)\\
	&+8c_1  n \mu^2  c^2 \alphai{3}   \cwo^5\cxo  \rho   \kc m^{0.5}  \rcc     \left(1-\frac{\eta \lambda}{2}\right)^{k/2}\|\umi{0} -\ymm \|_2.
	\end{split}
	\end{equation*}  
	
	\textbf{Step 4. Bound of $\left\|\Gmii{}{}(k) - \Gmii{}{}(0) \right\|_{2}$.} 
	
	By combining the above results and ignoring all constants for brevity, we can bound 
	\begin{equation*}
	\begin{split}
	\left\| \Gmii{}{}(k) -  \Gmii{}{}(0)  \right\|_2 \leq &\left\|\Gmbii{}{0}(k) - \Gmbii{}{0}(0) \right\|_{2} \!+\! \sum_{l=0}^{h-1} \sum_{s=0}^{l-1} \left\|\Gmii{}{ls}(k) -  \Gmii{}{ls}(0)\right\|_{2} \!+ \!\sum_{s=0}^{h-1} \left\| \Gmii{}{s}(k)-\Gmii{}{s}(0) \right\|_{2}\\ 
	\leq & c_2 c h  \mu \kc^{0.5} \cxo \taum \rcc n^{0.5} \left( \rho h \mu^2 \kc \cuo^2  \cwo    + \alphai{3}    c \rho h  \mu  \kc^{0.5}  \cwo^5     n^{0.5}\right)\\
	\end{split}
	\end{equation*}
	where $c=\left(1+\alphaii{}{2} + 2\alphaii{}{3}  \mu \sqrt{\kc} \cwo \right)^{h}$ and $c_2$ is a constant.  Considering \begin{equation*}
	\rcc= \frac{8\cxo\|\ymm-\um(0)\|_2}{\lambda  \sqrt{mn}} \max\left(1,  2 \left(1+\alphaii{}{2} + 2\alphaii{}{3} \mu \sqrt{\kc} \cwo \right)^{h} \alphai{3} \mu \sqrt{\kc} \cwo    \right) \leq \cwo,
	\end{equation*}
	to achieve
	\begin{equation*}
	\begin{split}
	\left\| \Gmii{}{}(k) -  \Gmii{}{}(0)  \right\|_2 \leq \frac{\lambda}{2},
	\end{split}
	\end{equation*}
	$m$ should be at the order of  
	\begin{equation*}
	\begin{split}
	m\geq \frac{c_3 \alphai{3}^2 \mu^2 \kc    \cxo^2 c^2 }{\lambda^2  n  }      ,
	\end{split}
	\end{equation*}
	where $c_3$ is a constant, $c=\left(1+\alphaii{}{2} + 2\alphaii{}{3}  \mu \sqrt{\kc} \cwo \right)^{h}$, $\alphaii{}{2}=\max_{s,l}\alphaii{(l)}{s,2}$ and $\alphaii{}{3}=\max_{s,l} \alphaii{(l)}{s,3}$.  The proof is completed. 
\end{proof}

\subsubsection{Proof of Lemma~\ref{mainconvergence2}}\label{proofofmainconvergence2}

\begin{proof}
	Lemma~\ref{mainconvergence} proves that when  $m=\Oc{\frac{\rho\kc^2 \cwo^2 \|\ymm-\um(0)\|_2^2}{\lambda^2   n} \left(1+ \alphaii{}{2}+2\alphaii{}{3}\mu\sqrt{\kc} \cwo   \right)^{2h} }$, then with probability at least $1-\delta/2$ we have 
	\begin{equation*}
	\begin{split}
	\|\ymm - \hmm(k)\|_2^2  \leq \left( 1 - \frac{\eta \lambda}{2}\right) \|\ymm - \hmm(k-1)\|_2^2 \leq \left( 1 - \frac{\eta \lambda}{2}\right)^{k}\|\ymm - \hmm(0)\|_2^2,
	\end{split}
	\end{equation*}
	where $\lambda$ is smallest eigenvalue of the Gram matrix $\Gm(t)\ (t=1,\cdots,k-1)$. 
	Lemma~\ref{boundofeigenvalue} shows that if 	$m$ satisfies 
	$m\geq \frac{c_3 \alphai{3}^2 \mu^2 \kc    \cxo^2 c^2 }{\lambda^2  n  }      $,
	where $c_3$ is a constant, $c=\left(1+\alphaii{}{2} + 2\alphaii{}{3}  \mu \sqrt{\kc} \cwo \right)^{h}$, $\alphaii{}{2}=\max_{s,l}\alphaii{(l)}{s,2}$ and $\alphaii{}{3}=\max_{s,l} \alphaii{(l)}{s,3}$, then we have 
	\begin{equation*}
	\begin{split}
	\left\| \Gmii{}{}(k) -  \Gmii{}{}(0)  \right\|_2 \leq\frac{  \lambda_{\min}\left( \Gm(0) \right)}{2},
	\end{split}
	\end{equation*}
	where $ \lambda_{\min}\left( \Gm(0) \right)$ is the smallest eigenvalue of $\Gmii{}{}(0)  $.  So we have 
	\begin{equation*}
	\begin{split}
	\lambda_{\min}( \Gmii{}{}(t))\geq \frac{  \lambda_{\min}\left( \Gm(0) \right)}{2}.
	\end{split}
	\end{equation*}
	
	So combining these results, we have 
	\begin{equation*}
	\begin{split}
	\|\ymm - \hmm(k)\|_2^2  \leq \left( 1 - \frac{\eta  \lambda_{\min}\left( \Gm(0) \right) }{4}\right) \|\ymm - \hmm(k-1)\|_2^2  \leq \left( 1 - \frac{\eta  \lambda_{\min}\left( \Gm(0) \right) }{4}\right)^{k}\|\ymm - \hmm(0)\|_2^2,
	\end{split}
	\end{equation*}
	when $m$ satisfies
	$m\geq  \frac{c_m'   \rho c^2  \kc^2  \cwo^2  \mu^2 }{\lambda^2 n} $
	and $	\eta\leq  \frac{c_\eta' \lambda}{\sqrt{m} \mu^4   h^{3}  \kc^2 c^4},$ where $c_m', c_\eta'$ are constants, $c=\left(1+\alphaii{}{2} + 2\alphaii{}{3}  \mu \sqrt{\kc} \cwo \right)^{h}$, $\alphaii{}{2}=\max_{s,l}\alphaii{(l)}{s,2}$ and $\alphaii{}{3}=\max_{s,l} \alphaii{(l)}{s,3}$. The proof is completed. 
\end{proof}

\subsection{Step 2  Lower Bound of Eigenvalue of Gram Matrix}\label{ProofofLowerbound}
Here we define some necessary notations for this subsection first. By Gaussian distribution $\mathcal{P}$ over a $q$-dimensional subspace $\W$, it means that for a basis $\{\emm_1, \emm_2, \cdots, \emm_q\}$ of $\W$ and $(v_1, v_2,\cdots, v_q)\sim\mathcal{N}(0,\Imm)$ such that $\sum_{i=1}^q v_i \emm_i\sim\mathcal{P}$. Then we equip one Gaussian distribution $\mathcal{P}^{(i)}$ with each linear subspace $\W$. Based on these, we define a transform $\W$ as 
\begin{equation*}
\Wii{ls}{tq}(\Km) =\begin{cases}
\EE_{\Wmii{(l)}{t}\sim\mathcal{P}} [\Wmii{(l)}{t}\Km(\Wmii{(l)}{t})^\top],\qquad\qquad \text{if} \ l=s \ \text{and}\ t=q\\
\EE_{\Wmii{(l)}{t}\sim\mathcal{P},\Wmii{(s)}{q}\sim\mathcal{P}} [\Wmii{(l)}{t}\Km(\Wmii{(s)}{q})^\top],\  \ \text{otherwise}
\end{cases},
\end{equation*}
where $\Km\in\Rs{p\times p}$ and $\Wmii{(l)}{t}$ denotes the parameters in convolution.

Then we define the population Gram matrix as follows. For brevity, let  $  \Xmb=
\Phi(\Xm) \in \Rs{\kc m\times p}.$  We first define the case where  $l=0$: 
\begin{equation*}
\begin{split}
&\bmi{i}^{(-1)} =\bm{0} \in \Rs{p},  \qquad\Kmii{(-1)}{ij} = \Xmi{i}^{\top} \Xmi{i} , \qquad\ \Qmii{(-1)}{ij} = \Xmbi{i}^{\top} \Xmbi{i} \in \Rs{p\times p},\\
& \Amii{(00)}{} = \begin{bmatrix}
\Wi{0}(\Qmii{(-1)}{ij}), \Wi{0}(\Qmii{(-1)}{ij})\\
\Wi{0}(\Qmii{(-1)}{ji}), \Wi{0}(\Qmii{(-1)}{jj})\\
\end{bmatrix}, \qquad  \qquad (\Mmii{(00)}{},\Nmii{(00)}{}) \sim \mathcal{N} \left( \bm{0}, \Amii{(00)}{}\right) \\
& \bmi{i}^{(0)} = \tau \EE_{\Mmii{(00)}{}} \sigmai{\Mmii{(00)}{}},\qquad \qquad\qquad\qquad \qquad \Kmii{(00)}{ij} =   \EE_{(\Mmii{(00)}{},\Nmii{(00)}{})} \left(\sigmai{\Mmii{(00)}{}} \sigmai{\Nmii{(00)}{}}^\top\right),\\
&\Qmii{(00)}{ij, ab} = \tr{\Kmii{(00)}{ij,\Sii{(l)}{a}, \Sii{(s)}{b}}},
\end{split}
\end{equation*}
where $\Wi{0}(\Km)=\EE_{\Wmii{(0)}{}\sim\mathcal{P}} [\Wmii{(0)}{}\Km(\Wmii{(0)}{})^\top],$  $\Qmii{(00)}{ij}  \in \Rs{p\times p}$, $\Kmii{(00)}{ij, ab} $ denotes the $(a,b)$-th entry in $\Kmii{(00)}{ij}$, and $\Sii{(0)}{a}=\{j\ | \ \Xmii{}{:,j} \in \text{the}\ a-\text{th patch for convolution}\}$. 

Then for $1\leq l\leq h, 1\leq s \leq l$, we can recurrently define 
\begin{equation*}
\begin{split}
& \Amii{(ls)}{tq} = \begin{bmatrix}
\Wii{ls}{tq}(\Qmii{(tq)}{ii}), \Wii{ls}{tq}(\Qmii{(tq)}{ij})\\
\Wii{ls}{tq}(\Qmii{(tq)}{ji}), \Wii{ls}{tq}(\Qmii{(tq)}{jj})\\
\end{bmatrix},\quad   (\Mmii{(ls)}{tq},\Nmii{(ls)}{tq}) \sim \mathcal{N} \left( \bm{0}, \Amii{(ls)}{tq} \right),\qquad (0\leq t,q \leq l-1), \\
& \bmi{i}^{(l)} = \sum_{t=1}^{l-1} \left(\alphaii{(l)}{t,2} \bmi{i}^{(t)}  + \tau \alphaii{(l)}{t,3}\EE_{\Mmii{(ll)}{tt}} \sigmai{\Mmii{(ll)}{tt}} \right);\\
&\Kmii{(ls)}{ij} =\sum_{t=1}^{l-1} \sum_{q=1}^{s-1} \left[ \alphaii{(l)}{t,2} \alphaii{(s)}{q,2}  \Kmii{(tq)}{ij} + \tau \EE_{(\Mmii{(ls)}{tq},\Nmii{(ls)}{tq})} \left(  \alphaii{(l)}{t,3} \alphaii{(s)}{q,2}  \sigmai{\Mmii{(ls)}{tq}} (\bmi{j}^{(q)})^\top   +   \alphaii{(l)}{t,2} \alphaii{(s)}{q,3}  \bmi{i}^{(t)} \sigmai{\Nmii{(ls)}{tq}}^\top \right.\right. \\
&\qquad\qquad \qquad \qquad \left. \left. +  \tau  \alphaii{(l)}{t,3} \alphaii{(s)}{q,3}   \sigmai{\Mmii{(ls)}{tq}} \sigmai{\Nmii{(ls)}{tq}}^\top\right) \right],\\
&\Qmii{(ls)}{ij, ab} = \tr{\Kmii{(ls)}{ij,\Sii{(l)}{a}, \Sii{(s)}{b}}},
\end{split}
\end{equation*}
where $\Kmii{(ls)}{ij}  \in \Rs{p\times p}$, $\Qmii{(ls)}{ij, ab} $ denotes the $(a,b)$-th entry in $\Qmii{(ls)}{ij}$, and $\Sii{(s)}{a}=\{j\ | \ \Xmii{(s-1)}{:,j} \in \text{the}\ a-\text{th patch for convolution}\}$. 
Finally, we define 
\begin{equation*}
\begin{split}
&\Am^{(s)}  = \begin{bmatrix}
\Wii{hh}{ss} (\Qmii{(ss)}{ii}), \Wii{hh}{ss}(\Qmii{(ss)}{ij})\\
\Wii{hh}{ss} (\Qmii{(ss)}{ji}), \Wii{hh}{ss}(\Qmii{(ss)}{jj})\\
\end{bmatrix},\\
&\Qmii{(s)}{ij, ab} = \Qmii{(ss)}{ij,ab} \EE_{( (\Mm,\Nm)\sim \bar{\Am}^{(s)})}  \sigma'\left(\Mm \right) \sigma'\left(\Nm \right)^\top, \qquad \Kmii{(s)}{ij, ab} = \tr{\Qmii{(s)}{ij}}, \ (s=0,h-1).
\end{split}
\end{equation*}

For brevity, we first define 
\begin{equation*}
\begin{split}
&\Kmtii{(ls)}{ij}= \frac{1}{m} \sum_{t=1}^m  \Xmi{i,t}^{(l)}  (\Xmi{j,t}^{(s)})^{\top}, \qquad \bmtii{(l)}{i}=\frac{1}{m} \sum_{t=1}^m  \Xmi{i,t}^{(l)}.
\end{split}
\end{equation*}

Then we prove that $\Kmti{s}$ is very close to the randomly generated gram matrix $\Kmtii{(ls)}{ij}$. 

\begin{lem}\label{boundofgrammartix}
	With probability at least $1-\delta$ over the convolution parameters $\Wm$  in each layer, then for $0\leq t\leq h, 0\leq s \leq h$, it holds 
	\begin{equation*}
	\left\|\frac{1}{m} \sum_{s=1}^{m} (\Xmi{i,s}^{(t)})^\top \Xmi{j,s}^{(q)} - \Kmii{(tq)}{ij} \right\|_{\infty} \leq C \sqrt{\frac{\log(n^2 p^2 h^2/\delta)}{m}},
	\end{equation*}
	and 
	\begin{equation*}
	\left\|\frac{1}{m} \sum_{s=1}^{m} \Xmi{i,s}^{(t)} - \bmi{i}^{(t)} \right\|_{\infty}  \leq C \sqrt{\frac{\log(n^2 p^2 h^2/\delta)}{m}},
	\end{equation*}
	where $C$ is a constant which depends on the activation function $\sigma(\cdot)$, namely $C\sim \sigma(0) +\sup_x \sigma'(x)$. 
\end{lem}
See its proof in Appendix~\ref{proofofboundofgrammartix}.

\begin{lem}\label{boundofdifference}
	Suppose Assumptions~\ref{activationassumption}, \ref{initilizationassumption} and~\ref{sampleassumption} hold.  Then if $m\geq \frac{c_4\mu^2 p^2 n^2\log(n/\delta)}{\lambda^2}$, we have 
	\begin{equation*}
	\begin{split}
	&\left\|\Gmii{}{hs}(0) -  (\alphaii{(h)}{s,3})^2\Kmii{(s)}{}  \right\|_{\op}\leq    \frac{\lambda}{4}\qquad (s=0,\cdots,h),\\
	\end{split}
	\end{equation*}
	where $c_4$ and $\lambda$ are constants. 
\end{lem}
See its proof in Appendix~\ref{proofofboundofdifference}.

\begin{lem}\label{sdaffdasfasfdafdreawsdfadad}
	Suppose Assumptions~\ref{activationassumption}, \ref{initilizationassumption} and~\ref{sampleassumption} hold. 	Suppose $\sigma$ is analytic and not a polynomial function. Consider data $\{\Xmi{i=1}^n\}_{i=1}^n$ are not parallel, namely $\vect{\Xmi{i}}\notin \text{span}(\vect{\Xmi{j}})$ for all $i\neq j$. Then if $m\geq \frac{c_4\mu^2 p^2 n^2\log(n/\delta)}{\lambda^2}$, it holds that with probability at least $1-\delta/2$, the smallest eigenvalue the matrix $\Gm$ satisfies 
	\begin{equation*}
	\begin{split}
	\lambda_{\min}\left(\Gmii{}{}(0)\right) 
	\geq \frac{3c_{\sigma}}{4} \sum_{s=0}^{h-1}(\alphaii{(h)}{s,3} )^2
	\left(\prod_{t=0}^{s-1}  (\alphaii{(s)}{t,2})^2\right)   
	\lambda_{\min}({\Km}).
	\end{split}
	\end{equation*}
	where $\lambda=3c_{\sigma} \sum_{s=0}^{h-1}(\alphaii{(h)}{s,3} )^2
	\left(\prod_{t=0}^{s-1}  (\alphaii{(s)}{t,2})^2\right)   
	\lambda_{\min}({\Km})$, $c_\sigma$ is a constant that only depends on $\sigma$ and the input data,  	 $\lambda_{\min}({\Km})=\min_{i,j} \lambda_{\min}(\Km_{ij}) $ is larger than zero  in which  $\lambda_{\min}(\Km_{ij}) $ is the the smallest eigenvalue of $\Km_{ij} =  \begin{bmatrix}
	\Xmi{i}^{\top} \Xmi{j}, \Xmi{i}^{\top} \Xmi{j}\\
	\Xmi{j}^{\top} \Xmi{i}, \Xmi{j}^{\top} \Xmi{j}\\
	\end{bmatrix}$ .

\end{lem}

See its proof in~\ref{proofofsdaffdasfasfdafdreawsdfadad}.

\subsubsection{Proof of Lemma~\ref{boundofgrammartix}}\label{proofofboundofgrammartix}

\begin{proof}
	We use mathematical induction to prove these results. For brevity, let  $  \Xmb=
	\Phi(\Xm) \in \Rs{\kc m\times p}$ and $\Xmi{i,s} =\Xmi{i,s:}^{\top} \in \Rs{p}$. For the first layer $(l=0)$, we have
	\begin{equation}
	\Xmi{i,s}^{(0)}=\tau \sigma\left(\sum_{t=1}^m \Wm^{(0)}_{ts} \Xmbi{i,t}^{}\right)
	\end{equation}
	Then let 
	\begin{equation}
	\Am^{(0)}_{i,s}=\sum_{t=1}^m \Wm^{(0)}_{ts} \Xmbi{i,t}^{}.
	\end{equation}
	Since the convolution parameter $\Wm$ satisfies Gaussian distribution, $ \Am^{(0)}_{i,s:}$ is a mean-zero Guassian variable with covariance matrix as follows
	\begin{equation*}
	\begin{split}
	\EE \left[(\Am^{(0)}_{i,s})^{\top} \! \Am^{(0)}_{j,q}\right] \!=\! \EE \sum_{t, t'} \!\Wm^{(0)}_{ts} \Xmbi{i,t}^{(0)} (\Xmbi{j,t'}^{})^T \!(\Wm^{(0)}_{t'q})^T \!=\! \delta_{st}  \W^{(0)}  \left( \sum_{t} \Xmbi{i,t}  \Xmbi{j,t}^\top \right) \!=\!\delta_{st} \W^{(0)} \! \left( \Qmii{(-1)}{ij} \right)\!,
	\end{split}
	\end{equation*}
	where $\delta_{st}$ is a random variable with $\delta_{st}= \pm 1$ with both probability 0.5. 
	Therefore, we have 
	\begin{equation*}
	\begin{split}
	\EE \left[  \frac{1}{m} \sum_{i=1}^m  \Xmi{i,t}^{(0)}  (\Xmi{j,t}^{(0)})^\top \right]=  \Kmii{(00)}{ij}, \quad  \EE \left[  \frac{1}{m} \sum_{i=1}^m  \Xmi{i,t}^{(0)}   \right] = \bmi{i}^{(0)}.
	\end{split}
	\end{equation*}
	In this way, following~\cite{du2018gradient} we can apply Hoeffding and Bernstein  bounds  and obtain the following results:
	\begin{equation*}
	\begin{split}
	\Pro\left( \max_{ij} \left\| \frac{1}{m} \sum_{t=1}^m  \Xmi{i,t}^{(0)}  (\Xmi{j,t}^{(0)})^T - \Kmii{(00)}{ij} \right\|_{\infty} \!\!\!\leq\! \sqrt{\frac{16 (1+2C_1^2/\sqrt{\pi})M^2 \log(4n^2p^2h^2/\delta))}{m}}\right) \!\geq\! 1- \frac{\delta}{h^2}\!,
	\end{split}
	\end{equation*}
	where we use $\|\Xmi{i,t}^{(0)}  (\Xmi{j,t}^{(0)})^\top\|_2 \leq\|\Xmi{i,t}^{(0)}  (\Xmi{j,t}^{(0)})^\top\|_F \leq 0.5( \|\Xmi{i,t}^{(0)} \|_F^2  +\|\Xmi{j,t}^{(0)})^\top\|_F^2) \led{172} \cxo^2$, 
	$M_1= 1+ 100\max_{i,j,s,t,l} |\W^{0}(\Qmii{(-1)}{ij})_{st}|$.  Here \ding{172} holds by using Lemma~\ref{boundofinitialization}. Similarly, we can prove 
	\begin{equation*}
	\begin{split}
	\Pro\left(  \left\| \frac{1}{m} \sum_{t=1}^m  \Xmi{i,t}^{(1)}   -  \bmi{i}^{(1)} \right\|_{\infty} \leq \sqrt{\frac{2C_1M \log(2n p h/\delta))}{m}}\right)\geq 1-\delta/h^2.
	\end{split}
	\end{equation*}
	Then we prove the results still hold when $l \geq 1, l\geq s \geq 0$. For brevity, we first define 
	\begin{equation*}
	\begin{split}
	&\Kmtii{(ls)}{ij}= \frac{1}{m} \sum_{t=1}^m  \Xmi{i,t}^{(l)}  (\Xmi{j,t}^{(s)})^\top, \qquad \bmtii{(l)}{i}=\frac{1}{m} \sum_{t=1}^m  \Xmi{i,t}^{(l)}.
	\end{split}
	\end{equation*}
	Suppose the results in our lemma holds for $0\leq l \leq k, 0\leq q \leq l$ with probability at least $1- \frac{k^2}{h^2}\delta$.  For $l=k+1$, we need to prove the results still hold with probability at least $1-\frac{2l-1}{h^2}\delta$. Toward this goal, we have 
	\begin{equation*}
	\Xmii{(l)}{i,s} =\sum_{0\leq q \leq l-1} \left[\Xmii{(q)}{i,s} + \tau \sigma\left(\sum_{t=1}^m \Wmii{(l)}{q,ts} \Xmbi{i,t}^{(q)}\right) \right],
	\end{equation*}
	where $\tau=\frac{1}{\sqrt{m}} $. Then let 
	\begin{equation*}
	\Am^{(lq)}_{i,s}= \sum_{t=1}^m \Wmii{(l)}{q,ts} \Xmbi{i,t}^{(q)}.
	\end{equation*}
	Similarly, we can obtain $ \Am^{(lq)}_{i,s}$ is a mean-zero Guassian variable with covariance matrix 
	\begin{equation*}
	\begin{split}
	\EE \left[ \Am^{(lq)}_{i,s} (\Am^{(lr)}_{i,s})^\top\right] =\delta_{st} \Wii{l}{qr}  \left(\sum_{t} \Xmbii{(q)}{i,t}  (\Xmbii{(q)}{j,t})^\top \right)=  \delta_{st} \Wii{l}{qr}  \left( \Qmtii{qr}{ij} \right).
	\end{split}
	\end{equation*}
	 
	Note that since for convolution networks, each element in the output involves several elements in the input (implemented by the operation $\Phi(\cdot)$), we need to consider this by combining the involved elements.     	Therefore, we can conclude
	\begin{equation*}
	\begin{split}
	\Qmtii{(ls)}{ij, ab}  = \tr{\Kmtii{(ls)}{ij,\Sii{(l)}{a}, \Sii{(s)}{b}}}\ (1\leq s\leq l)
	\end{split}
	\end{equation*}
	where $\Kmtii{(ls)}{ij, ab} $ denotes the $(a,b)$-th entry in $\Kmtii{(ls)}{ij}$, and $\Sii{(s)}{a}=\{j\ | \ \Xmii{(s-1)}{:,j} \in \text{the}\ a-\text{th patch}\}$. Moreover, we can easily obtain
	\begin{equation*}
	\begin{split}
	\EE\left[ \bmtii{(l)}{i} \right]=  \sum_{t=1}^{l-1} \left(\alphaii{(l)}{t,2} \bmtii{(t)}{i}+ \tau \alphaii{(l)}{t,3}\EE_{\Mmtii{(l)}{tt}} \sigmai{\Mmtii{(l)}{tt}} \right).
	\end{split}
	\end{equation*}
	In this way, we can further obtain
	\begin{equation*}
	\begin{split}
	& \Amtii{(l)}{tq} = \begin{bmatrix}
	\Wii{l}{tq}(\Qmtii{(tq)}{ii}), \Wii{l}{tq}(\Qmtii{(tq)}{ij})\\
	\Wii{l}{tq}(\Qmtii{(tq)}{ji}), \Wii{l}{tq}(\Qmtii{(tq)}{jj})\\
	\end{bmatrix},\quad   (\Mmtii{(l)}{tq},\Nmtii{(l)}{tq}) \sim \mathcal{N} \left( \bm{0}, \Amtii{(l)}{tq} \right),\qquad (0\leq t,q \leq l-1), \\
	&\EE\left[\Kmtii{(ls)}{ij} \right]=\sum_{t=1}^{l-1} \sum_{q=1}^{s-1} \left[ \alphaii{(l)}{t,2} \alphaii{(s)}{q,2}  \Kmtii{(tq)}{ij} + \tau \EE_{(\Mmtii{(l)}{tq},\Nmtii{(l)}{tq})} \left(  \alphaii{(l)}{t,3} \alphaii{(s)}{q,2}  \sigmai{\Mmtii{(l)}{tq}} (\bmtii{(q)}{j})^\top   +   \alphaii{(l)}{t,2} \alphaii{(s)}{q,3} \bmtii{(t)}{i}   \sigmai{\Nmtii{(l)}{tq}}^\top \right.\right. \\
	&\qquad\qquad \qquad \qquad \left. \left. +  \tau  \alphaii{(l)}{t,3} \alphaii{(s)}{q,3}   \sigmai{\Mmtii{(l)}{tq}} \sigmai{\Nmtii{(l)}{tq}}^\top\right) \right] \in \Rs{p\times p}. 
	\end{split}
	\end{equation*}
	Then we also apply the concentration inequality and obtain that for $1\leq s \leq l$ 
	\begin{equation*}
	\begin{split}
	\Pro\left( \max_{ij} \left\| \frac{1}{m} \sum_{t=1}^m  \Xmi{i,t}^{(l)}  (\Xmi{j,t}^{(s)})^T - \EE \Kmtii{(ls)}{ij} \right\|_{\infty} \leq \sqrt{\frac{16 (1+2C_1^2/\sqrt{\pi})M^2 \log(4n^2p^2h^2/\delta))}{m}}\right) \geq 1- \delta/h^2
	\end{split}
	\end{equation*}
	where  we use $\|\Xmi{i,t}^{(0)}  (\Xmi{j,t}^{(0)})^\top\|_2 \leq\|\Xmi{i,t}^{(0)}  (\Xmi{j,t}^{(0)})^\top\|_F \leq 0.5( \|\Xmi{i,t}^{(0)} \|_F^2  +\|\Xmi{j,t}^{(0)})^\top\|_F^2) \leq \cxo^2$, $M_1= 1+ 100\max_{i,j,s,t,l} |\W^{l}(\Kmbii{(l-1)}{ij})_{st}|$. Similarly, we can prove 
	\begin{equation*}
	\begin{split}
	\Pro\left(  \left\| \frac{1}{m} \sum_{t=1}^m  \Xmi{i,t}^{(l)}   - \EE \bmtii{(l)}{i} \right\|_{\infty} \leq \sqrt{\frac{2C_1M \log(2n p h/\delta))}{m}}\right)\geq 1- \delta/h^2.
	\end{split}
	\end{equation*}
	According to the definition
	\begin{equation*}
	\begin{split}
	&\Kmtii{(ls)}{ij}= \frac{1}{m} \sum_{t=1}^m  \Xmi{i,t}^{(l)}  (\Xmi{j,t}^{(s)})^\top, \qquad \bmtii{(l)}{i}=\frac{1}{m} \sum_{t=1}^m  \Xmi{i,t}^{(l)}. 
	\end{split}
	\end{equation*}
	we have 
	\begin{equation*}
	\begin{split}
	&\left\|  \frac{1}{m} \sum_{t=1}^m  \Xmi{i,t}^{(l)}  (\Xmi{j,t}^{(s)})^\top  - \Kmii{(ls)}{ij} \right\|_{\infty} \leq \left\|  \frac{1}{m} \sum_{t=1}^m  \Xmi{i,t}^{(l)}  (\Xmi{j,t}^{(s)})^\top  -\EE \Kmtii{(ls)}{ij}  \right\|_{\infty}  + \left\|  \EE \Kmtii{(ls)}{ij}   - \Kmii{(ls)}{ij}\right\|_{\infty},\\
	& \left\| \frac{1}{m} \sum_{t=1}^m  \Xmi{i,t}^{(l)}   -   \bmii{(l)}{i} \right\|_{\infty}  \leq  \left\| \frac{1}{m} \sum_{t=1}^m  \Xmi{i,t}^{(l)}   - \EE \bmtii{(l)}{i} \right\|_{\infty}  +  \left\|  \EE \bmtii{(l)}{i} - \bmii{(l)}{i} \right\|_{\infty} .
	\end{split}
	\end{equation*}
	
	Then we only need to bound 
	\begin{equation*}
	\begin{split}
	\left\|  \EE \Kmtii{(ls)}{ij}   - \Kmii{(ls)}{ij}\right\|_{\infty}\quad \text{and} \quad \left\|  \EE \bmtii{(l)}{i} - \bmii{(l)}{i} \right\|_{\infty} . 
	\end{split}
	\end{equation*}
	In the following content, we bound these two terms in turn. To begin with, we have  
	
	\begin{equation*}
	\begin{split}
	&\left\| \EE \Kmtii{(ls)}{ij}   - \Kmii{(ls)}{ij} \right\|_{\infty}
	=\left\| \tr{\Qmtii{(ls)}{ij,\Sii{(s)}{a}, \Sii{(ls)}{b}} }-  \tr{\Qmii{(ls)}{ij,\Sii{(s)}{a}, \Sii{(ls)}{b}} } \right\|_{\infty}
	\leq \left\| \Qmtii{(l)}{ij} - \Qmii{(l)}{ij} \right\|_{\infty}\\
	\leq & \sum_{t=1}^{l-1} \sum_{q=1}^{s-1} \left[ \alphaii{(l)}{t,2} \alphaii{(s)}{q,2} \left\| \Kmtii{(tq)}{ij} - \Kmii{(tq)}{ij} \right\|_{\infty} \right.\\
	&\qquad + \tau   \alphaii{(l)}{t,3} \alphaii{(s)}{q,2}  \left\|  \EE_{( (\Mmtii{(tq)}{},\Nmtii{(tq)}{}))}  \sigmai{\Mmtii{(tq)}{}} (\bmtii{(q)}{j})^\top - \EE_{( (\Mmii{(tq)}{},\Nmii{(tq)}{}))}  \sigmai{\Mmii{(tq)}{}} (\bmii{(q)}{j})^\top   \right\|_{\infty}  \\
	&\qquad +  \tau  \alphaii{(l)}{t,2} \alphaii{(s)}{q,3}   \left\|   \EE_{( (\Mmtii{(tq)}{},\Nmtii{(tq)}{}))}  \bmtii{(t)}{i} \sigmai{\Nmtii{(tq)}{}}^\top -  \EE_{( (\Mmii{(tq)}{},\Nmii{(tq)}{}))}  \bmii{(t)}{i} \sigmai{\Nmii{(tq)}{}}^\top \right\|_{\infty}  \\
	&\qquad   \left. +  \tau  \alphaii{(l)}{t,3} \alphaii{(s)}{q,3}   \left\|   \EE_{( (\Mmtii{(tq)}{},\Nmtii{(tq)}{}))} \sigmai{\Mmtii{(tq)}{}} \sigmai{\Nmtii{(tq)}{}}^\top  -  \EE_{( (\Mmii{(tq)}{},\Nmii{(tq)}{}))} \sigmai{\Mmii{(tq)}{}} \sigmai{\Nmii{(tq)}{}}^\top \right\|_{\infty}   \right]\\ 
	\end{split}
	\end{equation*}
	Then we bound   
	\begin{equation*}
	\begin{split}
	& \left\|  \EE_{( (\Mmtii{(tq)}{},\Nmtii{(tq)}{}))}  \sigmai{\Mmtii{(tq)}{}} (\bmtii{(q)}{j})^\top - \EE_{( (\Mmii{(tq)}{},\Nmii{(tq)}{}))}  \sigmai{\Mmii{(tq)}{}} (\bmii{(q)}{j})^\top   \right\|_{\infty} \\
	=  & \left\|  \EE_{( (\Mm,\Nm)\sim \Amtii{(tq)}{})}  \sigmai{\Mmii{}{}} (\bmtii{(q)}{j})^\top -  \EE_{( (\Mm,\Nm)\sim \Amii{(tq)}{})}  \sigmai{\Mmii{}{}} (\bmii{(q)}{j})^\top   \right\|_{\infty} \\
	\leq  & \left\|  \EE_{( (\Mm,\Nm)\sim \Amtii{(tq)}{})}  \sigmai{\Mmii{}{}} (\bmtii{(q)}{j} \!- \!\bmii{(q)}{j})^\top   \right\|_{\infty}  \!\!\!+ \!  \left\| \left[ \EE_{( (\Mm,\Nm)\sim \Amtii{(tq)}{})}  \sigmai{\Mmii{}{}} \!-\!  \EE_{( (\Mm,\Nm)\sim \Amii{(tq)}{})}  \sigmai{\Mmii{}{}}\right] (\bmii{(q)}{j})^\top   \right\|_{\infty} 
	\end{split}
	\end{equation*}

	Next, we bound the above inequality by bound each term:
	\begin{equation*}
	\begin{split}
	& \left\| \left[ \EE_{( (\Mm,\Nm)\sim \Amtii{(tq)}{})}  \sigmai{\Mmii{}{}} -  \EE_{( (\Mm,\Nm)\sim \Amii{(tq)}{})}  \sigmai{\Mmii{}{}}\right] (\bmii{(q)}{j})^\top   \right\|_{\infty} \\
	\leq& \max_{i}\|\bmii{(q)}{j}\|_{\infty} (\sigmai{0} + \sup_{x} \sigma'({x})) \|  \Amtii{(tq)}{}  -   \Amii{(tq)}{}\|_{\infty}\\
	\leq & c_1c_2 c_3\| \Qmtii{(tq)}{ij}  -  \Qmii{(tq)}{ij}  \|_{\infty}\\
	=&c_1c_2 c_3 \max_{a,b}\left\| \tr{\Kmtii{(ls)}{ij,\Sii{(l)}{a}, \Sii{(s)}{b}} }-  \tr{\Kmii{(ls)}{ij,\Sii{(l)}{a}, \Sii{(s)}{b}} } \right\|_{\infty}\\
	\leq &c_1c_2 c_3q \left\| \Kmtii{(l)}{ij} - \Kmii{(l)}{ij} \right\|_{\infty},\\
	\end{split}
	\end{equation*}
	where $c_1 = \max_{l}1+\|\Wii{l}{tq}\|_{L^{\infty}\rightarrow L^{\infty}}$, $c_2= \sigmai{0} + \sup_{x} \sigma'({x}) $, $c_3=\max_{i,q} \|\bmii{(q)}{i}\|_{\infty} $. Similarly, we can bound 
	\begin{equation*}
	\begin{split}
	\left\|  \EE_{( (\Mm,\Nm)\sim \Amtii{(tq)}{})}  \sigmai{\Mmii{}{}} (\bmtii{(q)}{j} - \bmii{(q)}{j})^\top   \right\|_{\infty} 
	\leq &  c_2 \sqrt{c_1 c_4} \| \bmi{j}^{(q)} - \bmbi{j}^{(q)}  \|_{\infty}
	\end{split}
	\end{equation*}
	where $c_4=\max_{ij} \|\Qmtii{(tq)}{ij}) \|_{\infty}\leq q\max_{ij} \|\Kmtii{(tq)}{ij}) \|_{\infty}\leq q\cxo^2$ and $1\leq q \leq l-1$.  Therefore we have 
	\begin{equation*}
	\begin{split}
	& \left\|  \EE_{( (\Mmtii{(tq)}{},\Nmtii{(tq)}{}))}  \sigmai{\Mmtii{(tq)}{}} (\bmtii{(q)}{j})^\top - \EE_{( (\Mmii{(tq)}{},\Nmii{(tq)}{}))}  \sigmai{\Mmii{(tq)}{}} (\bmii{(q)}{j})^\top   \right\|_{\infty} \\
	=  & (c_1c_2 c_3 q + c_2 \sqrt{c_1 c_4}  )\max \left(\| \Kmtii{(tq)}{ij}  -  \Kmii{(tq)}{ij}  \|_{\infty}, \| \bmi{j}^{(q)} - \bmbi{j}^{(q)}  \|_{\infty} \right). 
	\end{split}
	\end{equation*}
	By using the same method, we can upper bound 
	\begin{equation*}
	\begin{split}
	&\left\|   \EE_{( (\Mmtii{(tq)}{},\Nmtii{(tq)}{}))}  \bmtii{(t)}{i} \sigmai{\Nmtii{(tq)}{}}^\top -  \EE_{( (\Mmii{(tq)}{},\Nmii{(tq)}{}))}  \bmii{(t)}{i} \sigmai{\Nmii{(tq)}{}}^\top \right\|_{\infty} \\
	=  & (c_1c_2 c_3 q+ c_2 \sqrt{c_1 c_4}  )\max \left(\| \Kmtii{(tq)}{ij}  -  \Kmii{(tq)}{ij}  \|_{\infty}, \| \bmi{j}^{(q)} - \bmbi{j}^{(q)}  \|_{\infty} \right). 
	\end{split}
	\end{equation*}
	Next, we can upper bound 
	\begin{equation*}
	\begin{split}
	&\left\|   \EE_{( (\Mmtii{(tq)}{},\Nmtii{(tq)}{}))} \sigmai{\Mmtii{(tq)}{}} \sigmai{\Nmtii{(tq)}{}}^\top  -  \EE_{( (\Mmii{(tq)}{},\Nmii{(tq)}{}))} \sigmai{\Mmii{(tq)}{}} \sigmai{\Nmii{(tq)}{}}^\top \right\|_{\infty} \\
	=  & \left\|  \EE_{( (\Mm,\Nm)\sim \Amtii{(tq)}{})} \sigmai{\Mmii{(tq)}{}} \sigmai{\Nmii{(tq)}{}}^\top  -  \EE_{( (\Mm,\Nm)\sim \Amii{(tq)}{})}  \sigmai{\Mmii{(tq)}{}} \sigmai{\Nmii{(tq)}{}}^\top  \right\|_{\infty} \\
	\leq & c_{\sigma}\| \Amtii{(tq)}{}  -  \Amii{(tq)}{} \|_{\infty} 
	\leq   c_{\sigma} c_1 \| \Qmtii{(tq)}{ij})  -  \Qmbii{(tq)}{ij})  \|_{\infty} 	\leq   c_{\sigma} c_1 q\| \Kmtii{(tq)}{ij})  -  \Kmbii{(tq)}{ij})  \|_{\infty},
	\end{split}
	\end{equation*}
	where $c_{\sigma}$ is a constant that only depends on $\sigma$. 
	Combing all results yields
	\begin{equation*}
	\begin{split}
	&\left\| \EE \Kmtii{(ls)}{ij}   - \Kmii{(ls)}{ij} \right\|_{\infty} \\
	\leq & \sum_{t=1}^{l-1} \sum_{q=1}^{s-1} \left[ (\alphaii{(l)}{t,2} \alphaii{(s)}{q,2} +\tau^2  \alphaii{(l)}{t,3} \alphaii{(s)}{q,3} c_{\sigma} c_1 q ) \| \Kmtii{(tq)}{ij})  -  \Kmii{(tq)}{ij})  \|_{\infty}  \right.\\
	&\qquad \left. +  \tau  (\alphaii{(l)}{t,2} \alphaii{(s)}{q,2} +\alphaii{(l)}{t,3} \alphaii{(s)}{q,2} )  (c_1c_2 c_3q  + c_2 \sqrt{c_1 c_4}  )\max \left(\| \Kmtii{(tq)}{ij}  -  \Kmii{(tq)}{ij}  \|_{\infty}, \| \bmi{j}^{(q)} - \bmbi{j}^{(q)}  \|_{\infty} \right)    \right]\\
	\leq & c\max_{1\leq t\leq l-1, 1\leq q \leq l-1} \left(\| \Kmtii{(tq)}{ij}  -  \Kmii{(tq)}{ij}  \|_{\infty}, \| \bmi{j}^{(q)} - \bmbi{j}^{(q)}  \|_{\infty} \right)   \\
	\end{split}
	\end{equation*}
	where $c_{l}=\sum_{t=1}^{l-1} \sum_{q=1}^{s-1}  \left[ \alphaii{(l)}{t,2} \alphaii{(s)}{q,2} +\tau^2  \alphaii{(l)}{t,3} \alphaii{(s)}{q,3} c_{\sigma} c_1 q + \tau  (\alphaii{(l)}{t,2} \alphaii{(s)}{q,2} +\alphaii{(l)}{t,3} \alphaii{(s)}{q,2} )  (c_1c_2 c_3 q + c_2 \sqrt{c_1 c_4}  )\right]$. Since we have assumed that with probability $1-(l-1)^2\delta/h^2$ for $0\leq t\leq l-1, 0\leq s \leq l-1$, it holds  
	\begin{equation*}
	\max\left(\left\|\frac{1}{m} \sum_{s=1}^{m} (\Xmi{i,s}^{(t)})^\top \Xmi{j,s}^{(q)} - \Kmii{(tq)}{ij} \right\|_{\infty}, \left\|\frac{1}{m} \sum_{s=1}^{m} \Xmi{i,s}^{(t)} - \bmi{i}^{(t)} \right\|_{\infty} \right)\leq C_{l-1}\sqrt{\frac{\log(n^2 p^2 h^2/\delta)}{m}},
	\end{equation*}
	where $C$ is a constant. Then with probability $1-(l-1)^2\delta/h^2$, we have for all $0\leq s \leq l$
	\begin{equation*}
	\begin{split}
	\left\| \EE \Kmtii{(ls)}{ij}   - \Kmii{(ls)}{ij} \right\|_{\infty} \leq  c_{l} C_{l-1}\sqrt{\frac{\log(n^2 p^2 h^2/\delta)}{m}}.
	\end{split}
	\end{equation*}
	Thus, with probability $(1-(l-1)^2\delta/h^2)(1-\delta/h^2)\geq 1-l^2\delta/h^2\geq 1-\delta $, we have for all  for $0\leq t\leq h, 0\leq s \leq h$
	\begin{equation*}
	\left\|\frac{1}{m} \sum_{s=1}^{m} (\Xmi{i,s}^{(t)})^\top \Xmi{j,s}^{(q)} - \Kmii{(tq)}{ij} \right\|_{\infty} \leq C\sqrt{\frac{\log(n^2 p^2 h^2/\delta)}{m}},
	\end{equation*}
	where $C=C_0 \prod_{l=1}^{h} c_{l}$ is  a constant.

	Now we consider to bound   	
	\begin{equation*}
	\begin{split}
	&\left\|  \EE \bmtii{(l)}{i} - \bmii{(l)}{i} \right\|_{\infty}\\
	=& \left\|  \sum_{t=1}^{l-1} \left(\alphaii{(l)}{t,2} (\bmtii{(t)}{i} -\bmii{(t)}{i}) + \tau \alphaii{(l)}{t,3} \left(\EE_{\Mm\sim\Amtii{lt}{}} \sigmai{\Mm} - \EE_{\Mm\sim \Amii{lt}{}} \sigmai{\Mm}\right) \right) \right\|_{\infty}\\
	\leq & \sum_{t=1}^{l-1} \left(  \alphaii{(l)}{t,2} \left\|  \bmtii{(t)}{i} -\bmii{(t)}{i}\right\|_{\infty} + \tau \alphaii{(l)}{t,3} \left\|\left(\EE_{\Mm\sim\Amtii{(l-1)t}{}} \sigmai{\Mm} - \EE_{\Mm\sim \Amii{(l-1)t}{}} \sigmai{\Mm}\right)  \right\|_{\infty}   \right)\\
	\leq & \sum_{t=1}^{l-1} \left(  \alphaii{(l)}{t,2} \left\|  \bmtii{(t)}{i} -\bmii{(t)}{i}\right\|_{\infty} + \tau \alphaii{(l)}{t,3} c_{\sigma} \left\| \Amtii{(l-1)t}{}  - \Amii{(l-1)t}{}    \right\|_{\infty}   \right)\\
	\leq & \sum_{t=1}^{l-1} \left(  \alphaii{(l)}{t,2} \left\|  \bmtii{(t)}{i} -\bmii{(t)}{i}\right\|_{\infty} + \tau \alphaii{(l)}{t,3} c_{\sigma} \left\| \Qmtii{(l-1)t}{}  - \Qmii{(l-1)t}{}    \right\|_{\infty}   \right)\\
	\leq & \sum_{t=1}^{l-1} \left(  \alphaii{(l)}{t,2} + \tau \alphaii{(l)}{t,3} c_{\sigma} c_1 q \right) \max \left(\left\|  \bmtii{(t)}{i} -\bmii{(t)}{i}\right\|_{\infty} , \left\| \Kmtii{(l-1)t}{}  - \Kmii{(l-1)t}{}    \right\|_{\infty}   \right)\\
	\end{split}
	\end{equation*}
	where $c_l'=\sum_{t=1}^{l-1} \left(  \alphaii{(l)}{t,2} + \tau \alphaii{(l)}{t,3} c_{\sigma} c_1 q\right)$. 
	Then with probability $(1-(l-1)^2\delta/h)(1-\delta/h)\geq  1-\delta $, we have for all  for $0\leq t\leq h$
	\begin{equation*}
	\left\|\frac{1}{m} \sum_{s=1}^{m} \Xmi{i,s}^{(t)} - \bmi{i}^{(t)} \right\|_{\infty}   \leq C\sqrt{\frac{\log(n^2 p^2 h^2/\delta)}{m}},
	\end{equation*}
	where $C=C_0 \prod_{l=1}^{h} \max(c_{l}, c_l')$  is  a constant. The proof is completed. 
\end{proof}

\subsubsection{Proof of Lemma~\ref{boundofdifference}}\label{proofofboundofdifference}
\begin{proof}
	For brevity, here we just use $\Xmii{(s)}{i}$, $ \Wmii{(h)}{s}$, $\Umi{h}$, $\Xmbii{(s)}{i}) $ to respectively denote $Xmii{(s)}{i}(0)$
	$ \Wmii{(h)}{s}(0)$, $\Umi{h}(0)$, $\Phi(\Xmii{(s)}{i}) $, since here we only involve the initialization and does not update the variables.  Let $\Xmbii{(s)}{i,t}) =( \Xmbii{(s)}{i,:t})^{\top} $ and $\Zmi{i,tr} = (\Wmii{(h)}{s,: r})^\top \Xmbii{(s)}{i,t}) $. Firstly according to the definition, we have 
	\begin{equation*}
	\begin{split}
	& \Gmii{ij}{hs}(0)  =  \left\langle \frac{\partial \ell_i}{\partial \Wmii{(h)}{s}(0)},  \frac{\partial \ell_j}{\partial \Wmii{(h)}{s}(0)}\right\rangle\\
	= & (\alphaii{(h)}{s,3}\tau)^2\left\langle \Phi(\Xmii{(s)}{i})    \left(\sigma'\left( \Wmii{(l)}{s} \Phi(\Xmii{(s)}{i})\right) \odot \Umi{h}\right)^{\top},  \Phi(\Xmii{(s)}{j})    \left(\sigma'\left( \Wmii{(l)}{s} \Phi(\Xmii{(s)}{j})\right) \odot \Umi{h}\right)^{\top} \right\rangle\\
	= & (\alphaii{(h)}{s,3}\tau)^2 \sum_{t=1}^{p}  \sum_{q=1}^{p}  \Xmbii{(s)}{i,t})  (\Xmbii{(s)}{j,q})^{\top}   \sum_{r=1}^{m}   \Umi{h, tr} \Umi{h, qr} \sigma'\left( \Zmi{i,tr} \right)  \sigma'\left( \Zmi{j,qr} \right).
	\end{split}
	\end{equation*}
	Then by taking expectation on $\Wm\sim \mathcal{N}(0,\Imm)$ and $\Um\sim \mathcal{N}(0,\Imm)$, we have
	\begin{equation}\label{aqeqadqeqa}
	\begin{split}
	\Gmii{ij}{hs}(0)  
	= & (\alphaii{(h)}{s,3}\tau)^2 \sum_{t=1}^{p}  \sum_{q=1}^{p}  \Xmbii{(s)}{i,t})  (\Xmbii{(s)}{j,q})^{\top}   \sum_{r=1}^{m}  \EE_{\Umi{h}} \left[\Umi{h, tr} \Umi{h, qr}\right] \EE_{\Wmii{(h)}{s}} \left[\sigma'\left( \Zmi{i,tr} \right)  \sigma'\left( \Zmi{j,qr} \right)\right]\\
	= & (\alphaii{(h)}{s,3}\tau)^2 \sum_{t=1}^{p}  \Xmbii{(s)}{i,t})  (\Xmbii{(s)}{j,t})^{\top}   \sum_{r=1}^{m}   \EE_{\Wmii{(h)}{s}} \left[\sigma'\left( \Zmi{i,tr} \right)  \sigma'\left( \Zmi{j,qr} \right)\right]\\
	\end{split}
	\end{equation}
	where \ding{172} holds since $\EE_{\Umi{h}} \left[\Umi{h, tr} \Umi{h, qr}\right]=1$ if $t=q$ and $\EE_{\Umi{h}} \left[\Umi{h, tr} \Umi{h, qr}\right]=0$ if $t\neq q$.
	\begin{equation*}
	\Zmi{i,r} = \sum_{t=1}^m (\Wmii{(h)}{s,tr})^\top \Xmbii{(s)}{i,t}) .
	\end{equation*}
	
	Since the convolution parameter $\Wmii{(h)}{s}$ satisfies Gaussian distribution, $\Zmi{i,r}$ is a mean-zero Guassian variable with covariance matrix as follows
	\begin{equation}\label{qeqdadsad} 
	\begin{split}
	\EE \left[(\Zmi{i,r} )^{\top}  \Zmi{j,q} \right] = & \EE \sum_{t, t'} (\Wmii{(h)}{s,t})^\top \Xmbii{(s)}{i,t} (\Xmbii{(s)}{j,t'})^{\top}  (\Wmii{(h)}{s,t'q})^\top= \delta_{st}  \W^{(hs)}  \left( \sum_{t}  \Xmbii{(s)}{i,t} (\Xmbii{(s)}{j,t})^{\top}\right) \\
	=& \delta_{st} \W^{(hs)}  \left( \Qmtii{(s)}{ij} \right),
	\end{split}
	\end{equation}
	where $\delta_{st}$ is a random variable with $\delta_{st}= \pm 1$ with both probability 0.5, and 
	\begin{equation*}
	\begin{split}
	& \Kmtii{(ss)}{ij}= \frac{1}{m} \sum_{t=1}^m  \Xmii{(s)}{i,t} (\Xmii{(s)}{j,t})^\top, \qquad \qquad  \Qmtii{(ss)}{ij}= \frac{1}{m} \sum_{t=1}^m  \Xmbii{(s)}{i,t} (\Xmbii{(s)}{j,t})^\top.
	\end{split}
	\end{equation*}
	According to this definition, we actually have 
	\begin{equation*}
	\begin{split}
	&\Qmtii{(ss)}{ij, ab} = \tr{\Kmtii{(ss)}{ij,\Sii{(s)}{a}, \Sii{(s)}{b}}},
	\end{split}
	\end{equation*}
	where $\Kmtii{(ss)}{ij}  \in \Rs{p\times p}$, $\Qmtii{(ss)}{ij, ab} $ denotes the $(a,b)$-th entry in $\Qmtii{(ss)}{ij}$, and $\Sii{(s)}{a}=\{j\ | \ \Xmii{(s-1)}{:,j} \in \text{the}\ a-\text{th patch for convolution}\}$. 
	Then according to the following definitions 
	\begin{equation*}
	\begin{split}
	&\widehat{\Am}^{(s)}  = \begin{bmatrix}
	\Wii{h}{ss} (\Qmtii{(ss)}{ii}), \Wii{h}{ss}(\Qmtii{(ss)}{ij})\\
	\Wii{h}{ss} (\Qmtii{(ss)}{ji}), \Wii{h}{ss}(\Qmtii{(ss)}{jj})\\
	\end{bmatrix},\\
	&\Qmtii{(s)}{ij, ab} = \Qmtii{(ss)}{ij,ab} \EE_{( (\Mm,\Nm)\sim \widehat{\Am}^{(s)})}  \sigma'\left(\Mm \right) \sigma'\left(\Nm \right)^\top, \qquad \Kmtii{(s)}{ij, ab} = \tr{\Qmtii{(s)}{ij}}, \ (s=0,h-1).
	\end{split}
	\end{equation*}
	and Eqns.~\eqref{aqeqadqeqa} and~\eqref{qeqdadsad}, we have
	\begin{equation*}
	\begin{split}
	\EE \left[\Gmii{ij}{hs}(0)  \right] 
	=  (\alphaii{(h)}{s,3})^2  \Kmtii{(s)}{ij},\qquad \EE \left[\Gmii{}{hs}(0)  \right] 
	=  (\alphaii{(h)}{s,3})^2  \Kmtii{(s)}{}.
	\end{split}
	\end{equation*}

	In this way, we can apply the Hoeffding inequality and obtain that if $m\geq \Oc{\frac{n^2\log(n/\delta)}{\lambda^2}}$
	\begin{equation*}
	\begin{split}
	\left\|\Gmii{}{hs}(0) -  (\alphaii{(h)}{s,3})^2 \Kmtii{(s)}{}  \right\|_{\op}\leq \frac{\lambda}{8}.
	\end{split}
	\end{equation*}
	On the other hand, Lemma~\ref{boundofgrammartix} shows that  
	with probability at least $1-\delta$ 
	\begin{equation*}
	\left\|\Kmtii{(ss)}{ij}  - \Kmii{(ss)}{ij} \right\|_{\infty} \leq C \sqrt{\frac{\log(n^2 p^2 h^2/\delta)}{m}} \led{172} \frac{C_3\lambda}{n},
	\end{equation*}
	where \ding{172} holds by setting $m\geq \Oc{\frac{C_3^2n^2 \log(n^2 p^2 h^2/\delta)}{\lambda^2}}$. Moreover, Lemma~\ref{boundofinitialization} shows
	\begin{equation*}
	\begin{split}
	\frac{1}{\cxo} \leq \|\Xmii{(l)}{} (0)\|_F  \leq \cxo.  
	\end{split}
	\end{equation*}
	where $\cxo\geq 1$ is a constant. So $\|\Kmtii{(ss)}{ij}\|_{\infty}$ is upper bounded by $\cxo^2$.
	
	Next, Lemma~\ref{distributionbound}   shows if each diagonal entry in $\Am$ and $\Bm$ is upper bounded by c and lower upper bounded by $1/c$, then 
	\begin{equation*}
	|g(\Am) - g(\Bm)|\leq c\|\Am-\Bm\|_F \leq 2C_1\|\Am-\Bm\|_{\infty},
	\end{equation*}
	where $g(\Am)=\EE_{(u,v)\sim\mathcal{N}(0,\Am)}\sigma(u)\sigma(v)$, $C_1$ is a constant that only depends on $c$ and the Lipschitz and smooth parameter of $\sigma(\cdot)$. By applying this lemma, we can obtain
	\begin{equation*}
	\begin{split}
	&|\Qmtii{(ss)}{ij,rq} \EE_{(\Mm,\Nm) \sim \widehat{\Am}^{(s)} } \left[ \sigma'(\Mm_r))  \sigma'(\Nm_q)\right] - \Qmii{(ss)}{ij,rq} \EE_{(\Mm,\Nm) \sim \bar{\Am}^{(s)} } \left[ \sigma'(\Mm_r))  \sigma'(\Nm_q)\right]|\\
	\leq &|\Qmtii{(ss)}{ij,rq} \left(\EE_{(\Mm,\Nm) \sim \widehat{\Am}^{(s)} } \left[ \sigma'(\Mm_r))  \sigma'(\Nm_q)\right] -\EE_{(\Mm,\Nm) \sim \bar{\Am}^{(s)} } \left[ \sigma'(\Mm_r))  \sigma'(\Nm_q)\right] \right)|\\
	&+| (\Qmtii{(ss)}{ij,rq} -\Qmii{(ss)}{ij,rq} ) \EE_{(\Mm,\Nm) \sim \bar{\Am}^{(s)} } \left[ \sigma'(\Mm_r))  \sigma'(\Nm_q)\right]|\\
	\leq &C_1\cxo^2 |   \widehat{\Am}^{(s)}  - \Am^{(s)} | + \mu^2 | \Qmtii{(ss)}{ij,rq} -\Qmii{(ss)}{ij,rq} |\\
	\leq &C_1C_2 \cxo^2 \max_{i,j} |    \Qmtii{(ss)}{ij,rq} -\Qmbii{(ss)}{ij,rq} | + \mu^2 | \Qmtii{(ss)}{ij,rq} -\Qmii{(ss)}{ij,rq} |\\
	\leq &(C_1C_2 \cxo^2 + \mu^2) \| \Qmtii{(ss)}{ij} -\Qmii{(ss)}{ij} \|_{\infty}\\
	\leq &(C_1C_2 \cxo^2 + \mu^2) \max_{a,b}\left\|  \tr{\Kmtii{(ss)}{ij,\Sii{(s)}{a}, \Sii{(s)}{b}}} -  \tr{\Kmii{(ss)}{ij,\Sii{(s)}{a}, \Sii{(s)}{b}}} \right\|_{\infty}\\
	\leq &(C_1C_2 \cxo^2 + \mu^2) p \left\| \Kmtii{(ss)}{ij} - \Kmii{(ss)}{ij} \right\|_{\infty},
	\end{split}
	\end{equation*}
	where $C_2= 1+\|\Wii{h}{ss}\|_{L^{\infty}\rightarrow L^{\infty}}$. 
	
	Then we can bound 
	\begin{equation*}
	\begin{split}
	&\|\Kmtii{(s)}{} - \Kmbii{(s)}{} \|_{op}\leq \|\Kmtii{(s)}{} - \Kmbii{(s)}{} \|_{F}
	= \sqrt{\sum_{i=1}^n\sum_{j=1}^n \left[\tr{\Qmtii{(s)}{ij}} - \tr{\Qmii{(s)}{ij}}\right]^2}\\
	\leq & \sqrt{\sum_{i=1}^n\sum_{j=1}^n p\sum_{r=1}^{p} \left[ \Qmtii{(s)}{ij,rr}  -  \Qmii{(s)}{ij,rr} \right]^2}\\
	\leq & \! \sqrt{\sum_{i=1}^n\sum_{j=1}^n\!  p\sum_{r=1}^{p}\! \left[\Qmtii{(ss)}{ij,rr} \EE_{( (\Mm,\Nm)\sim \widehat{\Am}^{(s)})}  \sigma'\! \left(\Mm_r \right) \sigma'\left(\Nm_r \right)^\top  \!-\!  \Qmii{(ss)}{ij,rr} \EE_{( (\Mm,\Nm)\sim \bar{\Am}^{(s)})}  \sigma'\! \left(\Mm_r \right) \sigma'\! \left(\Nm_r \right)^\top \right]^2}\\
	\leq & \sqrt{\sum_{i=1}^n\sum_{j=1}^n p^2\sum_{r=1}^{p} (C_1C_2 \cxo^2 + \mu^2)^2 \| \Kmtii{(ss)}{ij} -\Kmbii{(ss)}{ij} \|_{\infty}^2}\\
	\leq & (C_1C_2 \cxo^2 + \mu^2) C_3p^2\lambda \\
	\led{172} & \frac{\lambda }{8},\\
	\end{split}
	\end{equation*}
	where \ding{172} holds by setting $C_3 \leq \frac{1}{(C_1C_2 \cxo^2 + \mu^2) p^2}$.  In this way, we have 
	\begin{equation*}
	\begin{split}
	\left\|\Gmii{}{hs}(0) -  (\alphaii{(h)}{s,3})^2 \Kmbii{(s)}{}  \right\|_{\op}\leq  \left\|\Gmii{}{hs}(0) - (\alphaii{(h)}{s,3})^2 \Kmtii{(s)}{}  \right\|_{\op} + (\alphaii{(h)}{s,3})^2 \left\|\Kmtii{(s)}{} -\Kmbii{(s)}{}    \right\|_{\op}  \leq \frac{\lambda}{4}.
	\end{split}
	\end{equation*}
	The proof is completed. 
\end{proof}

\subsubsection{Proof of Lemma~\ref{sdaffdasfasfdafdreawsdfadad}}\label{proofofsdaffdasfasfdafdreawsdfadad}
\begin{proof}
	To begin with, according to the definition, we have 
	\begin{equation*}
	\begin{split}
	&\Kmii{(ls)}{ij} - \bmi{i}^{(l)} (\bmi{i}^{(s)})^{\top}  =\sum_{t=1}^{l-1} \sum_{q=1}^{s-1} \left[ \alphaii{(l)}{t,2} \alphaii{(s)}{q,2} \left( \Kmii{(tq)}{ij} - \bmi{i}^{(t)} (\bmi{i}^{(q)})^{\top}  \right) \right.\\
	&\qquad \left. +  \tau^2  \alphaii{(l)}{t,3} \alphaii{(s)}{q,3} \left[ \EE_{(\Mmii{(ls)}{tq},\Nmii{(ls)}{tq})} \sigmai{\Mmii{(ls)}{tq}} \sigmai{\Nmii{(ls)}{tq}}^\top -  \EE_{\Mmii{(ls)}{tq}} \sigmai{\Mmii{(ls)}{tq}}  \EE_{\Nmii{(ls)}{tq}} \sigmai{\Nmii{(ls)}{tq}}^\top  \right] \right] .
	\end{split}
	\end{equation*}
	By defining  
	\begin{equation*}
	\begin{split}
	\Rm^{(ls)}_{tq} := &\EE_{(\Mmii{(ls)}{tq},\Nmii{(ls)}{tq})}  \! \! \!\begin{bmatrix}
	\sigmai{\Mmii{(ls)}{tq}} \sigmai{\Mmii{(ls)}{tq}}^{\top}, \sigmai{\Mmii{(ls)}{tq}}\sigmai{\Nmii{(ls)}{tq}}^\top\\
	\sigmai{\Nmii{(ls)}{tq}} \sigmai{\Mmii{(ls)}{tq}}^\top, \sigmai{\Nmii{(ls)}{tq}} \sigmai{\Nmii{(ls)}{tq}}^{\top}\\
	\end{bmatrix} \\
	&-  \EE_{(\Mmii{(ls)}{tq},\Nmii{(ls)}{tq})}   \! \! \begin{bmatrix}
	\sigmai{\Mmii{(ls)}{tq}} \\
	\sigmai{\Nmii{(ls)}{tq}}\\
	\end{bmatrix} \!\!  \EE_{(\Mmii{(ls)}{tq},\Nmii{(ls)}{tq})}  \begin{bmatrix}
	(\sigmai{\Mmii{(ls)}{tq}}^\top\!\!, \sigmai{\Nmii{(ls)}{tq}}^\top  
	\end{bmatrix} ,
	\end{split} 
	\end{equation*}
	we can further obtain
	\begin{equation*}
	\begin{split}
	& \begin{bmatrix}
	\Kmii{(ls)}{ii}, \Kmii{(ls)}{ij}\\
	\Kmii{(ls)}{ji}, \Kmii{(ls)}{jj}\\
	\end{bmatrix} - \begin{bmatrix}
	\bmi{i}^{(l)}  \\
	\bmi{j}^{(l)} \\
	\end{bmatrix}  \begin{bmatrix}
	( \bmi{i}^{(s)}  )^\top, ( \bmi{j}^{(s)} )^\top \\
	\end{bmatrix} \\
	=& \sum_{t=1}^{l-1} \sum_{q=1}^{s-1}    \left[ \alphaii{(l)}{t,2} \alphaii{(s)}{q,2} \left[\begin{bmatrix}
	\Kmii{(tq)}{ii}, \Kmii{(tq)}{ij}\\
	\Kmii{(tq)}{ji}, \Kmii{(tq)}{jj}\\
	\end{bmatrix} - \begin{bmatrix}
	\bmi{i}^{(t)} \\
	\bmi{j}^{(t)} \\
	\end{bmatrix}  \begin{bmatrix}
	(\bmi{i}^{q)} )^\top, (\bmi{j}^{(q)})^\top \\
	\end{bmatrix}   \right]  + \tau^2     \alphaii{(l)}{t,3} \alphaii{(l)}{q,3}  \Rm^{(ls)}_{tq}    \right].
	\end{split}
	\end{equation*}
	
	Let 
	\begin{equation*}
	\begin{split}
	\bar{\Rm}^{(ls)}_{tq} =\begin{bmatrix}
	\sigmai{\Mmii{(ls)}{tq}} \\
	\sigmai{\Nmii{(ls)}{tq}}\\
	\end{bmatrix}  -   \EE_{(\Mmii{(ls)}{tq},\Nmii{(ls)}{tq})}   \! \! \begin{bmatrix}
	\sigmai{\Mmii{(ls)}{tq}} \\
	\sigmai{\Nmii{(ls)}{tq}}\\
	\end{bmatrix}.
	\end{split} 
	\end{equation*}
	Then we have 
	\begin{equation*}
	\begin{split}
	\Rm^{(ls)}_{tq} =\EE_{(\Mmii{(ls)}{tq},\Nmii{(ls)}{tq})}  \left[ \bar{\Rm}^{(ls)}_{tq} (\bar{\Rm}^{(ls)}_{tq})^{\top}\right] \succeq \bm{0}.
	\end{split} 
	\end{equation*}
	Therefore, by induction, we can conclude 
	\begin{equation*}
	\begin{split}
	\begin{bmatrix}
	\Kmii{(ls)}{ii}, \Kmii{(ls)}{ij}\\
	\Kmii{(ls)}{ji}, \Kmii{(ls)}{jj}\\
	\end{bmatrix} - \begin{bmatrix}
	\bmi{i}^{(l)}  \\
	\bmi{j}^{(l)} \\
	\end{bmatrix}  \begin{bmatrix}
	( \bmi{i}^{(s)}  )^\top, ( \bmi{j}^{(s)} )^\top \\
	\end{bmatrix} 
	\succeq  & a \left[\begin{bmatrix}
	\Kmii{(-1)}{ii}, \Kmii{(-1)}{ij}\\
	\Kmii{(-1)}{ji}, \Kmii{(-1)}{jj}\\
	\end{bmatrix} - \begin{bmatrix}
	\bmi{i}^{(-1)} \\
	\bmi{j}^{(-1)} \\
	\end{bmatrix}  \begin{bmatrix}
	(\bmi{i}^{-1)} )^\top, (\bmi{j}^{(-1)})^\top \\
	\end{bmatrix}      \right]\\
	\succeq  & a \begin{bmatrix}
	\Kmii{(-1)}{ii}, \Kmii{(-1)}{ij}\\
	\Kmii{(-1)}{ji}, \Kmii{(-1)}{jj}\\
	\end{bmatrix} \overset{\text{\ding{172}}}{\succ} 0,
	\end{split}
	\end{equation*}
	where $a$ is a constant that depends on $ \alphaii{(l)}{t,2}\ (\forall l, t)$, \ding{172} holds by using Lemma~\ref{sdaffdasfreawsdf} which shows that $\Kmii{(00)}{ii} \succ 0$. Based on this result, we can estimate 
	\begin{equation*}
	\begin{split}
	& \begin{bmatrix}
	\Kmii{(ll)}{ii}, \Kmii{(ll)}{ij}\\
	\Kmii{(ll)}{ji}, \Kmii{(ll)}{jj}\\
	\end{bmatrix} - \begin{bmatrix}
	\bmi{i}^{(l)}  \\
	\bmi{j}^{(l)} \\
	\end{bmatrix}  \begin{bmatrix}
	( \bmi{i}^{(l)}  )^\top, ( \bmi{j}^{(l)} )^\top \\
	\end{bmatrix} \\
	=& \sum_{t=1}^{l-1} \sum_{q=1}^{l-1}    \left[ \alphaii{(l)}{t,2} \alphaii{(s)}{q,2} \left[\begin{bmatrix}
	\Kmii{(tq)}{ii}, \Kmii{(tq)}{ij}\\
	\Kmii{(tq)}{ji}, \Kmii{(tq)}{jj}\\
	\end{bmatrix} - \begin{bmatrix}
	\bmi{i}^{(t)} \\
	\bmi{j}^{(t)} \\
	\end{bmatrix}  \begin{bmatrix}
	(\bmi{i}^{q)} )^\top, (\bmi{j}^{(q)})^\top \\
	\end{bmatrix}   \right]  + \tau^2     \alphaii{(l)}{t,3} \alphaii{(l)}{q,3}  \Rm^{(ls)}_{tq}    \right]\\
	\succeq& \sum_{t=1}^{l-1}    \left[ (\alphaii{(l)}{t,2})^2 \left[\begin{bmatrix}
	\Kmii{(tt)}{ii}, \Kmii{(tt)}{ij}\\
	\Kmii{(tt)}{ji}, \Kmii{(tt)}{jj}\\
	\end{bmatrix} - \begin{bmatrix}
	\bmi{i}^{(t)} \\
	\bmi{j}^{(t)} \\
	\end{bmatrix}  \begin{bmatrix}
	(\bmi{i}^{t)} )^\top, (\bmi{j}^{(t)})^\top \\
	\end{bmatrix}   \right]  + \tau^2     (\alphaii{(l)}{t,3})^2    \Rm^{(ll)}_{tt}    \right]\\
	\succeq &
	\left(\prod_{t=1}^{l-1}  (\alphaii{(l)}{t,2})^2\right)  \left[\begin{bmatrix}
	\Kmii{(-1)}{ii}, \Kmii{(-1)}{ij}\\
	\Kmii{(-1)}{ji}, \Kmii{(-1)}{jj}\\
	\end{bmatrix} - \begin{bmatrix}
	\bmi{i}^{(-1)} \\
	\bmi{j}^{(-1)} \\
	\end{bmatrix}  \begin{bmatrix}
	(\bmi{i}^{-1)} )^\top, (\bmi{j}^{(-1)})^\top \\
	\end{bmatrix}      \right]\\
	\succeq & 
	\left(\prod_{t=1}^{l-1}  (\alphaii{(l)}{t,2})^2\right)  \begin{bmatrix}
	\Kmii{(-1)}{ii}, \Kmii{(-1)}{ij}\\
	\Kmii{(-1)}{ji}, \Kmii{(-1)}{jj}\\
	\end{bmatrix}.
	\end{split}
	\end{equation*}
	Then there must exit a constant $c$ such that 
	\begin{equation*}
	\begin{split}
	\lambda_{\min} (\Kmii{(ll)}{} )\geq 
	\left(\prod_{t=0}^{l-1}  (\alphaii{(l)}{t,2})^2\right)   
	\lambda_{\min}(\widehat{K}).
	\end{split}
	\end{equation*}
	where $\widehat{K} =  \begin{bmatrix}
	\Kmii{(-1)}{ii}, \Kmii{(-1)}{ij}\\
	\Kmii{(-1)}{ji}, \Kmii{(-1)}{jj}\\
	\end{bmatrix}.$
	On the other hand, we have 
	\begin{equation*}
	\begin{split}
	&\Qmii{(ll)}{ij, ab} = \tr{\Kmii{(ll)}{ij,\Sii{(l)}{a}, \Sii{(l)}{b}}},
	\end{split}
	\end{equation*}
	where $\Sii{(s)}{a}=\{j\ | \ \Xmii{(s-1)}{:,j} \in \text{the}\ a-\text{th patch for convolution}\}$.  This actually means that we can obtain $\Qmii{(ll)}{ij}$ by using (adding) linear transformation on $\Kmii{(ll)}{ij}$. Since for all $\Qmii{(ll)}{ij}$ we use the same linear transformation which means that   $\Qmii{(ll)}{}$ by using (adding) linear transformation on $\Kmii{(ll)}{}$. Since linear transformation does not change the eigenvalue property of a matrix, we can further obtain 
	\begin{equation*}
	\begin{split}
	\lambda_{\min} (\Qmii{(ll)}{} )\geq 
	\left(\prod_{t=0}^{l-1}  (\alphaii{(l)}{t,2})^2\right)   
	\lambda_{\min}(\widehat{K} ).
	\end{split}
	\end{equation*}
	Finally, let $\Qmii{}{}=\Bm\SSm\Bm^{\top}$ be the SVD of $\Qmii{}{}$ and $\Zm=\SSm^{1/2} \Bm^{\top}$ denotes $n$ samples (each column denotes one). Since $\Qmii{}{}$ is full rank, the samples in $\Zm$ are not parallel. In this way, we can apply Lemma~\ref{sdaffdasfreawsdf} and obtain that $\Qmii{(s)}{} $ which is defined below, is full rank
	\begin{equation*}
	\begin{split}
	&\Am^{(l)}  = \begin{bmatrix}
	\Wii{h}{ll} (\Qmii{(ll)}{ii}), \Wii{h}{ll}(\Qmii{(ll)}{ij})\\
	\Wii{h}{ll} (\Qmii{(ll)}{ji}), \Wii{h}{ll}(\Qmii{(ll)}{jj})\\
	\end{bmatrix},\\
	&\Qmii{(l)}{ij, ab} = \Qmii{(ll)}{ij,ab} \EE_{( (\Mm,\Nm)\sim \bar{\Am}^{(l)})}  \sigma'\left(\Mm \right) \sigma'\left(\Nm \right)^\top, \qquad \Kmii{(l)}{ij, ab} = \tr{\Qmii{(s)}{ij}}, \ (s=l,\cdots , h-1).
	\end{split}
	\end{equation*}
	Recall that Lemma~\ref{boundofinitialization} shows 
	\begin{equation*}
	\begin{split}
	\frac{1}{\cxo} \leq \|\Xmii{(l)}{} (0)\|_F  \leq \cxo.  
	\end{split}
	\end{equation*}
	where $\cxo\geq 1$ is a constant. Therefore, we have  $\Kmii{ll}{ii} =\langle \Xmii{(l)}{} (0), \Xmii{(l)}{} (0) \rangle \in[1/\cxo^2, \cxo^2]$ and thus $\Qmii{ll}{ii} =\langle \Phi(\Xmii{(l)}{} (0)), \Phi(\Xmii{(l)}{} (0) \rangle \geq  \langle \Xmii{(l)}{} (0), \Xmii{(l)}{} (0) \rangle \geq 1/\cxo^2$  and $\Qmii{ll}{ii} =\langle \Phi(\Xmii{(l)}{} (0)), \Phi(\Xmii{(l)}{} (0) \rangle \leq \kc  \langle \Xmii{(l)}{} (0), \Xmii{(l)}{} (0) \rangle \geq \kc/\cxo^2$.
	Then we have 
	\begin{equation*}
	\begin{split}
	\Qmii{(l)}{ij} =\Qmii{ll}{ij}  \EE_{( \Mm\sim \N{0,\Imm})}  \sigma'\left( \Mm \Zm_i \right) \sigma'\left(\Mm \Zm_j \right)^\top
	\end{split}
	\end{equation*}
	where  $\Zm=\SSm^{1/2} \Bm^{\top}$ and $\Zm_i=\Zm_{:i}$ in which  $\Qmii{ll}{}=\Bm\SSm\Bm^{\top}$ is the SVD of $\Qmii{ll}{}$. Since Since $\Qmii{ll}{}$ is full rank, the samples in $\Zm$ are not parallel. Then we can apply Lemma~\ref{sdaffdasfasfdafdreawsdf} and obtain 
	\begin{equation*}
	\begin{split}
	\lambda_{\min} (\Qmii{(l)}{} )\geq c_{\sigma}
	\left(\prod_{t=0}^{l-1}  (\alphaii{(l)}{t,2})^2\right)   
	\lambda_{\min}(\widehat{K} ),
	\end{split}
	\end{equation*}
	where $c_{\sigma}$ is a constant that only depends on $\sigma$ and input data.  Since 
	\begin{equation*}
	\begin{split}
	\Kmii{(s)}{ij, ab} = \tr{\Qmii{(s)}{ij}}, \ (s=0,h-1)
	\end{split}
	\end{equation*}
	which means that $\Kmii{(s)}{}$ can be obtained by using adding linear transformation on $\Qmii{(s)}{}$. So the eigenvalue of $\Kmii{(s)}{}$  also satisfies 
	\begin{equation*}
	\begin{split}
	\lambda_{\min} (\Kmii{(l)}{} )\geq c_{\sigma}
	\left(\prod_{t=0}^{l-1}  (\alphaii{(l)}{t,2})^2\right)   
	\lambda_{\min}(\widehat{K} ),
	\end{split}
	\end{equation*}

	In this way, we can further establish
	\begin{equation*}
	\begin{split}
	\lambda_{\min}\left(\Gmii{}{}(0)\right) \geq&  \sum_{s=0}^{h-1} \lambda_{\min}\left(\Gmii{}{hs}(0)\right)  \ged{172}    \sum_{s=0}^{h-1} (\alphaii{(h)}{s,3} )^2\lambda_{\min}   \left(\Kmii{(s)}{}(0)\right)  -\frac{\lambda }{4} \\
	\geq& \frac{3c_{\sigma}}{4} \sum_{s=0}^{h-1}(\alphaii{(h)}{s,3} )^2
	\left(\prod_{t=0}^{s-1}  (\alphaii{(s)}{t,2})^2\right)   
	\lambda_{\min}(\widehat{K} ),
	\end{split}
	\end{equation*}
	where \ding{172} holds since we set $\lambda=c_{\sigma} \sum_{s=0}^{h-1}(\alphaii{(h)}{s,3} )^2
	\left(\prod_{t=0}^{s-1}  (\alphaii{(s)}{t,2})^2\right)   
	\lambda_{\min}(\widehat{K})$ and Lemma~\ref{boundofdifference} shows 
	\begin{equation*}
	\begin{split}
	&\left\|\Gmii{}{hs}(0) -  (\alphaii{(h)}{s,3})^2  \Kmii{(s)}{}  \right\|_{\op}\leq    \frac{\lambda}{4}\qquad (s=0,\cdots,h).
	\end{split}
	\end{equation*}
	where $\lambda$ is a constant.  The proof is completed. 
\end{proof}

\section{Proofs of Results in Sec.~\ref{methodpart}}\label{proofofmethodpart}

\subsection{Proof of Theorem~\ref{lossproperty}}\label{Proofoflossproperty}
\begin{proof} We first prove the first result.  Suppose except one gate $\gii{(l)}{s,t}$, all remaining stochastic gates $\gii{(l')}{s',t}$  are fixed. Then we discuss the type of the gate $\gii{(l)}{s,t}$. Note  $\gii{(l)}{s,t}$ denotes one operation in the operation set $\Om\!=\!\{O_{t}\}_{t=1}^{s}$,  including zero operation, skip connection, pooling, and convolution with any kernel size,  between nodes $\Xmii{(s)}{}$ and $\Xmii{(l)}{}$. Now we discuss different kinds of operations. 
	
	If the gate $\gii{(l)}{s,t}$ is for zero operation, it is easily to check that the loss $F_{\mbox{val}}(\Wm^*(\betaii{}{}), \betaii{}{})$ in \eqref{dartsmodel}  will not change, since zero operation does not delivery any information to subsequent node $\Xmii{(l)}{}$. 
	
	If the gate $\gii{(l)}{s,t}$ is for skip connection, there are two cases. Firstly, increasing the weight  $\gii{(l)}{s,t}$ gives smaller loss. For this case, it directly obtain our result. Secondly, increasing  the weight  $\gii{(l)}{s,t}$ gives larger loss. For this case, suppose we increase $\gii{(l)}{s,t}$  to $\gii{(l)}{s,t} + \epsilon$.  Then node $\Xmii{(l)}{}$ will become $\Xmii{(l)}{}+\epsilon \Xmii{(s)}{}=\Xmii{(l)}{\mbox{conv}} + \Xmii{(l)}{\mbox{nonconv}}+\epsilon \Xmii{(s)}{}$ if we fix the remaining operations, where $\Xmii{(l)}{\mbox{conv}} $ denotes the output of convolution and $ \Xmii{(l)}{\mbox{nonconv}}$ denotes the sum of all remaining operations.  Now suppose the convolution operation between node $\Xmii{(l)}{}$ and $\Xmii{(s)}{}$ is $ \gii{(l)}{s,t} \convo{\Wmii{(l)}{s};\Xmii{(s)}{}}=\gii{(l)}{s,t}\sigmai{\Wmii{(l)}{s}\Phi(\Xmii{(s)}{})}$ where $t$ denotes the index of convolution in the operation set $\O$. Then  we consider a function
	\begin{equation}\label{afdasfcsafdcre2qd}
	\gii{(l)}{s,t}\sigmai{\Wmbii{(l)}{s}\Phi(\Xmii{(s)}{})}=-\epsilon \Xmii{(s)}{}.
	\end{equation} 
	Since for the almost activation functions are monotone increasing, this means that $\sigmai{}$ does not change the rank of $\Wmbii{(l)}{s}\Phi(\Xmii{(s)}{})$. At the same time, the linear transformation  $\Phi(\Xmii{(s)}{})$ has the same rank as $\Xmii{(s)}{}$. Then when $	\gii{(l)}{s,t}\neq 0$ there exist a $\Wmbii{(l)}{s}$ such that Eqn.~\eqref{afdasfcsafdcre2qd} holds.  On the other hand, we already have
	\begin{equation*}
	\gii{(l)}{s,t}\sigmai{\Wmii{(l)}{s}\Phi(\Xmii{(s)}{})}=\Xmii{(l)}{\mbox{conv}} .
	\end{equation*} 
	Since we assume the function $\sigmai{}$ is Lipschitz  and smooth	and the constant $\epsilon$ is sufficient small, then by using mean value theorem, there must exist $\gii{(l)}{s,t}\sigmai{\Wmtii{(l)}{s}\Phi(\Xmii{(s)}{})}=\Xmii{(l)}{\mbox{conv}}  -\epsilon \Xmii{(s)}{}.$ So the convolution can counteract the increment $\epsilon \Xmii{(s)}{}$ brought by increasing the weight of skip connection. In this way, the whole network remains the same, leading the same loss.  When the weight of convolution satisfies $\gii{(l)}{s,t} = 0$, we only need to increase $\gii{(l)}{s,t} $ to a positive constant, then we use the same method and can prove the same result. In this case, we actually increase the weights of skip connection and convolution at the same time, which also accords with our results in the Proposition~\ref{lossproperty}.
	
	If the gate $\gii{(l)}{s,t}$ is for pooling connection, we can use the same method for skip connection to prove our result, since pooling operation is also a linear transformation.
	
	If the gate $\gii{(l)}{s,t}$ is for convolution, then we increase it to  $\gii{(l)}{s,t} + \epsilon \gii{(l)}{s,t}$ and obtain the new output  $(1+\epsilon)\Xmii{(l)}{\mbox{conv}}$ because of $\gii{(l)}{s,t}\sigmai{\Wmii{(l)}{s}\Phi(\Xmii{(s)}{})}=\Xmii{(l)}{\mbox{conv}}$.  If the new feature map can lead to smaller loss, then we directly obtain our results. If   the new feature map can lead to larger loss we only need to find a new parameter $\Wmtii{(l)}{s}$ such that $\gii{(l)}{s,t}\sigmai{\Wmtii{(l)}{s}\Phi(\Xmii{(s)}{})}=\frac{1}{1+\epsilon}\Xmii{(l)}{\mbox{conv}}$. Since for most activation $\sigma(0)=0$, we have $\gii{(l)}{s,t}\sigmai{\Wmbii{(l)}{s}\Phi(\Xmii{(s)}{})}=0$ when $\Wmbii{(l)}{s}=0$. On the other hand,  we have $\gii{(l)}{s,t}\sigmai{\Wmii{(l)}{s}\Phi(\Xmii{(s)}{})}=\Xmii{(l)}{\mbox{conv}}$. Moreover since we assume the function $\sigmai{}$ is Lipschitz  and smooth	and the constant $\epsilon$ is sufficient small, then by using mean value theorem, there must exist $\Wmtii{(l)}{s}$ such that  $\gii{(l)}{s,t}\sigmai{\Wmtii{(l)}{s}\Phi(\Xmii{(s)}{})}=\frac{1}{1+\epsilon}\Xmii{(l)}{\mbox{conv}}$.

	Then we prove the results in the second part. From Theorem~\ref{mainconvergence3}, we know that for the $k$-th iteration in the search phase, increasing the weights  $\gii{(l)}{s,t_1}\ (l\neq h)$  of  skip connects and the weights $\gii{(h)}{s,t_2}$  of convolutions can  reduce the  loss $F_{\mbox{train}}(\Wm^*(\betaii{}{}), \betaii{}{})$ in \eqref{dartsmodel},  where $t_1$ and $t_2$ respectively denote the indexes of skip connection and convolution in the operation set $\Om\!=\!\{O_{t}\}_{t=1}^{s}$.  Specifically, Theorem~\ref{mainconvergence3} proves for the training loss 
	
	\begin{equation*}
	\begin{split}
	\|\ymm - \hmm(k)\|_2^2  \leq \left( 1 - \frac{\eta  \lambda }{4}\right)^{k}\|\ymm - \hmm(0)\|_2^2, 
	\end{split}
	\end{equation*} 
	where $\lambda 
	= \frac{3c_{\sigma}}{4} \lambda_{\min}(\widehat{\Km} )\sum_{s=0}^{h-1}(\alphaii{(h)}{s,3} )^2
	\prod_{t=0}^{s-1}  (\alphaii{(s)}{t,2})^2 
	$. Moreover, since $
	F(\Omegam) = \frac{1}{2n}\sum_{i=1}^{n} (\ui{i}-\ymi{i})^2= \frac{1}{2n}\|\um-\ymm\|_2^2 ,$ increasing the weights  $\gii{(l)}{s,t_1}\ (l\neq h)$  of  skip connects and the weights $\gii{(h)}{s,t_2}$  of convolutions can  reduce the  loss $F_{\mbox{train}}(\Wm^*(\betaii{}{}), \betaii{}{})$. Since  the samples for training and validation are drawn from the same distribution which means that $\EE [F_{\mbox{train}}(\Omegam)]=\EE[F_{\mbox{val}}(\Omegam)]$ , increasing  weights of skip connections and convolution  can reduce $F_{\mbox{val}}(\Omegam)$ in expectation. Then by using first-order extension, we can obtain 
	\begin{equation*}
	\EE\left[F_{\mbox{val}}(\gii{(l)}{s,t_1} + \epsilon) - F_{\mbox{val}}(\gii{(l)}{s,t_1} ) \right]=   \epsilon \EE\left[\nabla_{\gbii{(l)}{s,t_1}} F_{\mbox{val}}(\gii{(l)}{s,t_1} )\right] .
	\end{equation*}
	where $\gii{(l)}{s,t_1} \in \gbii{(l)}{s,t_1} \leq \gii{(l)}{s,t_1}+\epsilon$. 
	Since as above analysis, increasing the weights  $\gii{(l)}{s,t_1}\ (l\neq h)$  of  skip connects will reduce the current loss $F_{\mbox{val}}(\gii{(l)}{s,t_1}$ in expectation, which means that $\EE\left[\nabla_{\gii{(l)}{s,t_1}} F_{\mbox{val}}(\gii{(l)}{s,t_1} )\right]$ is positive. Since when the algorithm does not converge, we have  $0<C\leq \EE\left[\nabla_{\gii{(l)}{s,t_1}} F_{\mbox{val}}(\gii{(l)}{s,t_1} )\right]$. In this way, we have 
	\begin{equation*}
	\EE\left[F_{\mbox{val}}(\gii{(l)}{s,t_1} + \epsilon) - F_{\mbox{val}}(\gii{(l)}{s,t_1} ) \right]\geq C  \epsilon.
	\end{equation*}
	Similarly, for convolution we can obtain 
	\begin{equation*}
	\EE\left[F_{\mbox{val}}(\gii{(l)}{s,t_2} + \epsilon) - F_{\mbox{val}}(\gii{(l)}{s,t_2} ) \right]\geq C  \epsilon.
	\end{equation*}
	The proof is completed. 
\end{proof}

\subsection{Proof of Theorem~\ref{gateproperty}}\label{proofofgateproperty}

\begin{proof}
	For the results in the first part, it is easily to check according to the definitions. Now we focus on proving the results in the second part. When $\gtii{(l)}{s,t}\leq -\frac{a}{b-a}$, then $\gii{(l)}{s,t}=0$. Meanwhile, the cumulative distribution of $\gtii{(l)}{s,t}$ is $\Theta\big(\tau(\ln \delta - \ln(1-\delta)) - \betaii{(l)}{s,t} \big)$~\cite{louizos2017learning}. In this way, we can easily compute  
	\begin{equation*}
	\begin{split}
	\Pro\left(\gii{(l)}{s,t}\neq 0 \right)= & 1 -  \Pro\left(\gtii{(l)}{s,t}\leq -\frac{a}{b-a} \right) \\
	= & 1 -\Theta\left(\tau \left(\ln\big(-\frac{a}{b-a}\big) - \ln\big(1+\frac{a}{b-a}\big) \right)- \betaii{(l)}{s,t}\right)\\
	= &\Theta\left(\betaii{(l)}{s,t} - \tau \ln \frac{-a}{b} \right).
	\end{split}
	\end{equation*} 
	The proof is completed. 
\end{proof}

\subsection{Proof of Theorem~\ref{fasterconvergenceloss}}\label{Proofoffasterconvergenceloss}
\begin{proof}
	Here we first prove the convergence rate of the shallow network with two branches. The proof is very similar to Theorem~\ref{proofofmainconvergence3}. By using the totally same method, we can follow Lemma~\ref{mainconvergence2} to prove 
	\begin{equation*}
	\begin{split}
	\|\ymm - \hmm(k)\|_2^2  \leq \left( 1 - \frac{\eta  \lambda_{\min}\left( \Gm(0) \right)}{4}\right) \|\ymm - \hmm(k-1)\|_2^2.
	\end{split}
	\end{equation*} 
	Here $ \Gm(0)$ denotes the Gram matrix of the shallow network and have the same definition as the Gram matrix of deep network with one branch. Please refer to the definition of Gram matrix in Appendix~\ref{notations}. 
	
	The second step is to prove the smallest least eigenvalue of $\Gm(0)$ is lower bounded. For this step, the analysis method is also the same as the method to lower bounding smallest least eigenvalue of $\Gm(0)$ in DARTS.  Specifically,  by following Lemma~\ref{sdaffdasfasfdafdreawsdfadad}, we can obtain 
	\begin{equation*}
	\begin{split}
	\lambda_{\min}\left(\Gmii{}{}(0)\right) 
	\geq \frac{3c_{\sigma}}{4} \left[ \sum_{s=1}^{\frac{h}{2}-1}(\alphaii{(h/2)}{s,3} )^2
	\left(\prod_{t=0}^{s-1}  (\alphaii{(s)}{t,2})^2\right)  + \sum_{s=\frac{h}{2}}^{h-1}(\alphaii{h}{s,3} )^2
	\left(\prod_{t=0}^{s-1}  (\alphaii{(s)}{t,2})^2\right) \right]
	\lambda_{\min}(\Kmii{}{}).
	\end{split}
	\end{equation*}
	where  $c_\sigma$ is a constant that only depends on $\sigma$ and the input data, $\lambda_{\min}(\Kmii{}{})>0$ is given in Theorem~\ref{mainconvergence3}. 
	
	From Theorem~\ref{mainconvergence3}, we know that for deep cell with one branch, the loss satisfies 
	\begin{equation*}
	\begin{split}
	\|\ymm - \hmm(k)\|_2^2  \leq \left( 1 - \frac{\eta  \lambda }{4}\right)^{k}\|\ymm - \hmm(0)\|_2^2, 
	\end{split}
	\end{equation*} 
	where $\lambda 
	= \frac{3c_{\sigma}}{4} \lambda_{\min}(\Km)\sum_{s=0}^{h-1}(\alphaii{(h)}{s,3} )^2
	\prod_{t=0}^{s-1}  (\alphaii{(s)}{t,2})^2 
	$. 
	
	Since all weights $\alphaii{(l)}{s,t}$ belong to the range $[0,1]$, by comparison, the convergence rate $\lambda'$ of shallow cell with two branch  is large than the convergence rate $\lambda$ of shallow cell with two branch:
	\begin{equation*}
	\begin{split}
	\lambda'= & \frac{3c_{\sigma}}{4} \left[ \sum_{s=1}^{\frac{h}{2}-1}(\alphaii{(h/2)}{s,3} )^2
	\left(\prod_{t=0}^{s-1}  (\alphaii{(s)}{t,2})^2\right)  + \sum_{s=\frac{h}{2}}^{h-1}(\alphaii{h}{s,3} )^2
	\left(\prod_{t=0}^{s-1}  (\alphaii{(s)}{t,2})^2\right) \right]
	\lambda_{\min}(\Kmii{}{}) \\
	>&  \lambda = \frac{3c_{\sigma}}{4} \lambda_{\min}(\Km)\sum_{s=0}^{h-1}(\alphaii{(h)}{s,3} )^2
	\prod_{t=0}^{s-1}  (\alphaii{(s)}{t,2})^2 .
	\end{split}
	\end{equation*}
	This completes the proof. 
\end{proof}

\section{Proofs of Auxiliary Lemmas}\label{proofofAuxiliaryLemmas}

\subsection{Proof of Lemma~\ref{gradientcomputation}}\label{proofgradientcomputation}

\begin{proof}
	We use chain rule to obtain the following gradients:
	\begin{equation*} 
	\begin{split}
	&\frac{\partial \ell }{\partial \Xmii{(h-1)}{}}=  \taum (\ui{}  -\ymi{} )\Umi{h}\in\Rs{m\times p};\\
	&\frac{\partial \ell }{\partial \Xmii{(l)}{}}=   \taum(\ui{}  -\ymi{} ) \Umi{l} +\sum_{s=l+1}^{h}\frac{\partial \ell }{\partial \Xmii{(s)}{}} \frac{\partial \Xmii{(s)}{}}{\partial \Xmii{(l)}{}} \ (l=0,\cdots, h-2)\\
	&= \!\taum(\ui{} -\ymi{} ) \Umi{l} +\!\sum_{s=l+1}^{h}\!\left(\alphaii{(s)}{l,2}\frac{\partial \ell }{\partial \Xmii{(s)}{}}  \!+\! \alphaii{(s)}{l,3}\tau   \convb{(\Wmii{(s)}{l})^{\top} \!\left(\sigma'\left( \Wmii{(s)}{l} \Phi(\Xmii{(l)}{})\right) \odot \frac{\partial \ell }{\partial \Xmii{(s)}{}}\right)}  \right) \!\in\!\Rs{m\times p};\\
	&	\frac{\partial \ell }{\partial \Xmii{}{}}=   \frac{\partial \ell }{\partial \Xmii{(1)}{}} \frac{\partial \Xmii{(1)}{}}{\partial \Xmii{(0)}{}} =      \tau   \convb{(\Wmii{(0)}{})^{\top} \left(\sigma'\left( \Wmii{(0)}{} \Phi(\Xmii{}{})\right) \odot \frac{\partial \ell }{\partial \Xmii{(0)}{}}\right)} \in\Rs{m\times p},\\
	&\frac{\partial \ell }{\partial \Wmii{(l)}{s}}=  \frac{\partial \ell }{\partial \Xmii{(l)}{}} \frac{\partial \Xmii{(l)}{}}{\partial \Wmii{(l)}{s}} =  \alphaii{(l)}{s,3}\tau  \Phi(\Xmii{(s)}{})    \left(\sigma'\left( \Wmii{(l)}{s} \Phi(\Xmii{(s)}{})\right) \odot \frac{\partial \ell }{\partial \Xmii{(l)}{}}\right)^{\top}  \in\Rs{m\times p} \\
	&\qquad  \qquad\qquad\qquad \qquad\qquad\qquad\qquad\qquad\qquad\qquad\qquad\qquad\qquad\ (1\leq l \leq h, 1 \leq s \leq l-1);\\
	&\frac{\partial \ell }{\partial \Wmii{(0)}{}}=  \frac{\partial \ell }{\partial \Xmii{(0)}{}} \frac{\partial \Xmii{(0)}{}}{\partial \Wmii{(0)}{}} =   \tau  \Phi(\Xmii{}{})    \left(\sigma'\left( \Wmii{(0)}{} \Phi(\Xmii{}{})\right) \odot \frac{\partial \ell }{\partial \Xmii{(0)}{}}\right)^{\top}  \in\Rs{m\times p},\\
	&\frac{\partial \ell }{\partial \Umi{s}}=   \taum (\ui{} -\ymi{} ) \Xmii{(l)}{} \in\Rs{m\times p},\\
	\end{split}
	\end{equation*}
	where $\odot$ denotes the dot product. 
\end{proof}

\subsection{Proof of Lemma~\ref{gradientcomputation2}}\label{proofgradientcomputation2}

\begin{proof}
	We use chain rule to obtain the following gradients:
	\begin{equation*} 
	\begin{split}
	&\frac{\partial  \ui{} }{\partial \Xmii{(h-1)}{}}=  \taum  \Umi{h-1}\in\Rs{m\times p};\\
	&\frac{\partial  \ui{} }{\partial \Xmii{(l)}{}}= \taum  \Umi{l} +\sum_{s=l+1}^{h}\frac{\partial  \ui{} }{\partial \Xmii{(s)}{}} \frac{\partial \Xmii{(s)}{}}{\partial \Xmii{(l)}{}} \ (l=0,\cdots, h-2)\\
	&=\taum \Umi{l} +\sum_{s=l+1}^{h}\left(\alphaii{(s)}{l,2}\frac{\partial  \ui{} }{\partial \Xmii{(s)}{}}  + \alphaii{(s)}{l,3}\tau   \convb{(\Wmii{(s)}{l})^{\top} \left(\sigma'\left( \Wmii{(s)}{l} \Phi(\Xmii{(l)}{})\right) \odot \frac{\partial  \ui{} }{\partial \Xmii{(s)}{}}\right)}  \right) \in\Rs{m\times p};\\
	& \qquad  \qquad \qquad \qquad \qquad \qquad \qquad \qquad \qquad \qquad \qquad \qquad	\ (0\leq l \leq h-1, 0 \leq s \leq l-1),\\
	&	\frac{\partial\ui{}  }{\partial \Xmii{}{}}=  \frac{\partial  \ui{} }{\partial \Xmii{(1)}{}} \frac{\partial \Xmii{(1)}{}}{\partial \Xmii{(0)}{}} =      \tau   \convb{(\Wmii{(0)}{})^{\top} \left(\sigma'\left( \Wmii{(0)}{} \Phi(\Xmii{}{})\right) \odot \frac{\partial \ui{} }{\partial \Xmii{(0)}{}}\right)} \in\Rs{m\times p},\\
	&\frac{\partial  \ui{} }{\partial \Wmii{(l)}{s}}=  \frac{\partial  \ui{} }{\partial \Xmii{(l)}{}} \frac{\partial \Xmii{(l)}{}}{\partial \Wmii{(l)}{s}} =  \alphaii{(l)}{s,3}\tau  \Phi(\Xmii{(s)}{})    \left(\sigma'\left( \Wmii{(l)}{s} \Phi(\Xmii{(s)}{})\right) \odot \frac{\partial  \ui{} }{\partial \Xmii{(l)}{}}\right)^{\top}  \in\Rs{m\times p}\\
	&\qquad  \qquad\qquad\qquad \qquad\qquad\qquad\qquad\qquad\qquad\qquad\qquad\qquad\qquad\ (0\leq l \leq h-1, 1 \leq s \leq l-1);\\
	&\frac{\partial  \ui{} }{\partial \Wmii{(0)}{}}=  \frac{\partial  \ui{} }{\partial \Xmii{(0)}{}} \frac{\partial \Xmii{(0)}{}}{\partial \Wmii{(0)}{}} =   \tau  \Phi(\Xmii{}{})    \left(\sigma'\left( \Wmii{(0)}{} \Phi(\Xmii{}{})\right) \odot \frac{\partial  \ui{} }{\partial \Xmii{(0)}{}}\right)^{\top}  \in\Rs{m\times p},\\
	\end{split}
	\end{equation*}
	where $\odot$ denotes the dot product. 
\end{proof}

\subsection{Proof of Lemma~\ref{boundofinitialization}}\label{proofofboundofinitialization}
\begin{proof}
	We each layer in turn. Our proof follows the proof framework in~\cite{du2018gradient}. Note for notation simplicity, we have assumed that the input $\Xm$ is of size $m\times p$ in Sec.~\ref{notations}. To begin with, we look at the first layer. For brevity, let $\Hm=\Phi(\Xm)$. According to the definition, we have 
	\begin{equation*}
	\begin{split}
	\EE\left[ \|\Xmii{(0)}{}(0)\|_F^2 \right] =& \tau^2 \EE\left[ \| \sigmai{\Wmii{(0)}{}(0) \Phi(\Xm)}\|_F^2 \right]
	= \tau^2 \sum_{i=1}^{m} \sum_{j=1}^{p}\EE\left[  \sigma^2(\Wmii{(0)}{i:}(0) \Hm_{:j}) \right] \\
	\lee{172} &    \sum_{j=1}^{p}\EE_{\omega \sim \Nn(0,1)}\left[  \sigma^2( \|\Hm_{:j}\|_F  \omega) \right] 
	\ged{173}     \EE_{\omega \sim \Nn(0,1)}\left[  \sigma^2( \|\Hm_{:j'}\|_F  \omega) \right] \\
	\geq  &   \EE_{\omega \sim \Nn(0,\frac{1}{\sqrt{p}})}\left[  \sigma^2(  \omega) \right]:=c >0,
	\end{split}
	\end{equation*}
	where \ding{172} holds since $\tau=1/\sqrt{m}$ and the entries in $\Wmii{(0)}{}(0)$ obeys i.i.d. Gaussian distribution which gives $\sum_{i=1}^n a_i \omega_i \sim \N(0,  \sum_{i=1}^{n} a_i^2) $ with $\omega_i \sim\N(0,1)$;  \ding{173} holds since $\|\Xm\|=1$ which means there must exist one $j'$ such that  $\|\Hm_{:j'}\|_F \geq \frac{1}{\sqrt{p}}$. 
	
	Next, we can bound the variance
	\begin{equation*}
	\begin{split}
	&\Var\left[ \|\Xmii{(0)}{}(0)\|_F^2 \right] \\
	=& \tau^4\Var\left[ \|\sigmai{\Wmii{(0)}{}(0) \Phi(\Xm)}\|_F^2 \right] = \tau^4 \Var\left[\sum_{i=1}^{m} \sum_{j=1}^{p}\EE\left[  \sigma^2(\Wmii{(0)}{i:}(0) \Hm_{:j}) \right]  \right] \\  
	\lee{172}&  \tau^2 \Var\left[  \sum_{j=1}^{p}\EE\left[  \sigma^2(\Wmii{(0)}{i:} (0)\Hm_{:j}) \right]  \right]   
	\led{173}  \tau^2\EE_{\omega \sim \Nn(0,1)}\left[ \left( \sum_{j=1}^{p}  \left(\sigma(0) + \|\Hm_{:j}\| |\omega| \right)^2   \right)^2 \right] \\ 
	\leq  &  \frac{p^2}{m} c_1, \\ 
	\end{split}
	\end{equation*}
	where \ding{172} holds since $\tau=1/\sqrt{m}$ and the entries in $\Wmii{(0)}{}(0)$ obeys i.i.d. Gaussian distribution, \ding{173} holds since  $\Var{(x)} \leq \EE[x^2] - [\EE(x)]^2$,  \ding{174} holds since $ \|\Hm_{:j}\|\leq 1$ and $c_1 = \sigma^4(0) + 4  |\sigma^3(0)| \mu \sqrt{2/\pi} + 8|\sigma(0)| \mu^3 \sqrt{2/\pi} +32 \mu^4   $. Then by using Chebyshev's inequality in Lemma~\ref{Chebyshevinequality}, we have
	\begin{equation*}
	\begin{split}
	\Pro\left( |\|\Xmii{(0)}{}(0)\|_F^2-\EE[\|\Xmii{(0)}{}(0)\|_F^2]|\geq \frac{c}{2} \right) \leq \frac{4\Var(\|\Xmii{(0)}{}(0)\|_F^2)}{c^2} \leq  \frac{4p^2}{m c^2} c_1.
	\end{split}
	\end{equation*}
	By setting $m\geq \frac{4c_1 n p^2}{c^2\delta}$, we have with probability at least $1- \frac{\delta}{n}$,
	\begin{equation*}
	\begin{split}
	\|\Xmii{(0)}{}(0)\|_F^2 \geq \frac{c}{2} .
	\end{split}
	\end{equation*}
	Meanwhile, we can upper bound $ \|\Xmii{(0)}{}(0)\|_F^2$ as follows:
	\begin{equation*}
	\begin{split}
	\|\Xmii{(0)}{}(0)\|_F^2 \leq \tau^2   \| \sigmai{\Wmii{(0)}{}(0) \Phi(\Xm)}\|_F^2  \leq \tau^2 \mu^2  \| \Wmii{(0)}{}(0) \Phi(\Xm) \|_F^2 \led{172}   \mu^2 \cwo^2 \| \Phi(\Xm) \|_F^2 \led{173} \kc \mu^2 \cwo^2,
	\end{split}
	\end{equation*}
	where \ding{172} holds since $\|\Wmii{(l)}{s}(0)\|_2\leq \sqrt{m} \cwo$, and \ding{173} uses $\| \Phi(\Xm) \|_F^2 \leq \kc\|\Xm\|_F^2$. 
	
	Next we consider the cases where $l\geq 1$.  According to the definition, we can obtain
	\begin{equation*}
	\begin{split}
	\|\Xmii{(l)}{} (0)\|_F =& \left\|\sum_{s=0}^{l-1}   \left(\alphaii{(l)}{s,2}  \Xmii{(s)}{}(0)  + \alphaii{(l)}{s,3} \tau \sigmai{\Wmii{(l)}{s}(0)  \Phi(\Xmii{(s)}{}(0))}  \right) \right\|_F \\
	\leq&  \sum_{s=0}^{l-1}   \left( \alphaii{(l)}{s,2}  \|\Xmii{(s)}{}(0) \|_F  + \alphaii{(l)}{s,3} \tau \|\sigmai{\Wmii{(l)}{s}(0)  \Phi(\Xmii{(s)}{}(0))}  \|_F \right)\\
	\led{172}&   \left( \alphaii{(l)}{s,2}  + \alphaii{(l)}{s,3} \sqrt{\kc} \mu \cwo    \right)\sum_{s=0}^{l-1}  \|\Xmii{(s)}{}(0) \|_F \\
	\led{173} & \frac{c_2^{l+1} -1}{c_2-1} c_2  \sqrt{\kc} \mu \cwo,
	\end{split}
	\end{equation*}
	where \ding{172} uses the fact that $\|\sigmai{\Wmii{(l)}{s}(0)  \Phi(\Xmii{(s)}{}(0))}  \|_F \leq \mu \|\Wmii{(l)}{s}(0)  \Phi(\Xmii{(s)}{}(0))\|_F \leq \sqrt{m} \mu \cwo\|  \Phi(\Xmii{(s)}{}(0))\|_F \leq \sqrt{m} \mu \sqrt{\kc} \cwo\|\Xmii{(s)}{}(0)\|_F$, \ding{173} holds by setting $c_2=  \alphaii{(l)}{s,2}  + \alphaii{(l)}{s,3} \sqrt{\kc} \mu \cwo$. Similarly, we can obtain
	\begin{equation*}
	\begin{split}
	\|\Xmii{(l)}{} (0)\|_F =& \left\|\sum_{s=0}^{l-1}   \left(\alphaii{(l)}{s,2}  \Xmii{(s)}{}(0)  + \alphaii{(l)}{s,3} \tau \sigmai{\Wmii{(l)}{s}(0)  \Phi(\Xmii{(s)}{}(0))}  \right) \right\|_F \\
	\geq&  \min_{0\leq s\leq l-1}  \left| \alphaii{(l)}{s,2}  \|\Xmii{(s)}{}(0) \|_F  - \alphaii{(l)}{s,3} \tau \|\sigmai{\Wmii{(l)}{s}(0)  \Phi(\Xmii{(s)}{}(0))}  \|_F \right|\\
	\geq &   \min_{0\leq s\leq l-1}  \left|\alphaii{(l)}{s,2}  - \alphaii{(l)}{s,3} \sqrt{\kc} \mu \cwo    \right|  \|\Xmii{(s)}{}(0) \|_F \\
	\geq  & \left|\alphaii{(l)}{s,2}  - \alphaii{(l)}{s,3} \sqrt{\kc} \mu \cwo    \right|^{l-1} \sqrt{\kc} \mu \cwo>0.
	\end{split}
	\end{equation*}
	Therefore, we can obtain that there exists a constant $\cxo$ such that for all $l\in[0,1,\cdots,h-1]$,
	\begin{equation*}
	\begin{split}
	\frac{1}{\cxo} \leq \|\Xmii{(l)}{} (0)\|_F  \leq \cxo.
	\end{split}
	\end{equation*}
	The proof is completed. 
\end{proof}

\subsection{Proof of Lemma~\ref{boundofmidoutput}}\label{Proofofboundofmidoutput}

\begin{proof}
	For this proof, we will respectively  bound  each layer. We first consider the first layer, namely $l=1$.  
	
	\textbf{Step 1. Case where $l=0$: upper bound of  $\|\Xmii{(0)}{} (k)-\Xmii{(0)}{} (0)\|_F$}. According to the definition, we have $\Xmii{(0)}{} (k)=\tau \sigmai{\Wmii{(0)}{}(k)\Phi(\Xm)} $ which yields 
	\begin{equation*}
	\begin{split}
	\|\Xmii{(0)}{} (k)-\Xmii{(0)}{} (0)\|_F = & \tau \|   \sigmai{\Wmii{(0)}{}(k)\Phi(\Xm)}  -    \sigmai{\Wmii{(0)}{}(k)\Phi(\Xm)} \|_F\\
	\led{172}  & \tau\mu \| \Wmii{(0)}{}(k) \Phi(\Xm) - \Wmii{(0)}{}(0) \Phi(\Xm)   \|_F\\
	\led{173}  &  \tau \mu \sqrt{\kc}\|  \Wmii{(0)}{}(k)  - \Wmii{(0)}{}(0)     \|_F\\
	\led{174}  &   \mu \taum \sqrt{\kc} \rc,
	\end{split}
	\end{equation*} 
	where \ding{172} uses the $\mu$-Lipschitz of $\sigma(\cdot)$, \ding{173} uses $ \|\Phi(\Xm)\| \leq \sqrt{\kc} \|\Xm\|\leq \sqrt{\kc}  $, \ding{174} uses the assumption $\|\Wmii{(0)}{}(k)  - \Wmii{(0)}{}(0)\|_2 \leq \taum \sqrt{\m} \rc$.

	\textbf{Step 2. Case where $l\geq 1$: upper bound of  $\|\Xmii{(l)}{} (k)-\Xmii{(l)}{} (0)\|_F$}. 	According to the definition, we have 
	\begin{equation*} 
	\begin{split}
	&\|\Xmii{(l)}{} (k)-\Xmii{(l)}{} (0)\|_F \\
	= & \left\|  \sum_{s=0}^{l-1}   \left[\alphaii{(l)}{s,2}  \left(\Xmii{(s)}{}(k)-\Xmii{(s)}{}(0) \right)  + \alphaii{(l)}{s,3} \tau \left( \sigmai{\Wmii{(l)}{s}(k)  \Phi(\Xmii{(s)}{}(k) )} -  \sigmai{\Wmii{(l)}{s}(0)  \Phi(\Xmii{(s)}{}(0) )} \right) \right]    \right\|_F\\
	= &  \sum_{s=0}^{l-1}  \left[\alphaii{(l)}{s,2}    \left\| \Xmii{(s)}{}(k)-\Xmii{(s)}{}(0) \right\|_F + \alphaii{(l)}{s,3} \tau \left\| \sigmai{\Wmii{(l)}{s}(k)  \Phi(\Xmii{(s)}{}(k) )} -  \sigmai{\Wmii{(l)}{s}(0)  \Phi(\Xmii{(s)}{}(0) )}   \right\|_F\right]\\
	\leq &  \sum_{s=0}^{l-1}  \left[\alphaii{(l)}{s,2}    \left\| \Xmii{(s)}{}(k)-\Xmii{(s)}{}(0) \right\|_F + \alphaii{(l)}{s,3} \tau \mu \left\| \Wmii{(l)}{s}(k)  \Phi(\Xmii{(s)}{}(k) ) -  \Wmii{(l)}{s}(0)  \Phi(\Xmii{(s)}{}(0) )   \right\|_F\right]\\
	\end{split}
	\end{equation*} 
	Then we first bound the second term as follows:
	\begin{equation*}
	\begin{split}
	&\left\| \Wmii{(l)}{s}(k)  \Phi(\Xmii{(s)}{}(k) ) -  \Wmii{(l)}{s}(0)  \Phi(\Xmii{(s)}{}(0) )   \right\|_F\\
	\leq &\left\| \Wmii{(l)}{s}(k)  \Phi(\Xmii{(s)}{}(k) ) -  \Wmii{(l)}{s}(k)  \Phi(\Xmii{(s)}{}(0) )   \right\|_F+ \left\| \Wmii{(l)}{s}(k)  \Phi(\Xmii{(s)}{}(0) ) -  \Wmii{(l)}{s}(0)  \Phi(\Xmii{(s)}{}(0) )   \right\|_F\\
	\leq & \|\Wmii{(l)}{s}(k) \| \left\| \Phi(\Xmii{(s)}{}(k) )  -  \Phi(\Xmii{(s)}{}(0) )  \right\|_F + \left\|  \Wmii{(l)}{s}(k)   -  \Wmii{(l)}{s}(0)    \right\|_F \|  \Phi(\Xmii{(s)}{}(0) )\|_F\\
	\leq & \sqrt{\kc} \|\Wmii{(l)}{s}(k) \| \left\| \Xmii{(s)}{}(k)   - \Xmii{(s)}{}(0)   \right\|_F + \sqrt{\kc} \left\|  \Wmii{(l)}{s}(k)   -  \Wmii{(l)}{s}(0)    \right\|_F \|  \Xmii{(s)}{}(0) \|_F\\
	\led{172} &\sqrt{\kc}\sqrt{m} \left( \taum \rc + \cwo\right) \left\| \Xmii{(s)}{}(k)   - \Xmii{(s)}{}(0)   \right\|_F +  \taum \sqrt{\kc m}\cxo \rcc ,
	\end{split}
	\end{equation*} 
	where in \ding{172} we use $\|\Wmii{(l)}{s}(k) \|_F \leq \|\Wmii{(l)}{s}(k)  -\Wmii{(l)}{s}(0)\|_F+\| \Wmii{(l)}{s}(0) \|_F \leq \sqrt{m}(\taum \rc + \cwo)$, $\left\|  \Wmii{(l)}{s}(k)   -  \Wmii{(l)}{s}(0)    \right\|_F\leq  \taum \sqrt{m} \rcc$, and the results in Lemma~\ref{boundofinitialization} that $\frac{1}{\cxo} \leq \|\Xmii{(l)}{} (0)\|_F  \leq \cxo$.  Plugging this result into the above inequality gives
	\begin{equation}\label{afcsafwfaws}
	\begin{split}
	&\|\Xmii{(l)}{} (k)-\Xmii{(l)}{} (0)\|_F \\
	\leq &  \sum_{s=0}^{l-1}  \left[\alphaii{(l)}{s,2}    \left\| \Xmii{(s)}{}(k)-\Xmii{(s)}{}(0) \right\|_F + \alphaii{(l)}{s,3} \tau \mu \left\| \Wmii{(l)}{s}(k)  \Phi(\Xmii{(s)}{}(k) ) -  \Wmii{(l)}{s}(0)  \Phi(\Xmii{(s)}{}(0) )   \right\|_F\right]\\
	\leq &  \sum_{s=0}^{l-1}  \left[ \left(\alphaii{(l)}{s,2} +\alphaii{(l)}{s,3}  \mu\sqrt{\kc} \left( \taum \rc + \cwo\right)  \right)\left\| \Xmii{(s)}{}(k)-\Xmii{(s)}{}(0) \right\|_F + \alphaii{(l)}{s,3} \taum  \mu \sqrt{\kc }\cxo \rcc \right]\\
	\leq &  \sum_{s=0}^{l-1}  \left[ \left(\alphaii{(l)}{s,2} +\alphaii{(l)}{s,3}  \mu\sqrt{\kc} \left( \taum \rc + \cwo\right)  \right)\left\| \Xmii{(s)}{}(k)-\Xmii{(s)}{}(0) \right\|_F + \alphaii{(l)}{s,3}  \taum \mu \sqrt{\kc }\cxo \rcc \right]\\
	\leq &  \sum_{s=0}^{l-1}  \left[ \left(\alphaii{}{2} +\alphaii{}{3}  \mu\sqrt{\kc} \left(\taum \rc + \cwo\right)  \right)\left\| \Xmii{(s)}{}(k)-\Xmii{(s)}{}(0) \right\|_F + \alphaii{(l)}{s,3} \taum \mu \sqrt{\kc }\cxo \rcc \right]\\
	\leq &  \left(1+ \alphaii{}{2} +\alphaii{}{3}  \mu\sqrt{\kc} \left( \taum\rc + \cwo\right)  \right) \|\Xmii{(l-1)}{} (k)-\Xmii{(l-1)}{} (0)\|_F\\
	\leq &  \left(1+ \alphaii{}{2} +\alphaii{}{3}  \mu\sqrt{\kc} \left(\taum \rc + \cwo\right)  \right)^{l} \|\Xmii{(0)}{} (k)-\Xmii{0)}{} (0)\|_F\\
	\leq &  \left(1+ \alphaii{}{2}+\alphaii{}{3}\mu\sqrt{\kc} \left( \taum\rc + \cwo\right)  \right)^{l} \taum \mu \sqrt{\kc} \rc,
	\end{split}
	\end{equation} 
	where $\alphaii{}{2}=\max_{s,l}\alphaii{(l)}{s,2}$ and $\alphaii{}{3}=\max_{s,l} \alphaii{(l)}{s,3}$.
	
	By using Eqn.~\eqref{afcsafwfaws}, we have 
	\begin{equation*}
	\begin{split}
	\left\| \Wmii{(l)}{s}(k)  \Phi(\Xmii{(s)}{}(k) ) -  \Wmii{(l)}{s}(0)  \Phi(\Xmii{(s)}{}(0) )   \right\|_F \leq \frac{1}{\alphai{3}}
	\left(1+ \alphaii{}{2}+\alphaii{}{3}\mu\sqrt{\kc} \left( \taum \rc + \cwo\right)  \right)^{l}  \taum \sqrt{\kc m} \rc,
	\end{split}
	\end{equation*} 
	The proof is completed. 
\end{proof}

\subsection{Proof of Lemma~\ref{boundgradient}}\label{proofofboundgradient}
 
\begin{proof}
	According to definition, we have
	\begin{equation}\label{adadcasfcsaf}
	\begin{split}
	\frac{1}{n}\sum_{i=1}^{n} \left\|\frac{\partial \ell }{\partial \Xmii{(h)}{i}(t)}\right\|_F
	=  \frac{1}{n}\sum_{i=1}^{n}\taum \left\|(\ui{i}(t) -\ymi{i} ) \Umi{h} (t) \right\|_F 
	\led{172}  \frac{1}{\sqrt{n}}  \taum \left\|\um(t)-\ymm  \right\|_F \left\| \Umi{l}(t) \right\|_F \led{173} \taum c_y c_u,
	\end{split}
	\end{equation}	
	where  \ding{172} holds since $\sum_{i=1}^{n}|\ui{i} -\ymi{i}|  \leq \sqrt{n} \|\um-\ymm\|_2= \sqrt{n} \sqrt{\sum_i (\ui{i} -\ymi{i})^2}$, \ding{173} holds by assuming $ \frac{1}{\sqrt{n}}   \left\|\um(t)-\ymm  \right\|_F = c_y$ and $ \left\| \Umi{h}(t) \right\|_F  \leq  c_u$. 
	
	Then for $0\leq l <h$, we have 
	\begin{equation*}
	\begin{split}
	& \frac{1}{n}\sum_{i=1}^{n} \left\|\frac{\partial \ell }{\partial \Xmii{(l)}{i}(t)}\right\|_F 
	=  \frac{1}{n}\sum_{i=1}^{n} \left\|\taum (\ui{i}(t) -\ymi{i} ) \Umi{l}(t) \right. \\ &\left.+\sum_{s=l+1}^{h-1}\left(\alphaii{(s)}{l,2}\frac{\partial \ell }{\partial \Xmii{(s)}{i}(t) }  + \alphaii{(s)}{l,3}\tau   \convb{(\Wmii{(s)}{l}(t) )^{\top} \left(\sigma'\left( \Wmii{(s)}{l}(t)  \Phi(\Xmii{(l)}{i}(t) )\right) \odot \frac{\partial \ell }{\partial \Xmii{(s)}{i}(t) }\right)}  \right) \right\|_F \\
	\leq &  \frac{1}{n}\sum_{i=1}^{n} \taum \left\|(\ui{i}(t)  -\ymi{i} ) \Umi{l}(t)   \right\|_F\\ &+\sum_{s=l+1}^{h-1} \frac{1}{n}\sum_{i=1}^{n} \left\| \alphaii{(s)}{l,2}\frac{\partial \ell }{\partial \Xmii{(s)}{i}(t) }  + \alphaii{(s)}{l,3}\tau   \convb{(\Wmii{(s)}{l}(t) )^{\top} \left(\sigma'\left( \Wmii{(s)}{l} (t) \Phi(\Xmii{(l)}{i}(t) )\right) \odot \frac{\partial \ell }{\partial \Xmii{(s)}{i}(t) }\right)}   \right\|_F \\
	\end{split}
	\end{equation*}	
	The main task is to bound 
	\begin{equation*}
	\begin{split}
	&\left\| \alphaii{(s)}{l,2}\frac{\partial \ell }{\partial \Xmii{(s)}{i}(t) }  + \alphaii{(s)}{l,3}\tau   \convb{(\Wmii{(s)}{l}(t) )^{\top} \left(\sigma'\left( \Wmii{(s)}{l} (t) \Phi(\Xmii{(l)}{i}(t) )\right) \odot \frac{\partial \ell }{\partial \Xmii{(s)}{i}(t) }\right)}   \right\|_F \\
	\leq &\alphaii{(s)}{l,2}\left\| \frac{\partial \ell }{\partial \Xmii{(s)}{i}(t) }  \right\|_F+\alphaii{(s)}{l,3}\tau   \left\|  \convb{(\Wmii{(s)}{l}(t) )^{\top} \left(\sigma'\left( \Wmii{(s)}{l} (t) \Phi(\Xmii{(l)}{i}(t) )\right) \odot \frac{\partial \ell }{\partial \Xmii{(s)}{i}(t) }\right)}     \right\|_F \\
	\led{172} &\alphaii{(s)}{l,2}\left\| \frac{\partial \ell }{\partial \Xmii{(s)}{i}(t) }  \right\|_F +\alphaii{(s)}{l,3}\tau \mu \sqrt{\kc} \|\Wmii{(s)}{l}(t)\|_F \left\|   \frac{\partial \ell }{\partial \Xmii{(s)}{i}(t)}   \right\|_F \\
	\led{172} &\left(\alphaii{(s)}{l,2} + \alphaii{(s)}{l,3}  \mu \sqrt{\kc}  ( \cwo + \taum   r) \right)\left\| \frac{\partial \ell }{\partial \Xmii{(s)}{i}(t) }  \right\|_F,
	\end{split}
	\end{equation*}	
	where \ding{172} holds since $\|\convb{\Xm}\|_F \leq \sqrt{\kc} \|\Xm\|_F$ and  the activation function $\sigmai{\cdot}$  is $\mu$-Lipschitz, \ding{173} holds since $\|\Wmii{(s)}{l}(t)\|_F  \leq \|\Wmii{(s)}{l}(t) - \Wmii{(s)}{l}(0)\|_F  + \|\Wmii{(s)}{l}(0)\|_F  \leq \sqrt{m}(\cwo +\taum r)$. Similar to \eqref{adadcasfcsaf}, we can prove 
	\begin{equation*} 
	\begin{split}
	\frac{1}{n}\sum_{i=1}^{n} \taum \left\|(\ui{i}(t)  -\ymi{i} ) \Umi{l}(t)   \right\|_F 
	\leq   \frac{1}{\sqrt{n}}  \taum \left\|\um(t)-\ymm  \right\|_F \left\| \Umi{l}(t) \right\|_F \leq \taum c_y c_u,
	\end{split}
	\end{equation*}	
	Combining the above results yields  
	\begin{equation*}
	\begin{split}
	\frac{1}{n}\sum_{i=1}^{n} \left\|\frac{\partial \ell }{\partial \Xmii{(l)}{i}(t)}\right\|_F 
	\leq &  \taum c_y c_u +\sum_{s=l+1}^{h-1}\left(\alphaii{(s)}{l,2} + \alphaii{(s)}{l,3} \mu \sqrt{\kc} ( \cwo +\taum r) \right)  \frac{1}{n}\sum_{i=1}^{n} \left\| \frac{\partial \ell }{\partial \Xmii{(s)}{i}(t) }  \right\|_F\\
	\led{172} &  \taum c_y c_u +\sum_{s=l+1}^{h-1}\left(\alphaii{}{2} + \alphaii{}{3} \mu \sqrt{\kc} ( \cwo +\taum  r)\right)  \frac{1}{n}\sum_{i=1}^{n} \left\| \frac{\partial \ell }{\partial \Xmii{(s)}{i}(t) }  \right\|_F\\
	\leq&  \left(1+\alphaii{}{2} + \alphaii{}{3} \mu \sqrt{\kc} ( \cwo +\taum  r) \right)  \frac{1}{n}\sum_{i=1}^{n} \left\| \frac{\partial \ell }{\partial \Xmii{(l-1)}{i}(t) }  \right\|_F\\
	\leq&  \left(1+\alphaii{}{2} + \alphaii{}{3} \mu \sqrt{\kc} ( \cwo +\taum  r)\right)^{l} \frac{1}{n}\sum_{i=1}^{n} \left\| \frac{\partial \ell }{\partial \Xmii{(0)}{i}(t) }  \right\|_F\\
	\leq&  \left(1+\alphaii{}{2} + \alphaii{}{3} \mu \sqrt{\kc} ( \cwo +\taum r) \right)^{l}  \taum c_y c_u,
	\end{split}
	\end{equation*}	
	where \ding{172} uses $\alphaii{}{2}=\max_{s,l}\alphaii{(l)}{s,2}$ and $\alphaii{}{3}=\max_{s,l} \alphaii{(l)}{s,3}$. The proof is completed. 
\end{proof}

\subsection{Proof of Lemma~\ref{boundofallweights}}\label{proofofboundofallweights}

\begin{proof} Here we use mathematical induction to prove these results in turn. We first consider $t=0$. The following results hold:
	\begin{equation} \label{boundsaaasss}
	\begin{split}
	\|\Wmii{(l)}{s}(t) - \Wmii{(l)}{s}(0)\|_F  \leq \taum \sqrt{\m}  \rcc,\quad \|\Umi{s}(t) - \Umi{s}(0)\|_F  \leq \taum \sqrt{\m} \rcc.
	\end{split}
	\end{equation}	
	Now we assume \eqref{boundsaaasss} holds for $t=1,\cdots,k$. We only need to prove it hold for $t+1$.  According to the definitions, we can establish
	
	\begin{equation*} \label{boundsaaa}
	\begin{split}
	\|\Wmii{(l)}{s}(t+1) - \Wmii{(l)}{s}(t)\|_F= &\eta \alphaii{(l)}{s,3}\tau    \left\| \frac{1}{n}\sum_{i=1}^{n} \Phi(\Xmii{(s)}{i}(t))    \left(\sigma'\left( \Wmii{(l)}{s} (t)\Phi(\Xmii{(s)}{i}(t))\right) \odot \frac{\partial \ell }{\partial \Xmii{(l)}{i}(t)}\right)^{\top}  \right\|_F\\
	\leq  &\eta \alphaii{(l)}{s,3}\tau  \frac{1}{n}\sum_{i=1}^{n}  \left\|  \Phi(\Xmii{(s)}{i}(t))    \left(\sigma'\left( \Wmii{(l)}{s} (t)\Phi(\Xmii{(s)}{i}(t))\right) \odot \frac{\partial \ell }{\partial \Xmii{(l)}{i}(t)}\right)^{\top}  \right\|_F\\
	\led{172} &\eta \alphaii{(l)}{s,3}\tau  \sqrt{\kc} \frac{1}{n}\sum_{i=1}^{n} \|\Xmii{(s)}{i}(t)\| \left\|   \sigma'\left( \Wmii{(l)}{s}(t) \Phi(\Xmii{(s)}{i}(t))\right) \odot \frac{\partial \ell }{\partial \Xmii{(l)}{i}(t)}  \right\|_F\\
	\led{173} &2 \eta \alphaii{(l)}{s,3}\tau  \sqrt{\kc} \cxo \frac{1}{n}\sum_{i=1}^{n} \left\|   \sigma'\left( \Wmii{(l)}{s}(t) \Phi(\Xmii{(s)}{i}(t))\right) \odot \frac{\partial \ell }{\partial \Xmii{(l)}{i}(t)}  \right\|_F\\
	\end{split}
	\end{equation*}			
	where \ding{172} holds since $ \|\Phi(\Xmii{(s)}{})\|_F\leq \sqrt{\kc}\|\Xmii{(s)}{}\|_F$; \ding{173} holds since in Lemma~\ref{boundofmidoutput} and Lemma~\ref{boundofinitialization}, we have 
	\begin{equation}\label{adafcsadfcaswre3qwaf}
	\begin{split}\|\Xmii{(l)}{}(t)\| \leq  &
	\|\Xmii{(l)}{} (t)-\Xmii{(l)}{} (0)\|_F + \|\Xmii{(l)}{} (0)\|_F \\
	\leq &\cxo + \left(1+ \alphaii{}{2}+\alphaii{}{3}\mu\sqrt{\kc} \left( \taum \rc + \cwo\right)  \right)^{l} \taum  \mu \sqrt{\kc} \rc\\ \led{172} &2\cxo,
	\end{split}
	\end{equation}	
	where $\alphaii{}{2}=\max_{s,l}\alphaii{(l)}{s,2}$ and $\alphaii{}{3}=\max_{s,l} \alphaii{(l)}{s,3}$, and $\cxo\geq 1$ is given in Lemma~\ref{boundofinitialization}. The inequality holds by setting $\rc$ small enough, namely $\rc \leq \min(\frac{\cxo}{\left(1+ \alphaii{}{2}+2\alphaii{}{3}\mu\sqrt{\kc}  \cwo  \right)^{l}  \mu \sqrt{\kc}}, \cwo)$. This condition will be satisfied by setting enough large $m$ and will be discussed later. 
	
	Since the activation function $\sigmai{\cdot}$  is $\mu$-Lipschitz, we have 
	\begin{equation*}  
	\begin{split}
	\left\|   \sigma'\left( \Wmii{(l)}{s}(t) \Phi(\Xmii{(s)}{}(t))\right) \odot \frac{\partial \ell }{\partial \Xmii{(l)}{}(t)}  \right\|_F \leq  \mu \left\|  \frac{\partial \ell }{\partial \Xmii{(l)}{}(t)}  \right\|_F.
	\end{split}
	\end{equation*}		
	So the remaining task is to upper bound $\left\|  \frac{\partial \ell }{\partial \Xmii{(l)}{}(t)}  \right\|_F.$  Towards this goal, we have $\frac{1}{\sqrt{n}  } \left\|\um(t)-\ymm  \right\|_F \leq  c_y=\frac{1}{\sqrt{n}  }(1-\frac{\eta \lambda}{2})^{t/2} \|\ymm-\um(0)\|_2$,  $ \left\| \Umi{h}(t) \right\|_F \leq  \left\| \Umi{h}(t) - \Umi{h}(0)  \right\|_F +  \left\| \Umi{h}(0) \right\|_F \leq  c_u=\sqrt{m} (\taum \rcc + \cwo)$, $ \|\Wmii{(s)}{l}(t) - \Wmii{(s)}{l}(0)\|_F \leq \taum \sqrt{m} r$, and $ \|\Wmii{(s)}{l}(0)\|_F  \leq\cwo$. In this way, we can use Lemma Lemma~\ref{boundgradient} and obtain
	\begin{equation*}
	\begin{split}
	\frac{1}{n}\sum_{i=1}^{n} \left\|\frac{\partial \ell }{\partial \Xmii{(l)}{i}(t)}\right\|_F 
	\leq  c_1  \taum c_y c_u= \frac{c_1 \taum ( \taum \rcc + \cwo)}{\sqrt{n}} \left(1-\frac{\eta \lambda}{2}\right)^{t/2} \|\ymm-\um(0)\|_2,
	\end{split}
	\end{equation*}	
	where $c_1= \left(1+\alphaii{}{2} + \alphaii{}{3}\tau \mu \sqrt{\kc}(\taum \rcc +\cwo) \right)^{l}$ with $\alphaii{}{2}=\max_{s,l}\alphaii{(l)}{s,2}$ and $\alphaii{}{3}=\max_{s,l} \alphaii{(l)}{s,3}$.  
	
	By combining the above results, we can directly obtain
	\begin{equation*}  
	\begin{split}
	\|\Wmii{(l)}{s}(t+1) - \Wmii{(l)}{s}(t)\|_F\leq& \frac{2c_1\taum \eta \alphaii{(l)}{s,3} \mu \sqrt{\kc} \cxo (\taum \rcc + \cwo) }{\sqrt{n}}\left\|\um(t)-\ymm  \right\|_F\\	 
	\leq & \frac{2 c_1 \taum \eta \alphaii{(l)}{s,3} \mu \sqrt{\kc} \cxo  (\taum \rcc + \cwo)}{\sqrt{n}} \left(1-\frac{\eta \lambda}{2}\right)^{t/2} \|\ymm-\um(0)\|_2.
	\end{split}
	\end{equation*}			
	Therefore, we have 
	\begin{equation*}  
	\begin{split}
	\|\Wmii{(l)}{s}(t+1) - \Wmii{(l)}{s}(0)\|_F\leq &  \|\Wmii{(l)}{s}(t+1) - \Wmii{(l)}{s}(t)\|_F + \|\Wmii{(l)}{s}(t) - \Wmii{(l)}{s}(0)\|_F \\
	\leq & \frac{8 c_1  \taum  \alphaii{(l)}{s,3} \mu \sqrt{\kc} \cxo  (\taum \rcc + \cwo)  }{\lambda \sqrt{n}} \|\ymm-\um(0)\|_2\led{172} \sqrt{m} \taum \rcc,
	\end{split}
	\end{equation*}		
	where \ding{172} holds by setting   $\rcc=\frac{16 \left(1+\alphaii{}{2} + 2\alphaii{}{3} \mu \sqrt{\kc} \cwo \right)^{l} \alphaii{(l)}{s,3} \mu \sqrt{\kc} \cxo \cwo  }{\lambda \sqrt{mn}} \|\ymm-\um(0)\|_2 \leq \cwo$. 
	By using the same way, we can prove 
	\begin{equation*}  
	\begin{split}
	&\|\Wmii{(0)}{}(t+1) - \Wmii{(0)}{}(t)\|_F\leq \frac{2 c_1  \taum \eta   \mu \sqrt{\kc} \cxo  (\rcc + \cwo)}{\sqrt{n}} \left(1-\frac{\eta \lambda}{2}\right)^{t/2} \|\ymm-\um(0)\|_2,\\
	&\|\Wmii{(l)}{s}(t+1) - \Wmii{(l)}{s}(0)\|_F\leq   \sqrt{m} \taum \rcc.
	\end{split}
	\end{equation*}

	Then similarly, we can obtain
	\begin{equation*} \label{boundsaaa}
	\begin{split}
	\|\Umi{s}(t+1) - \Umi{s}(t)\|_F= &\eta  \taum  \left\| \frac{1}{n}\sum_{i=1}^{n}  (\ui{i} -\ymi{i})  \Xmii{(s)}{i}(t) \right\|_F 
	\leq \eta  \taum \frac{1}{n}\sum_{i=1}^{n} |\ui{i}(t) -\ymi{i}|  \left\|  \Xmii{(s)}{i}(t) \right\|_F\\ \led{172} & \frac{2 \taum \eta \cxo}{ \sqrt{n}} \|\um(t)-\ymm\|_2 \leq  \frac{2 \taum \eta \cxo}{ \sqrt{n}}\left(1-\frac{\eta \lambda}{2}\right)^{t/2} \|\ymm-\um(0)\|_2,
	\end{split}
	\end{equation*}			
	where  \ding{172} holds since $ \sum_{i=1}^{n}|\ui{i} -\ymi{i}|  \leq \sqrt{n} \|\um-\ymm\|_2$, and $\left\|  \Xmii{(s)}{i}(t) \right\|_F\leq 2\cxo$ in \eqref{adafcsadfcaswre3qwaf}. Then we establish
	\begin{equation*}  
	\begin{split}
	\|\Umi{s}(t+1) - \Umi{s}(0)\|_F\leq &  \|\Umi{s}(t+1) - \Umi{s}(t)\|_F + \|\Umi{s}(t) - \Umi{s}(0)\|_F \\
	\leq & \frac{ 8 \taum  \cxo \|\ymm-\um(0)\|_2}{\lambda \sqrt{n} } \led{172} \sqrt{m} \taum \rcc,
	\end{split}
	\end{equation*}	
	where \ding{172} holds by setting $\rcc=\frac{ 8  \cxo   \|\ymm-\um(0)\|_2}{\lambda  \sqrt{m n}} $. Finally, combining the value of $\rcc$, we have $\rcc=\max\left(\frac{ 8  \cxo\|\ymm-\um(0)\|_2}{\lambda  \sqrt{mn}}, \frac{16 \left(1+\alphaii{}{2} + 2\alphaii{}{3} \mu \sqrt{\kc} \cwo \right)^{l} \alphaii{(l)}{s,3} \mu \sqrt{\kc} \cxo \cwo }{\lambda \sqrt{mn}} \|\ymm-\um(0)\|_2 \right) \leq \cwo$. Under this setting, we have
	\begin{equation*}  
	\begin{split}
	\|\Wmii{(l)}{s}(t+1) - \Wmii{(l)}{s}(t)\|_F\leq & \frac{4 c \taum  \eta \alphaii{(l)}{s,3} \mu   \cxo    \cwo \sqrt{\kc }}{\sqrt{n}}\left\|\um(t)-\ymm  \right\|_F\\
	\leq& \frac{4 c \taum \eta \alphaii{(l)}{s,3} \mu   \cxo    \cwo \sqrt{\kc }}{\sqrt{n}} \left(1-\frac{\eta \lambda}{2}\right)^{t/2} \|\ymm-\um(0)\|_2,\\
	\|\Wmii{(0)}{}(t+1) - \Wmii{(0)}{}(t)\|_F\leq & \frac{4 c \taum \eta  \mu   \cxo    \cwo \sqrt{\kc }}{\sqrt{n}}\left\|\um(t)-\ymm  \right\|_F\\
	\leq& \frac{4 c  \taum \eta \mu   \cxo    \cwo \sqrt{\kc }}{\sqrt{n}} \left(1-\frac{\eta \lambda}{2}\right)^{t/2} \|\ymm-\um(0)\|_2,
	\end{split}
	\end{equation*}		
	where $c= \left(1+\alphaii{}{2} + 2\alphaii{}{3} \mu \sqrt{\kc} \cwo  \right)^{l}$ with $\alphaii{}{2}=\max_{s,l}\alphaii{(l)}{s,2}$ and $\alphaii{}{3}=\max_{s,l} \alphaii{(l)}{s,3}$.  
	The proof is completed. 
\end{proof}

\subsection{Proof of Lemma~\ref{boundedtempoutpsss}}\label{proofofboundedtempoutpsss}
 
\begin{proof}
	We use mathematical induction to prove the results. We first consider $h=0$. According to the definition, we have 
	\begin{equation*}
	\begin{split}
	\left\|\Xmii{(0)}{}(k+1) - \Xmii{(0)}{}(k)\right\|_F  =&  \tau \left\|\sigmai{\Wmii{(0)}{}(k+1) \Phi(\Xm)} - \sigmai{\Wmii{(0)}{}(k) \Phi(\Xm)}\right\|_F\\
	\leq &  \tau \mu \left\| \Wmii{(0)}{}(k+1) -  \Wmii{(0)}{}(k) \right\|_F \| \Phi(\Xm) \|_F\\
	\led{172} &  \tau \taum  \mu \sqrt{\kc} \left\| \Wmii{(0)}{}(k+1) -  \Wmii{(0)}{}(k) \right\|_F  \\
	\led{173} & \frac{  4 c  \taum   \tau \eta  \mu^2   \cxo    \cwo  \kc }{\sqrt{n}}\left\|\um(k)-\ymm  \right\|_F,
	\end{split}
	\end{equation*}
	where \ding{172} uses $\| \Phi(\Xm) \|_F\leq \sqrt{\kc}\|\Xm\|_F\leq \sqrt{\kc}$ where the sample $\Xm$ obeys $\|\Xm\|_F=1$; \ding{173} uses the result in Lemma~\ref{boundofallweights} that $	\|\Wmii{(0)}{}(t+1) - \Wmii{(0)}{}(t)\|_F \leq  \frac{4 c \taum  \eta  \mu   \cxo    \cwo \sqrt{\kc }}{\sqrt{n}}\left\|\um(t)-\ymm  \right\|_F$. 
	
	Then we first consider $h\geq 1$.  
	\begin{equation*}
	\begin{split}
	&\left\|\Xmii{(l)}{}(k+1) - \Xmii{(l)}{}(k)\right\|_F \\
	=& \!   \left\|\sum_{s=0}^{l-1} \!\!  \left(\alphaii{(l)}{s,2}  (\Xmii{(s)}{}(k\!+\!1) \!-\! \Xmii{(s)}{}(k))   \!+\! \alphaii{(l)}{s,3} \tau \left( \sigmai{\Wmii{(l)}{s}(k\!+\!1)  \Phi(\Xmii{(s)}{}(k\!+\!1) )} \!-\!\sigmai{\Wmii{(l)}{s}(k)  \Phi(\Xmii{(s)}{}(k) )} \right)\right) \right\|_F\\
	\leq & \!\sum_{s=0}^{l-1}\!\left[    \alphaii{(l)}{s,2} \left\| \Xmii{(s)}{}(k\!+\!1)\!-\! \Xmii{(s)}{}(k))\right\|_F  \! +\! \alphaii{(l)}{s,3} \tau  \left\| \sigmai{\Wmii{(l)}{s}(k\!+\!1)  \Phi(\Xmii{(s)}{}(k\!+\!1) )}\! -\!\sigmai{\Wmii{(l)}{s}(k)  \Phi(\Xmii{(s)}{}(k) )}   \right\|_F \right]\\
	\leq &\! \sum_{s=0}^{l-1}\!\left[    \alphaii{(l)}{s,2} \left\| \Xmii{(s)}{}(k\!+\!1)\!-\! \Xmii{(s)}{}(k))\right\|_F  \! +\! \alphaii{(l)}{s,3} \tau  \mu\left\|  \Wmii{(l)}{s}(k\!+\!1)\!  \Phi(\Xmii{(s)}{}(k\!+\!1) )\! -\! \Wmii{(l)}{s}(k)  \Phi(\Xmii{(s)}{}(k) )   \right\|_F \right]\\\!
	\end{split}
	\end{equation*}
	Then we bound the second term carefully:
	\begin{equation*}
	\begin{split}
	&\left\|  \Wmii{(l)}{s}(k+1)  \Phi(\Xmii{(s)}{}(k+1) ) - \Wmii{(l)}{s}(k)  \Phi(\Xmii{(s)}{}(k) )   \right\|_F  \\
	=&\left\|  \Wmii{(l)}{s}(k+1)  (\Phi(\Xmii{(s)}{}(k+1) ) - \Phi(\Xmii{(s)}{}(k) ) ) \right\|_F  + \left\|  (\Wmii{(l)}{s}(k+1)   - \Wmii{(l)}{s}(k) ) \Phi(\Xmii{(s)}{}(k) )   \right\|_F\\
	\leq & \sqrt{\kc}\left\|  \Wmii{(l)}{s}(k+1) \right\|_F \left\|\Xmii{(s)}{}(k+1) - \Xmii{(s)}{}(k) ) \right\|_F  + \sqrt{\kc}\left\|   \Wmii{(l)}{s}(k+1)   - \Wmii{(l)}{s}(k) \right\|_F \left\| \Xmii{(s)}{}(k)   \right\|_F\\
	\end{split}
	\end{equation*}
	By using Lemma~\ref{boundofmidoutput} and Lemma~\ref{boundofinitialization}, we have 
	\begin{equation*}\label{adafcsadfcaswre3qwaf}
	\begin{split}\|\Xmii{(s)}{}(k)\| \leq  &
	\|\Xmii{(l)}{i} (k)-\Xmii{(l)}{i} (0)\|_F + \|\Xmii{(l)}{i} (0)\|_F \\
	\leq &\cxo + \left(1+ \alphaii{}{2}+\alphaii{}{3}\mu\sqrt{\kc} \left( \rcc + \cwo\right)  \right)^{l}  \mu \sqrt{\kc} \rcc \led{172} 2\cxo,
	\end{split}
	\end{equation*}	
	where $\alphaii{}{2}=\max_{s,l}\alphaii{(l)}{s,2}$ and $\alphaii{}{3}=\max_{s,l} \alphaii{(l)}{s,3}$, and $\cxo\geq 1$ is given in Lemma~\ref{boundofinitialization}. \ding{172} holds since in Lemma~\ref{boundofallweights}, we set $m$ large enough such that $\rcc$ is enough small. 
	
	Besides,   Lemma~\ref{adafcsadfcaswre3qwaf} shows that  $$\|\Wmii{(l)}{s}(k+1) - \Wmii{(l)}{s}(k)\|_F \leq  \frac{4 c \taum  \eta \alphaii{(l)}{s,3} \mu   \cxo    \cwo \sqrt{\kc }}{\sqrt{n}}\left\|\um(k)-\ymm  \right\|_F,$$ 
	where $c= \left(1+\alphaii{}{2} + 2\alphaii{}{3} \mu \sqrt{\kc} \cwo  \right)^{l}$ with $\alphaii{}{2}=\max_{s,l}\alphaii{(l)}{s,2}$ and $\alphaii{}{3}=\max_{s,l} \alphaii{(l)}{s,3}$.   
	Combing all results yields
	\begin{equation*}
	\begin{split}
	&\left\|  \Wmii{(l)}{s}(k+1)  \Phi(\Xmii{(s)}{}(k+1) ) - \Wmii{(l)}{s}(k)  \Phi(\Xmii{(s)}{}(k) )   \right\|_F  \\
	\leq & 2\sqrt{\kc m} \cwo  \left\|\Xmii{(s)}{}(k+1) - \Xmii{(s)}{}(k) ) \right\|_F  +  \frac{8 c\taum  \eta \alphaii{(l)}{s,3} \mu   \cxo^2    \cwo \kc }{\sqrt{n}}\left\|\um(k)-\ymm  \right\|_F.
	\end{split}
	\end{equation*}
	
	Thus, we can further obtain
	\begin{equation*}
	\begin{split}
	&\left\|\Xmii{(l)}{}(k+1) - \Xmii{(l)}{}(k)\right\|_F \\
	\leq & \sum_{s=0}^{l-1}\left[    (\alphaii{(l)}{s,2} +2\sqrt{\kc } \cwo   \alphaii{(l)}{s,3}   \mu ) \left\| \Xmii{(s)}{}(k+1)- \Xmii{(s)}{}(k))\right\|_F   +     \frac{8 \tau \taum  c \eta (\alphaii{(l)}{s,3})^2 \mu^2   \cxo^2    \cwo \kc }{\sqrt{n}}\left\|\um(k)-\ymm  \right\|_F\right]\\
	\led{172} & \sum_{s=0}^{l-1}\left[    (\alphaii{}{2} +2\sqrt{\kc } \cwo   \alphaii{}{3}   \mu ) \left\| \Xmii{(s)}{}(k+1)- \Xmii{(s)}{}(k))\right\|_F   +     \frac{8 \tau\taum   c \eta (\alphaii{}{3})^2 \mu^2   \cxo^2    \cwo \kc }{\sqrt{n}}\left\|\um(k)-\ymm  \right\|_F\right]\\
	\leq &   \left(1+\alphaii{}{2} +2\sqrt{\kc } \cwo   \alphaii{}{3}  \mu \right)^l \left( \left\| \Xmii{(0)}{}(k+1)- \Xmii{(0)}{}(k))\right\|_F   +     \frac{8 \tau \taum  c \eta (\alphaii{}{3})^2 \mu^2   \cxo^2    \cwo \kc }{(\alphaii{}{2} +2\sqrt{\kc } \cwo   \alphaii{}{3} \mu )\sqrt{n}}\left\|\um(k)-\ymm  \right\|_F \right) \\
	\leq & \left(1+\alphaii{}{2} +2\sqrt{\kc } \cwo  \alphaii{}{3}   \mu \right)^l \left( \frac{  4 c   \tau \eta  \mu^2   \cxo    \cwo  \kc }{\sqrt{n}}   +     \frac{8 \tau \taum  c \eta (\alphaii{}{3})^2 \mu^2   \cxo^2    \cwo \kc }{(\alphaii{}{2} +2\sqrt{\kc } \cwo   \alphaii{}{3} \mu )\sqrt{n}}  \right) \left\|\um(k)-\ymm  \right\|_F\\
	\leq&  \left(1+\alphaii{}{2} +2\sqrt{\kc } \cwo   \alphaii{}{3}   \mu \right)^l \left(1 +     \frac{2  (\alphaii{}{3})^2     \cxo  }{(\alphaii{}{2} +2\sqrt{\kc } \cwo   \alphaii{}{3}   \mu )\sqrt{n}}  \right)  \frac{  4 c  \taum   \tau \eta  \mu^2   \cxo    \cwo  \kc }{\sqrt{n}}    \left\|\um(k)-\ymm  \right\|_F.
	\end{split}
	\end{equation*}
	The proof is completed. 
\end{proof}

\subsection{Proof of Lemma~\ref{boundedtempoutpsss222}}\label{proofofboundedtempoutpsss222}

\begin{proof}
	In Lemma~\ref{boundofallweights}, we have show  
	\begin{equation} \label{conditionasdadada}
	\begin{split}
	\max\left(\|\Wmii{(0)}{}(t) - \Wmii{(0)}{}(0)\|_F , \|\Wmii{(l)}{s}(t) - \Wmii{(l)}{s}(0)\|_F , \|\Umi{s}(t) - \Umi{s}(0)\|_F \right) \leq \sqrt{\m} \taum \rcc\leq \sqrt{\m} \taum \cwo.
	\end{split}
	\end{equation}	
	Note $\taum =\frac{1}{\sqrt{m}}$.	In this way, from Lemma~\ref{boundofallweights}, we have 
	\begin{equation*}  
	\begin{split}
	&	\left\| \Wmii{(0)}{}(t)\right\|_F \leq  \left\| \Wmii{(0)}{}(t)-   \Wmii{(0)}{}(0) \right\|_F  +  \left\| \Wmii{(0)}{}(0) \right\|_F\leq 2\sqrt{m} \cwo ,	\\
	&	\left\| \Wmii{(l)}{s}(t)\right\|_F \leq  \left\| \Wmii{(l)}{s}(t)-   \Wmii{(l)}{s}(0) \right\|_F  +  \left\| \Wmii{(l)}{s}(0) \right\|_F\leq 2\sqrt{m} \cwo ,	\\
	& \left\| \Umi{h}(t) \right\|_F \leq  \left\| \Umi{h}(t) -   \Umi{h}(0)  \right\|_F  +  \left\| \Umi{h}(0) \right\|_F\leq 2\sqrt{m} \cwo 	
	\end{split}
	\end{equation*}	
	In Lemma~\ref{boundofinitialization}, we show that when Eqn.~\eqref{conditionasdadada} holds which is proven in Lemma~\ref{boundofallweights}, then $\|\Xmii{(l)}{i} (0)\|_F\leq \cxo$.  Under Eqn.~\eqref{boundofinitialization}, Lemma~\ref{boundofmidoutput}  shows 
	\begin{equation*} 
	\|\Xmii{(l)}{i} (k)-\Xmii{(l)}{i} (0)\|_F  
	\leq  \left(1+ \alphaii{}{2}+2\alphaii{}{3}\mu\sqrt{\kc} \cwo   \right)^{l}  \mu \sqrt{\kc} \taum \rcc \led{172}  \taum \cxo,
	\end{equation*}
	where \ding{172} holds since in Lemma~\ref{boundofallweights}, we set $m=\Oc{\frac{\kc^2 \cwo^2 \|\ymm-\um(0)\|_2^2}{\lambda^2   n} \left(1+ \alphaii{}{2}+2\alphaii{}{3}\mu\sqrt{\kc} \cwo   \right)^{4h} }$  such that 
	\begin{equation*}
	\begin{split}
	\rcc=& \frac{8\cxo\|\ymm-\um(0)\|_2}{\lambda  \sqrt{mn}} \max\left(1,  2 \left(1+\alphaii{}{2} + 2\alphaii{}{3} \mu \sqrt{\kc} \cwo \right)^{l} \alphaii{(l)}{s,3} \mu \sqrt{\kc} \cwo    \right) \\
	\leq& \frac{\cxo}{\left(1+ \alphaii{}{2}+2\alphaii{}{3}\mu\sqrt{\kc} \cwo   \right)^{l}  \mu \sqrt{\kc} }.
	\end{split}
	\end{equation*}	
	Therefore, we have 
	\begin{equation*}  
	\begin{split}
	\left\| \Xmii{(l)}{i} (k)\right\|_F \leq  \|\Xmii{(l)}{i} (k)-\Xmii{(l)}{i} (0)\|_F    +  \left\| \Xmii{(l)}{i} (0) \right\|_F\leq 2\cxo. 
	\end{split}
	\end{equation*}	
	The proof is completed. 
\end{proof}

\subsection{Proof of Lemma~\ref{boundX}} \label{proofpfboundX}
 
\begin{proof}
	We first consider $l=0$. Specifically, we have 
	\begin{equation*} 
	\begin{split}
	\| \Xmii{(0)}{i}(k)  - \Xmii{(0)}{i}(0) \|_F= & \tau \left\| \sigmai{\Wmii{(0)}{}(k) \Phi(\Xmi{i})} - \sigmai{\Wmii{(0)}{} (0)\Phi(\Xmi{i})}\right\|_F\\
	\leq  & \tau\mu \left\|  \Wmii{(0)}{}(k)  -  \Wmii{(0)}{} (0) \right\|_F \| \Phi(\Xmi{i})\|_F\\
	\led{172}  &\taum \tau\mu  \sqrt{\kc}\left\|  \Wmii{(0)}{}(k)  -  \Wmii{(0)}{} (0) \right\|_F \\
	\led{173} & \taum\mu \sqrt{\kc }\rcc,
	\end{split}
	\end{equation*}
	where \ding{172} holds since $\| \Phi(\Xmi{i})\|_F\leq\sqrt{\kc} \|\Xmi{i}\|_F\leq \sqrt{\kc}$ and the results in Lemma~\ref{boundofallweights} that $\left\|  \Wmii{(0)}{}(k)  -  \Wmii{(0)}{} (0) \right\|_F\leq \sqrt{m} \taum \rcc$. 
	
	Then we consider $l\geq 1$. According to the definition, we have 
	\begin{equation*} 
	\begin{split}
	&\| \Xmii{(l)}{i}(k)  - \Xmii{(l)}{i}(0) \|_F\\
	= &   \left\| \sum_{s=0}^{l-1}   \left(\alphaii{(l)}{s,2}  (\Xmii{(s)}{i}(k) -\Xmii{(s)}{i}(0) ) + \alphaii{(l)}{s,3} \tau \left( \sigmai{\Wmii{(l)}{s}(k) \Phi(\Xmii{(s)}{i}(k))}- \sigmai{\Wmii{(l)}{s}(0) \Phi(\Xmii{(s)}{i}(0))} \right)\right) \right\|_F\\
	\leq  &   \sum_{s=0}^{l-1} \left[\alphaii{(l)}{s,2} \left\| \Xmii{(s)}{i}(k) -\Xmii{(s)}{i}(0) \right\|_F + \alphaii{(l)}{s,3} \tau \left\|\sigmai{\Wmii{(l)}{s}(k) \Phi(\Xmii{(s)}{i}(k))}- \sigmai{\Wmii{(l)}{s}(0) \Phi(\Xmii{(s)}{i}(0))} \right\|_F\right]\\
	\leq  &   \sum_{s=0}^{l-1} \left[\alphaii{(l)}{s,2} \left\| \Xmii{(s)}{i}(k) -\Xmii{(s)}{i}(0) \right\|_F + \alphaii{(l)}{s,3} \tau \mu \left\| \Wmii{(l)}{s}(k) \Phi(\Xmii{(s)}{i}(k))- \Wmii{(l)}{s}(0) \Phi(\Xmii{(s)}{i}(0)) \right\|_F\right].
	\end{split}
	\end{equation*}
	Then we bound 
	\begin{equation*} 
	\begin{split}
	& \left\| \Wmii{(l)}{s}(k) \Phi(\Xmii{(s)}{i}(k))- \Wmii{(l)}{s}(0) \Phi(\Xmii{(s)}{i}(0))  \right\|_F\\
	\leq  &  \left\| (\Wmii{(l)}{s}(k) -\Wmii{(l)}{s}(0)) \Phi(\Xmii{(s)}{i}(k)) \right\|_F + \left\| \Wmii{(l)}{s}(0) (\Phi(\Xmii{(s)}{i}(k))  -\Phi(\Xmii{(s)}{i}(0)) ) \right\|_F \\
	\leq &  \left\| \Wmii{(l)}{s}(k) -\Wmii{(l)}{s}(0) \right\|_F \left\| \Phi(\Xmii{(s)}{i}(k)) \right\|_F + \left\| \Wmii{(l)}{s}(0) \right\|_F \left\|\Phi(\Xmii{(s)}{i}(k))  -\Phi(\Xmii{(s)}{i}(0)) \right\|_F \\
	\led{172} &  2\sqrt{\kc m} \cxo\taum \rcc   + 2\sqrt{\kc m}\cwo  \left\|\Xmii{(s)}{i}(k) -\Xmii{(s)}{i}(0) \right\|_F,
	\end{split}
	\end{equation*}
	where \ding{172} holds since   Lemma~\ref{boundofallweights} shows  $\left\|  \Wmii{(0)}{}(k)  -  \Wmii{(0)}{} (0) \right\|_F\leq \sqrt{m} \taum \rcc$ and Lemma~\ref{boundedtempoutpsss222} shows $\left\| \Xmii{(s)}{i}(k) \right\|_F\leq 2\cxo$ and $\left\| \Wmii{(l)}{s}(0) \right\|_F \leq 2\sqrt{m}\cwo$.
	
	In this way, we have 
	\begin{equation*} 
	\begin{split}
	&\| \Xmii{(l)}{i}(k)  - \Xmii{(l)}{i}(0) \|_F\\
	\leq  &   \sum_{s=0}^{l-1} \left[ \left(\alphaii{(l)}{s,2} +  2 \alphaii{(l)}{s,3}  \mu    \sqrt{\kc}\cwo\right) \left\| \Xmii{(s)}{i}(k) -\Xmii{(s)}{i}(0) \right\|_F  +2 \alphaii{(l)}{s,3}   \mu \sqrt{\kc } \cxo \taum \rcc  \right]\\
	\led{172}  &   \sum_{s=0}^{l-1} \left[ \left(\alphai{2} +  2 \alphai{3}   \mu    \sqrt{\kc}\cwo\right) \left\| \Xmii{(s)}{i}(k) -\Xmii{(s)}{i}(0) \right\|_F  +2 \alphai{3}   \mu \sqrt{\kc } \cxo\taum \rcc  \right]\\
	\led{173}  &    c \left[   \left\| \Xmii{(0)}{i}(k) -\Xmii{(s)}{i}(0) \right\|_F  +2 \alphai{3}   \mu \sqrt{\kc } \cxo\taum \rcc  \right]\\
	= &    c ( 1 +2 \alphai{3}     \cxo   ) \mu \sqrt{\kc } \taum\rcc\\
	\end{split}
	\end{equation*}
	where \ding{172} and \ding{173} hold by using $c= \left(1+\alphaii{}{2} + 2\alphaii{}{3} \mu \sqrt{\kc} \cwo  \right)^{l}$ with $\alphaii{}{2}=\max_{s,l}\alphaii{(l)}{s,2}$ and $\alphaii{}{3}=\max_{s,l} \alphaii{(l)}{s,3}$.  The proof is completed. 
\end{proof}

\subsection{Proof of Lemma~\ref{boundfinaloutput}} \label{proofofboundfinaloutput}

\begin{proof}
	For this proof, we need to use the results in other lemmas. Specifically, Lemma~\ref{boundofallweights}
	\begin{equation} \label{adasfcsac}
	\begin{split}
	&\|\Wmii{(0)}{}(t) - \Wmii{(0)}{}(0)\|_F  \leq \sqrt{\m} \taum \rcc,\ \|\Wmii{(l)}{s}(t) - \Wmii{(l)}{s}(0)\|_F  \leq \sqrt{\m} \taum \rcc,\ \|\Umi{s}(t) - \Umi{s}(0)\|_F  \leq \sqrt{\m} \taum \rcc,
	\end{split}
	\end{equation}	
	where $c= \left(1+\alphaii{}{2} + 2\alphaii{}{3} \mu \sqrt{\kc} \cwo  \right)^{l}$ with $\alphaii{}{2}=\max_{s,l}\alphaii{(l)}{s,2}$ and $\alphaii{}{3}=\max_{s,l} \alphaii{(l)}{s,3}$. Based on this,  	Lemma~\ref{boundedtempoutpsss222} further shows  
	\begin{equation}\label{vfdsgveswfgewrgf}
	\begin{split}
	\left\|\Wmii{(0)}{}(k)\right\|_F  \leq 2\sqrt{m} \cwo, \ \	\left\|\Wmii{(l)}{s}(k)\right\|_F  \leq2\sqrt{m} \cwo, \ \ \left\|\Umi{s}(k)\right\|_F  \leq 2\sqrt{m} \cwo, \ \  \left\|\Xmii{(l)}{i}(k)\right\|_F  \leq 2\cxo.
	\end{split}
	\end{equation}
	Next, Lemma~\ref{boundX} also proves $$
	\| \Xmii{(l)}{i}(k)  - \Xmii{(l)}{i}(0) \|_F
	\leq      c ( 1 +2 \alphai{3}     \cxo   ) \mu \sqrt{\kc } \taum \rcc.
	$$
	
	Then we can easily obtain our result: 
	\begin{equation*}
	\begin{split}
	|   \ui{i} (k) - \ui{i}(0)| =& \left|  \sum_{s=1}^{h}  \langle \Umi{s}(k), \Xmii{(l)}{i}(k) \rangle -\langle \Umi{s}(0), \Xmii{(l)}{i}(0) \rangle  \right|\\
	\leq & \sum_{s=1}^{h}  \left|  \langle \Umi{s}(k) - \Umi{s}(0), \Xmii{(l)}{i}(k) \rangle +\langle \Umi{s}(0), \Xmii{(l)}{i}(k)  - \Xmii{(l)}{i}(0) \rangle  \right|\\
	\leq & \sum_{s=1}^{h} 2 \sqrt{m}\rcc \cxo + 2\sqrt{m}\cwo c ( 1 +2 \alphai{3}     \cxo   ) \mu \sqrt{\kc }\rcc \\
	=&   2 \sqrt{m} h  \left( \cxo +\cwo c ( 1 +2 \alphai{3}     \cxo   ) \mu \sqrt{\kc } \right) \rcc.
	\end{split}
	\end{equation*}
	
	Then we look at the second part. We first look at $l=h$:
	\begin{equation}\label{afdsarweqfdcsa}
	\begin{split}
	\left\| \frac{\partial \ell }{\partial \Xmii{(l)}{i}(k)} - \frac{\partial \ell }{\partial \Xmii{(l)}{i}(0)}   \right\|_F
	=&\taum \left\| (\ui{i}(k) -\ymi{i} ) \Umi{l}(k)   - (\ui{i}(0) -\ymi{i} ) \Umi{l}(0)   \right\|_F  \\
	= & \taum |\ui{i}(k) -\ymi{i} | \left\|   \Umi{l}(k) \right\|_F  +\taum  |\ui{i}(0) -\ymi{i}| \left\| \Umi{l}(0)   \right\|_F  \\
	\leq &\taum   \left\| (\ui{i}(k) -\ui{i}(0)) \Umi{l}(k) \right\|_F  +\taum  \left\|(\ui{i}(0) -\ymi{i} ) (\Umi{l}(k)-\Umi{l}(0))   \right\|_F  \\
	\leq & \taum  |\ui{i}(k) -\ui{i}(0)| \left\| \Umi{l}(k) \right\|_F  +\taum  |\ui{i}(0) -\ymi{i} | \left\| (\Umi{l}(k)-\Umi{l}(0))   \right\|_F  \\
	\leq &  4\sqrt{m} \taum \rcc \left( \cwo    \sqrt{m} h  \left( \cxo +\cwo c ( 1 +2 \alphai{3}     \cxo   ) \mu \sqrt{\kc } \right)    + |\ui{i}(0) -\ymi{i} |  \right).
	\end{split}
	\end{equation} 
	Then we consider $l<h$. According to the definitions in Lemma~\ref{gradientcomputation}, we have  	
	\begin{equation*}
	\begin{split}
	\frac{\partial \ell }{\partial \Xmii{(l)}{}}
	= \taum  (\ui{} -\ym ) \Umi{l} +\sum_{s=l+1}^{h}\left(\alphaii{(s)}{l,2}\frac{\partial \ell }{\partial \Xmii{(s)}{}}  + \alphaii{(s)}{l,3}\tau   \convb{(\Wmii{(s)}{l})^{\top} \left(\sigma'\left( \Wmii{(s)}{l} \Phi(\Xmii{(l)}{})\right) \odot \frac{\partial \ell }{\partial \Xmii{(s)}{}}\right)}  \right).
	\end{split}
	\end{equation*}
	In this way, we can upper bound 
	\begin{equation*}
	\begin{split}
	& \left\| \frac{\partial \ell }{\partial \Xmii{(l)}{i}(k)} - \frac{\partial \ell }{\partial \Xmii{(l)}{i}(0)}   \right\|_F\\
	=&\taum \left\|(\ui{i}(k) \!-\!\ymi{i} ) \Umi{l}(k) \!-\!(\ui{i}(0) \!-\!\ymi{i} ) \Umi{l} (0) \right\|_F  \!+\!\sum_{s=l+1}^{h} \! \alphaii{(s)}{l,2} \left\| \frac{\partial \ell }{\partial \Xmii{(s)}{i}(k) } \!-\! \frac{\partial \ell }{\partial \Xmii{(s)}{i}(k) } \right\|_F \!+\! \sum_{s=l+1}^{h}\! \alphaii{(s)}{l,3}\tau \sqrt{\kc} D,
	\end{split}
	\end{equation*} 	
	where $D=\left\| \Am_k^{\top} (\Bm_k\odot \Cm_k) - \Am_0^{\top} (\Bm_0\odot \Cm_0) \right\|_F$ in which  $ \Am_k=\Wmii{(s)}{l}(k), \Bm_k= \sigma'\left( \Wmii{(s)}{l}(k)  \Phi(\Xmii{(l)}{i}(k))\right), \Cm_k=  \frac{\partial \ell }{\partial \Xmii{(s)}{i}(k) }$.  Similar to Eqn.~\eqref{afdsarweqfdcsa}, we have 
	\begin{equation*}
	\begin{split}
	& \left\| (\ui{i}(k) -\ymi{i} ) \Umi{l}(k)   - (\ui{i}(0) -\ymi{i} ) \Umi{l}(0)   \right\|_F  \\
	&\qquad \qquad \qquad\qquad \leq   4\sqrt{m} \rcc \left( \cwo    \sqrt{m} h  \left( \cxo +\cwo c ( 1 +2 \alphai{3}     \cxo   ) \mu \sqrt{\kc } \right)    + |\ui{i}(0) -\ymi{i} |  \right).
	\end{split}
	\end{equation*} 
	Then, we can bound $D$ as follows:
	\begin{equation*}
	\begin{split}
	D= &\left\|  (\Am_k - \Am_0)^{\top} (\Bm_0\odot \Cm_0) \right\|_F + \left\|  \Am_k^{\top} (\Bm_k\odot \Cm_k-\Bm_0\odot \Cm_0) \right\|_F\\
	\leq & \| \Am_k - \Am_0\|_F \| \Bm_0\odot \Cm_0  \|_F +\|\Am_k\|_F \|\Bm_k\odot \Cm_k-\Bm_0\odot \Cm_0 \|_F\\
	\led{172} & \mu \sqrt{m}\rcc  \|  \Cm_{0}  \|_{2} +2\sqrt{m}\cwo \|\Bm_k\odot \Cm_k-\Bm_0\odot \Cm_0 \|_F\\
	\end{split}
	\end{equation*} 
	where \ding{172} uses the results in Eqns.~\eqref{vfdsgveswfgewrgf} and \eqref{adasfcsac}.  The remaining work is to bound 
	\begin{equation*}
	\begin{split}
	\|\Bm_k\odot \Cm_k-\Bm_0\odot \Cm_0 \|_F = &  \|\Bm_k\odot (\Cm_k-\Cm_0) \|_F + \|(\Bm_k-\Bm_0)\odot \Cm_0  \|_F \\
	\leq  & \mu \| \Cm_k-\Cm_0 \|_F + \rho \left\| \Wmii{(s)}{l}(k)  \Phi(\Xmii{(l)}{i}(k))-\Wmii{(s)}{l}(0)  \Phi(\Xmii{(l)}{i}(0)) \right\|_F \|  \Cm_0  \|_{\infty} \\
	\end{split}
	\end{equation*} 
	where \ding{172} uses the assumption that the activation function $\sigmai{\cdot}$  is $\mu$-Lipschitz  and $\rho$-smooth. Note $\|  \Cm_0  \|_\infty$ is a constant, since it is the gradient norm at the initialization which does not involves the algorithm updating.   Recall Lemma~\ref{boundofmidoutput} shows  
	\begin{equation*}
	\begin{split} 
	&\left\| \Wmii{(l)}{s}(k)  \Phi(\Xmii{(s)}{}(k) ) -  \Wmii{(l)}{s}(0)  \Phi(\Xmii{(s)}{}(0) )   \right\|_F \leq \frac{1}{\alphai{3}}
	\left(1+ \alphaii{}{2}+\alphaii{}{3}\mu\sqrt{\kc} \left( \taum \rc + \cwo\right)  \right)^{l}  \taum \sqrt{\kc m} \rcc,
	\end{split}
	\end{equation*} 
	where $\alphaii{}{2}=\max_{s,l}\alphaii{(l)}{s,2}$ and $\alphaii{}{3}=\max_{s,l} \alphaii{(l)}{s,3}$, and $\cxo\geq 1$ is given in Lemma~\ref{boundofinitialization}.
	Then we upper bound 
	\begin{equation*}
	\begin{split}
	&\left\| \Wmii{(s)}{l}(k)  \Phi(\Xmii{(l)}{i}(k))-\Wmii{(s)}{l}(0)  \Phi(\Xmii{(l)}{i}(0)) \right\|_F
	\leq    \frac{1}{\alphai{3}}
	\left(1+ \alphaii{}{2}+\alphaii{}{3}\mu\sqrt{\kc} \left( \taum \rc + \cwo\right)  \right)^{l}  \taum \sqrt{\kc m} \rcc.
	\end{split}
	\end{equation*} 
	Therefore, we have 
	\begin{equation*}
	\begin{split}
	D	 \leq  & \mu \sqrt{m}\rcc  \|  \Cm_0  \|_2 \!+\!2\sqrt{m}\cwo \left( \mu \| \Cm_k\!-\!\Cm_0 \|_F \!+\!  \frac{\rho  \|  \Cm_0  \|_{\infty}}{\alphai{3}}
	\left(1\!+ \!\alphaii{}{2}\!+\!\alphaii{}{3}\mu\sqrt{\kc} \left( \taum \rc \!+\! \cwo\right)  \right)^{l}  \taum \sqrt{\kc m} \rcc\right)\\
	\end{split}
	\end{equation*} 
	
	By combining the above results, we have 
	\begin{equation*}
	\begin{split}
	&	\left\| \frac{\partial \ell }{\partial \Xmii{(l)}{i}(k)} - \frac{\partial \ell }{\partial \Xmii{(l)}{i}(0)}   \right\|_F\\
	\leq &c_1+   \sum_{s=l+1}^{h} \left[   \left(\alphaii{(s)}{l,2} +  2\alphaii{(s)}{l,3} \sqrt{\kc} \mu\cwo\right) \left\| \frac{\partial \ell }{\partial \Xmii{(s)}{i}(k) } - \frac{\partial \ell }{\partial \Xmii{(s)}{i}(k) } \right\|_F+ c_2 \right]\\
	\leq &c_1+   \sum_{s=l+1}^{h} \left[   \left(\alphai{2} +  2\alphai{3} \sqrt{\kc} \mu\cwo\right) \left\| \frac{\partial \ell }{\partial \Xmii{(s)}{i}(k) } - \frac{\partial \ell }{\partial \Xmii{(s)}{i}(k) } \right\|_F+ c_3 \right]\\
	\leq &  \left(1+ \alphai{2} +  2\alphai{3} \sqrt{\kc} \mu\cwo\right)^{l} \left[  	\left\| \frac{\partial \ell }{\partial \Xmii{(h)}{i}(k)} - \frac{\partial \ell }{\partial \Xmii{(h)}{i}(0)}   \right\|_F+ c_3 \right]\\
	\end{split}
	\end{equation*} 	
	where $c_1=4\sqrt{m} \rcc \left( \cwo    \sqrt{m} h  \left( \cxo +\cwo c ( 1 +2 \alphai{3}     \cxo   ) \mu \sqrt{\kc } \right)    + |\ui{i}(0) -\ymi{i} |  \right) $,  $c_2=\alphaii{(s)}{l,3}  \left(  \mu \rcc  \|  \Cm_0  \|_2 +2\cwo \frac{\rho  \|  \Cm_0  \|_{\infty}}{\alphai{3}}
	\left(1+ \alphaii{}{2}+\alphaii{}{3}\mu\sqrt{\kc} \left( \taum \rc + \cwo\right)  \right)^{l}  \taum \sqrt{\kc m} \rcc\right) $ and $c_3=\alphai{3}   \left(  \mu \rcc  \|  \Cm_0  \|_2 +2\cwo \frac{\rho  \|  \Cm_0  \|_{\infty}}{\alphai{3}}
	\left(1+ \alphaii{}{2}+\alphaii{}{3}\mu\sqrt{\kc} \left( \taum \rc + \cwo\right)  \right)^{l}  \taum \sqrt{\kc m} \rcc\right) $. Consider $\|  \Cm_0  \|_2 =\Oc{\sqrt{m}}$,  for brevity, we ignore constants and obtain
	
	\begin{equation*}
	\begin{split}
	\left\| \frac{\partial \ell }{\partial \Xmii{(l)}{i}(k)} - \frac{\partial \ell }{\partial \Xmii{(l)}{i}(0)}   \right\|_F
	\leq    c_1 c\alphai{3}   \cwo^2\cxo  \rho   \kc m \taum \rcc,
	\end{split}
	\end{equation*} 	
	where $c=\left(1+ \alphai{2} +  2\alphai{3} \sqrt{\kc} \mu\cwo\right)^{l}$ and $c_1$ is a constant. 
	The proof is completed. 
\end{proof}

\subsection{Proof of Lemma~\ref{initialbound}}\label{proofoinitialbound}
\begin{proof}
	By Assumption~\ref{initilizationassumption}, each entry for the initial parameter $\Wmii{(l)}{s} (0)$ obeys Gaussian distribution $\mathcal{N}(0,1)$. Then $\|\Wmii{(l)}{s} (0)\|_F^2$ is chi-square variable with freedom degree $\kc p m$. In this way, by using Lemma~\ref{chisquare}, we have 
	\begin{equation*}
	\Pro\left(\|\Wmii{(l)}{s} (0)\|_F^2 -\kc p m\geq  2 \sqrt{\kc p m t} +2t \right) \leq \exp(-t).
	\end{equation*}
	Therefore, with probability at least  $1- \frac{\delta}{2h(h+3)}$, we can obtain
	\begin{equation*}
	\|\Wmii{(l)}{s} (0)\|_F \leq  \sqrt{\kc p m + 2 \sqrt{\kc p m \log(2h(h+3)/\delta) } +2\log(2h(h+3)/\delta)} \leq \sqrt{m}\cwo,
	\end{equation*}
	where $\cwo \sim \sqrt{\kc p} $ is a constant. Note here we focus on $m$ more than $p$ and $\kc$, since $m$ is much larger than $p$ and $\kc$ which is introduced in subsequent analysis. 
	
	By using the same method, we can prove that with probability at least  $1- \frac{\delta}{2h(h+3)}$, 
	\begin{equation*}
	\|\Wmii{0}{} (0)\|_F  \leq \sqrt{m}\cwo \quad \text{and} \quad \|\Umi{s}(0)\|_F  \leq \sqrt{m}\cwo 
	\end{equation*}
	In this way, with probability at least $\left(1-   \frac{\delta}{2h(h+3)}\right)^{\frac{h(h+3)}{2}}\geq  1- \frac{\delta}{2h(h+3)} {\frac{h(h+3)}{2}}=1-\delta/4$, these results hold at the same time. The proof is completed.  
\end{proof}

\end{document}

%% file: MACPdef-ams_manu.tex
\newcommand{\op}{\mbox{op}}

\newcommand{\traceop}{\operatorname{Tr}}
\newcommand{\tr}[1]{\ensuremath{\traceop\left(#1\right)}}

\newtheorem{thm}{Theorem}
\newtheorem{lem}{Lemma}
\newtheorem{assum}{Assumption}


%% file: MACPdef_manu.tex
\DeclareMathOperator*{\argmin}{argmin}

\newcommand{\led}[1]{\overset{\text{\ding{#1}}}{\leq}}

\newcommand{\lee}[1]{\overset{\text{\ding{#1}}}{=}}
\newcommand{\ged}[1]{\overset{\text{\ding{#1}}}{\geq}}

\newcommand{\Rs}[1]{{\mathbb{R}^{#1}}}

\newcommand{\Sii}[2]{S^{#1}_{#2}}

\newcommand{\bmi}[1]{\bm{b}_{#1}}

\newcommand{\umi}[1]{\bm{u}^{#1}}

\newcommand{\LL}{\mathcal{L}}

\newcommand{\vect}[1]{{\textsf{vec}\left(#1\right)}}
\newcommand{\Deltamii}[2]{{\bm{\Delta}_{#1}^{#2}}}

\newcommand{\zeroo}[1]{\textsf{zero}{(#1)}}
\newcommand{\skipo}[1]{\textsf{skip}{(#1)}}
 
\newcommand{\convs}{\textsf{conv}}
\newcommand{\convo}[1]{\textsf{conv}{(#1)}}
\newcommand{\convb}[1]{\Psi{\left(#1\right)}}

\newcommand{\taum}{}

\newcommand{\cwo}{c_{w0}}
\newcommand{\cxo}{c_{x0}}
\newcommand{\cuo}{c_{u0}}
\newcommand{\cgo}{c_{u0}}

\newcommand{\alphai}[1]{\bm{\alpha}_{#1}}
\newcommand{\alphaii}[2]{\bm{\alpha}^{#1}_{#2}}
\newcommand{\alphanii}[2]{\bm{\alpha}^{{\scriptscriptstyle (}#1{\scriptscriptstyle )}}_{\mathsmaller{#2}}}
\newcommand{\betaii}[2]{\bm{\beta}^{#1}_{#2}}
\newcommand{\betanii}[2]{\bm{\beta}^{{\scriptscriptstyle (}#1{\scriptscriptstyle )}}_{\mathsmaller{#2}}}
\newcommand{\gii}[2]{\bm{g}^{#1}_{#2}}
\newcommand{\gnii}[2]{\bm{g}^{{\scriptscriptstyle (}#1{\scriptscriptstyle )}}_{#2}}
\newcommand{\gbii}[2]{\bar{\bm{g}}^{#1}_{#2}}
\newcommand{\gnbii}[2]{\bar{\bm{g}}^{{\scriptscriptstyle (}#1{\scriptscriptstyle )}}_{#2}}
\newcommand{\gtii}[2]{\tilde{\bm{g}}^{#1}_{#2}}
\newcommand{\gntii}[2]{\tilde{\bm{g}}^{{\scriptscriptstyle (}#1{\scriptscriptstyle )}}_{#2}}
\newcommand{\alpham}{\bm{\alpha}}

\newcommand{\betam}{\bm{\beta}}

\newcommand{\sigmai}[1]{\sigma{(#1)}}
\newcommand{\m}{m}
\newcommand{\p}{p}

\newcommand{\kc}{k_c}

\newcommand{\sco}{s_c}

\newcommand{\pc}{p_c}
\newcommand{\ui}[1]{u_{#1}}

\newcommand{\hii}[1]{u_{#1}}
\newcommand{\Omegam}{\bm{\Omega}}
\newcommand{\Omegami}[1]{\bm{\Omega}_{#1}}
\newcommand{\Nn}{\mathcal{N}}
\newcommand{\Var}{\textsf{Var}}

\newcommand{\emm}{\bm{e}}

\newcommand{\xm}{\bm{x}}
\newcommand{\ym}{\bm{y}}
\newcommand{\ymm}{\bm{y}}
\newcommand{\hmm}{\bm{u}}

\newcommand{\um}{\bm{u}}

\newcommand{\ymi}[1]{y_{#1}}
\newcommand{\hmi}[1]{u_{#1}}

\newcommand{\Am}{\bm{A}}
\newcommand{\Bm}{\bm{B}}
\newcommand{\Cm}{\bm{C}}

\newcommand{\Hm}{\bm{H}}
\newcommand{\Imm}{\bm{I}}
\newcommand{\Mm}{\bm{M}}
\newcommand{\Nm}{\bm{N}}
\newcommand{\Gm}{\bm{G}}
\newcommand{\Xm}{\bm{X}}

\newcommand{\Zm}{\bm{Z}}
\newcommand{\Um}{\bm{U}}

\newcommand{\Wm}{\bm{W}}
\newcommand{\Km}{\bm{K}}

\newcommand{\SSm}{\bm{S}}
\newcommand{\Rm}{\bm{R}}

 \newcommand{\rc}{r}
  \newcommand{\rcc}{\widetilde{r}}

\newcommand{\Ami}[1]{\bm{A}_{#1}}
\newcommand{\Bmi}[1]{\bm{B}_{#1}}

\newcommand{\Hmi}[1]{\bm{H}_{#1}}

\newcommand{\Xmb}{\bar{\bm{X}}}
\newcommand{\Xmi}[1]{\bm{X}_{#1}}
\newcommand{\Xmbi}[1]{\bar{\bm{X}}_{#1}}
\newcommand{\Xmbii}[2]{\bar{\bm{X}}^{#1}_{#2}}
\newcommand{\Kmii}[2]{\bm{K}^{#1}_{#2}}
\newcommand{\Qmii}[2]{\bm{Q}^{#1}_{#2}}
\newcommand{\Gmii}[2]{{\bm{G}_{#1}^{#2}}}
\newcommand{\Gmbii}[2]{{\bar{\bm{G}}_{#1}^{#2}}}
\newcommand{\Xmii}[2]{\bm{X}^{#1}_{#2}}
\newcommand{\Xnii}[2]{\bm{X}^{{\scriptscriptstyle (}#1{\scriptscriptstyle )}}_{\mathsmaller{#2}}} 

\newcommand{\Wmii}[2]{\bm{W}^{#1}_{#2}}
\newcommand{\Wmi}[1]{\bm{W}_{#1}}
\newcommand{\Wnii}[2]{\bm{W}^{{\scriptscriptstyle (}#1{\scriptscriptstyle )}}_{\mathsmaller{#2}}}
\newcommand{\Mmii}[2]{\bm{M}^{#1}_{#2}}
\newcommand{\Nmii}[2]{\bm{N}^{#1}_{#2}}
\newcommand{\Amii}[2]{\bm{A}^{#1}_{#2}}
\newcommand{\Amtii}[2]{\widehat{\bm{A}}^{#1}_{#2}}
\newcommand{\Mmtii}[2]{\widehat{\bm{M}}^{#1}_{#2}}
\newcommand{\Nmtii}[2]{\widehat{\bm{N}}^{#1}_{#2}}
\newcommand{\Kmbii}[2]{\bar{\bm{K}}^{#1}_{#2}}
\newcommand{\Kmtii}[2]{\widehat{\bm{K}}^{#1}_{#2}}
\newcommand{\Qmbii}[2]{\bar{\bm{Q}}^{#1}_{#2}}
\newcommand{\bmtii}[2]{\widehat{\bm{b}}^{#1}_{#2}}

\newcommand{\bmii}[2]{{\bm{b}}^{#1}_{#2}}
\newcommand{\Qmtii}[2]{\widehat{\bm{Q}}^{#1}_{#2}}

\newcommand{\Umi}[1]{\bm{U}_{#1}}

\newcommand{\Zmi}[1]{\bm{Z}_{#1}}

\newcommand{\Kmti}[1]{\bm{K}^{(#1)}}

\newcommand{\bmbi}[1]{\widehat{\bm{b}}_{#1}}

\newcommand{\Oc}[1]{\mathcal{O}\left(#1\right)}

\newcommand{\W}{\mathcal{W}}
\newcommand{\N}{\mathcal{N}}

\newcommand{\D}{\mathcal{D}}
\newcommand{\B}{\mathcal{B}}
\newcommand{\Om}{\mathcal{O}}

\newcommand{\EE}{\mathbb{E}}

\newcommand{\Pro}{\mathbb{P}}

\newcommand{\Wi}[1]{\mathcal{W}^{(#1)}}
\newcommand{\Wii}[2]{\mathcal{W}^{(#1)}_{#2}}

\newcommand{\Wmbii}[2]{\bar{\bm{W}}^{#1}_{#2}}
\newcommand{\Wmtii}[2]{\widetilde{\bm{W}}^{#1}_{#2}}